\documentclass[12pt,reqno]{amsart}

\usepackage[left=1in,right=1in,top=1in,bottom=1in]{geometry}
\usepackage{amsmath, amssymb, graphicx, url, mathtools}
\usepackage[round]{natbib}
\usepackage{xcolor}
\usepackage{wrapfig}
\usepackage{tikz}
\usepackage{mathptmx}
\usepackage{tikz}
\usepackage{multirow}
\usepackage{xr}
\usepackage{mathtools}

\makeatletter
\newcommand*{\addFileDependency}[1]{% argument=file name and extension
  \typeout{(#1)}
  \@addtofilelist{#1}
  \IfFileExists{#1}{}{\typeout{No file #1.}}
}
\makeatother

\newcommand{\ubar}[1]{\underaccent{\bar}{#1}}

\newcommand{\vertiii}[1]{{\left\vert\kern-0.25ex\left\vert\kern-0.25ex\left\vert #1 
    \right\vert\kern-0.25ex\right\vert\kern-0.25ex\right\vert}}
    
\usepackage{natbib}
\usepackage{graphicx}
\usepackage{subcaption}
\usepackage{color}
\usepackage{wrapfig}
\usepackage{mathtools}
\usepackage{placeins}
\usepackage[framemethod=TikZ]{mdframed}
\usepackage{xcolor}
\usepackage{tikz}
\usetikzlibrary {positioning}

\usepackage{bm}
\usepackage{paralist}
\usepackage{accents}
\usepackage{enumerate}
\usepackage{caption}

\renewcommand{\qed}{\hfill{\tiny \ensuremath{\blacksquare} }}%
\renewcommand{\Pr}{{\mathrm{P}}}

\newcommand{\Ep}{\mathrm{E}}

\newcommand{\ENT}{{\mathbb{E}_{NT}}}
\newcommand{\GNT}{{\mathbb{G}_{NT}}}

\newcommand{\G}{\mathbb{G}}

\newcommand{\E}{\mathbb{E}}
\usepackage{nccmath}
		{\end{pmatrix}\end{medsize}}%

\newtheorem{theorem}{Theorem}
\newtheorem{assumption}{Assumption}

\newtheorem{corollary}{Corollary}
\newtheorem{lemma}{Lemma}
\newtheorem{algorithm}{Algorithm}

\theoremstyle{definition}
\newtheorem{example}{Example}

\newtheorem{remark}{Remark}

\newtheorem{definition}{Definition}[section]

\renewcommand{\thesection}{\arabic{section}}
\renewcommand{\theequation}{\arabic{section}.\arabic{equation}}
\renewcommand{\thetheorem}{\arabic{section}.\arabic{theorem}}
\renewcommand{\theassumption}{\arabic{section}.\arabic{assumption}}

\renewcommand{\thecorollary}{\arabic{section}.\arabic{corollary}}
\renewcommand{\thelemma}{\arabic{section}.\arabic{lemma}}

\renewcommand{\theremark}{\arabic{section}.\arabic{remark}}

\DeclareMathOperator{\eig}{eig}

\begin{document}

\setcounter{page}{1}

\title[Inference in High-Dimensional Dynamic Panels
]{Inference on Heterogeneous Treatment Effects in High-Dimensional Dynamic Panels under Weak Dependence
}%\thanks{}

% VS: 

\author{ Vira Semenova}\thanks{University of California, Berkeley, semenovavira@gmail.com} 
\author{Matt Goldman}\thanks{Meta, mattgoldman5850@gmail.com} 
\author{Victor Chernozhukov}\thanks{MIT, vchern@mit.edu} 
\author{Matt Taddy}\thanks{Amazon,  mataddy@gmail.com }
%\today
\date{\today; Initial ArXiv Submission: December 2017. The authors are grateful to Michael Jansson, Pat Kline,  Sylvia Klosin,  Chris Taber, and two anonymous referees whose comments helped improve the paper.}
\makeatletter
\def\thanks#1{\protected@xdef\@thanks{\@thanks
        \protect\footnotetext{1}}}
\makeatother

\maketitle

\begin{abstract}
This paper provides estimation and inference methods for conditional average treatment effects (CATE) characterized by a high-dimensional parameter in both homogeneous cross-sectional and unit-heterogeneous dynamic panel data settings.   In our leading example, we model CATE by interacting the base treatment variable with explanatory variables.  The first step of our procedure is orthogonalization, where we partial out the controls and unit effects from the outcome and the base treatment and take the cross-fitted residuals.  This step uses a novel generic cross-fitting method we design for weakly dependent time series and panel data. This method  "leaves out the neighbors" when fitting nuisance components, and we theoretically power it by using Strassen's coupling.
 As a result, we can rely on any modern machine learning method
 in the first step, provided it learns the residuals well enough.  Second, we construct an orthogonal (or residual) learner of CATE -- the Lasso CATE -- that regresses the outcome residual on the vector of interactions of the residualized treatment with explanatory variables. If  the complexity of CATE function is simpler than that of the first-stage regression, the orthogonal learner converges faster than the single-stage regression-based learner.   Third, we perform simultaneous inference on parameters of the CATE function using debiasing.  We also can use ordinary least squares in the last two steps when CATE is low-dimensional.   In heterogeneous panel data settings, we model the unobserved unit heterogeneity as a weakly sparse deviation from \cite{mundlack:1978}'s model of correlated unit effects as a linear function of time-invariant covariates and make use of L1-penalization to estimate these models.   We demonstrate our methods by estimating price elasticities of groceries based on scanner data.  We note that our results are new even for the cross-sectional (i.i.d) case. \\
 
Keywords: Orthogonal Learning, Residual Learning, CATE, High-dimensional, Cross-section Data, Dynamic Panel Data, Weakly Dependent Time Series,  Neighbours-Left-Out, Cross-fitting.

\end{abstract}
% , Awesome. :)

\newpage

\section{Introduction}

Inference on heterogeneous treatment effects is an important problem (see, e.g, \cite{AtheyImbens},  \cite{chernozhukovdemirer}, \cite{AtheyWager}, \cite{DavisHeller}, \cite{Banerjeeetal}). Estimating these effects involves an inherent trade-off between flexibility and precision.  On the one hand, discovering heterogeneity requires flexible models for the effects (e.g., by considering many groups). 
 On the other hand,  flexible models produce noisy estimates that are not useful for making decisions (e.g., the noise can result from having too few observations per group).  To resolve this trade-off,  empiricists decide how to create groups after making multiple attempts, a subjective, labor-intensive method that is prone to erroneous inference.
 
 This paper contributes a method for the estimation and inference of heterogeneous treatment effects in a panel data set with many potential controls and unobserved unit heterogeneity, which addresses many of the challenges listed above.  Our key results are new even to the cross-sectional settings.  Thus, we first consider the cross-sectional case and the following leading example as our model to explain the key ideas.  Suppose $Y$ is an outcome, and $P$ is a vector of treatment variables (for example, polynomials in price), and $X$ are controls.  Suppose that the conditional expectation function is partially linear in $P$, as in \cite{robinson:88}, namely
$$
Y = e_0(X) + \beta_0(X)'P +  U, \quad \Ep[U \mid P, X] =0.
$$

Here $e_0(X)$ is the conditional average outcome in the ``untreated" state ($P=0)$, and $\beta_0(X)$ describes the conditional average treatment effect under the standard conditional ignorability/exogeneity conditions.  We can \textit{orthogonalize} the outcome and treatment with respect to controls $X$,
obtaining the residuals $\widetilde Y = Y - \Ep[Y \mid X]$ and $\widetilde P = P -
\Ep[P|X],$ and then observe that the CATE function satisfies the equation:
$$
\widetilde Y = \beta_0(X)'\widetilde P + U.
$$
Therefore we can learn the CATE function from this regression equation if  
we can learn the residuals $\widetilde Y$ and $\widetilde D$ well using modern machine learning methods.  In fact, under
certain conditions, we prove that our rate of learning the CATE function would be the same as if we knew the true residuals, which is an oracle-type property.  Moreover, we show this under both high-level and low-level regularity conditions.

Our approach consists of approximating $\beta_0(X)$ by a linear combination of terms of a dictionary of transformation $K=K(X)$ of $X$, which includes a constant of $1$,
$$
\beta_{0}(X) \approx K'\beta_0,
$$
If dimension $d = \dim (\beta_0)$  is low, that is, $d$ is much smaller than the sample size $n$, we can learn $\beta_0$ using least squares at the rate $\sqrt{d/n}$ provided the expectation functions $\Ep[Y \mid X]$ and $\Ep [P \mid X]$ are learnable at fast enough rates.  If $\beta_0$ is high-dimensional and sparse, we will rely on Lasso to learn the CATE function at the rate $\sqrt{s \log d/n}$ where $s = \| \beta_0 \|_0$ is the number of non-zero entries of $\beta_0$. Finally, we will use debiased Lasso methods to perform Gaussian inference on the components of $\beta$, including constructing simultaneous confidence bands using fast (Gaussian) bootstrap methods.  We call these new approaches above the Orthogonal Lasso and Debiased Orthogonal Lasso.  In addition, we also explore the use of grouped Lasso methods to enforce the exclusion or inclusion of groups of variables.

Our paper considers the dynamic panel data setting arising in many empirical applications.  This setup makes the problem a lot more challenging.  First, all variables above will be doubly-indexed by unit $i=1,\ldots,N$ and time $t=1,\ldots,T$, and controls can include lagged values of outcomes, for example, and we will need to introduce unit-specific effects in the model above judiciously.  We add the unit-specific effects to the conditional expectations of $Y_{it}$ and $P_{it}$, and we model individual effects as linear functions of time-invariant covariates plus fixed effects that are approximately sparse.  This constriction allows for the overall individual effect to be ``dense" while providing enough convenience to make estimation results work.  The strategy above is motivated by Mundlack's and Chamberlain's approach to viewing and modeling fixed effects as correlated random effects.

Our construction uses cross-fitting (CF) to estimate nonparametric reduced forms on a subset of data and construct the residuals (or scores) on another subset.  In the i.i.d.  setting, CF removes the overfitting biases that can arise from using complex nonparametric methods such as machine learning methods (see, e.g., \cite{bch:2010,zheng:laan,chernozhukov2016double} for recent examples and \cite{hasminskii:debiased,schick1986asymptotically} for early, classical uses).  For regular CF methods to work in a  time-series or unit-heterogeneous dynamic panel data,  the number of  periods $T$ must be very large relative to the number of units $N$.  We choose an alternative path and introduce a "neighbors-left-out" (NLO) cross-fitting method that applies to weakly dependent data.  The NLO approach ensures that the first-stage and the second-stage samples are approximately independent.  We provide exact bounds on the approximation error via Strassen's coupling.  These results are of independent interest and apply beyond our context.

We use our method to estimate heterogeneous price elasticities on grocery data as an empirical application.   This dataset consists of textual descriptions of the products, prices, and daily aggregate sales for each (store, product, distribution channel) combination.  We posit a partially linear specification where the (log) sales are the dependent variable, and lags of log prices and log sales and current product characteristics are the control variables.  Assuming that the residual between the price tomorrow and its expectation today is exogenous, we use this variation to identify price elasticities.  The approximate sparsity assumption helps us to rule out implausible values of price elasticities.  Our estimates are broadly consistent with findings in \cite{AER2003}.

All of the above constitute the principal contributions of the paper.  In what follows, we describe the relations to the literature and mention some additional extensions and contributions.  First, the paper contributes to modern literature on estimation and model selection in high-dimensional settings using debiased (orthogonal) machine learning (e.g., \cite{hasminskii:debiased,schick1986asymptotically,bch:2010,zheng:laan,belloni2011inference,BCK:BMKA,ZhangZhang,geer,chernozhukov2016double} and references therein) by considering the high-dimensional CATE function as the focus of inference.  Prior literature has mainly focused inference on low-dimensional or many target parameters without leveraging the model to help residualization (e.g., \cite{BCK:BMKA,BCCW}).  While our results are new, even for cross-sectional settings, our results cover the dynamic panel data settings.  

We provide general theoretical guarantees for Orthogonal Lasso methods that apply to any case where residuals 
are learned well enough in a preliminary step using general machine learning methods.  In cross-sectional settings, this automatically allows a wide range of high-quality machine learning tools with rigorous guarantees.  In panel data settings, we rely on Lasso and verify that we can learn the residuals well using lasso-based methods with weakly sparse individual effects, relying here upon in \cite{KockTang}.  We expect that other machine learning methods are potentially amenable to handling dynamic panel data settings, which is the subject of future work.  In this work, we abstract away from clustering (\cite{Chiang} and \cite{chiang2019multiway}), but it would be good to extend the present results in this direction.

In a related paper to ours, \cite{NieWager} establishes that the oracle rate of learning of CATE function is possible in a cross-sectional setting, proposing a similar residual regression approach.  Our paper is independent, and we circulated the paper around the same time as theirs (both in December of 2017 in ArXiv).  Moreover, we provide not only the oracle learning rates but also statistical inference results and also cover the dynamic panel data setting.  On the other hand, rate results of \cite{NieWager} apply to nonlinear learners of the CATE function.  A more recent work than ours is \cite{OprescuWu}, which develops orthogonal forest methods.  Specifically, they apply generalized random forest to regress outcome residual on treatment residual interacted with a forest function of controls.  They also provide some inferential results.  Finally, alternative approaches to handling heterogeneous and/or continuous treatment effects are discussed in  \cite{Ura}, \cite{AtheyWager},  \cite{SemCher}, \cite{Hardle2020}, \cite{Lieli},  \cite{ZimLech},  \cite{Colangelo}, \cite{Klosin}.

To conduct inference on high-dimensional parameters of the CATE function,  we combine the approach of  \cite{ZhangZhang}  and \cite{geer} with the \cite{CLIME}'s approach to matrix inversion.   The inference step can also be carried out by the methods of \cite{JM} and the double lasso method (\cite{BelCherHan,CHS}), but we focus on the former.  We rely on fast (Gaussian) bootstrap to perform simultaneous inference based on many debiased Lasso estimators, relying upon (\cite{CCKAS, CCKAP,CCK}) and suitably extending some results to our settings.  Finally, we build on (\cite{mundlack:1978}, \cite{Chamberlain:1982}) and contribute to the panel data literature that develops various approaches to handling heterogeneity,  e.g., \cite{Kock}, \cite{Manresa3}, \cite{SuLu2}, \cite{SuShi}, \cite{WeidnerMoonShum}, \cite{KockTang}, \cite{Manresa2,Manresa4}, \cite{GaoLi}, \cite{WeidnerChen},  \cite{SuLu}) among many others, see \cite{WeidnerARE} for a recent overview.  %Finally, this paper contributes to a literature (\cite{CJN},  \cite{Lei}, \cite{KlineSolvtsen}, \cite{Solvtsen}) on estimation and heteroscedasticity-robust inference 
%in the moderate-dimensional regime. 

\subsection*{Structure of the paper.} Section \ref{sec:section2} introduces the model and outlines the strategy.  Section \ref{sec:overview} gives definitions of estimators and outlines some theoretical results.  Section \ref{sec:theory} states our theoretical results under general high-level conditions about the first stage.  Section \ref{sec:fs} verifies the high level conditions focusing on the panel data settings.  Section \ref{sec:empirical} gives an empirical application, and Section \ref{sec:extension} concludes.

\section{The Set Up}
\label{sec:section2}
Here we present the model, explain how we handle unit level heterogeneity, and outline the overall inferential strategy.
\setcounter{equation}{0}
\setcounter{definition}{0}

\subsection*{Model}
Our starting point is the structural equation model \begin{equation}
\label{eq:APLM-1}
Y_{it} = \beta_0({X}_{it}, P_{it}) + e_{0}({X}_{it}) + \xi_i^E + U_{it},
\end{equation}
where $i = 1,2, \dots, N$ and $t=1,2, \dots, T$. Here, \begin{itemize}
    \item $Y_{it}$ is a scalar outcome of unit $i$ at time $t$;
    \item $P_{it} \in \mathrm{R}^{d_p}$ is a vector of treatment or policy variables; 
    \item ${X}_{it} \in  \mathrm{R}^{d_X}$ is a vector  of \textit{predetermined} variables, including possibly the lags of $P_{it}$ and $Y_{it}$;
    \item $ \xi_i^E$ is an  unobserved outcome unit fixed effect;  
    \item $M_i = \{M_{it}\}_{t=1}^T$ is a a collection fixed variables;
    \item ${X}_{it}$ can include known functions $M_i$, e.g. time averages $\bar M_i$ of $M_i$.
    \end{itemize}
The stochastic shock $U_{it}$ is assumed to  satisfy  the following sequential conditional exogeneity condition
 \begin{equation}
 \label{eq:condexog}
  \Ep  [U_{it} \mid P_{it},  {X}_{it}, \Phi_{it}]=0, \quad \forall (i, t)
  \end{equation}
 where  the filtration 
 \begin{equation}
 \Phi_{it}=\{  ({X}_{it'},  P_{it'},Y_{it'})_{t'=1}^{t-1}\}
 \end{equation}
 is the filtration that consists of  predetermined variables for unit $i$ prior to period $t$.  Here we view $M_i=\{M_{it}\}_{t=1}^T$'s as a fixed realization of strictly exogenous variables that can be time-varying. These variables are strictly exogenous, meaning that their entire trajectory has been pre-determined relative to all other variables in the model and relative to stochastic shocks $U_{it}$'s.

\begin{remark}[Fixed Effects]
Throughout the paper we assume that
$$
\{M_i, \xi_i\}_{i=1}^N \text{ are fixed.}
$$
We view this approach as (essentially) equivalent to treating these variables as random initially and then performing the analysis conditional on their realized values.\footnote{As in the standard panel data modes with fixed effects, we do not formalize this conditioning to reduce presentation complexity.}
\end{remark}

\begin{remark}[Important Notation Remark] Note that below we will be reassigning notation $X_{it} \leftarrow \mathsf{t}({X}_{it})$ to denote variables that have been obtained as transformations of the original variables ${X}_{it}$ via some mapping $\mathsf{t}$. Examples of transformations include powers and their interactions. We then shall make other modeling assumptions to model the observable unit-level heterogeneity.
\end{remark}

The structural function $p \mapsto \beta_0(x,p)$ encodes the conditional average treatment effects (CATE). Therefore, we will simply call this function the CATE function.  Indeed consider the intervention policy that fixes $P_{it} = p$ in the structural equation (\ref{eq:APLM-1}), inducing the potential outcome:\footnote{Here, as in \cite{Haavelmo},  we assume that the structural equation remains invariant under the intervention. Judea Pearl  refers to the fact that the structural model implies potential outcomes as the first law of causal inference.}
$$
Y_{it}(p) :=
\beta_0(X_{it}, p) + e_{0}(X_{it}) + \xi_i^E + U_{it}.
$$
Then we have that
$$
\beta_0(X_{it}, p_1)- \beta_0(X_{it}, p_0)= \Ep[ Y_{it}(p_1) \mid X_{it}]- \Ep[ Y_{it}(p_0) \mid X_{it}]
$$
is the CATE resulting from changing policy value from $p_0$ to $p_1$.

In what follows, we will assume that  $\beta_0(X_{it}, P_{it})$ is well approximated by a linear combination of terms of a dictionary $$ D_{it}:= D(X_{it}, P_{it})$$ of transformations of $X_{it}$ and $P_{it}$ so that
$$
\beta_0(X_{it}, P_{it}) = D_{it}'\beta_0.
$$
Putting things together, we arrive at the partially linear model:
\begin{equation}
\label{eq:APLM}
Y_{it} = D_{it}' \beta_{0} + e_{0}(X_{it}) + \xi_i^E + U_{it},
\end{equation}
where the key parameter $\beta_0$ is interpretable as a causal or treatment effect parameter. We will refer to $P_{it}$ and  $D_{it}$ as \textit{base} and \textit{technical} treatment vectors, respectively.

 In this paper, we focus on a practical case when the complexity of the control function $e_0(X_{it})$ substantially exceeds the complexity of CATE function (see Remark \ref{rm:improvement} for the formal comparison of complexities).
 
\subsection*{Reduced Forms and Orthogonalized Equations} To learn the CATE function at its fastest possible rate, we need to partial out controls from treatments and outcome. Consider the treatment equation:
\begin{equation}
D_{it} = d_{i0}(X_{it})  + V_{it},  \quad \Ep[ V_{it} \mid X_{it}, \Phi_{it} ]=0,
\label{eq:dtreat}
\end{equation}
which keeps track of confounding. We assume that the \textit{unit-specific treatment reduced form} takes the form:
\begin{align}
\label{eq:reduced_form}
\Ep[D_{it}\mid X_{it}, \Phi_{it}] =:d_{0i}(X_{it}) =d_0(X_{it}; \xi_i),
\end{align}
where $\xi=(\xi_1, \xi_2, \dots, \xi_N)$ denotes a fixed vector of unit-specific fixed treatment-selection effects.  A special case $d_{0i}(X_{it}):=d_0(X_{it})$ corresponds to no unobserved unit heterogeneity in treatment. Furthermore, if the function $d_0(X_{it})$ is constant itself, there is no confounding. 

Proceeding further, we model  \textit{the unit-specific outcome reduced form} as 
\begin{align}
\label{eq:reduced_form:out}
 \Ep[Y_{it}\mid X_{it}, \Phi_{it}] =: l_{i0} (X_{it}) = d_{i0}(X_{it})'\beta_0 + e_0(X_{it}) + \xi_i^E,
\end{align}
where $\xi^E=(\xi^E_1, \xi^E_2, \dots, \xi^E_N)$ denotes a fixed vector of unit-specific outcome effects.  

Given the outcome and treatment reduced forms, we define the treatment and outcome residuals
\begin{align}
\label{eq:resid}
V_{it} &:= D_{it} - d_{i0}(X_{it}), \quad \widetilde Y_{it}:= Y_{it} - l_{i0}(X_{it}).
\end{align} 
Equations \eqref{eq:APLM}-\eqref{eq:condexog} imply the following orthogonolized regression equation:
\begin{align}
\label{eq:lm}
\widetilde{Y}_{it} &=V_{it}' \beta_0 + U_{it}, \qquad \Ep [U_{it} \mid V_{it},X_{it}, \Phi_{it}] = 0.
\end{align}
This equation identifies $\beta_0$ as the coefficient of the best linear projection of $\widetilde Y_{it}$ on $V_{it}$.

\begin{example}[Linear in Treatment Base Treatment Structure]\label{HTE} Define
\begin{align}
 \label{eq:lineartreat}
 D_{it} = P_{it}K_{it},
 \end{align}
where $K_{it}:=K(X_{it})$ is a collection of transformations of a subset of variables in $X_{it}$, including a constant of $1$. Suppose there exists a  low-dimensional ``base" treatment variable $P_{it}$ whose reduced form is 
\begin{eqnarray}
 P_{it} = p_{0}(X_{it})+ \xi_i  + V_{it}^P, \quad \Ep [ V_{it}^P \mid X_{it}, \Phi_{it} ] =0.
 \label{eq:priceout}
\end{eqnarray}
Then, \eqref{eq:dtreat}, \eqref{eq:lineartreat}, and \eqref{eq:priceout}  imply 
\begin{align*}
    d_{i0}(X_{it}) = K(X_{it})' (p_{0}(X_{it})+ \xi_i), \quad 
V_{it} = K(X_{it}) V_{it}^P.
\end{align*}
As we will show later, the interactive structure \eqref{eq:lineartreat} simplifies estimation of treatment residuals. 
\end{example}

\subsection*{Unit-Level Effects}  A standard approach to unit-level additive heterogeneity is demeaning or differencing. Because these operations introduce \cite{Nickel1981} bias in dynamic panels, it requires an identification  strategy based on instrumental variables (e.g., \cite{ArellanoBond}). Furthermore,  differencing out time-invariant covariates may lead to an efficiency loss. 

In this paper, we take a fixed effect approach, in which  we approximate the vector of unobserved components of unit effects $\xi^E=(\xi^E_i)_{i=1}^N$ by  a weakly sparse vector. Informally, the weak sparsity assumption requires  $\xi^E$ to be well approximated by a sparse vector whose number of non-zero components is small. The sparsity assumption allows us to use Lasso methods to consistently estimate them \citep{KockTang}.

The weak sparsity assumption may appear restrictive at the first sight. However, it does allow for rich forms of overall unit-level effects driven by time-invariant covariates and the "residual" unit effects $\xi_i$. To explain this better, consider the following Mundlack-style model:
\begin{align}
\label{eq:cre}
e_0(X_{it}) + \xi^E_i = \bar X_{it}'\delta^X_0 +  \underbracket{\bar{M}_{i}'\delta^E_{M0} + \xi^E_{i}}_{a^E_i},
\end{align}
where $\bar X_{it}$ are time-varying pre-determined covariates and $\bar{M}_{i}= \frac{1}{T} \sum_{t=1}^T M_{it}$ is time average of fixed  covariates.  The important difference with Mundlack's approach is that we consider $\xi_i$'s to be weakly sparse and to condition on the realizations of $(\bar M_i, \xi_i)_{i=1}^N$. \footnote{ It would be interesting to consider non-weakly sparse $\xi$ in \text{dynamic} panel models, where $\xi$'s follow some known distribution (which is generally not compatible with the weak sparsity assumption). We leave this important direction to future research. Note that \cite{kock:2016} developed such results for non-dynamic panel data models, which could provide a starting point for such extension.}

\begin{figure} 
 	\begin{center}
        \includegraphics[scale = 1]{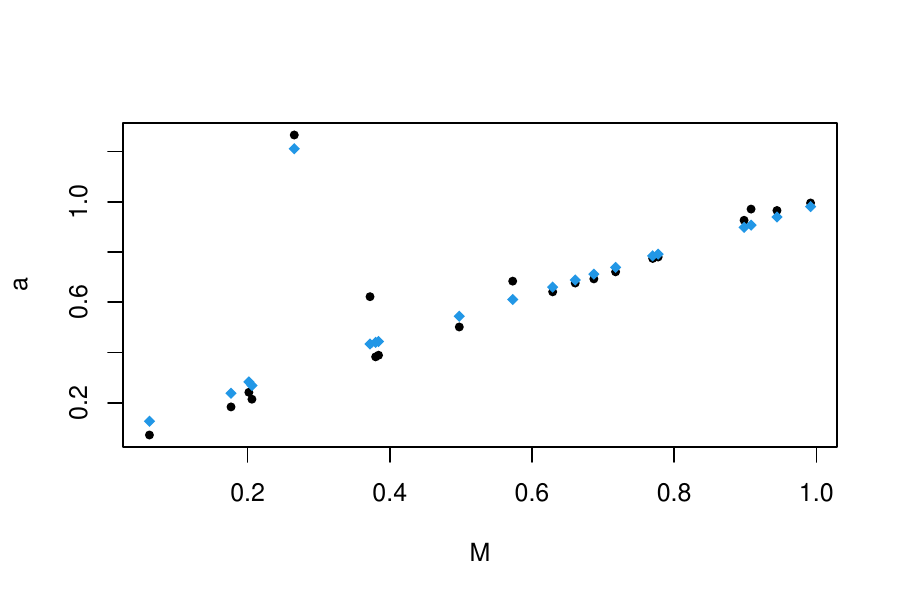}
\end{center}
    \caption{Lasso Approximation of a Correlated Random Effects Model with Approximately Sparse Deviations.}
	\caption*{\small \textbf{Notes:}
	Black dotes indicate $(M_i,a_i)$, with the horizontal axis showing values  of $\bar{M}_{i}$ and vertical axis the values of  $a_i = M_i'\delta_0^M + \xi_i$. The Lasso estimated unit effects are shown by blue rombi $(M_{i}, \widehat a_i)$, where $\widehat a_i:=M_{i}  \widehat{\delta}_M + \widehat \xi_i$. The time-invariant controls $M_{i}$ are generated as i.i.d draws from $U[0,1]$; the sparse deviations are $\xi_i = 1/i^2, i=1,2,\dots, N=20, T=1$; and $\delta^M_0=1$. Here we show the realization just for one experiment.}
	\label{fig:Lasso}
\end{figure}

We note that while the residual effects $\xi_i$'s are required to be weakly sparse, the overall unit effect $a_i$ can actually be \textit{dense}. Finally, the decomposition $a^E_i = \bar{M}_{i}' \delta^E_{M0} + \xi^E_i$ may not be unique. However, our analysis shows that this non-uniqueness does not prevent the overall $a^E_i$'s be consistently estimated, as we illustrate in Figure \ref{fig:Lasso}, as long as there exist at least one decomposition with $\delta^E_{M0}$ and $\xi^E$ being sufficiently sparse. We provide a more technical explanation in Remark \ref{remark:Identification}.

 For the unit-level heterogeneity in treatment, we can proceed similarly.  This strategy works especially well in conjunction with linear structures such as Example \ref{HTE}, where the same approach as above applies, swapping $\xi^E_i$ for $\xi_i$, so that
 \begin{align}
\label{eq:cre-treatment}
p_0(X_{it}) + \xi_i = \bar X_{it}'\delta^P_0 +  \underbracket{\bar{M}_{i}'\delta^P_{M0} + \xi_{i}}_{a^P_i},
\end{align}
where $a^P_i$ is the overall unit-level effect, consisting of a dense part $\bar{M}_{i}'\delta^P_{M0}$ plus a weakly sparse  deviation  $\xi_i$ from it.
 
 Additive unit heterogeneity works well for linear models such as in Example \ref{HTE}.   On the other hand, purely additive fixed effects are not well-suited for binary or discrete treatments.\footnote{For example, in the binary case, the conditional expectation function of $P_{it}$ is naturally bounded by 0 and 1, but the additive fixed effects model does not naturally respect this range.} In the latter case empirical researchers may proceed as follows:  supposing $P_{it}$ is binary, we model  $$
 P_{it} = \Lambda (\xi_i + \bar X_{it}'\beta+ \bar{M}_{i}'\delta)+ V_{it}^P, \quad
 \Ep[V_{it}^P \mid X_{it}, \Phi_{it}] = 0,
 $$
 where $z \rightarrow \Lambda(z)$ is the link function such as logit that forces the logical range restriction on the conditional expectation function.  The fixed effects here are naturally non-additive (though additive inside the link function).  Then here one can still impose approximate sparsity on $\xi = (\xi_i)_{i=1}^N$ and apply Lasso-penalized logistic regression to estimate such models in practice. We expect that the results of \cite{KockTang} extend to this case, but this requires its own formal analysis that we leave to future work.

 The above discussion is still somewhat abstract. We thus present the following concrete example that illustrates the flexibility of the proposed framework.  In this example the Lasso-based methods are particularly helpful for both estimation of reducted forms and residuals. We will use this example to illustrate the plausibility of regularity conditions that we invoke later in the paper.

\begin{example}[Linear Panel Vector Autoregression with High-Dimensional Controls and Unit Effects]\label{ex:VAR}
\label{rm:subgauss}
The following example is the special case of Example \ref{HTE}:
   \begin{align}
    Y_{it} & = P_{it}K_{it}' \beta_0 + e_0(X_{it}) + \xi^E_i + U_{it}
    \nonumber  \\
     &=  P_{it} K_{it}'\beta_0 +\sum_{l=1}^L  Y_{i,t-l} \delta^{EE}_{0l} + \sum_{l=1}^L  P_{i,t-l} \delta^{EP}_{0l} + \bar{X}_{it}'\bar{\delta}^E_0 +{ \bar{M}_{i}'\delta^E_{M0} + \xi^E_i} + U_{it},  \nonumber  \\
      &= P_{it} K_{it}'\beta_0 + X_{it}'\delta^E_0 + \xi^E_i + U_{it},  \label{eq:main1} \\
     P_{it} &= p_0(X_{it}) + \xi_i + V_{it}^P  \nonumber  \\
    &= \sum_{l=1}^L  P_{i,t-l} \delta^{PP}_{0l} + \sum_{l=1}^L  Y_{i,t-l} \delta^{PE}_{0l} + \bar{X}_{it}'\bar{\delta}^P_0 + \bar{M}_{i}'\delta^P_{M0} + \xi_i  + V_{it}^P \nonumber  \\
    & = X_{it}'\delta^P_0 + \xi_i + V_{it}^P.   \label{eq:main2}
    \end{align}
In this example the outcome responds to the current and past values of the treatment as well as past values of outcomes; a set of covariates and unit effects provide further shifts. Likewise, the treatment is assigned in response to  current and past values of the treatment as well as past values of outcomes; and a set of covariates and unit effects provide further shifts.
    \end{example}

\subsection*{Estimation and Inference Strategy} 
 We are primarily interested in \textit{high-dimensional sparse} regime, where the number of technical treatments $d$ is large
$$
d = \dim(D_{it}) = \dim (\beta_0) \gg NT,
$$
but only a small number  $s \ll NT$ of them has non-zero effect\footnote{We can relax this exact sparsity assumption  to approximate/weak sparsity as in \cite{Belloni_2014}. We chose a simpler assumption given the complexity of the rest of the analysis.} 
:
\begin{align}
    \label{eq:sparsity}
    \| \beta_0 \|_0 = s,
\end{align}
Importantly, the identity and the number of the non-zero coefficients is unknown. 

\begin{algorithm} In this high-dimensional regime,  our  estimation and inference approach has the following steps:
\begin{enumerate} 
\item[(1)] Estimate the residuals $\widetilde {Y}_{it}$ and $V_{it}$ using machine learning with cross-fitting.

\item[(2)] Estimate the CATE function by Lasso-penalized regression of estimated  $\widetilde Y_{it}$ on $V_{it}$.

\item[(3)] Perform Gaussian inference parameters of the CATE function using Debiased Lasso.

\end{enumerate}
\end{algorithm}

In the last step, more specifically, we are performing inference on classes of the linear functionals of parameters $\beta_0$ of the CATE function $D_{it}'\beta_0$. In cross-sectional settings, a wide variety of machine learning methods provably apply to carry out step 1. In panel data settings, carrying out step 1 requires a judicious mix of modeling structures and machine learning methods that can handle fixed effects.  Structures such as Example \ref{HTE} and Lasso with penalized fixed effects work provably well for this purpose. Other methods potentially apply, but this remains to be proven.
 Moreover, for step 1 we have to design cross-fitting to respect the panel data structure. The last step uses debiasing most similar to that of \cite{geer}, but other methods such as double lasso can also be used to carry out debiasing. The next section provides formal definitions of estimation steps focusing on the dynamic panel data case.

If $D_{it}$ is low-dimensional, namely
$$
d \ll NT,
$$
the sparsity assumption is not required. The steps (2) and (3) are replaced by linear regression estimated by ordinary least squares.

\begin{algorithm} If $d \ll NT$, we perform the following steps.
\begin{enumerate} 
\item[(1$'$)] Estimate the residuals $\widetilde {Y}_{it}$ and $V_{it}$ using machine learning with cross-fitting.
\item[(2$'$)] Estimate the CATE function by linear regression of estimated  $\widetilde Y_{it}$ on $V_{it}$.
\item[(3$'$)] Perform Gaussian inference on parameters of the CATE function using OLS.
\end{enumerate}
\end{algorithm}

This covers many practical cases, and is a very attractive applied option.  We point  our, however, that even in this regime, if the sparsity condition $s \ll d$ holds,  the Orthogonal Lasso methods can outperform OLS in terms of accuracy of estimating the CATE function. Furthermore, we also note that the OLS method  is not designed to handle the model selection problem. Indeed, typically researchers combine OLS 
with prior model selection step, which can lead to well-known inferential problems \cite{LeebPotcher}. The Debiased Orthogonal Lasso explicitly addresses the model selection issue and provides rigorous theoretical guarantees for inference.

\section{Cross-Fitting with Time Series, Estimation, and Inference}

\setcounter{example}{0}

\label{sec:overview}
In this section we introduce neighbor-excluding cross-fitting method that is generally applicable to weakly dependent time series or panel data. We also provide a key theoretical support for this method using Strassen's coupling. In this section we also write down details of some estimators, focusing on the dynamic panel data case. We also informally  preview theoretical results.  

\subsection*{Notation} In the remainder we use  notation $W_{it}$ to refer to the data vector on unit $i$ at time $t$;  $W_{\cdot,t}$ the data vector on all units at time $t$, and so on.
The data and all other random elements are defined on the underlying probability space $(\Omega, \mathcal{F}, \Pr)$ that has been enriched to carry an independent standard uniform random variable. We assume that all random variables are random vectors in Euclidean spaces (the coupling result below applies to random variables taking values in Polish space). We denote the total variation of a signed measure $v$ the space $(\Omega, \mathcal{F})$ as $$
\|v\|_{TV}:=\sup v(A)-v\left(A^c\right),$$
where the supremum is taken over all measurable sets $A$.

\subsection{Cross-fitting for Weakly Dependent Data} 
\label{sec:fs1}  

Cross-fitting (CF) reduces overfitting biases from fitting the model's nonparametric components via machine learning.  In the i.i.d.  settings, CF uses a one subsample to estimate nonparametric components (e.g., expectation functions) and its complement to compute the sample average of i.i.d residuals (depending on these functions).  As a result, CF plays an essential role in modern debiased inference in semi-parametric models; see, e.g., \cite{bch:2010,zheng:laan,chernozhukov2016double} for recent examples and \cite{hasminskii:debiased} and \cite{schick1986asymptotically} for early, classical uses of simpler sample-splitting methods for debiased inference. 

In cross-sectional cases, we create partitions into folds and their complements by sampling folds randomly from the data. In the time series case, there is only one dimension to split on.  In the  unit-heterogeneous  panel settings, we can only split by time dimension so that unit-specific effects are estimated for all units on every partition.   In both cases, two contiguous time splits may not be independent\footnote{For example, if we cut a panel into two halves, we end up with two dependent data blocks. We can, in principle, make this approach work under beta-mixing by recognizing that the dependence has a vanishing effect on statistics that are sample averages of a.s. bounded random variables. However, such approach would require the number of time periods (i.e., the length of the panel) to be sufficiently large (i.e., $\log N/T = o ( (NT)^{-1/2})$). We avoid assuming these additional unpleasant conditions by using the NLO approach.}.  Here, we introduce a "neighbors-left-out" (NLO) cross-fitting method that applies to weakly dependent data.  Whenever the data are weakly dependent, the NLO approach ensures that the first- and the second-stage samples are approximately independent.  We give exact bounds on the approximation error  by independent blocks via Strassen's coupling below.

\begin{definition}[Folds and Their Quasi-Complements for Weakly Dependent Data]
\label{splitting:panel} 
Consider partition of $\{1,\ldots,T\}$ into adjacent blocks $\{\mathcal{M}_k\}_{k=1}^K$, 
$$\{1,\ldots,T\} = \{\mathcal{M}_1,\ldots, \mathcal{M}_K\},$$
where each block has length  $T_k \geq T_{\text{block}}:=\lfloor T/(K-1) \rfloor$ for each $k$, such that $K \geq 3$. Let $\mathcal{N}(k)$ denote $k$
and its immediate neighbors in $\{1,..,K\}$.  Define the quasi-complement of $\mathcal{M}_k$ as $\mathcal{M}^{\text{qc}}_k =  \{\mathcal{M}_1,\ldots, \mathcal{M}_K\} \setminus  \{\mathcal{M}_{l}: l \in \mathcal{N}(k)\}$, and the corresponding data blocks  $B_k  = \{ W_{\cdot, t}: {t \in  \mathcal{M}_{k} }\}$ and
$B^{\text{qc}}_k  = \{ W_{\cdot, t}: {t \in  \mathcal{M}^{\text{qc}}_{k} }\}$.\end{definition}

The construction creates quasi-complementary sets with the left-out-neighbors.  Since we use the quasi-complementary sets to fit the nonparametric nuisance functions,  we recommend $K\geq 10$, to ensure that at least $70\%$ of data is used for this task.  To clarify the construction further, consider the following example:  suppose we have $K=10$ blocks $\{\mathcal{M}_k\}_{k=1}^{10}$ of adjacent time stamps $t \in \{1,...,T\}$, each of size $T_{\text{block}}$, so that $T= K T_{\text{block}}$. Then, the first quasi-complementary set
$\mathcal{M}^{\text{qc}}_1$ consists of $\{\mathcal{M}_k\}_{k=3}^{10}$, the second set $\mathcal{M}^{\text{qc}}_2$ consists of $\{\mathcal{M}_k\}_{k=4}^{10}$, 
the third set $\mathcal{M}^{\text{qc}}_3$ consists of $\{\mathcal{M}_1\} \cup \mathcal{M}_{k=5}^{10}$, the fourth set $\mathcal{M}^{\text{qc}}_4$ consists of $\{\mathcal{M}_k\}_{k=1}^2 \cup \mathcal{M}_{k=6}^{10}$, \ldots , and the final set $\mathcal{M}^{\text{qc}}_{10}$ consists of $\{\mathcal{M}_k\}_{k=1}^8$.

The following is the application of the NLO cross-fitting method above in our context.

\begin{algorithm}[NLO Cross-Fitted Residuals]
\noindent (1) Construct blocks $(B_k, B^{\text{qc}}_k)$ for $k=1,\ldots,K$ using Definition \ref{splitting:panel}; (2) For each $k$, compute estimators of reduced forms using quasi-complementary sets, namely
	\begin{align*}
	    \widehat d_{ik} (\cdot)= \widehat d_{ik}( \cdot,  B^{\text{qc}}_k), \ \ 
	      \widehat l_{ik} (\cdot) = \widehat l_{ik}( \cdot,  B^{\text{qc}}_k), \quad i=1,2,\dots, N.
	\end{align*}
\noindent (3) Obtain the estimated residuals 
\begin{align}\widehat {\widetilde{Y}}_{it} := Y_{it} -\widehat l_{ik} (X_{it}), \ \  \widehat {V}_{it} := D_{it} -\widehat d_{ik} (X_{it}), \quad i=1,2,\dots, N.
\end{align}
\end{algorithm} 

In the case of the base treatment structure of Example \ref{HTE}, the last step reduces to
$$ 
 \widehat {V}^P_{it} = P_{it} - \widehat p_{ik} (X_{it}), \text{ and } \widehat {V}_{it} := K(X_{it}) \widehat {V}^P_{it}, \quad i=1,2,\dots, N,$$
since we first construct  $\widehat p_{ik}  (X_{it})$ and then set $\widehat d_{ik}(X_{it}) = K(X_{it})\widehat p_{ik}  (X_{it})$.

\subsection{Theoretical Support for the NLO Cross-Fitting Method}
To explain the benefits of the construction, we define some notation.  Suppose $X$ and $Y$ are random elements on the same Polish space.  Define their dependence coefficient (the beta-mixing coefficient) as $$
\gamma(X, Y)=\frac{1}{2}\left\|P_{X, Y}-P_X \times P_Y\right\|_{TV},
$$
where $P_V$ denotes the distribution of the random element $V$.  The dependence coefficient vanishes if and only if $X$ and $Y$ are independent. 

We also make use of the following coupling result of \cite{strassen65} for underlying spaces being Polish:
\begin{equation}\label{schwarz}
\min\{ \Pr(X \neq Y): X \sim P_X, Y \sim P_Y\} = \frac{1}{2} \| P_X - P_Y\|_{TV}.
\end{equation}
(Note that the problem above is the optimal transportation problem for 0-1 cost; see \cite{villani:OT} for discussion).

The following result follows from the application of Strassen's coupling (\ref{schwarz}) and Lemma 2.11 of \cite{dudley:philipp}.
\footnote{Note that Strassen's couping also underlies the Berbee coupling, \cite{Berbee}, for real-valued random variables. We extend Berbee coupling to random vectors or, more general, random variables taking values in complete, separable metric spaces in the Appendix, and then use it to obtain concentration results.} 

\begin{lemma}[Independent Coupling for NLO Data Blocks via Strassen] By suitably enriching probability space, we can construct  $\widetilde{B}_k$ and $\widetilde{B}^{\text{qc}}_k$
that are independent of each other and that have the same marginal distributions as $B_k$
and $B^{\text{qc}}_k$ such that
$$
\Pr \left \{ (B_k, B^{\text{qc}}_k) \neq 
(\tilde B_k, \tilde B^{\text{qc}}_k) \right \}
= \frac{1}{2} \left \| P_{B_k, B^{\text{qc}}_k}- P_{B_k} \times P_{B^{\text{qc}}_k} \right\|_{TV} =: \gamma ( B_k, B^{\text{qc}}_k),
$$
where $P_{B_k, B^{\text{qc}}_k}$ is the distribution of $(B_k, B^{\text{qc}}_k)$ and $P_{B_k} \times P_{B^{\text{qc}}_k}$ is the distribution of $(\tilde B_k, \tilde B^{\text{qc}}_k)$. 
\end{lemma}

If the data sequence $(W_{\cdot, t}: t \geq 1)$ is beta-mixing in $t$,  we have that $\gamma ( B_k, B^{\text{qc}}_k) \to 0$, since the blocks are separated by $\lfloor T/(K-1) \rfloor \to \infty$ periods as $T \to \infty$. Thus, under beta-mixing, by using the NLO-cross-fitting, we can replace each block and its quasi-complement with independent blocks, with the probabilistic error determined by the speed of mixing of the weakly dependent time series.   Since we generally obtain the nuisance parameter estimates using quasi-complements, NLO-CF
allows us to treat these estimates as if (essentially) independent from the data used to compute semi-parametric scores (residuals in our context). \\

 \subsection{First-Stage Estimators for Learning Residuals in Panel Data}

In this section, we give examples of the first-stage reduced form estimators for dynamic panel data, focusing on the models with base treatment structure \eqref{eq:lineartreat}.  We rely heavily on the results of \cite{KockTang} in this stage.

\begin{example}[First-Stage Treatment Lasso and Reduced Form]
\label{ex:kocktang}
Consider the model \eqref{eq:priceout} with a single base treatment. Suppose
\begin{align}
\label{eq:pricep}
p_{i0}(X_{it}) = X_{it}' \delta^P_0 + \xi_i.
\end{align}
Here, to save notation, we reassign  $X_{it}$ to denote the dictionary of transformations of original controls $X_{it}$, that is, $X_{it} \leftarrow \mathsf{t}(X_{it})$, where the map $\mathsf{t}(\cdot)$ generates the dictionary and $X_{it} \in \mathrm{R}^{d_X}$. 

For the \textit{first-stage treatment} penalty level $\lambda_P = C_P \sqrt{NT\log^3 (d_X +N)} $ for some constant $C_P$, define the $k$-fold specific estimator:
\begin{align}
\label{eq:KockTangprice}
(\widehat{\delta}_k^P,\widehat{\xi}_k) = \arg \min_{\delta^P, \xi} \sum_{i=1}^N \sum_{t \in \mathcal{M}^{\text{qc}}_{k}} (P_{it} - X_{it}'\delta^P  - \xi_i)^2  +2 \lambda_P \| \delta^P\|_1 +  2\frac{\lambda_P}{\sqrt{N}}  \| \xi \|_1, \quad k=1,\ldots,K.
\end{align}
(Note that here and below the subscript index $k$ in $\hat \xi_k$ serves to indicate the $k$-specific estimator of the vector of fixed effects $\xi$, which is not to be confused with the index $i$ that enumerates the elements of the vector $\xi$).

Then, for   any $t \in \mathcal{M}_k$ and any $i=1,2,\dots, N$, the base treatment reduced form estimate is
\begin{align*}
\widehat{p}_{ik}(X_{it}) = X_{it}' \widehat{\delta}_k^P + \widehat{\xi}_{i,k}, \quad k=1,..,K.
\end{align*}

The properties of this estimator under weak sparsity assumptions on $\delta^P$ and $\xi$ follow from \cite{KockTang}.

\end{example}

\begin{example}[First-Stage Outcome Lasso and Reduced Form]
\label{ex:kocktangoutcome} Consider the outcome model:
\begin{align}
\label{eq:pricee}
  Y_{it}  = D_{it}' \beta_0 + X_{it}'\delta_0^P + \xi^E_i + U_{it}.
\end{align}
Here, to save notation, we reassign  $X_{it}$ to denote the dictionary of transformations of original controls $X_{it}$, that is,  $X_{it} \leftarrow \mathsf{t}(X_{it})$, where the map $\mathsf{t}(\cdot)$ generates the dictionary. 

For the \textit{first-stage outcome} penalty level $\lambda_E = C_E \sqrt{NT\log^3 (d_X +N)} $ for some constant $C_E$, define  the $k$-fold specific estimator:
\begin{align}
\label{eq:KockTangoutcome}
(\check \beta_k, \widehat{\delta}_k^E,\widehat{\xi}^E_k) = \arg \min_{\beta, \delta^E,\xi^E} \sum_{i=1}^N \sum_{t \in \mathcal{M}^{\text{qc}}_{k}} (Y_{it} - D_{it}'\beta  - 
X_{it}'\delta^E-\xi^E_i)^2 + 2 \lambda_E \| (\beta, \delta^E) \|_1  +  2\frac{\lambda_E}{\sqrt{N}} \| \xi^E \|_1.
\end{align}
Then, for   any $t \in \mathcal{M}_k$,  the  outcome reduced form estimate is
\begin{align}
\widehat{l}_{ik} (X_{it}) = \widehat{d}_{ik} (X_{it})'\check{\beta}_k + X_{it}' \widehat{\delta}_k^E + \widehat{\xi}_{ik}^E,
\end{align}
where $\widehat{d}_{ik}(\cdot)$ is the treatment reduced form estimate.   In what follows, we refer to $\check{\beta}_k$ as the \textit{preliminary}, or the \textit{one-stage} estimator of $\beta_0$.  \qed
\end{example}

Lemma \ref{lem:kocktang2} establishes properties of this estimator under weak sparsity assumptions on  $\delta^E$ and $\xi^E$, based upon \cite{KockTang}'s analysis of dynamic panel data Lasso with weakly sparse unit effects.

\subsection{The Second Stage: Estimating CATE Functions  } 
\label{sec:ss}
Here we describe the second stage estimators.

When $D_{it}$ is low-dimensional, we can apply ordinary least squares to residuals.

\begin{definition}[Orthogonal Least Squares]
Define
\begin{align} 
	\label{eq:OLS}
	 \widehat{\beta}_{OLS} & := \arg \min_{\beta \in \mathrm{R}^d} \frac{1}{NT} \sum_{i=1}^N \sum_{t=1}^T (\widehat{\widetilde{Y}}_{it}  - \widehat{V}_{it}' \beta)^2.
	\end{align}
\label{OLS}
\end{definition}
Appendix \ref{sec:lowd} in Online Supplement  establishes estimation and inference results for Orthogonal Least Squares.  The rate of convergence is $\sqrt{d/NT}$.

\begin{definition}[Orthogonal Lasso]
Let $\lambda_\beta=C_{\beta} \sqrt{\log d /NT}$ and $C_{\beta}$ be a penalty parameter.   Define
	\begin{align}
	\label{eq:OL}
	 \widehat{\beta}_{L} & := \arg \min_{\beta \in \mathrm{R}^d} \frac{1}{NT} \sum_{i=1}^N \sum_{t=1}^T (\widehat{\widetilde{Y}}_{it}  - \widehat{V}_{it}' \beta)^2  + \lambda_{\beta} \sum_{j=1}^d | \beta_j |.
	\end{align}
\label{OL}
\end{definition}

Theorem \ref{thrm:ortholasso} provides the near-oracle rates of convergence $$\sqrt{s \log d/ NT}$$ for the CATE function. 
We notice that Orthogonal Lasso outperforms Orthogonal Least Squares even in low-dimensional settings when $s \log d \ll d \ll NT$, that is, when effective dimension $s$ of $\beta_0$ is much smaller than its nominal dimension $d$.

\begin{remark}[\textbf{Key Point of Orthogonalization}]
By working with estimated residuals, we attain the quasi-oracle rates of convergence -- the rates that result if we knew the true residuals exactly and used them instead. As a result, Orthogonal Lasso also outperforms the single-stage Outcome Regression estimators when the CATE function is much simpler and, therefore, easier to learn the overall regression function. For example, Orthogonal Lasso outperforms the first-stage outcome Lasso when the CATE function is more sparse than the overall regression function, so the near-oracle rate is much better than the overall rate. Remark \ref{rm:improvement} below provides a formal statement.   The phenomenon our paper points out is more general and is not specific to using Lasso methods used in the final stage. More recent works than our paper use orthogonalization procedures like ours to learn the CATE functions for other choices of the final stage estimator; see, e.g.,  Kennedy (2020).\end{remark}

\begin{remark}[Data-Adaptive Penalty Levels]
We choose the penalty level for first and stage estimators of the stated simple form above to simplify theoretical arguments.  Rigorous and data-adaptive choices of penalty levels $\lambda $'s, in particular of the constants $C_P$ and $C_P$, as well as the generalization of $\ell_1$-penalty to its weighted analog,   are discussed in e.g., \cite{BCCH} and \cite{Program} for cross-sectional case, and implemented in \url{hdm} $R$ package by \cite{hdm}. Their choices likely carry over to the dynamic panel data settings under the conditional sequential exogeneity condition.
\end{remark}

\subsection{The Third Stage: Debiased Inference on Parameters of CATE Functions.}  
\label{sec:inference}
Here we describe the third stage that performs debiased inference. Due to the bias induced by $\ell_1$-shrinkage, penalized estimators  cannot be used for inference based on the standard Gaussian approximation. We construct a debiased estimator based on a variant of \cite{geer} and \cite{ZhangZhang} with a new choice of debiasing matrix.

Consider the covariance matrix of residuals 
\begin{align}
\label{eq:covmatq}
    Q = \dfrac{1}{NT} \sum_{i=1}^N \sum_{t=1}^{T} \Ep V_{it} V_{it}'
\end{align}
and its inverse $Q^{-1}$.  Define the sample covariance matrix of the residuals as
\begin{align}
    \label{eq:widehatq}
    \widehat{Q}:= \frac{1}{NT}  \sum_{i=1}^N \sum_{t=1}^{T} \widehat{V}_{it} \widehat{V}_{it}'.
\end{align}
Estimate approximate inverse of $\widehat{Q}$ by 
\begin{align}
\label{eq:prelimclime}
    \widehat{\Omega} = \arg \min_{\Omega \in \mathrm{R}^{d \bigtimes d}} \| \Omega \|_1  :
     \| \widehat{Q} \Omega  - I_d \|_{\infty} \leq  \lambda_Q,
\end{align}
where
\begin{align}
\label{eq:kappant}
  \lambda_Q := C_Q \kappa_{NT}, \quad
\kappa_{NT}:= \sqrt{\log^3 (d^2 \log (NT)) \log NT /NT },
\end{align}
where $C_Q$ is a tuning constant.  Finally, symmetrize the approximate inverse  $\widehat{\Omega}$ as \begin{equation}
\label{eq:finalclime}
\widehat{\Omega}^{\text{CLIME}}= (\widehat{\omega}_{ij}^{\text{CLIME}}), \quad \widehat{\omega}_{ij}^{\text{CLIME}}= \widehat{\omega}_{ij}   1_{\{ | \widehat{\omega}_{ij} | < | \widehat{\omega}_{ji} | \}} +  \widehat{\omega}_{ji}   1_{\{ | \widehat{\omega}_{ij} | > | \widehat{\omega}_{ji} | \}}.
\end{equation}
In other words, between $\widehat{\omega}_{ij} $ and $\widehat{\omega}_{ji}$, we take the one with smaller absolute value to obtain a symmetric matrix $\widehat{\Omega}^{\text{CLIME}}$, as in \cite{CLIME}.

\begin{definition}[Debiased  Orthogonal Lasso]
\label{DOL}
Define
\begin{align}
\label{eq:DOL1}
\widehat{\beta}_{DL} :&= \widehat{\beta}_{L} +\widehat{\Omega}^{\text{CLIME}} \frac{1}{NT}  \sum_{i=1}^N \sum_{t=1}^{T}  \widehat{V}_{it} (\widehat{\widetilde{Y}}_{it}  - \widehat{V}_{it}'  \widehat{\beta}_{L} ).
\end{align}
\end{definition}
Theorems \ref{thrm:DOL} and \ref{thrm:manycoef} show that $\sqrt{NT} (\widehat {\beta}_{DL} - \beta_0)$ is approximately distributed as $ N(0, \Sigma) $ over rectangular regions. The  covariance matrix 
\begin{align}
\label{eq:sigma}
    \Sigma := Q^{-1} \Gamma Q^{-1} = Q^{-1} \dfrac{1}{NT} \sum_{i=1}^N \sum_{t=1}^T \Ep V_{it}V_{it}' U_{it}^2 Q^{-1}
\end{align}
is estimated by its sample analog
\begin{align}
    \label{eq:sigmahathds}
    \widehat{\Sigma}(\widehat{\beta}_L) := \widehat{Q}^{-1}  \frac{1}{NT} \sum_{i=1}^N \sum_{t=1}^{T} \widehat{V}_{it} \widehat{V}_{it}' (\widehat{\widetilde{Y}}_{it}  - \widehat{V}_{it}' \widehat{\beta}_L)^2 \widehat{Q}^{-1}=: \widehat{Q}^{-1} \widehat{\Gamma}(\widehat{\beta}_L) \widehat{Q}^{-1}.
\end{align}
This method allows for  constructing  componentwise and simultaneous confidence intervals for all components of $\beta_0$. This method also allows for performing inference on linear functionals $a'\beta_0$ of $\beta_0$ (provided that the $\ell_1$-norm of $a$ is bounded.)

\section{Theoretical Results on Orthogonal Lasso}
\label{sec:theory}
\setcounter{remark}{0}
\setcounter{equation}{0}
\setcounter{definition}{0}

\subsection{Consistency of Orthogonal Lasso}
\label{sec:mainrest}
The following assumptions impose regularity conditions on weak dependence, tail behavior, and  the reduced form estimators. 

\begin{assumption}[Sampling and Asymptotics]
\label{ass:sampling}
(1) The data sequence $\{ \{W_{it}\}_{t=1}^{T}\}_{i=1}^{N}$ obeys the model \eqref{eq:APLM} of Section \ref{sec:section2}. (2) The data on units $W_{i, \cdot}$ are independent across $i$,
and beta-mixing at geometric speed with respect to time $t$, uniformly in $i$:
\begin{align}
     \gamma(q):= \sup_{\bar t \leq T, i \leq N} \gamma 
     \Big ( \{W_{i t}\}_{t \leq \bar t}, \{W_{i t}\}_{t \geq \bar t+q} 
     \Big)
     \leq  C_{\kappa} \exp(-\kappa q) \label{eq:expmix}
\end{align}
 for all $q \geq 1$, and for some constants $C_{\kappa} \geq 0$ and $\kappa>0$. (3) The number of time periods $T$ is large enough, $ T^{-1} \log  (N) = o(1)$.

 \end{assumption}

 Assumption \ref{ass:sampling} limits the data dependence across time periods with exponential mixing step. It  is a standard weak dependence condition in the literature (\cite{HahnKuersteiner}, \cite{ValLee}).    We incur it to ensure the validity of inference based on the panel cross-fitting of Definition \ref{splitting:panel}.  Note that the requirement on $N$ comes from the relation
 \begin{equation}
\gamma(\{W_{\cdot, t}\}_{t \leq \bar t}, \{W_{\cdot, t}\}_{t \geq \bar t+q} )
\leq N \max_{i \leq N} \gamma(\{W_{i, t}\}_{t \leq \bar t}, \{W_{i, t}\}_{t \geq \bar t+q} ),
\end{equation}
which can be found using the union bound.

The next condition ensures identification of the coefficients of the CATE function.
\begin{assumption}[Identification]
	\label{ass:identification}
	Let $Q= (NT)^{-1} \sum_{i=1}^N \sum_{t=1}^{T} \Ep V_{it} V_{it}^\top $ denote the population covariance matrix of treatment residuals.  Assume that there exist constants $C_{\text{min}},C_{\text{max}}$  such that $0<C_{\min} \leq \min \eig(Q) \leq  \max \eig(Q) \leq C_{\text{max}}<\infty$.
\end{assumption}

A collection of centered random variables $\{ X_j \} \in \mathrm{R}$ is said to be uniformly $\sigma^2$-sub-Gaussian if
\begin{align}
\label{eq:subg}
\Ep \exp({\lambda X_j})  \leq \exp({\lambda^2 \sigma^2/2}), \quad \forall \lambda \in \mathrm{R} \quad \forall j.
\end{align}

\begin{assumption}[sub-Gaussian Tails]
\label{ass:subgauss}
The following conditions hold for some constants $0 < \ubar \sigma^2 < \bar \sigma^2 < \infty$. (1) For $j=1,2,\dots, d$, $(V_{it})_j$ are $\bar \sigma^2$-sub-Gaussian conditional on $X_{it},\Phi_{it}$. (2) $U_{it}$ is $\bar \sigma^2$-sub-Gaussian conditional on $V_{it}, X_{it},\Phi_{it}$. (3) $U_{it}$ is conditionally non-degenerate,  namely 
 $\inf_{it} \E[ U_{it}^2 | V_{it},X_{it}, \Phi_{it}] \geq \ubar{\sigma}^2$ with probability 1.
\end{assumption}

\begin{assumption}[Additional Regularity Conditions]
\label{ass:addreg}
 We suppose that the true parameter vector has bounded $\ell_1$-norm: (a) $$\|\beta_0\|_1 \leq \bar C_\beta$$ for some finite constant $\bar C_\beta$;  
(b) and that the number of non-zero coefficients does not increase too quickly: 
$$ (s \vee 1) \kappa_{NT}  =  (s \vee 1)  \sqrt{\log^3 (d^2 \log (NT)) \log NT /NT }= o(1).$$
(c) The tuning constants  $C_\beta$ and $C_Q$ in the penalty levels $\lambda_\beta$ and $\lambda_Q$ are sufficiently large. (d) The number $d \rightarrow \infty$.

\end{assumption}

Let $1 \leq i \leq N$ be a unit index, and let $1 \leq j \leq d_P$ be a component index. We define a generic nuisance function to be $$\mathbf{g}(\cdot)=\{g_{ij} (\cdot) \}  : \ \mathcal{X} \rightarrow  \mathrm{R}^{N \times d_P},$$ 
a generic $N\times d_P$-matrix, and  we let 
$$\mathbf{g}_0(\cdot) = \{ g_{ij0}(\cdot)\} : \  \mathcal{X} \rightarrow  \mathrm{R}^{N \times d_P}$$
be its true value; here $\mathcal{X}$ is a subset of 
$\mathrm{R}^{N \times d_X}$. Let $G_{NT}$  be a sequence of neighborhoods around $\mathbf{g}_0(\cdot)$ containing realizations of its generic machine learning estimators $\widehat{\mathbf{g}}(\cdot)$ with probability approaching 1.  As the sample size $NT$ increases, we expect the sets $G_{NT}$ to converge towards $\mathbf{g}_0$ in suitable norms.  We denote rate of convergence in the mean square norm as:
 \begin{align}
 \label{eq:prate}
\textbf{g}_{NT} &:= \max_{1 \leq j \leq d_P} \sup_{\mathbf{g} \in G_{NT}} \left( (NT)^{-1} \sum_{t=1}^{T }  \sum_{i=1}^N \Ep  (g_{ij}(X_{it}) - g_{ij0}(X_{it}))^2    \right)^{1/2}  
\end{align}
and let $ \textbf{g}_{NT,\infty}$ be a sequence of nonnegative constants such that with probability $1-o(1)$:
\begin{align}
\label{eq:suprate}
 \max_{1 \leq j \leq d_P} \sup_{\mathbf{g} \in G_{NT}}  \sup_{it} | g_{ij}(X_{it}) - g_{ij0}(X_{it}) | \leq \textbf{g}_{NT,\infty}.
 \end{align}

In particular, we specialize the notion as follows:

\begin{enumerate}
    \item[1.] For the technical treatment reduced form, we replace the letters $g$ with $d$:
\begin{itemize}
\item  $\mathbf{d}(\cdot)$ denotes the parameter and $\mathbf{d}_0(\cdot)$ the true reduced form for treatment ;
\item $D_{NT}$ denotes the set containing first-stage estimates $\widehat{\mathbf{d}}(\cdot)$  of $\mathbf{d}_0(\cdot)$  w.p. $1-o(1)$
\item $\textbf{d}_{NT}$ and  $\textbf{d}_{NT,\infty}$ are rates of convergence of $D_{NT}$ to $\textbf{d}_0(\cdot)$; \\
\end{itemize}

\item[2.] For the outcome reduced form, we replace the letters $g$ with $l$:
\begin{itemize}
\item  $\mathbf{l}(\cdot)$ denotes the parameter and $\mathbf{l}_0(\cdot)$ the true value reduced form for outcome;
\item $L_{NT}$ denotes the set containing first-stage estimates $\widehat{\mathbf{l}}(\cdot)$  of $\mathbf{l}_0(\cdot)$  w.p. $1-o(1)$
\item $\textbf{l}_{NT}$ and  $\textbf{l}_{NT,\infty}$ are rates of convergence of $L_{NT}$ to $\textbf{l}_0(\cdot)$.
\end{itemize}

\end{enumerate}

The key assumption on  the quality of reduced form estimators is as follows.

\begin{assumption}[Regularity Conditions and Convergence Rates for Residual Learners]
We suppose that the reduced form estimators obey: $\widehat{\mathbf{l}}(\cdot) \in L_{NT}$ and $\widehat{\mathbf{d}}(\cdot) \in D_{NT}$  such that   
\label{ass:smallbiashds}
$\textbf{d}_{NT}$,  $\textbf{d}_{NT,\infty}$, $\textbf{l}_{NT}, \textbf{l}_{NT,\infty}$  decay sufficiently fast:
\begin{align}
  & \textbf{d}_{NT}^2  +  \textbf{d}_{NT}\textbf{l}_{NT}  = o((NT)^{-1/2} ), \label{eq:rate} \\
      &  \textbf{l}_{NT,\infty}  = o(\log^{-1/2} (dNT)), 
\quad       \textbf{d}_{NT,\infty}  = o(\log^{-1/2} (dNT)).  \label{eq:suprate2}
\end{align}

\end{assumption}

As we discuss in Section \ref{sec:fs}, this assumption is plausible for the Lasso-based first stage estimators that we consider and may be plausible for others too.  The conditon of bounded $\ell_1$-norm can be relaxed to allow for an increasing norm at the cost of somewhat more complicated regularity condtions, as can be seen from the proofs.

Theorem \ref{thrm:ortholasso} establishes the  convergence rate of Orthogonal Lasso in $\ell_2$ and $\ell_1$ norm.

\begin{theorem}[Oracle Rates for Orthogonal Lasso]
Suppose Assumptions  \ref{ass:sampling}--\ref{ass:smallbiashds} hold. Then, the Orthogonal Lasso possesses the following oracle rate guarantees:
\begin{align}
\label{eq:lassoboundmain}
    \left(\E_{NT} (V_{it}' (\widehat{\beta}_L - \beta_0))^2  \right)^{1/2} = O_P \left( \sqrt{\dfrac{s\log d}{NT}} \right), \quad \| \widehat{\beta}_{L} - \beta_0 \|_{1} =O_P \left( \sqrt{\dfrac{s^2\log d}{NT}} \right).
    \end{align}
     \label{thrm:ortholasso}
\end{theorem}

Theorem \ref{thrm:ortholasso} is our first main result. It establishes the convergence rate of Orthogonal Lasso. This rate coincides with the oracle convergence rate, where oracle knows  the first-stage function $e_0(\cdot)$  and the  unobserved unit effects $\{\xi_i^E\}_{i=1}^N$  in the model \eqref{eq:APLM}, and therefore, knows the residuals and uses Lasso on these true residuals.

\subsection{Estimation of $Q^{-1}$ and $\Sigma$ in High-Dimensional Setting}

To perform statistical inference we will need to assume approximate sparsity for the inverse $Q^{-1}$ of the covariance matrix of the residuals $Q$. We will use the following notation. For a square matrix $A=(a_{ij})$, denote
$$
\|A \|_{1,\infty} = \max_{1 \leq j \leq d} \sum_{i=1}^d |a_{ij}|, \qquad \|A \|_{\infty,1} = \max_{1 \leq i \leq d} \sum_{j=1}^d |a_{ij}|.
$$

 \begin{assumption}[Regularity for Estimating $Q^{-1}$]
\label{ass:approx}  (a) Let $A_Q$ and $a_Q>1$ be finite constants such that for any column $j$
 \begin{align*}
     (Q^{-1}_{mj})^{*} \leq A_Q m^{-a_Q}, \quad m,j=1,2,\dots,d
 \end{align*}
 where $(Q^{-1}_j)^{*}$ is a non-increasing rearrangement of $(|Q^{-1}_{mj}|)_{m=1}^d$.  Furthermore, for $\lambda_Q = C_Q \kappa_{NT}$, 
\begin{align}
\label{eq:debiasing}
 \lambda_Q^{1-1/a_Q} = o (s^{-1} \log^{-1/2} d).
\end{align}
  \end{assumption}
Assumption \ref{ass:approx}(a) ensures that the CLIME estimator of  $Q^{-1}$  defined in the equation \eqref{eq:finalclime} converges sufficiently fast.  If $d \gg NT$, it requires $Q^{-1}$ to be approximately sparse so that it can be consistently estimated  with only $NT$ observations. Examples of  high-dimensional matrices $Q$ with an approximately sparse inverse  include block diagonal, Toeplitz, and band matrices.

The following lemma establishes the rate bound for the CLIME estimator of $Q^{-1}$ for $\widehat Q$ in \eqref{eq:widehatq}. The result holds under mixing dependence and approximate sparsity of $Q$, which may be of independent interest.

\begin{lemma}[Consistency of the CLIME estimator]
\label{lem:climerate}
Suppose Assumptions \ref{ass:sampling}--\ref{ass:approx} hold.  The CLIME estimator converges in $\ell_{\infty}$-norm and $\ell_{\infty,1}$-norm
\begin{align}
& \|   \widehat{\Omega}^{\text{CLIME}} - Q^{-1} \|_{\infty}= \|   \widehat{\Omega} - Q^{-1} \|_{\infty} = O_P ( \lambda_Q) \label{eq:omegaqinv1} \\
&   \| \widehat{\Omega}^{\text{CLIME}} - Q^{-1} \|_{1, \infty} = \| \widehat{\Omega}^{\text{CLIME}} - Q^{-1}  \|_{\infty,1} = O_P (  {\lambda_Q}^{1-1/a_Q} ) \label{eq:debiasing1}  \\
& \|  I_d-  \widehat{\Omega}^{\text{CLIME}}\widehat{Q}  \|_{\infty} =  \|  I_d- \widehat{Q} \widehat{\Omega}^{\text{CLIME}}  \|_{\infty} = O_P (   {\lambda_Q}^{1-1/a_Q}). \label{eq:debiasing2} 
\end{align}
\end{lemma}

\subsection{Pointwise Gaussian Inference with Debiased Orthogonal Lasso}
\label{sec:mainrinf}

 The following theorem establishes validity of Gaussian inference for parameters $\alpha'\beta_0$, where $\alpha$ is a fixed vector with bounded $\ell_1$ norm.  This is our second main result.

 \begin{theorem} Let $K_{\alpha}$ be a finite constant. Suppose Assumptions \ref{ass:sampling}--\ref{ass:approx} hold, and  the Lindeberg condition holds for each $m>0$: $$\limsup_{NT \to \infty}\sup_{\|\alpha\|_2 =1, \| \alpha\|_1 \leq K_{\alpha}} (NT)^{-1} \sum_{i=1}^N \sum_{t=1}^T \Ep [(\alpha' V_{it} U_{it})^2  1 \{  |\alpha' V_{it} U_{it}|   > m \sqrt{NT}\}  = 0.$$
Then, the Debiased Lasso estimator  is asymptotically Gaussian:
    \begin{align*}
       \lim_{N T \rightarrow \infty} \sup_{\| \alpha\|_2 = 1, \| \alpha\|_1 \leq K_{\alpha}} \sup_{t \in \mathrm{R}} \bigg| \Pr \bigg(\frac{ \sqrt{NT} \alpha'(\widehat{\beta}_{DL}- \beta_{0})}{ \sqrt{\alpha' \Sigma \alpha}} < t \bigg) - \Phi(t) \bigg| = 0,
    \end{align*}
    where $\Phi(t)$ is the CDF of $N(0,1)$. Moreover, the result continues to hold when $\Sigma$ is replaced by $\widehat{\Sigma}$ such that $\| \widehat{\Sigma} - \Sigma\|_\infty = o_P(1).$

\label{thrm:DOL}     
\end{theorem}

The following lemma establishes consistency of the high-dimensional covariance matrix for the approximate Gaussian distribution of the Debiased Lasso estimator.  Notably, the dimension of this matrix can exceed the sample size.  Define
\begin{align}
\label{eq:gammant}
\gamma_{NT}:=(NT)^{-1/4} + \mathbf{l}_{NT} + \sqrt{s  \log d/NT} + \mathbf{l}^2_{NT} \log (d^2 NT). 
\end{align}

\begin{assumption}[Conditions for $\Sigma$ Estimation]\label{ass:sigma} We suppose that the following condition on the growth of the dimension $d$ holds
\begin{align}
\label{eq:kappad2nt}
\kappa_{NT} \log^2 (d^2 NT) = o(1).
\end{align}
and $\gamma_{NT} = o(1)$.
\end{assumption}

\begin{lemma}[Consistency of Variance Matrix Estimator]
\label{lem:sigmaolsbound}
Suppose Assumptions \ref{ass:sampling}--\ref{ass:sigma} hold.  Then, the estimator $\widehat{\Sigma}(\widehat{\beta}_L)$ converges in $\ell_{\infty}$ norm
\begin{align}
    \| \widehat{\Sigma}(\widehat{\beta}_L) - \Sigma \|_{\infty} = O_P \left(\gamma_{NT}  + \lambda_Q^{1-1/a_Q}\right) =: O_P (\zeta_{NT}) \label{eq:zetant} = o_P(1).
\end{align}
\end{lemma}

\subsection{Simultaneous Inference}
\label{sec:mainrsim}

We next present theoretical results on simultaneous inference on many structural coefficients.

Define the following rates
\begin{align}
\rho_{NT} & :=    \sqrt{\log (dNT)/NT} (\textbf{d}_{NT,\infty}  + \textbf{l}_{NT,\infty} ) 
+ r_{NT} \label{eq:rhont}\\
r_{NT} & :=   \textbf{d}_{NT}^2 + \textbf{d}_{NT}\textbf{l}_{NT}
+ ( \textbf{d}_{NT,\infty}^2 +
\textbf{d}_{NT,\infty}\textbf{l}_{NT,\infty}) \sqrt{ (NT)^{-1}  \log (NT) \log d  }\label{eq:rnterror} 
\end{align}

\begin{assumption}[Regularity Conditions for Simultaneous Inference on Many Coefficients]
\label{ass:manycoef} (1) There exists a sequence $\pi_{NT}^{UV} \geq 1$ so that $\sup_{it} \| V_{it} U_{it} \|_{\infty} \leq \pi_{NT}^{UV} \text{ a.s. } $  and
$$
0< \min_{it} \| \Ep V_{it} V_{it}' \|_{\infty} < \infty.
$$
(2) For some constant $c_2: 0<c_2< 1/4$, the following rate conditions hold
\begin{align}
\label{eq:theoreme2}
\pi_{NT}^{UV} \log d \log (NT)  \log^{7/2} (dNT)  &\lesssim (NT)^{1/2-2c_2} 
\end{align} 
and $\log^4 d \log^2 (NT) = o ( \sqrt{NT})$.
(3) The following rate conditions holds
\begin{align}
\label{eq:o1tcond}
   \sqrt{NT} \rho_{NT} + \lambda^{1-1/a_Q}_Q s \log^{1/2} d = o ( \log^{-1/2} d \log^{-1/2} NT). 
\end{align}
\end{assumption}

Theorem \ref{thrm:manycoef}  establishes high-dimensional Gaussian approximation for a  treatment effect vector $\beta_0$ and allows to conduct simultaneous inference on its coefficients. 
\begin{theorem}[Simultaneous Inference on Many Coefficients]
Suppose Assumptions  \ref{ass:sampling}-\ref{ass:manycoef} with $\Sigma$  as in \eqref{eq:sigma} and $\widehat{\Sigma}$ as in \eqref{eq:sigmahathds}. Then, the following Gaussian approximation result holds for $\widehat{\beta}_{DL}$
\begin{align}
\label{eq:rbound}
\sup_{R \in \mathcal{R}} |\Pr ( (\mathrm{ diag } \ \Sigma)^{-1/2}\sqrt{NT} (\widehat{\beta}_{DL} - \beta_0) \in R) - \Pr ( Z \in R )| \to 0,
\end{align}
where $Z \sim N(0, \mathcal{C})$ is a centered Gaussian random vector with the covariance matrix  $$ \mathcal{C}= ( \mathrm{ diag } \ \Sigma)^{-1/2} \Sigma (\mathrm{ diag } \ \Sigma)^{-1/2}$$ and $\mathcal{R}$ denotes a collection of cubes in $\mathbb{R}^d$ centered at the origin.  In addition, if 
\begin{align}
    \gamma_{NT} + \lambda^{1-1/a_Q}_Q = o (\log^{-2} d \log^{-1} NT), \label{eq:zetantcond}
\end{align}
then, replacing $\mathcal{C}$ with $\mathcal{\widehat C}$ $=$ $(\mathrm{ diag } \ \widehat{\Sigma})^{-1/2} \widehat{\Sigma} (\mathrm{ diag } \ \widehat{\Sigma})^{-1/2}$,
we also have  for $\widehat{Z} \mid \mathcal{\widehat C}   \sim N(0,\widehat {\mathcal{C}})$,
\begin{align}
\label{eq:rbound2}
\sup_{R \in \mathcal{R}} |\Pr (  (\mathrm{ diag } \ \widehat \Sigma)^{-1/2}  \sqrt{NT} (\widehat{\beta}_{DL} - \beta_0) \in R) - \Pr ( \widehat{Z}  \in R\mid \mathcal{\widehat C} )| \to_P 0.
\end{align}
Consequently, for the $c_{1 - \xi} = (1-\xi)$-quantile of $\| \widehat{Z} \|_\infty \mid \widehat {\mathcal{C}}$, we have
\begin{equation*}  \Pr (\beta_{0,j} \in [\widehat{\beta}_{DL,j} \pm c_{1 - \xi}  \widehat{\Sigma}_{jj}^{1/2} (NT)^{-1/2}], j = 1,2,\ldots,d) \to (1- \xi).\end{equation*}

\label{thrm:manycoef}
\end{theorem}
Theorem \ref{thrm:manycoef} is our third main result.  It extends  the high-dimensional Gaussian approximations of \cite{CCKBoot, Wu,CCK} to a panel setting. 

% \textsc{[Here we do need some rate for estimating $\Sigma$, but it is not clear from the regularity conditions what the rate is. ]}

\subsection{Orthogonal Group Lasso}
\label{sec:grouplasso}

In this section, we focus on  Example \ref{HTE} with a linear control function $e_0(\cdot)$ in \eqref{eq:APLM}.   Applied economists fitting \eqref{eq:APLM} often would like to include   the variable $(K_{it})_j$ whenever the  interaction of the base treatment $P_{it}$ and the control $(K_{it})_j$ is selected in the second stage. However, in case of Orthogonal Lasso, the sets of controls selected in the stage $2$ may not be a subset of the controls selected in the stage $1$. To address this concern, we \textit{group} the main and interaction effects of controls $K_{it}$ to attain the desired model selection pattern. 

Decompose the covariate vector
$$
X_{it} = (K_{it}, Z_{it}),
$$
where $K_{it}$ is a vector of heterogeneity-relevant controls and $Z_{it}$ is its complement. The function can be written as
$$
e_0(X_{it}) = K_{it}'\rho_0 +  Z_{it}' \delta^E_{0Z}
$$
and the linear model \eqref{eq:APLM} 
\begin{align}
\label{eq:APLM3}
 Y_{it} = \underbracket{(P_{it} K_{it}, K_{it})'}_{D_{it}} \underbracket{(\beta_0, \rho_0)}_{\bar{\beta}_0} + Z_{it}' \delta^E_{0Z} + \xi_i^E +U_{it}.
\end{align}
% VS: Forgot fixed effects!
Assuming both the interaction effect $\beta_0$ and the main effect $\rho_0$ are $s$-sparse, the  vector $\bar{\beta}_0$ obeys \textit{group sparsity assumption}
\begin{align*}
\| \beta_0, \rho_0 \|_{2, 0} := \sum_{j=1}^d 1 \{ (\beta_{0j}, \rho_{0j}) \neq (0,0) \} \leq \sum_{j=1}^d 1 \{ \beta_{0j} \neq 0 \} +  \sum_{j=1}^d 1 \{ \rho_{0j} \neq 0 \}  = 2s \ll d.
\end{align*}
The unit-specific treatment reduced form is
$$
D_{it} = d_{i0} (Z_{it}) = d_0 (Z_{it}; \xi_i)
$$
and the unit-specific outcome reduced form is
$$
\Ep [Y_{it} \mid Z_{it}] = d_{i0} (Z_{it})' \bar \beta_0 + Z_{it}' \delta^E_{0Z} + \xi^E_i.
$$
The residualized form is 
\begin{align*}
    \bar {\widetilde{Y}}_{it} = \bar V_{it}' \bar{\beta}_0 + U_{it},
\end{align*}
where  
\begin{align}
    \label{eq:vitgroup}
   \bar  {\widetilde{Y}}_{it} := Y_{it} - {\Ep} [Y_{it} \mid Z_{it}], \bar {V}_{it} := D_{it} - {\Ep} [D_{it} \mid Z_{it}] =:D_{it} - d_{i0}(Z_{it})
\end{align}
The  \textit{Orthogonal Group Lasso} estimator is the first component $ \widehat{\beta}_{GL} $ of the following minimization problem
\begin{align}
	\label{eq:OGL}
	 \widehat{\beta}_{GL} & := \arg \min_{\bar{\beta} =(\beta, \rho) \in \mathrm{R}^{2d}} \frac{1}{NT} \sum_{i=1}^N \sum_{t=1}^T (\bar{\widehat{\widetilde{Y}}}_{it}  - \widehat{\bar{V}}_{it}' \bar{\beta})^2  + \lambda_{ \beta} \sum_{j=1}^d \| (\beta_j, \rho_j) \|_2.
\end{align}
The \textit{Debiased Orthogonal Group Lasso} estimator is
\begin{align}
\label{eq:DOL2}
\widehat{\beta}_{DGL} :&= \widehat{\beta}_{GL} +\widehat{\Omega}^{\text{CLIME}} \frac{1}{NT}  \sum_{i=1}^N \sum_{t=1}^{T}  \widehat{V}_{it} (\widehat{\widetilde{Y}}_{it}  - \widehat{V}_{it}'  \widehat{\beta}_{GL} ).
\end{align}
%In this proposal, the   group penalty forces the main effect $(K_{it})_j$ and interaction effect $(P_{it} K_{it})_j$ to be included together and to attain the desired model selection pattern. 

\begin{lemma}[Orthogonal Group Lasso]
\label{lem:extended}
Under Assumptions \ref{ass:sampling}--\ref{ass:smallbiashds} for $\bar{V}_{it}$  and $d_{i0}(Z_{it})$  as in \eqref{eq:vitgroup}, the Orthogonal Group Lasso  attains the following rate:
\begin{align}
\label{eq:lassoboundgroup}
    \left(\E_{NT} (\bar{V}_{it}' (\widehat{\beta}_{GL} - \beta_0))^2  \right)^{1/2} = O_P \left( \sqrt{\dfrac{s\log d}{NT}} \right), \quad \| \widehat{\beta}_{GL} - \beta_0 \|_{1} =O_P \left( \sqrt{\dfrac{s^2\log d}{NT}} \right).
    \end{align}Furthermore,  the statements of Theorems \ref{thrm:DOL} and \ref{thrm:manycoef} hold for the Debiased Orthogonal Group Lasso.  \qed  
\end{lemma}

\section{Verification of Assumptions on the First Stage Estimators of Residuals}\label{sec:fs}
\setcounter{remark}{0}
\setcounter{lemma}{0}
The purpose of this section is to verify Assumption \ref{ass:smallbiashds} in i.i.d. and panel data settings. For the Lasso-based methods of Examples \ref{ex:kocktang}-\ref{ex:kocktangoutcome}, we give examples of  nuisance parameter estimates, nuisance realization sets, and the low-level assumptions that suffice for  Assumption \ref{ass:smallbiashds} to hold. Unless proven immediately, all numbered Remarks are formally proven in the Online Supplement, Appendix \ref{sec:proofs:fs}.

\subsection{No Unobserved Unit Heterogeneity: General ML} 
Suppose the unit-specific vector function in \eqref{eq:priceout} obeys $p_{i0}(\cdot) =  p_{j0}(\cdot), 1 \leq i, j \leq N$. Let $p_0(\cdot) = p_{i0}(\cdot) =  p_{j0}(\cdot)$ be the single coordinate of the $N$-vector $\mathbf{p}_0(\cdot)$ entering in \eqref{eq:prate}. If the covariates $X_{it}$ are i.i.d. over $i$ and $t$, the mean square rate  $\textbf{p}_{NT}$ reduces to 
\begin{align*}
\textbf{p}_{NT} = \sup_{p \in P_{NT}} (NT)^{-1} \sum_{i=1}^N \sum_{t=1}^T (\Ep  (p(X_{it}) - p_0(X_{it}))^2)^{1/2} =  \sup_{p \in P_{NT}} (\Ep (p(X) - p_0(X))^2)^{1/2}.
\end{align*}
Furthermore,  one can split by unit index to construct independent partitions and use regular cross-fitting instead of the NLO one. In this case, the condition (2) of Assumption \ref{ass:sampling} redundant.

The upper bound on $\textbf{p}_{NT}$ are available for i.i.d. data (across time) for many regularized  methods under structured assumptions on the functions $p_0(x)$ and $e_0(x)$,  such as random forest, neural networks, or boosting.    Specifically, the bound on $\textbf{p}_{NT}$ is achievable by $\ell_1$ penalized methods in sparse models (\cite{geer}, \cite{orthogStructural}, \cite{belloni2013}),   $\ell_2$ boosting  in sparse linear models (\cite{Luo}),  neural networks (\cite{Schmidt}, \cite{Farrell}), and random forest in small  (\cite{WagerWalther}) and  high (\cite{fastRF}) dimensions with the sparsity structure.  
While most of these results are established in an i.i.d. setting, we conjecture  that similar rates could be established under weak dependence, by relying on Berbee coupling, exponential mixing conditions and/or the martingale-difference property of regression errors.

\subsection{Unobserved Unit Heterogeneity: Lasso}    In this section, we verify Assumption \ref{ass:smallbiashds} for the first stage Lasso estimator.

We will make use of weak sparsity assumptions on fixed effects.  Weak sparsity generalizes the exact sparsity restriction to accommodate  small deviations from sparsity.  
Given a constant $\nu: 0 < \nu <1$,  the vector $u \in \mathrm{R}^{d_u}$ is  said to be $(\nu,S)$-weakly sparse (\cite{Negahban}) if there exists a bound $S=S(N,T)$, that may depend on $N,T$, such that
\begin{equation}\label{eq:weaksparse} \sum_{j=1}^{d_u} |u_j|^{\nu} \leq S. 
\end{equation}
Lemma A.1 in \cite{KockTang} gives examples of distributions that generate weakly sparse draws with probability $1-o(1)$. For one example, if $\xi_i$
are independent Gaussian draws
\begin{align}
\label{eq:effectdraws}
\xi_i \sim N(0, \sigma_i^2), \quad \max_{1 \leq i \leq N} \sigma^2_i = O (\log^3 (d_X +N)/(N^{1/\nu} T)), \quad i=1,2,\dots, N.
\end{align}
Then, the vector $\xi=(\xi_i)_{i=1}^N$ obeys \eqref{eq:weaksparse} with probability $1-o(1)$ with $S^P= \sqrt{N \log^{3\nu} (d_X +N) /T^{\nu}}$.

In what follows,  we will rely on  regressors and treatments being sub-Gaussian. In dynamic models where regressors include lagged values of outcomes and treatments, this assumption is non-trivial. We verify it from the model primitives in Remark \ref{rm:subgaussPO} below.

\begin{remark}[Plausibility of sub-Gaussian Assumption on Treatments and Outcomes and Their Lags] \label{rm:subgaussPO}
Consider Example \ref{ex:VAR}.     Substituting the treatment equation into the outcome equation gives
\begin{align}
\left( \begin{array}{c}
     Y_{it}  \\
     P_{it}
\end{array} \right)
= \sum_{l=1}^L A_{l, it}
\left( \begin{array}{c}
     Y_{i,t-l}  \\
     P_{i,t-l}
\end{array} \right) +
T_{it},
\end{align}
where \begin{align}
& A_{l, it} = 
\left[ \begin{array}{cc}
   \delta_{0l}^{EE} + K_{it}'\beta_0 \delta_{0l}^{PE}  &  \delta_{0l}^{EP} + 
   K_{it}'\beta_0 \delta_{0l}^{PP}\\
    \delta_{0l}^{PE} &   \delta_{0l}^{PP}
\end{array} \right], \\
& T_{it} := \left[ \begin{array}{l}\bar{X}_{it}'\bar{\delta}^E_0 +{ \bar{M}_{i}'\delta^E_{M0} + \xi^E_i+ U_{it}} + K_{it}'\beta_0(\bar{X}_{it}'\bar{\delta}^P_0 + {\bar{M}_{i}'\delta^P_{M0} + \xi_i+ V^p_{it}}) \\
\bar{X}_{it}'\bar{\delta}^P_0 + {\bar{M}_{i}'\delta^P_{M0} + \xi_i+ V^p_{it}}
\end{array} \right]
\end{align}
We can represent the reduced form as the canonical vector auto-regression of order 1:
\begin{align*}
& F_{it} = \Pi_{it} F_{i,t-1} + \varphi_{it}, \quad  F_{it}= [(Y_{i,t-l}, P_{i,t-l})_{l=0}^{L-1}]',
\end{align*}
where 
$$
\underset{(2L \times 2L)}{\Pi_{it}}:= \left[ \begin{array}{ccccc}
    A_{1,it} &  A_{1,it} & \ldots  & A_{L-1,it} & A_{L,it} \\
    I_2 &  0_{2} & \ldots &  0_2 & 0_2 \\
    0_2 &  I_2 &  0_2 & \ldots  &0_2 \\
    \vdots &  \ddots  & \ddots  & \ddots & \vdots \\
    0_2 & \ldots & 0_2 & I_2 & 0_2
\end{array} \right], \qquad  \underset{(2L \times 1)}{ \varphi_{it} }:= \left[ \begin{array}{c}
   T_{it} \\
    0_2   \\
    0_2  \\
    \vdots  \\
    0_2  
\end{array} \right].
$$

 We note that there are no restrictions on $L$ here, but our conditions implicitly restrict $K_{it}'\beta_0$ to be bounded. Assume that the following conditions holds uniformly in $(i,t)$: (1) The initial condition $F_{i,0}$ and $T_{it}$'s are  $\bar \sigma^2$-sub-Gaussian vectors. (2) The singular values $\lambda(\Pi_{it})$ of $\Pi_{it}$ obey $\| \lambda(\Pi_{it})\|_\infty \leq 1- \delta$
for some constant $\delta>0$.Then,  $\|F_{it}\|$ is $ A \bar \sigma^2/(1-\delta)$- sub-Gaussian, for some numerical constant $A$.
\end{remark}

Consider the following condition for learning the treatment reduced form:

\begin{quote}
(FS-TL) 
Consider the model 
$$
  P_{it}= X_{it}'\delta^P_0 + \xi_i +  V_{it}^P,
$$
for each $i$, the residuals $V^P_{it}$ are a martingale difference sequence with respect to filtration $\Phi_{it}$. Suppose  (a) the vectors $\delta^P_0$ and $\xi$ are $(\nu,S)$-weakly sparse with  $S=S^P$ and $S=N^{-\nu/2} S^P$, respectively, and  $S^P$ $=$ $O (N^{1/2} \log^{3 \nu/2} (d_X + N) T^{-\nu/2})$; $\| \delta^P_0 \|_1$ is bounded;
(b) The Gram matrix $$\Psi_{X} :=(NT)^{-1} \sum_{i=1}^N \sum_{t=1}^T{\Ep}_{} [X_{it} X_{it}']$$ has all of its eigenvalues bounded from above and below by $B_{\text{max}}$ and $B_{\text{min}}$, where  $0<B_{\min} \leq B_{\text{max}}$ are finite constants; (c) Each element of $X_{it}$ and $V^P_{it}$ is $\bar \sigma^2$-sub-Gaussian, where $\bar \sigma^2$ is a finite constant; (d) $\log (d_X NT)/ N = o(1)$.
\end{quote}

All the constants are understood to be independent of $(N,T)$. The condition FS-TL summarizes Assumptions A1-A3 in \cite{KockTang}.  Plugging $\lambda_P = C_P \sqrt{NT \log^3 (d_X +N)}$ into the stochastic bounds in Corollary A.1, p. 332, \cite{KockTang} results in the lemma below, which establishes the properties of the first stage treatment Lasso. 

\begin{lemma}[First-Stage Treatment Lasso]\label{lem:kocktang1}
Under Condition (FS-TL), the Lasso estimator in Example \ref{ex:kocktang} with $\lambda_P = C_P \sqrt{NT \log^3 (d_X +N)}$ large enough obeys the following bounds wp $\to 1$
\begin{align}\label{eq:boundmain}
    \| \widehat \delta^P - \delta^P_0 \|_1 \leq  N^{-1/2} \zeta_{NT, \infty}, \quad
    \| \widehat \xi - \xi_0 \|_1 & \leq \zeta_{NT, \infty}, 
\end{align}
where, for some large enough constant $\bar C_P$,
\begin{align}
\label{eq:zetantp}
 \zeta_{NT, \infty} &= \bar C^P S^P\left(  T^{-1/2 } \log^{3/2} (d_X +N)\right)^{(1-\nu)}.
\end{align}
\end{lemma}

\begin{remark}[Time-Invariant Covariates]\label{remark:Identification}
Consider Example \ref{ex:kocktang}. Suppose the condition FS-TL holds with time-invariant fixed covariates $(\bar M_i)_{i=1}^N$.  There are (infinitely) many ways to decompose the  total effect vector $a=(a_1,a_2,\dots,a_N)$ into the  observable part $M_i'\delta$ and remainder part $\xi_i$:
\begin{align*}
    a_i = M_i' \delta + \xi_i, \quad \forall i: \quad i=1,2,\dots, N.
\end{align*}
Given the sparsity parameters $(\nu, S^P)$, the minimizer $(\widehat \delta, \widehat \xi)$ of the Lasso optimization problem \eqref{eq:KockTangprice}  obeys the bound \eqref{eq:boundmain} expressed in terms of $(\nu, S^P)$. Thus, we are free to choose $(\delta, \xi)$  whose sparsity parameters  $\nu,S$ imply the tightest bound on $\zeta_{NT, \infty}$ in \eqref{eq:zetantp}. 

Multiple sparse decompositions must be equivalent in the following sense. Consider two possible decompositions 
\begin{align*}
    a_i = M_i' \delta^1 + \xi^1_i = M_i' \delta^2 + \xi^2_i, \quad \forall i,
\end{align*}
with $\zeta^1_{NT, \infty}$ and $\zeta^2_{NT, \infty}$ rates, respectively, determined by the weak sparsity parameters of $(\delta^1,\xi^1)$ and $(\delta^2,\xi^2)$.  Let $\widehat \xi$ be any given minimizer to Lasso problem. Then, these decompositions must be equivalent in the following sense:
\begin{align*}
    \| \xi^1 - \xi^2 \|_1 \leq  \| \xi^1 - \widehat \xi \|_1 + \|  \widehat \xi - \xi^2 \|_1 \leq 2 \max (\zeta^1_{NT, \infty}, \zeta^2_{NT, \infty}).
\end{align*}

Further, consider the following example with exact sparsity. Suppose $\xi$ is $s$-exactly sparse and  $M$ is a one-dimensional covariate drawn from Bernoulli distribution.  $M_i \sim Bern(p_M)$ such that $s \ll N p_M$. Consider 
\begin{align*}
    a_i = M_i \delta^1 + \xi^1_i = M_i \delta^2 + \xi^2_i, \quad \forall i,
\end{align*}
such that $\xi_i^1 \neq \xi_i^2$ for at least one $i$ (i.e., $\xi^1 \neq \xi^2$ as vectors). Then, $\delta^1 \neq \delta^2$, and the vector $M_i (\delta^1-\delta^2) \neq 0$ for at least $N p_M/2$ entries, w.p. $1-o(1)$. Since $s \ll Np_M$, it cannot be the case that $\xi^1$ and $\xi^2$ are $s$-sparse at the same time. In this case, there exists a single decomposition $(\delta,\xi)$ obeying sparsity assumption.

 \end{remark}

Consider the following condition:
\begin{quote}
(FS-OL). Consider the model of Example \ref{ex:kocktangoutcome}:
$$
  Y_{it}  = D_{it}' \beta_0 + X_{it}'\delta_0^P + \xi^E_i + U_{it},
$$
where the residuals $U_{it}$ are an m.d.s with respect to the filtration $\Phi_{it}$.  Suppose
(a)  $\xi^E = (\xi^E_i)_{i=1}^N \in \mathrm{R}^N$ and $(\delta^E_0, \beta_0) \in \mathrm{R}^{d_{DX}}$ are $(\nu^E,S^E)$ and $(\nu^E,N^{-\nu^E/2} S^E)$ 
-weakly sparse vectors with   $S^E = O (N^{1/2} \log^{3 \nu^E/2} (d_X +N) T^{-\nu^E/2})$; $\| (\beta_0, \delta^E_0) \|_1$ is bounded (b) The Gram matrix $$\Psi_{DX} :=(NT)^{-1} \sum_{i=1}^N \sum_{t=1}^T{\Ep}_{} [(D_{it}, X_{it}) (D_{it},X_{it})']$$ has all of its eigenvalues bounded from above and below by $B_{\text{max}}$ and $B_{\text{min}}$, where  $0<B_{\min} \leq B_{\text{max}} < \infty$, w.p. $1-o(1)$. (c) Each element of $D_{it}$, $X_{it}$ and $U_{it}$ is $\bar \sigma$-sub-Gaussian, where $\bar \sigma$ is a finite constant. 
\end{quote}

The following lemma follows from \cite{KockTang}, and establishes the properties of the first stage outcome Lasso.

\begin{lemma}[First-Stage Outcome  Lasso]\label{lem:kocktang2} Under the condition (FS-OL), the estimator $(\check{\beta}, \widehat{\delta}^E, \widehat{\xi}^E)$ defined in Example \ref{ex:kocktangoutcome}  obeys the following bounds wp $1-o(1)$:
\begin{align}
     \label{eq:boundout}
    \| (\check \beta, \widehat \delta^E) - (\beta_0, \delta^P_E) \|_1  \leq N^{-1/2} \zeta^E_{NT, \infty}, \quad  \| \widehat \xi^E - \xi^E_0 \|_1 \leq \zeta^E_{NT, \infty},
\end{align}
where, for some sufficiently large constant $\bar C_E$, 
\begin{align}\label{eq:zetante}
\zeta^E_{NT, \infty}=\bar C_E S^E\left( T^{-1/2}  \log^{3/2} (d_X +N) \right)^{(1-\nu^E)}.
\end{align}
\end{lemma}

 Next, we proceed to the construction of nuisance realization sets for the treatment and outcome models.

\begin{remark}[Realization Sets for Reduced Form for Base Treatment] \label{rm:realizationTR}
 Define $P_{NT}$ as a collection of $N$-vector functions
\begin{align}\label{eq:boundsp2}
    P_{NT} = \left \{ 
    \begin{array}{l} \mathbf{p}(x)= \{p_i(x_i)\}_{i=1}^N = 
    \{x_i' \delta^P + \xi_i\}_{i=1}^N: \\   \| \delta^P -  \delta^P_0 \|_1 \leq   N^{-1/2} \zeta_{NT, \infty}  , \| \xi - \xi_0 \|_1 \leq  \zeta_{NT, \infty}  \\\end{array} \right \}.
\end{align}
Under Condition (FS-TL), the mean square rate $\textbf{p}_{NT}$ in \eqref{eq:prate} obeys $\textbf{p}_{NT}  = O(N^{-1/2} \zeta_{NT, \infty})$, and the sup-rate upper bound  chosen as $\textbf{p}_{NT,\infty}:=2\zeta_{NT, \infty}$
satisfies the sup-rate condition \eqref{eq:suprate}.
\end{remark}

In the context of Example \ref{HTE}, define $D_{NT}$ as a collection of $N$-vector functions
\begin{align*}
    D_{NT} = \{ \mathbf{d}(x) = \{d_i (x_i)\}_{i=1}^N = \{p_i(x_i) K(x_i)\}_{i=1}^n:  \quad \mathbf{p}(x)=\{p_i(x_i)\}_{i=1}^N \in P_{NT}
    \}.
\end{align*}. The following remark helps verifying Assumption \ref{ass:smallbiashds} in models with base treatment structure.

\begin{remark}[Deducing Rates in Base Treatment Cases]
\label{rm:worst}
Consider the model \eqref{eq:APLM}-\eqref{eq:dtreat} with single base treatment structure  \eqref{eq:lineartreat}. Suppose the matrix $K(\cdot)$ in \eqref{eq:lineartreat} has a.s. bounded entries. Suppose the base treatment reduced form  vector $\mathbf{p}_{0} (\cdot)$ in \eqref{eq:priceout} converges at rates $ \textbf{p}_{NT}$ and $ \textbf{p}_{NT, \infty}$. Then, the worst-case rates  $\textbf{d}_{NT} $ and  $\textbf{d}_{NT,\infty}$ of the technical treatment reduced form  in \eqref{eq:lineartreat}  obey  $\textbf{d}_{NT} =O( \textbf{p}_{NT})$ and $\textbf{d}_{NT,\infty} =O( \textbf{p}_{NT, \infty})$.  This follows from the  Cauchy-Schwartz inequality.
\end{remark}

\begin{remark}[Realization Sets for Reduced Form for Outcome]
\label{rm:realization2}

Suppose  the matrix $$\Psi_{D} :=(NT)^{-1} \sum_{i=1}^N \sum_{t=1}^T{\Ep}_{} [ (P_{it} K_{it}' \beta_0)^2 X_{it} X_{it}'] $$
has all of its eigenvalues bounded from above and below by $B_{\text{max}}$ and $B_{\text{min}}$, where  $0<B_{\min} \leq B_{\text{max}} < \infty$, w.p. $1-o(1)$ and suppose $\| K (X_{it}) \|_{\infty} \leq \bar{K} < \infty \text{ a.s.} $ for some  constant $\bar{K}$. Define the outcome nuisance realization set \begin{align}
    L_{NT} =
     \left \{ 
    \begin{array}{l} 
    \mathbf{l}(x)=(l_i(x_i))_{i=1}^N
    =\{d_i(x_i)'\beta + x_i' \delta^E + \xi^E_i\}_{i=1}^N:  \mathbf{d}(x) \in D_{NT}, \\
    \| \beta - \beta_0\|_1+ \| \delta^E -  \delta^E_0 \|_1 \leq   N^{-1/2} \zeta^E_{NT, \infty}  ,  \| \xi^E - \xi^E_0 \|_1 \leq  \zeta^E_{NT, \infty}\\\end{array} \right \}.
\end{align}
Under Condition (FS-OL), the mean square rate $\textbf{l}_{NT}$ in \eqref{eq:prate} obeys 
$$\textbf{l}_{NT}  = O(N^{-1/2} (\zeta_{NT, \infty} + \zeta^E_{NT, \infty})),$$
and the sup-rate upper bound chosen as
$\textbf{l}_{NT,\infty}:=2(\bar{K} \| \beta_0 \|_1 \zeta_{NT, \infty}+\zeta^E_{NT, \infty})$
satisfies the sup-rate condition \eqref{eq:suprate}.
\end{remark}

Combining the results from Remarks \ref{rm:subgaussPO}--\ref{rm:realization2}, we provide sufficient conditions to verify Assumption \ref{ass:smallbiashds}.

\begin{remark}[Verification of Assumption \ref{ass:smallbiashds} for  First-Stage Lasso Estimators]
\label{rm:verification}
Consider the setup of Remarks \ref{rm:subgaussPO}--\ref{rm:realization2} with  $\nu,\nu^E <1$. Suppose the scales $S$ and $S^E$ are not too big, namely,
\begin{align*}
 (S^P)^2 N^{-1/2} T^{\nu-1/2}    \log^{3 (1-\nu)} (d_X +N)   &= o (1) \\
 S^P \cdot S^E  N^{-1/2} T^{(\nu + \nu^E)/2 -1} \log^{3 ( 1 - (\nu + \nu^E)/2 )} (d_X +N) &= o(1).
\end{align*}
Adding the equations above and multiplying by $(NT)^{1/2}$ gives
$$
\sqrt{NT} (\mathbf{p}^2_{NT} + \mathbf{p}_{NT} \mathbf{l}_{NT} ) = o(1),
$$
which suffices for  \eqref{eq:rate}. Likewise,  assuming
\begin{align*}
\zeta^P_{NT, \infty} = \bar C^P S^P\left(  T^{-1/2 } \log^{3/2} (d_X +N)\right)^{(1-\nu)} = o ( \log^{-1/2} (d NT)), \\ \zeta^E_{NT, \infty} =  \bar C^E S^E\left(  T^{-1/2 } \log^{3/2} (d_X +N)\right)^{(1-\nu_E)} = o ( \log^{-1/2} (d NT))
\end{align*}
directly verifies \eqref{eq:suprate2}. \qed
\end{remark}

Orthogonal Lasso achieves an oracle rate for CATE estimation, which can be strictly better than the  non-orthogonal approach. The comparison is provided in terms of the upper bounds on the rates. 

\begin{remark}[Improvement of Orthogonal Lasso upon One-Stage Lasso]
\label{rm:improvement}
Suppose the treatment effect vector $\beta_0$ is `'less complex'' than the first-stage parameter $(\delta^E_0, \xi^E_0)$, that is
\begin{align}
    \label{eq:above}
    s \log d \ll   (S^E)^2 T^{\nu^E} \log^{3(1-\nu^E)} (d_X +N) 
\end{align}
Then, under Assumption \ref{ass:smallbiashds}, the upper bound on the oracle Lasso rate is attained. Diving \eqref{eq:above} by $NT$ and taking square root gives
\begin{align*}
\sqrt{s \log d /NT}  = o (N^{-1/2} \zeta^E_{NT, \infty}),
\end{align*} 
where $N^{-1/2} \zeta^E_{NT, \infty}$ is the mean square rate bound of the preliminary non-orthogonal estimator $\check \beta$ in \eqref{eq:KockTangoutcome}.
\end{remark}

\section{Empirical application}
\label{sec:empirical}

\setcounter{equation}{0}
\setcounter{definition}{0}

To show the immediate usefulness of the method, we consider the problem of inference on demand elasticities for grocery products. Our transactional data come from a major European food distributor that sells to retailers.  The identifier of each observation consists of  the cross-sectional index -- the product code, the store location,  the distribution channel (i.e., Collection or Delivery) -- and the timestamp. For each value of the index, we compute weekly averages of  the price and the quantity sold. Overall, we have  $1, 163$ unique products, sold at $8$ site locations via $2$ delivery channels, at $T=208$ time periods (weeks). In addition to the transactional data, we have access to the product catalog, which classifies products into a tree. For example, the product code \textit{Vanilla Soft Scoop Ice Cream 4 Ltr package} is classified into a hierarchy whose \url{Level 1} is Sweets, \url{Level 2} is Ice Cream $\&$ Shakes  $\&$ Syrups, and  \url{Level 3} is Ice Cream.  We filter out observations whose either price or sales is zero, which constitutes less than $5 \%$ of the sample.

The next step is to convert the categorical covariates representing classification into a vector of binary covariates.  For each node $j$ and the product $i$, the binary indicator for the node $j$ is equal to one if the product $i$ belongs to the node $j$ and zero otherwise.  Since a binary indicator for a parent and all its children creates a linearly dependent covariate set, we exclude one child category for each parent.  In the absence of any restrictions, different excluded categories   yield  numerically equivalent results.   Under the sparsity assumption  \eqref{eq:sparsity}, this is no longer the case.  The sparsity assumption requires that most siblings have similar treatment effects, albeit for a small number of  exceptions whose identities are unknown. To obtain the sparse treatment modification effect, one has to exclude category that belongs to the majority (i.e., is not an exception). We assume that the store brand belongs to the majority, and can be taken as the baseline (excluded) category.

\begin{figure} 
 	\begin{center}
	\begin{subfigure}[b]{.9\textwidth}
        \includegraphics[scale = 0.085]{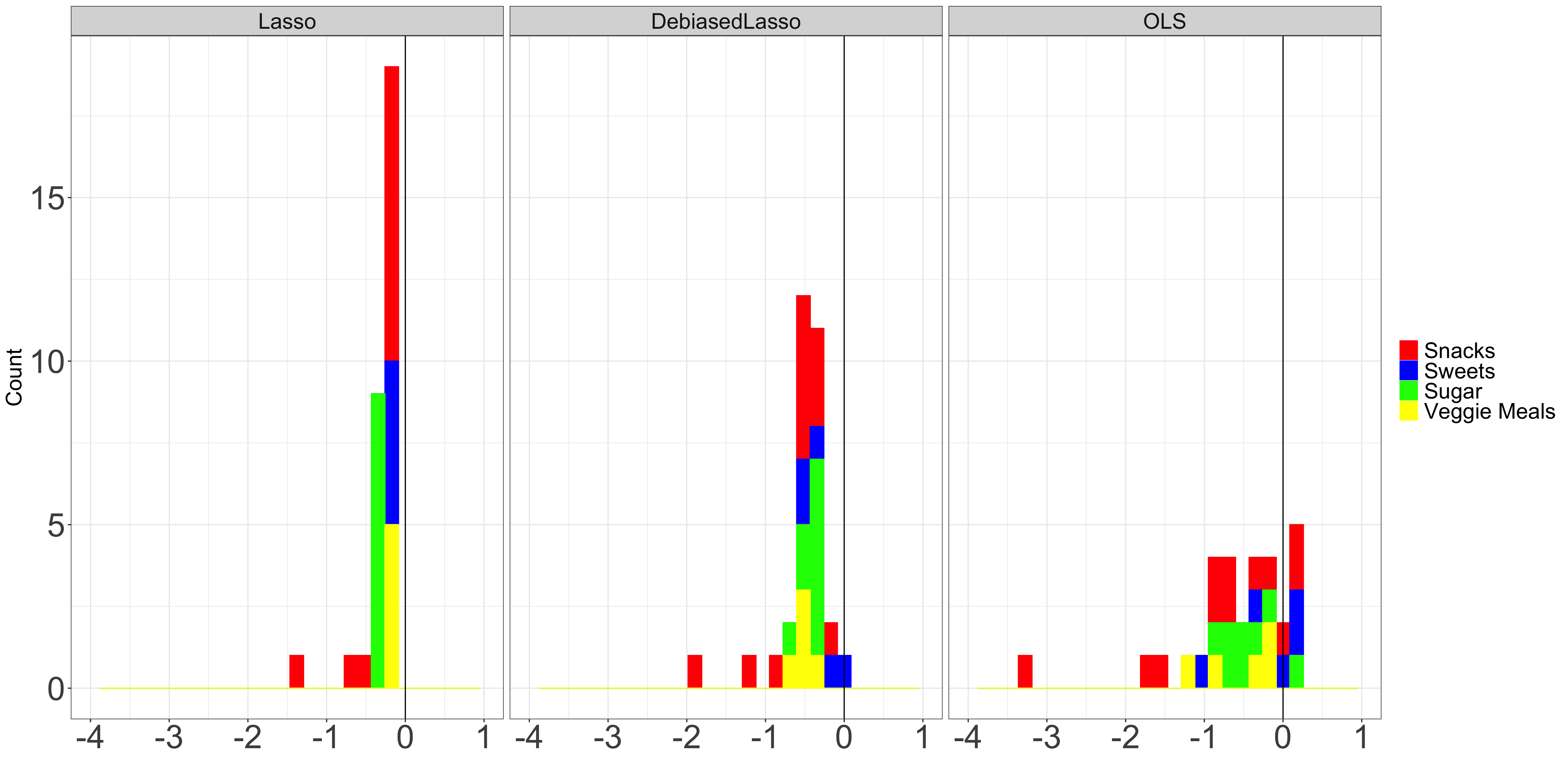}
		\caption{Level 2 groups whose total is $d=31$}
		\label{fig:Level2Snacks}
	\end{subfigure}
	\begin{subfigure}[b]{.9\textwidth}
        \includegraphics[scale = 0.085]{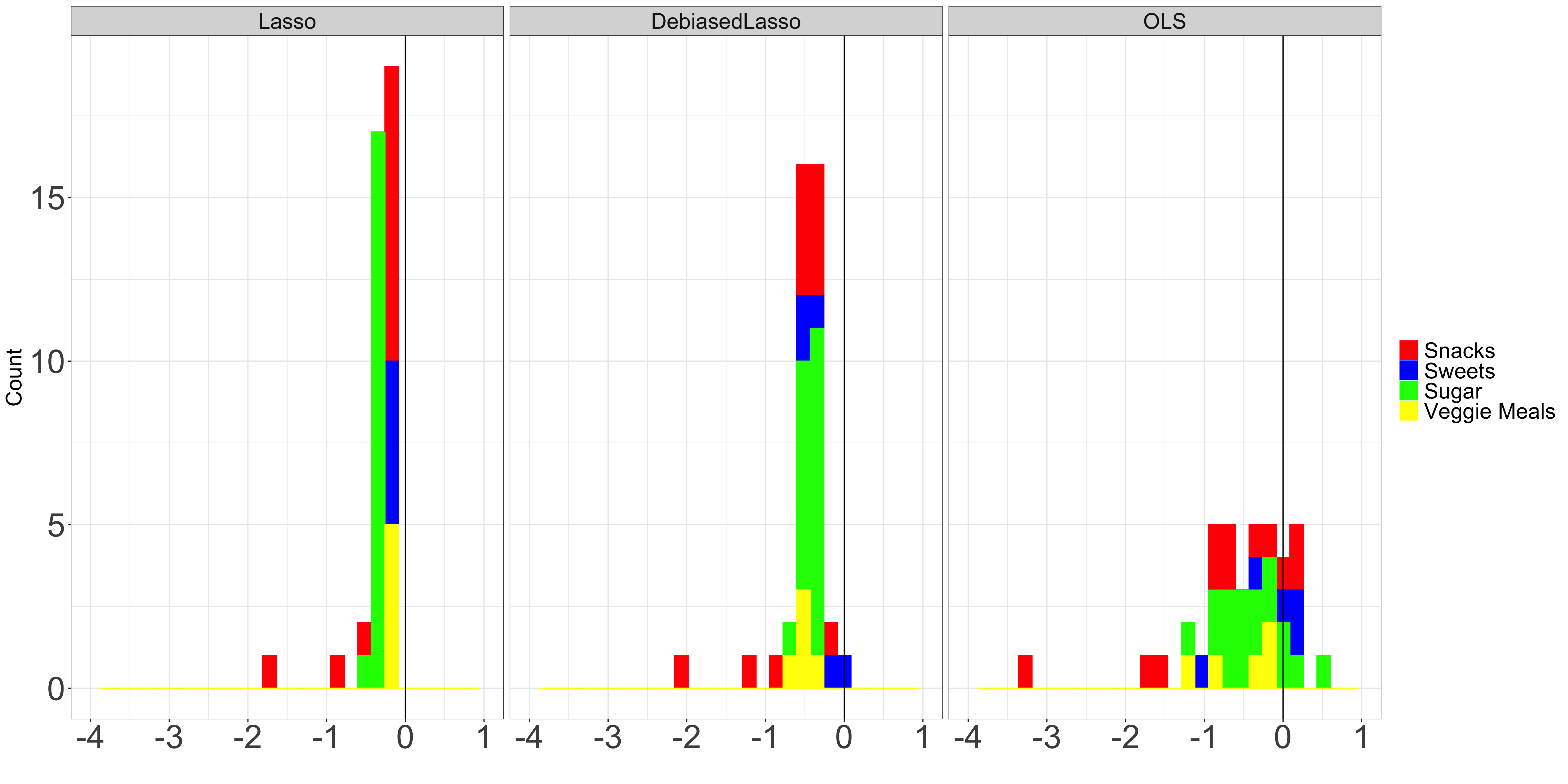}
		\caption{Level 3 groups whose total is $d=40$}
		\label{fig:Level3Snacks}
	\end{subfigure}

\end{center}
	\caption{Histogram of estimated price elasticities for each category of Level 2 (Figure \ref{fig:Level2Snacks}) and Level 3 (Figure \ref{fig:Level3Snacks}) for Snacks. Estimates: Orthogonal Lasso (left panel), Debiased Orthogonal Lasso (middle panel),  OLS  (right panel).   See text for details. }
	\label{fig:HistSnacks}
\end{figure}

We postulate a partially linear dynamic panel model for weekly $\log$ demand
\begin{align}
\label{eq:logsales}
&\log (Q)_{it} = \log (\text{P})_{it} \cdot \bigg(\sum_{h \in \mathcal{H} }1\{ h(i) = h \} \cdot \beta_{0h}\bigg)  \\
&+ ( \log (\text{P})_{it-1} ,\log (Q)_{it-1})' (\alpha^P_1, \alpha^S_1) +   \gamma^E_{h(i)} + a_{pc(i)}^E + \rho^E_{s(i)} + \zeta^E_{c(i)} +  \gamma_{m(t)}^E + U_{it}, \nonumber
\end{align}
where $i=1,2,\dots, N$ and $t=1,2,\dots, T$ with $T=208$ time periods (weeks).  The cross-sectional unit index  $i$ indicates  the combination of the product  $pc(i)$ at the store $s=s(i)$ offered via $c=c(i)$ channel. The outcome variable $Y_{it}:=\log (Q)_{it}$ is total log demand for unit $i$, and  the   base treatment $P_{it} := \log (\text{P})_{it}$ is the log price.  The hierarchy depth $\mathcal{H}$ varies between Level 2 (Figure \ref{fig:Level2Snacks}) and Level 3 (Figure \ref{fig:Level3Snacks}),  and notation $h(i)$ denotes the hierarchical encoding of the product code $pc(i)$.

The model  is a special case of Example \ref{HTE}. Here,  the interaction covariates are the hierarchy fixed effects 
 $$
 K_i = \cup_{h \in \mathcal{H}} 1 \{ h(i) = h\}
 $$
and  the parameter $\beta_0$ is
$$\beta_0 = (\cup_{h \in \mathcal{H}} \beta_{0h}),$$
which results in the CATE function
 $$
 \epsilon_i(\beta_0) = K_i' \beta_0 = \sum_{h \in \mathcal{H} }1\{ h(i) = h \} \cdot \beta_{0h}
 $$
being equal to the heterogeneous elasticity $ \epsilon_i(\beta_0) $ of unit $i$.  In addition to the hierarchy fixed effects $K_i$, the first-stage controls include the product, store, channel fixed effects
 $$
 Z_i = (\cup_{pc \in PC} 1\{ pc(i) = pc \},  \cup_{s \in S} 1\{ s(i) = s \}, \cup_{c \in C} 1\{ c(i)=c \})
 $$
 and the time effects $M_t = \cup_{m=1}^{12} \{ m(t) = m \}$. Thus, the first-stage controls are  
 \begin{align}
\label{eq:xeit}
X_{it}^E= (\log (\text{P})_{it-1} ,\log (Q)_{it-1}, K_i, Z_i, M_t ).
 \end{align}
 Therefore, \eqref{eq:logsales} is a special case of \eqref{eq:APLM} with  $X_{it} = X^E_{it}$ in \eqref{eq:xeit} and $\xi^E_i = 0 \quad \forall i$:
 $$
 Y_{it} = P_{it} \cdot (K_{i}'\beta_0) +  (X^E_{it})' \delta^E_0 + 0 + U_{it}. 
 $$
 To estimate  $(\beta_0, \delta^E_0)$,  we invoke   the Lasso estimator of Example \ref{ex:kocktangoutcome} with the outcome $Y_{it} =  \log (\text{Q})_{it}$ and the covariate vector $X_{it} = X^E_{it}$ in \eqref{eq:xeit}, restricting $\xi^E = (\xi^E_i)_{i=1}^N$ to be equal to zero. 

The price equation takes the form
\begin{align}
\label{eq:logprice}
 \log (\text{P})_{it}  &=    \log (\text{P})_{it-1}  \cdot \left(\sum_{h \in \mathcal{H}} h (i) \cdot \zeta^P_{0h}\right) + \gamma^P_{h(i)} + a_{pc(i)}^P + \rho^P_{s(i)} + \zeta^P_{c(i)} +  \gamma_{m(t)}^E +    V^P_{it}.
\end{align}
Taking
 \begin{align}
\label{eq:xpit}
X^P_{it} =  (\log (\text{P})_{it-1} \cdot \cup_{h \in \mathcal{H}} 1 \{ h (i) = h \},  K_i, Z_i, M_t)
\end{align}
and $\xi_i = 0$ gives 
$$
P_{it} = (X^P_{it})' \delta^P_0 + 0 + V^P_{it}.
$$
To estimate $\delta^P_0$, we invoke  the Lasso estimator of Example \ref{ex:kocktang} with the outcome $P_{it}$ and the covariate vector $X^P{it}$ as in \eqref{eq:xpit}, restricting $\xi = (\xi_i)_{i=1}^N$ to be equal to zero.

In the second stage, we interact the first-stage price residuals $ \widehat{V}^P_{it} $ with hierarchy fixed effects $K_i$ to obtain treatment residual
$$
\widehat {V}_{it}  = \widehat{V}^P_{it} K_i.
$$
Next, we regress the outcome residual $\widetilde{Y}_{it}=\widetilde{\log (Q)}_{it} $ onto  $V_{it}$
$$
\widetilde{Y}_{it} = V_{it}'\beta_0  + U_{it}.
$$
The Lasso estimator  $\widehat{\beta}_L$ is  as in Definition \ref{OL} with $\lambda_{\beta}$  chosen by cross-validation. To simplify computation, the  debiasing matrix  is taken to be the Ridge inverse, which has similar properties to the CLIME estimator in the moderate-dimensional case.  For each estimate $\widehat{\beta} \in \{\widehat{\beta}_{OLS},\widehat{\beta}_L,  \widehat{\beta}_{DL} \}$, we report a $d$-vector of distinct  heterogeneous elasticities $(\cup_{h \in \mathcal{H}} \widehat{\epsilon}_h (\widehat{\beta}))'$, that is,
 $$
\widehat{\epsilon}_h (\widehat{\beta})= \sum_{\bar{h} \in \mathcal{H}} 1 \{ \bar{h}  = h \}  \cdot \widehat{\beta}_{h}, \quad h \in \mathcal{H}.
$$
We consider two choices of the partition $\mathcal{H}$:  Level 2 partition (Figure \ref{fig:Level2Snacks}) with $d=31$ and Level 3 partition (Figure \ref{fig:Level3Snacks}) with $d=40$, respectively.

Figure \ref{fig:HistSnacks} qualitatively summarizes our results. On each panel, the histogram shows estimated heterogeneous elasticities.   The total number of points is equal to the total number of heterogeneous groups (i.e., the cardinality of $\mathcal{H}$). It is $d=31$ for Figure \ref{fig:Level2Snacks} and $d=40$ for Figure \ref{fig:Level3Snacks}.  A single vertical bar represents a collection of  heterogeneous groups with the same value of estimated elasticity, and its height shows the number of such groups. The distinct parts of the bar are grouped by Level 1 (Snacks, Sweets, Sugar and Veggie Meals), as marked by color.   A small number of distinct bars on Figure \ref{fig:HistSnacks}(b, left plot) indicates that the vector of heterogeneous elasticities' difference $\widehat{\beta}_{h}$ has many zeroes.  
As expected, the Lasso elasticity estimate $\widehat{\beta}_L$ is sparse, which makes the histogram of $\widehat{\epsilon}_h (\widehat{\beta}_L)$ very  concentrated. In contrast, the Debiased Lasso  $ \widehat{\beta}_{DOL}$ is not sparse, and the histogram of  $\widehat{\epsilon}_h (\widehat{\beta}_{DOL})$ is more dispersed. 

We find the Snacks category to be relatively homogenous. For example, Lasso estimates suggest that all Sugar  products  (Figure \ref{fig:HistSnacks} b, left panel, green bar) have the same elasticity value regardless of sugar type or packaging.  As a result, all  $d=40$ heterogeneous groups  can be pooled into $s=7$ distinct ones.  Second-stage shrinkage helps to reduce noise, which proves useful to obtain elasticities consistent with economic theory. For example,  $7$ out of $d=40$ groups have positive OLS estimates, while neither Lasso nor Debiased Lasso have any. We find the debiased Lasso elasticity estimates to be the ones most  consistent with economic theory predictions.

%\textsc{[This section misses th big picture conclusion: Debiased Lasso at Level 3 appears to give more sensible estimates of elasticities. ]}

%\textsc{[Is Appendix an Online Appendix?]}

\section{Extensions}

\label{sec:extension}

The following extensions are not formally covered by theoretical framework of Section \ref{sec:theory}. Nevertheless, we expect the results would extend to these settings with suitable treatment of clustering, given the recent developments of  \cite{chiang2019multiway}. 

\subsection*{Heterogeneous Own and Cross-Price Elasticities with Many Heterogeneous Products}
\label{pricing}
Consider a firm that makes a pricing decision about a large number $N$  of heterogeneous  goods. Let $\mathcal{C}:\{1,2,\dots, N\} \rightarrow \{1,2,\dots G\}$ be a known partition of the products into the set of $G$ independent clusters. The notation $ \mathcal{C}(i)$ stands for all members  of the $i$'th cluster. For any two products $i$ and $j$ from distinct clusters $\mathcal{C}(i) \neq \mathcal{C}(j)$, the cross-price elasticity between $i$ and $j$ is assumed to be zero.  Define the average leave-$i$-out price of products in the $i$'th cluster as
\begin{align}
\label{pminusi}
P_{-it}:= \dfrac{\sum_{j \in \mathcal{C}(i), j \neq i}  P_{jt} }{|\mathcal{C}(i)-1|}.
\end{align}
Suppose that in the short  term the realizations of prices and sales can be approximated by the following partially linear model:
\begin{align}
Y_{it} &=  D_{it}' \beta_0 + e_{0}(X_{it})+ \xi_i^E + U_{it}, \quad \Ep [U_{it}|  (X_{jt}, P_{jt}, \Phi_{jt})_{j \in \mathcal{C}(i) } ] = 0, \label{eq:APLM2}  \\
 D_{it} &= [ K_{it}'P_{it}, K_{it}'P_{-it} ],  \label{eq:dtreat2}  \\
 P_{it} &=  p_{0}(X_{it}) + \xi_i + V_{it}^P, \quad \Ep [ V^{P}_{it}| X_{it}, \Phi_{it}] = 0, \label{eq:price}
\end{align}
where $Y_{it}$ is the log sales of product $i$  at time $t$, $P_{it}$ is the  log price, $X_{it}$ is a $p_{X}$-vector of the observable product characteristics, the lagged realizations of market quantities $Y_{it}, P_{it}$, and the demand-side variables used for strategic price setting by the firm. The symbol $\Phi_{it}$ denotes the full information set  available for unit $i$ prior to period $t$, spanned by lagged realizations of the demand system. The controls $X_{it}$ affect  the price variable $P_{it}$ through   $p_{0}(\cdot)$ and the sales through $e_{0}(\cdot)$.

The technical treatment $D_{it}$ is  formed by interacting  $P_{it}$ and $P_{-it}$ with the observable product characteristics  $K_{it}$ such that $\E K_{it}=0$. The parameter $\beta_0$ stands for the vector of own and cross-price elasticities. In order to assign a causal interpretation to $\beta_0$, we assume that, after conditioning on all  pre-determined variables, the sales shock $U_{it}$ is mean independent  of all the information $(X_{jt}, \Phi_{jt}, P_{jt})_{j \in \mathcal{C}(i)}$ about members of the $i$'th cluster (i.e., $U_{it}$ is dissociated from  $(X_{jt}, \Phi_{jt}, P_{jt})_{j \in \mathcal{C}(i)}$, \cite{chiang2019multiway}).  The asymptotic results should be  clustered at the level of independent clusters $G$ rather than individual products $N$.

Equation \eqref{eq:priceout} defines the price effect of interest  $$\beta_0= (\beta_0^{own}, \beta_0^{cross}),$$  where $\beta_0^{own}$ and $\beta_0^{cross} $ are $d/2$ dimensional vectors of the own- and the cross-price effect, respectively.  A change in the own price $\Delta P_{it}$ affects  demand via
$$ \Delta D_{it}'\beta_0 = \Delta P_{it} K_{it}' \beta_0^{own}, $$ and a change in the average price $\Delta P_{-it}$ affects  demand via $$\Delta D_{it}'\beta_0 = \Delta P_{-it} K_{it}' \beta_0^{cross}.$$
Let $$\beta_0^{own}  := (\alpha^{own} _0, \gamma^{own} _0) \text{ and  } \beta_0^{cross} := (\alpha^{cross} _0, \gamma^{cross} _0).$$ We see that
\begin{itemize}
\item $\alpha^{own} _0$ is the Baseline Own Elasticity, and $K_{it}'\gamma^{own} _0$ is the Heterogeneous Own Elasticity;
\item $\alpha^{cross} _0$ is the Baseline Cross-Price Elasticity, and $K_{it}'\gamma^{cross} _0$ is the Heterogeneous Cross-Price Elasticity.
\end{itemize}

\appendix

\newpage

\setcounter{page}{1}

    \begin{abstract}
    Appendix \ref{sec:tools:strassen} presents and results on independence couplings.  Appendix \ref{sec:tools:concentration} develops concentration results for weakly-dependent panel data. Appendix C presents the results for high-dimensional CLT for weakly dependent data.    Appendix \ref{sec:app:proofs} contains proofs for Section \ref{sec:theory}, and Appendix \ref{sec:proofs:fs} for Section \ref{sec:fs}.  Appendix \ref{sec:toolslows} contains tail bounds for empirical rectangular matrices in operator norm. Appendix \ref{sec:lowd}
contains the analysis of OLS used in stage 3 of our inference procedure. 

    \end{abstract}

\maketitle

\tableofcontents

\setcounter{definition}{0}    
\setcounter{section}{0}    
\setcounter{assumption}{0}    
\setcounter{remark}{0}

\renewcommand{\theequation}{A.\arabic{equation}}
\renewcommand{\thelemma}{A.\arabic{lemma}}
\renewcommand{\thecorollary}{A.\arabic{corollary}}
\renewcommand{\thetable}{A.\arabic{table}}

\renewcommand{\theremark}{A.\arabic{remark}}
\renewcommand{\theassumption}{A.\arabic{assumption}}
\renewcommand{\thesection}{A}

\appendix

\newpage
\section*{Notation}

\subsubsection*{Notation.} We use the standard notation for numeric and stochastic dominance. For two numeric sequences $\{a_{n}\}_{n \geq 1}$ and $\{ b_{n} \}_{n \geq 1}$,  $ a_{n} \lesssim  b_n$ stands for $a_{n} = O (b_{n})$. For two sequences of random variables  $\{a_{n}\}_{n \geq 1}$ and $\{ b_{n} \}_{n \geq 1}$,   $a_{n} \lesssim_{P}  b_n $ stands for $  a_{n} \lesssim_P (b_{n})$. For a random vector $V$, let $V^{0}:= V - \Ep [V]$ be the demeaned vector.  Let $[N]:=\{1,2,\dots, N\}$, $[T]:=\{1,2,\dots,  T\}$ and $[j] :=\{1,2,\dots,d\}$.  Let $a \wedge b = \min \{ a, b\}$ and  $a \vee b = \max \{ a, b\} $.

\subsubsection*{Matrix and Vector Norms.}  For a vector $v \in \mathrm{R}^d$, denote the $\ell_2$-norm  of $v$ as $\| v\|_2 := \sqrt{\sum_{j=1}^d  v_j^2}$,  the $\ell_1$-norm of $v$ as $\|v\|_1:= \sum_{j=1}^d |v_j|$, the $\ell_{\infty}$-norm of $v$ as $\|v\|_{\infty}:= \max_{1 \leq j \leq d} |v_j|$, and the $\ell_0$-"norm" of $v$ as $\|v\|_0:= \sum_{j=1}^d 1_{\{ v_j \neq 0\}}$. Denote a unit sphere in $\mathrm{R}^d$ as $\mathcal{S}^{d-1} = \{ \alpha \in \mathrm{R}^d: \|\alpha \|=1\}$. For a matrix $A=(a_{ij}) \in  \mathrm{R}^{d \bigtimes d}$,  let its operator norm be  $\|A\|_2 = \sup_{\alpha \in \mathcal{S}^{d-1}} \| A \alpha \|_2$, the elementwise norm be $\| A\|_{\infty} = \max_{1\leq i, j \leq d} | a_{ij} |$,  and the maximal $\ell_1$-row-norm:
$$\|A \|_{1,\infty} = \max_{1 \leq j \leq d} \sum_{i=1}^d |a_{ij}|.
$$

\subsubsection*{Empirical Process Notation} In what follows, we use the standard empirical process notation. For a generic measurable function $f: \mathcal{W} \to \mathrm{R}$ and a generic sample $\{\{ W_{it}\}_{t=1}^T\}_{i=1}^N$,
where $W_{it}$'s take values in $\mathcal{W}$, define the empirical expectation \begin{align*}
\mathbb{E}_{NT} f(W_{it}) = \frac{1}{NT} \sum_{i=1}^N \sum_{t=1}^{T}  f(W_{it})
\end{align*}
and the empirical process: %average as
\begin{align*}
\G_{NT} f(W_{it}) = 
{\sqrt{NT}}  \mathbb{E}_{NT}  [f(W_{it}) - \Ep_{W_{it}} f(W_{it})].
\end{align*}

\newpage

\section{Tools: Strassen and Berbee Couplings. Implications for Cross-Fitting}
\label{sec:tools:strassen}

\subsection{Strassen's Coupling: Weak and Strong Form via Dudley-Philipp} 

Consider the following setup.

Let $S$ be a Polish space and $P_{Z,W}$ be a law on $S \times S$, with marginal laws $P_Z$ on $S$ and $P_W$ on $S$.
Let $(\Omega, \mathcal{B}, \Pr)$ be a probability space and $Z$ be a random variable on $\Omega$ with values in $S$ and law $\mathcal{L}(Z)=P_{Z}$. Assume that $(\Omega, \mathcal{B}, \Pr)$ has been extended to carry a random variable $U$ on $\Omega$, independent of $Z$, with values in $[0,1]$ and law $U(0,1)$. The total variation norm of a signed measure $\nu$ on on the Polish space $T$ is defined as
$$
\|\nu\|_{TV}=\sup_{F \text{closed} } \nu(F).
$$
The total variation distance between laws $P$ and $Q$ defined on the Polish space $T$ is defined by taking $\nu = P-Q$ in the definition above.

We also make use of the following Strassen's weak coupling result (e.g, \cite{villani:OT}, p.7):
\begin{equation}\label{strassen1}
\min_{Z^*, W^*} \{ \Pr(Z^* \neq W^*): \ \mathcal{L}(Z^*) =P_Z, \ \mathcal{L}(W^*) = P_W\} = \frac{1}{2} \| P_Z - P_W\|_{TV},
\end{equation}
where minimization is done over space of random variables $Z^*$ and $W^*$ defined on the probability space $(\Omega, \mathcal{B}, \Pr)$. Note that the problem above is the optimal transportation problem for 0-1 cost; see \cite{villani:OT} for discussion. The above is a special case of Strassen's original result; \cite{Schwarz} (Theorem 1) provides another proof of (\ref{strassen1}).

We now recall the following result.

\begin{lemma}[Strong Coupling; Lemma 2.11, \cite{dudley:philipp}] Let $S$ and $T$ be Polish spaces and $Q$ a law on $S \times T$, with marginal $P_Z$ on S. Let $(\Omega, \mathcal{B}, P)$ be a probability space and $Z$ be a random variable on $\Omega$ with values in $S$ and law $\mathcal{L}(Z)=P_{Z}$. Assume that there is a random variable $U$ on $\Omega$, independent of $Z$, with values in a separable metric space $R$ and law $\mathcal{L}(U)$ on $R$ having no atoms. Then there exists a random variable $W$ on $\Omega$ with values in $T$ and law $\mathcal{L}( (Z, W))=Q$.
\end{lemma}

This result is quoted with minor adaptation of notation. This lemma implies the strong form of Strassen's weak coupling (\ref{strassen1}) as stated in the following lemma.

\begin{lemma}[Strong Form of Strassen's Coupling]\label{strassen:strong}  Given the setup above with a given random variable $Z$, there exists  a random variable $W$ taking values in $S$, defined on the same probability space, and having law $\mathcal{L}(W)=P_W$ such that:
\begin{equation}
\Pr(Z \neq W) = \frac{1}{2} \| P_Z - P_W\|_{TV}.
\end{equation}
\end{lemma}

Proof. Strassen's weak coupling implies that there is a pair of random variables $(Z^*, W^*)$ with law $Q$ and marginals 
$P_Z$ and $P_W$ such that:
$$
\Pr(Z^* \neq W^*) = \frac{1}{2} \| P_Z - P_W\|_{TV}.
$$
Application of the Dudley-Philipp lemma with $S=T$ and $U$ taken to be uniform random variable  implies that for the given $Z$ there is a pairing random variable $W$, such that law of $(Z,W)$ is $Q$. Therefore,
$$
\Pr(Z \neq W) = \Pr(Z^* \neq W^*) = \frac{1}{2} \| P_Z - P_W\|_{TV}.
$$
\qed.

%The following result follows from the application of Strassen's coupling (\ref{schwarz}) and Lemma 2.11 of \cite{dudley:philipp}
%\footnote{Note that Strassen's couping also underlies the Berbee coupling, \cite{Berbee}, which we also use extensively in the proofs to establish tails bounds for averages.} 

\subsection{Independence Coupling}
Consider a special case of the setup above with $S=S_1 \times S_2$ and $T= S$, where $S_1$ and $S_2$ are Polish spaces, and where $Z= (X, Y)$ is a pair of random variables such that $\mathcal{L}(X) = P_{X}$ on $S_1$ and  $\mathcal{L}(Y) = P_{Y}$ on $S_2$, and $\mathcal{L}(X,Y)= P_{X, Y}$.

\begin{lemma}[Strong Coupling with Independence via Strassen-Dudley-Philip]\label{strassen:strong2} Consider the setup above. We can construct  $\widetilde{Y}$ and $\widetilde{X}$ that are independent of each other with laws $\mathcal{L}(X) = P_X$ and $\mathcal{L}(Y) = P_Y$ such that
$$
\Pr \left \{ (X,Y) \neq 
(\tilde X, \tilde Y) \right \}
= \frac{1}{2} \left \| P_{X, Y}- P_{X} \times P_{Y} \right\|_{TV}.
$$
%where $P_{X} \times P_{Y}$ is the law of $(\tilde X, \tilde Y)$. 
\end{lemma}

Proof.  In the previous lemma take
$Z= (X,Y)$ and $W=(\tilde X, \tilde Y)$, and note that  $P_W = P_{X} \times P_{Y}$. 
 \qed

\subsection{Berbee Coupling Extended}
Let $(X,Y)$ be a pair of random variables taking values in the Polish space $S_1 \times S_2$ as in the setup above. Define their coefficient of dependence as:
$$
\gamma(X, Y)=\frac{1}{2}\left\|P_{X, Y}-P_X \times P_Y\right\|_{TV} .
$$
This coefficient vanishes if and only if $X$ and $Y$ are independent. 
%We can rewrite
%$$
%\gamma(X, Y)=\Ep \frac{1}{2}\left\|P_{Y %\mid X}-P_Y\right\|_{TV}
%$$
%where $P_{X \mid Y}$ is the (regular) %conditional law of $X$ given $Y$. Hence if $Y^{\prime}$ is $Y$ measurable
%$$
%\beta\left(X, Y^{\prime}\right) \leq %\gamma(X, Y)
%$$

The following lemma is a minor extension
of Lemma 2.1 of Berbee from real-valued to Polish-space valued random variables.

\begin{lemma}[Berbee Coupling on Polish Spaces]\label{lem:originalberbee}
Let $X=(X_i)_{i=1}^n$ be a collection of random variables taking values on the Polish space $S=(S_1 \times ... \times S_n)$, and 
defined on the same probability space $(\Omega, \mathcal{B}, \Pr)$. Define for $1 \leq i<n$
$$
\gamma^{(i)}=\gamma\left(X_i,\left(X_{i+1}, \ldots, X_n\right)\right) .
$$
The probability space can be extended so that there exist a collection of random variables $\tilde X=(\tilde{X}_i)_{i=1}^n$ that are mutually independent, such that each $\tilde X_i$ has the same law as $X_i$ and
$$
\Pr\left(X \neq \tilde X\right) \leq \gamma^{(1)}+\ldots+\gamma^{(n-1)}.
$$
\end{lemma}

Proof. Assume that $(\Omega, \mathcal{B}, \Pr)$ has been extended to carry a random variable $U$ on $\Omega$, independent of $X$, with values in $[0,1]$ and law $U(0,1)$.

%It is well-known (see e.g. Schwarz 1980) that 

1. Application of strong form of Strassen's coupling in Lemma \ref{strassen:strong} implies that one can construct $\tilde{X}$ as in the statement of the lemma such that
$$
\Pr(X \neq \tilde{X})=\frac{1}{2}\left\|P_X-P_{\tilde{X}}\right\| .
$$

2. (Identical to Berbee). To prove the claim of the lemma, we have to estimate the right hand side.
If $X, Y$ and $\tilde{Y}$ are random variables, with $Y$ and $\tilde{Y}$ having values in the same space, then
$$
\begin{aligned}
\left\|P_{X, Y}-P_{X, \tilde{Y}}\right\| & \leqq\left\|P_{X, Y}-P_X \times P_Y\right\|_{TV}+\left\|P_X \times P_Y-P_X \times P_{\tilde{Y}}\right\|_{TV} \\
&=2 \gamma(X, Y)+\left\|P_Y-P_{\tilde{Y}}\right\|_{TV}
\end{aligned}
$$
 Applying this rule successively one obtains
$$
\begin{aligned}
&\frac{1}{2}\left\|P_{X_1, \ldots, X_n}-P_{X_1} \times \ldots \times P_{X_n}\right\| \\
&\quad \leq \gamma^{(1)}+\left\|P_{X_2, \ldots, X_n}-P_{X_2} \times \ldots \times P_{X_n}\right\| \\
&\quad \leq \ldots \leq \gamma^{(1)}+\ldots+\gamma^{(n-1)}
\end{aligned}
$$
\qed

\begin{corollary}[Berbee's Coupling for Panel Data]
\label{lem:berbee}
Let $ \{X_{i1}, X_{i2}, \dots, X_{iL} \}_{i=1}^N$ be real random matrices  of possibly distinct dimensions. Suppose the sequences 
$(X_{i1}, X_{i2}, \dots, X_{iL})$ are independent over $i$. For each $i$, $(X_{i1}, X_{i2}, \dots, X_{iL})$ is $\beta$-mixing whose coefficients are bounded as
\begin{align}
\label{eq:separation}
\sup_{1 \leq i \leq N} \sup_{1 \leq l \leq L-1} \gamma( (W_{i,1}, \dots, W_{i,l-1}), 
(W_{i,l}, \dots, W_{i,L}))  \leq \epsilon.
\end{align}
The probability space can be extended with random variables $X_{il}^{*}$ distributed as $X_{il}$ such that $X_{il}^{*}$ are independent over $i,l$ and 
\begin{align}
\label{eq:berbee}
\Pr \left (X_{il} \neq X^{*}_{il} \text{ for some }  i,l \right) \leq  N (L-1)  \epsilon.
\end{align}
\end{corollary}
This follows immediately from the union bound and Lemma \ref{lem:originalberbee}. \qed

\subsection{Applications to Cross-Fitting}
Here we recall the setup induced by the NLO construction given in the main text. Let $\mathcal{M}_k$ and $ \mathcal{M}^{\text{qc}}_k$ be two NLO subsets of time indices $\{1,2,\dots, T\}$, for $k=1,...,K$. Define the data blocks
\begin{align}
\label{eq:chunks}
B_k &=   \cup_{i=1}^N B_{ik},  \quad B_{ik}= \{W_{it} \}_{t \in \mathcal{M}_k}; \\
B^{\text{qc}}_k &=   \cup_{i=1}^N B^{\text{qc}}_{ik}, \quad B^{\text{qc}}_{ik}= \{W_{it} \}_{t \in \mathcal{M}^{\text{qc}}_k}. \nonumber
\end{align}
By construction, the time periods in $\mathcal{M}_k$ and $\mathcal{M}^{\text{qc}}_k$ are separated by at least $T_k \geq T_{\text{block}}:=\lfloor T/(K-1) \rfloor$ time periods.

%\vspace{.1in}
%\begin{remark}[Notational Simplification]To simplify notation,  we assume that each $\mathcal{M}_k$ has cardinality $q$ for each $k=1,...,K$.
%\end{remark}

\begin{lemma}[Approximate Independence of Separated Chunks]
Suppose Assumption \ref{ass:sampling} holds with 
\begin{align}
     \gamma(q):= \sup_{\bar t \leq T, i \leq N} \gamma 
     \Big ( \{W_{i t}\}_{t \leq \bar t}, \{W_{i t}\}_{t \geq \bar t+q} 
     \Big)
     \leq  C_{\kappa} \exp(-\kappa q) \label{eq:expmix2}
\end{align}
and $\log N/ T_{\text{block}} = o(1)$. Then,   there exist random elements $B^*_k$ and $B^{{qc}*}_k$ such that (1) $B^*_k$ and $B_k$ are equal in law,  $B^{{qc}*}_k$ and $B^{\text{qc}}_k$  are equal in law, (2) $B^*_k$ and $B^{{qc}*}_k$ are independent, and (3) the event
\begin{align}
\label{eq:berbeeevent}
 \mathcal{E}_{\text{berbee}} &:=  \{ (B_k, B^{\text{qc}}_k) =
 B^*_k, B^{\text{qc}*}_k), \text{ for all } k=1,...,K \}
\end{align}
holds with probability $1-o(1), NT \rightarrow \infty$.

\end{lemma}

\begin{proof}
Invoking Lemma \ref{strassen:strong2} shows that the required variables exist and obey
$$
\Pr \Big ((B_k, B^{\text{qc}}_k) \neq 
 (B^*_k, B^{\text{qc}*}_k) \Big ) \leq \gamma(B_k, B^{\text{qc}}_k) \leq N C_{\kappa} \exp(-\kappa T_{\text{block}}).
$$
Invoking union bound over the partitions $k=1,2,\dots, K$ gives 
$$
\Pr \Big ((B_k, B^{\text{qc}}_k) \neq 
 (B^*_k, B^{\text{qc}*}_k), \text{ for some } k =1,...,K \Big ) \leq K N C_{\kappa} \exp(-\kappa T_{\text{block}}).
$$
Since $K$ is finite and $\log N/ T_{\text{block}} = o(1)$ gives $K N C_{\kappa} \exp(-\kappa T_{\text{block}})=o(1)$.
 \end{proof}

\begin{corollary}[Convenient Rate Implications]
\label{lem:cond}
Consider the setup above. Suppose there exists a sequence $V_{NT}$ such that for some  $ \psi( B^*_k, B^{\text{qc}*}_k) $ is $O_P(V_{NT})$ for some measurable function $\psi$. Then, $\psi(  B_k  ,  B^{\text{qc}}_k)$ is  $O_P(V_{NT})$.
\end{corollary}

\begin{proof}[Proof of Corollary \ref{lem:cond}]
Consider any sequence of constants such that $\ell_{NT} \rightarrow \infty$.
Then 
\begin{align*}
\Pr ( \psi(B_k,B^{\text{qc}}_k)  > \ell_{NT} V_{NT}) &\leq^{i} \Pr ( \psi(B_k,B^{\text{qc}}_k)  > \ell_{NT} V_{NT} \cap   \mathcal{E}_{\text{berbee}}  ) + \Pr ( \mathcal{E}_{\text{berbee}}^c) \\
&\leq^{ii}    \Pr (  \psi(  B^*_k  ,  B^{\text{qc}*}_k) > \ell_{NT} V_{NT}   ) + \Pr ( \mathcal{E}_{\text{berbee}}^c)\\
&=^{iii}o(1),
\end{align*}
where (i) follows from union bound, (ii) holds since $ \psi(  B^*_k  ,  B^{\text{qc}*}_k) =\psi(B_k,B^{\text{qc}}_k) $ on $ \mathcal{E}_{\text{berbee}} $, and (iii) is assumed in the statement of Lemma.\end{proof}

%follows from Lemma 6.1 in \cite{chernozhukov2016double} for independent $B^*_k$ and $B^{{qc}*}_k$.

Consider the following setup.  We assume  all spaces to be separable and complete. Consider the parameter space $\mathcal{T}$ with elements $\eta$, typically a space of functions.   Consider also  a measurable function (the estimation map) $b^{qc} \mapsto \bar \eta (b^{qc})$ that maps  $\mathcal{W}^{T-q+1}$ to $\mathcal{T}$.  Here $\mathcal{W}$ is the metric space containing realizations 
of $W_{it}$ for all $i$ and $t$. Let $\widehat{\eta}_{k}= \bar \eta (B^{\text{qc}}_k)$ denote an estimator constructed on the data  $B^{\text{qc}}_k$. Let $b \mapsto \phi(b; \eta)$ be another measurable mapping, indexed by $\eta$, that maps $\mathcal{W}^q$ to $\mathrm{R}^{d_{\phi}}$. We assume that the composition map $(b, b^{qc}) \mapsto \phi(b; \bar \eta(b^{qc}))$ is measurable.\footnote{Otherwise, can use outer probability measures to work with the bounds below.} 

\begin{corollary}
\label{cor:firstorder}
 Suppose there exists a sequence of sets $\{\mathcal{\bar{T}}_{N,T}\} \subseteq \mathcal{\bar{T}}$ obeying the conditions as $NT \rightarrow \infty$: (A) $\Pr (\widehat{\eta}_{k}\in \mathcal{\bar{T}}_{N,T}) = 1-o(1)$ and (B) for any sequence $\{\eta_{NT}\} \in \mathcal{\bar{T}}_{N,T}$, $\phi (B_k,\eta_{NT}) = O_P (V_{NT})$. Then, $\phi(B_k, \widehat{\eta}_{k}) =  O_P(V_{NT})$. 
\end{corollary}

\begin{proof}[Proof of Corollary \ref{cor:firstorder}]
Invoke Lemma \ref{lem:cond} with $$\psi(b,b^{qc}):= \phi(b, \bar \eta(b^{qc})) 1_{\{ \bar \eta(b^{qc}) \in  \mathcal{\bar{T}}_{N,T} \}}.$$ Union bound implies
\begin{align*}
      \Pr (\phi(B_k, \widehat{\eta}_{k}) \geq \ell_{NT} V_{NT}) 
      &\leq \Pr (  \phi(B_k, \widehat{\eta}_{k})  \geq \ell_{NT} V_{NT} \cap \widehat{\eta}_{k}\in \mathcal{\bar{T}}_{N,T}  ) + \Pr ( \widehat{\eta}_{k}\not \in \mathcal{\bar{T}}_{N,T} )  \\
      &= \Pr (  \psi(B_k, B^{\text{qc}}_k)  \geq \ell_{NT} V_{NT} \cap \widehat{\eta}_{k}\in \mathcal{\bar{T}}_{N,T}  ) + \Pr ( \widehat{\eta}_{k}\not \in \mathcal{\bar{T}}_{N,T} )\\
      &\leq \Pr ( \psi(B_k, B^{\text{qc}}_k) \geq \ell_{NT} V_{NT} ) + o(1), 
\end{align*} 
where the last inequality holds by condition A.  We have that
$$
\Pr \Big ( \psi(B_k, B^{\text{qc}}_k) \geq \ell_{NT} V_{NT} \Big )
\leq  \Pr \Big ( \psi(B^*_k, B^{\text{qc}*}_k) \geq \ell_{NT} V_{NT} \Big ) + o(1),
$$
from the previous proof. By Condition B, 
$$
\Pr ( \psi (B^*_k,B^{{qc}*}_k) > \ell_{NT} V_{NT} \mid B^{\text{qc}*}_k) = 
\Pr ( \phi (B^*_k,\bar{\eta} (B^{{qc}*}_k)) 1\{ \bar{\eta} (B^{{qc}*}_k) \in \bar{\mathcal{T}}_{N,T} \} > \ell_{NT} V_{NT} \mid B^{\text{qc}*}_k)  = o_P(1).
$$
Therefore, using LIE
$$
 \Pr \Big ( \psi(B^*_k, B^{\text{qc}*}_k) \geq \ell_{NT} V_{NT} \Big )
 =  \Ep\left [\Pr \Big (\psi(B^*_k, B^{\text{qc}*}_k) \geq \ell_{NT} V_{NT} \ \Big \vert\  B^{\text{qc}*}_k \Big ) \right]  = o(1),
 $$
 where the final conclusion holds  by the boundedness (and therefore uniform integrability) of the integrand. \end{proof}

\begin{lemma}[Bounds on Cross-Fit Sample Averages]
\label{cor:remterms}
Let $w \mapsto A (w,\eta)$ be a generic (measurable) matrix-valued function defined on $\mathcal{W}$, indexed by the parameter $\eta \in \mathcal{\bar{T}}$.  Define
\begin{align}
\label{eq:genericbias}
    B_{Ak} (\eta) :&= (NT_k)^{-1} \sum_{i=1}^N \sum_{t \in\mathcal{M}_k} {\Ep}_{W_{it}} A (W_{it},\eta)  \\
    V_{Ak} (\eta):&=(NT_k)^{-1} \sum_{i=1}^N \sum_{t \in\mathcal{M}_k}  [A (W_{it},\eta)  - {\Ep}_{W_{it}} A (W_{it},\eta)]      \label{eq:genericdem} 
\end{align}
Suppose there exist sequences of constants $\zeta^B_{NT}$ and $\zeta^V_{NT}$ so that as $NT \rightarrow \infty$ for each $k=1,...,K$:
\begin{itemize} 
\item[(1)] $\Pr (\widehat{\eta}_{k}\in \mathcal{\bar{T}}_{N,T}) = 1-o(1)$.  
\item[(2)] for any sequence $\{\eta_{NT}\} \in \mathcal{\bar{T}}_{N,T}$ and any norm $\vertiii \cdot$, $$ \vertiii{B_{Ak} (\eta_{NT})} = O (\zeta^B_{NT}), \quad \vertiii{ V_{Ak} (\eta_{NT}) } \lesssim_P (\zeta^V_{NT}).
$$
\end{itemize}
Then,
$$
\vertiii { (NT)^{-1} \sum_{i=1}^N \sum_{k=1}^K \sum_{t \in \mathcal{M}_k}  A (W_{it},\hat \eta_k) } \lesssim_P (\zeta^V_{NT}+\zeta^B_{NT}).
$$
\end{lemma}

In our case, we will either use $\vertiii \cdot   = \| \cdot\|_\infty$ (sup-norm) or $\vertiii \cdot  = \| \cdot \|_2$ (operator norm).

\begin{proof}[Proof of Lemma \ref{cor:remterms}]
We invoke   Corollary \ref{cor:firstorder} with $\phi(B_k,\eta):= B_{Ak}(\eta) +   V_{Ak}(\eta). $ The conditions (A) and (B) are directly assumed in  Lemma \ref{cor:remterms} as conditions (1) and (2), respectively. Therefore, 
for each $k \leq K$,
$$ \vertiii{B_{Ak} (\widehat \eta_{k}) + V_{Ak} (\widehat \eta_{k})} \lesssim_P (\zeta^V_{NT}+\zeta^B_{NT}).$$
We next note that with probability converging to one,
$$
(NT)^{-1} \sum_{i=1}^N \sum_{k=1}^K \sum_{t \in \mathcal{M}_k}  A (W_{it},\hat \eta)
=  \frac{T_k}{T}  \sum_{k=1}^K [B_{Ak} (\widehat \eta_{k}) + V_{Ak} (\widehat \eta_{k})]. 
$$
Since $T_k \asymp T$, the claim holds by the triangle inequality and the union bound.
\end{proof}

\newpage

\setcounter{definition}{0}    
\setcounter{section}{0}    
\setcounter{lemma}{0}    
\setcounter{equation}{0}
\setcounter{remark}{0}
\setcounter{theorem}{0}

\renewcommand{\theequation}{B.\arabic{equation}}
\renewcommand{\thelemma}{B.\arabic{lemma}}
\renewcommand{\thecorollary}{B.\arabic{corollary}}
\renewcommand{\thetable}{B.\arabic{table}}

\renewcommand{\theassumption}{B.\arabic{assumption}}
\renewcommand{\thesection}{B}

\section{Tools: Tail Bounds for Maxima of Sums for Weakly Dependent Panels}
Here we collect and develop some useful lemmas, some of which can be of interest.
\label{sec:tools:concentration}

\subsection{Properties of Products of Sub-Gaussians}
A random variable $\xi$ is $(\sigma^2, \alpha)$-sub-Exponential if  \begin{align}
\label{eq:subexp}
{\Ep} {e}^{\lambda \xi} \leq {e}^{\lambda^2 \sigma^2/2} \text{ a.s. } \quad \forall \lambda: |\lambda| \leq \alpha^{-1}.
\end{align}
A $(\sigma^2, 0)$-sub-Exponential is $\sigma^2$-sub-Gaussian. Lemma \ref{lem:mds1} states concentration inequality for a sub-Exponential martingale difference sequence (m.d.s.).

\begin{lemma}[Properties of sub-Gaussian random variables]
\label{lem:subgauss}
(1) Let $\sigma_X, \sigma_Y>0$. If $X$ is $\sigma^2_X$-sub-Gaussian and $Y$ is $\sigma^2_Y$-sub-Gaussian, then $X+Y$ is $(\sigma_X+\sigma_Y)^2$-sub-Gaussian. (2)  Let  $\{X_m\}_{m = 1}^M$ be a sequence of $\sigma^2$-sub-Gaussian random variables.  Then, (2a) $\sum_{m=1}^M X_m$ is $(M^2\sigma^2)$-sub-Gaussian and (2b)  $\max_{1 \leq m \leq M}  X_m \lesssim_P (\sigma \sqrt{\log d})$.  
(3) Furthermore, $\sum_{m=1}^M X_m$ is $(M\sigma^2)$-sub-Gaussian if $\{ X_m\}_{m=1}^M$ are independent. (4) If $Y \in [-B, B] \text{ a.s. }$, $Y$ is $B^2$-sub-Gaussian. (5) If $X$ is $\sigma^2_{X}$-sub-Gaussian conditional on $Y$, and $Y \in [-B, B] \text{ a.s. }$, then $X\cdot Y$ is $\sigma^2_{X}B^2$-sub-Gaussian. (6) If $X_{mn}$ are $\bar{\sigma}^2$-sub-Gaussian for $n=1,2,\dots, \bar{N}$  ($\bar{N}$ finite) and $m=1,2,\dots, M$, then $\max_{1 \leq m \leq M} \prod_{n=1}^{\bar{N}} |X_{mn}| \lesssim_P ((2\bar{\sigma})^{\bar{N}} \log^{\bar{N}/2} (M\bar{N}) )$.  
\end{lemma}

\begin{proof}[Proof of Lemma \ref{lem:subgauss}]
We prove (1).  By Holder inequality, for any $p,q$ in $[1, \infty)$ such that $1/p +1/q=1$,
\begin{align}
\label{eq:subgproof}
\Ep {e} ^{\lambda (X+Y)} \leq (\Ep {e} ^{\lambda p X})^{1/p} (\Ep {e} ^{\lambda q Y})^{1/q} \leq {e}^{\lambda^2/ 2 (p \sigma^2_X + q \sigma^2_Y)}.
\end{align}
Plugging $p = (\sigma_Y + \sigma_X)/\sigma_X$ and $q=(\sigma_Y + \sigma_X)/\sigma_Y$ into \eqref{eq:subgproof} gives \eqref{eq:subg} with $\sigma^2=(\sigma_X+\sigma_Y)^2$.
We prove (2a) by induction over $M$. The statement holds for $M=1$. The inductive step follows from (1) with $\sigma_X = (M-1)\sigma$ and $\sigma_Y = \sigma$. (2b) is Theorem 1.14 in \cite{RigHutter}.
The statements (3) and (4) are Theorem 1.6 and Lemma 1.8 in  \cite{RigHutter}, respectively. To see (5), observe that $\Ep[ X \mid Y ] =0$ a.s. by assumption. LIE gives
$$
\Ep_{X,Y} [ X \cdot Y  ] ={\Ep}_{Y} \Ep[ X \mid Y ] Y = 0.
$$
Furthermore,
$$
\Ep_{Y} \Ep [{e} ^{\lambda XY} \mid Y] \leq {\Ep}_{Y} {e}^{\lambda^2 \sigma^2 Y^2/2} \leq {e}^{\lambda^2 \sigma^2 B^2/2},
$$
which gives the result. (6) Invoking union bound for any $t>0$
\begin{align*}
    \Pr \left( \max_{1 \leq m \leq M}  \prod_{n=1}^{\bar{N}} |X_{mn}| > t\right) & \leq  \sum_{m=1}^M \Pr \left (| \prod_{n=1}^{\bar{N}} X_{mn}| > t \right) \\ & \quad \leq \sum_{m=1}^M \sum_{n=1}^{\bar{N}} \Pr \left (| X_{mn}| > t^{1/\bar{N}} \right)  \leq  2 M \bar{N} e^{-t^{2/\bar{N}}/2 \bar{\sigma}^2}.
\end{align*}
Taking $t:= \bar{C} (2\bar{\sigma})^{\bar{N}} \log^{\bar{N}/2} (M\bar{N})$ and setting $\bar{C} \rightarrow \infty$ makes the R.H.S above $o(1)$.  Conclude that $$\max_{1 \leq m \leq M} \prod_{n=1}^{\bar{N}} |X_{mn}| \lesssim_P ((2\bar{\sigma})^{\bar{N}} \log^{\bar{N}/2} (M\bar{N})).$$

\end{proof}

\subsection{Tails Bounds for Maxima of Sums of Martingale Differences}

\begin{lemma}[Martingale Difference Sequences; Theorem 2.19 in \cite{Wainwright}]
\label{lem:mds1}
(1) Let $\{ (\xi_m, \Phi_m) \}_{m=1}^M$ be an m.d.s.  obeying  $${\Ep} [{e}^{\lambda \xi_m} \mid \Phi_{m-1} ] \leq {e}^{\lambda^2 \sigma^2/2} \text{ a.s. }$$
for any $\lambda$ such that $|\lambda| \leq \alpha^{-1}$. Then, the following bounds hold:
(1) The sum  $\sum_{m=1}^M \xi_m$ is $(\sigma^2 M, \alpha)$-sub-Exponential and satisfies concentration inequality
\begin{align*}
    \Pr \bigg( \bigg|  \sum_{m=1}^M \xi_m \bigg| \geq t \bigg) \leq \begin{cases} 2 {e}^{-t^2 /(2 M\sigma^2 )}, \quad  & 0 \leq t \leq  M\sigma^2/\alpha \\
    2 {e}^{-t /(2\alpha)}, \quad & t>  M\sigma^2/\alpha
    \end{cases}
\end{align*}
(2) For each $j: 1 \leq j \leq d$, let $\{ (\xi_{mj}, \Phi_m) \}_{m=1}^M$ be an m.d.s obeying the conditions above with the same parameters $(\sigma^2, \alpha)$.  Then, 
\begin{align}
\label{eq:maxineqsubexp}
    \Pr \bigg(  \| M^{-1} \sum_{m=1}^M \xi_m  \|_{\infty} > t \bigg)  \leq \begin{cases} 2  {e}^{\log d -t^2 M/(2 \sigma^2 )}, \quad  & 0 \leq t \leq  \sigma^2/\alpha \\
    2  {e}^{\log d -t M/(2\alpha)}, \quad & t>  \sigma^2/\alpha
    \end{cases}
\end{align}

%In particular, if  $\{ \xi_m \}$ are uniformly $\sigma^2$-sub-Gaussian, $\| M^{-1} \sum_{m=1}^M \xi_m \|_{\infty} \lesssim_P ( \sigma \sqrt{\log d/M})$.
\end{lemma}
\begin{proof}[Proof of Lemma \ref{lem:mds1}]
Lemma \ref{lem:mds1} is essentially  Theorem 2.19 in \cite{Wainwright}. Replacing $\xi$ by $c \cdot \xi$ in \eqref{eq:subexp} shows that 
$c \cdot \xi$ is $(c^2 \sigma^2, c \alpha)$-sub-Exponential.
\end{proof}

\begin{lemma}
\label{lem:mds}
Let $1 \leq i \leq N$ and $1 \leq t \leq T$ be the unit and the time indices. Denote the index $m$ as
\begin{align}
\label{eq:mit}
m=m(i,t) = T(i-1) + t.
\end{align}
Consider a sequence \begin{align} \label{eq:ksim} \xi_m =V_{it} U_{it},\quad  m=1,2,\dots, M=NT.\end{align}  Under Assumption \ref{ass:subgauss},
\begin{enumerate}
\item[A] \label{item:a} $\{ \xi_m \}_{m=1}^M$ is a martingale difference sequence with respect to  natural filtration:
\begin{align*}
\Ep[ \xi_m \mid \Phi_{m-1} ] := \Ep[ \xi_m \mid \xi_1, \dots, \xi_{m-1} ] = 0, \quad \forall m=1,2,\dots M.
\end{align*}
\item[B] Given a large enough constant $C_{VU}$ large enough, there exists $NT$ large enough such that the maximal norm of the empirical moment vector obeys:\label{item:b} \begin{equation} \label{eq:mdseq} \Pr \Big (\| \ENT V_{it} U_{it} \|_{\infty} > C_{VU} \sqrt{\log d /NT} \Big) \leq 2/d=o(1). \end{equation} 
%As a result, $\| \ENT V_{it} U_{it} %\|_{\infty} \lesssim_P (\sqrt{\log d /NT})$.
\end{enumerate}

\end{lemma}
\begin{proof}[Proof of Lemma \ref{lem:mds}]
By conditional sequential exogeneity \eqref{eq:APLM} and independence over $i$
 $$\Ep [ U_{it} \mid \cup_{t<t'}  V_{it'} U_{it'}, \cup_{j \neq i} \{ (V_{jt}, U_{jt})_{t=1}^T \} ] = 0 \quad \forall i,t. $$ Therefore, the martingale difference property A holds. Union bound and Assumption \ref{ass:subgauss}  imply
 $$
 \Pr (| V_{it,j} U_{it}| > t) \leq  \Pr (| V_{it,j} | > \sqrt{t}) + \Pr (| U_{it} |> \sqrt{t}) \leq 2 {e}^{-t/2\bar \sigma^2}.
 $$
By Theorem 2.13 in \cite{Wainwright}, $V_{it,j} U_{it}$ is $(\sigma^2, \alpha)$-sub-Exponential for some $\sigma, \alpha>0$ that do not depend on $j$, $N$ or $T$.  Since the cutoff point $\sigma^2/\alpha$ in \eqref{eq:maxineqsubexp} does not depend on $N,T$, for $C_{VU}$ large enough  and sample size $NT$ $$t:= C_{VU} \sqrt{\log d /NT}  \leq \sigma^2/\alpha.$$ The bound follows
\begin{align*}
     \Pr \Big (\| \ENT V_{it} U_{it} \|_{\infty} > C_{VU} \sqrt{\log d /NT} \Big)  \leq 2/d=o(1).
\end{align*}
\end{proof}

\subsection{Tail Bounds for Maxima of Sums of Sub-Gaussian products}

%Sums of sub-Gaussian Errors  under Weak Dependence
\begin{lemma}[Tail Bounds  for Weakly Dependent Matrices, $\ell_{\infty}$-norm]
\label{lem:matrixerrorinf}
Suppose Assumption \ref{ass:sampling}(1)  holds. For each $j=1,2,\dots, d$, let $\phi_{j}(W_{it})$ be centered $\sigma^2$-sub-Gaussian random variable for all $i,t$ where $\sigma = \sigma(N,T)$ and $\phi_{j}(W_{it})$ can depend on $N,T$. Then, 
\begin{align}
\label{eq:bound}
\| S \|_{\infty}:=\max_{1 \leq j \leq d} \left |(NT)^{-1} \sum_{i=1}^N \sum_{t=1}^T \phi_{j}(W_{it})  \right | \lesssim_P \sigma \sqrt{\log (NT) \log d/NT},
\end{align}
\end{lemma}

\begin{remark}[Triangular Arrays]
Note that all variables and $\sigma$ can be indexed by $(N,T)$, but we omit the indexing to keep the notation light. Thus, this lemma and all other lemmas stated below apply to triangular arrays.
\end{remark}

\begin{proof}[Proof of Lemma \ref{lem:matrixerrorinf}]
Let $q$ be the block size such that $1 \leq q \leq T/2$ and let $L = [T/2q]$. Define  the odd blocks
\begin{align}
\label{eq:oddblock}
B_{i(2l-1)}:= [W_{i,(2l-2)q+1}, W_{i,(2l-2)q+2}, \dots, W_{i,(2l-2)q+q}], \quad l=1,2,\dots, L
\end{align}
the even blocks
\begin{align}
\label{eq:evenblock}
B_{i(2l)}:= [W_{i,(2l-1)q+1}, W_{i,(2l-1)q+2}, \dots, W_{i,(2l-1)q+q}], \quad l=1,2,\dots, L
\end{align}
and the remainder block, which can be empty, as 
\begin{align}
\label{eq:remblock}
B_{ir}:=  [W_{i,2Lq+1}, W_{i,2Lq+2}, \dots,  W_{i, T}].
\end{align}
Note that $\{B_{i(2l-1)}\}_{l=1}^L$ obeys \eqref{eq:separation} and   $\{B_{i(2l)}\}_{l=1}^L$ obeys \eqref{eq:separation} with $\epsilon = \gamma(q)$. Let $B_{i (2l-1)}^{*}$ be the Berbee copy of $B_{i(2l-1)}$. Define the Berbee event
\begin{align*}
\mathcal{I}_1 := \{ B_{i (2l-1) }^{*} = B_{i (2l-1)} \text{ for all } i,l \}.
\end{align*}
Likewise,  let $B^{*}_{i(2l)}$ and $\mathcal{I}_2$ be the analogs of $B^{*}_{i(2l-1)}$ and $\mathcal{I}_1$ for even indices. Define  the blockwise sum
\begin{align}
& \phi(B^{*}_{i (2l-1)}):= \sum_{t=(2l-2)q+1}^{t=(2l-2)q+q} \phi(W^{*}_{it}) \label{eq:soddq2} \\
& S^{*}_{\text{odd}}(q):=(NT)^{-1} \sum_{i=1}^N \sum_{l=1}^L \phi (B^{*}_{i (2l-1)}). \label{eq:soddq}
\end{align}
Let $S^{*}_{\text{even}}(q)$ be the analog of $S^{*}_{\text{odd}}(q)$ for even indices. If $T \neq 2Lq$, the remainder block is non-empty, in which case define
\begin{align}
\label{eq:remblockq}
&\phi(B_{ir}) := \sum_{t=2Lq+1}^{T} \phi(W_{it}) \\
&S_{\text{rem}}(q):= (NT)^{-1} \sum_{i=1}^N \phi(B_{ir}).
\end{align}
On the event $\mathcal{I}_1 \cap \mathcal{I}_2$, the union bound gives
\begin{align}
\label{eq:odd}
    \| S \|_{\infty} \leq \| S^{*}_{\text{odd}}(q) \|_{\infty} + \| S^{*}_{\text{even}}(q) \|_{\infty} +\| S_{\text{rem}}(q) \|_{\infty}.
\end{align}
Thus, 
\begin{align}
\label{eq:gammaquniond}
    &\Pr ( \| S \|_{\infty} \geq 3t)  \leq \Pr ( \| \sum_{i=1}^N \sum_{l=1}^L \phi (B^{*}_{i(2l-1)}) \|_{\infty} \geq t)   \nonumber\\
    &+ \Pr ( \| \sum_{i=1}^N \sum_{l=1}^L \phi (B^{*}_{i(2l)}) \|_{\infty} \geq t) + \Pr ( \| \sum_{i=1}^N \phi (B_{ir}) \|_{\infty} \geq t) +2 NL \gamma(q).  \nonumber
\end{align}

For each $j$,  $S^{*}_{\text{odd}j}(q)$ is $ (NT)^{-2} (NL)q^2\sigma^2 \leq  (q/NT)\sigma^2$-sub-Gaussian by Lemma \ref{lem:subgauss};  similarly,for each $j$,  $S^{*}_{\text{even}j}(q)$ is $(q/NT)\sigma^2$-sub-Gaussian.  Note that here the dependency on $L$ is linear and not square, because the Berbee blocks are independent.
For the remainder block, for each $j$, $S_{\text{rem}j}(q)$ is $(NT)^{-2} (N)q^2\sigma^2$ $\leq  (q/NT)\sigma^2$-sub-Gaussian since $q \leq T$ by Lemma \ref{lem:subgauss}, where we use only independence across $i$.Since  $S^{*}_{\text{odd}j}(q)$ is $  (q/NT)\sigma^2$-sub-Gaussian for each $j$,  $\| S^{*}_{\text{odd}}(q) \|_{\infty}\lesssim_P \sigma \sqrt{q \log d/NT}$ by Lemma \ref{lem:subgauss} (2b). Likewise, $\| S_{\text{rem}}(q) \|_{\infty}\lesssim_P \sigma \sqrt{q \log d/NT}$ by Lemma \ref{lem:subgauss} (2b).  Given the parameter $\kappa$ in mixing coefficient \eqref{eq:expmix}, we set block size $q$ to be:
\begin{equation}\label{eq:defq}
q = \lfloor (2/ \kappa) \log (NT) \rfloor. 
\end{equation}
Invoking the bound \eqref{eq:expmix} in  Assumption \ref{ass:sampling} gives
\begin{align*}
\Pr ( \mathcal{I}_1^c ) + \Pr ( \mathcal{I}_2^c) &\leq 2N (L-1)\gamma(q) \leq (2NT/q) \gamma(q) \\
&\leq 2(NT/q) (NT)^{-2} = o ( (NT)^{-1}) = o(1), \quad NT \rightarrow \infty,
\end{align*} 
which implies \eqref{eq:bound}. 

\end{proof}

The following is an extension/clarification of a useful lemma due to  \cite{KockTang}.

\begin{lemma}[Concentration of Products of sub-Gaussian Random Variables with Independent Blocks]
\label{lem:proposition3kocktang}
Suppose the random variables $Z_{n,m,v,j}$
are uniformly $\bar \sigma^2_n$-sub-Gaussian as in \eqref{eq:subg} for $n=1,2,\dots, \bar{N}$, ($\bar{N} \geq 2$ is fixed and finite), $m=1,2,\dots, M$, $v=1,2,\dots, V$, and $j=1,2,\dots, d$. Suppose $Z_{n_1,m_1,v_1,j_1}$ and $Z_{n_2,m_2,v_2,j_2}$ are independent as long as $m_1 \neq m_2$ regardless of the values of other subscripts. 
Then,
\begin{align*}
    \max_{j, v, m} \Ep \left |  \prod_{n=1}^{\bar{N}} Z_{n,m,v,j} \right | \leq C_A \prod_{n=1}^{\bar N} \bar{\sigma}_n,
\end{align*}
for some positive constant $C_A$
that depends on $\bar N$ and with probability approaching 1,
\begin{align*}
    \max_{1 \leq j \leq d} \bigg| (MV)^{-1} \sum_{m=1}^M \sum_{v=1}^{V} \bigg(\prod_{n=1}^{\bar{N}} Z_{n,m,v,j} - {\Ep} \prod_{n=1}^{\bar{N}} Z_{n,m,v,j} \bigg) \bigg| \leq C_V \sqrt{\log^{\bar{N}+1} (dV)/M} \prod_{n=1}^{\bar N} \bar{\sigma}_n,
\end{align*}
for some positive constant $C_V$ that depends only on $\bar N$. 
\end{lemma}
%Note that this lemma exploits independence across the cross-sectional dimension of a panel. The lemma is a consequence of a slightly more general Lemma \ref{KT:extend} proven in the next subsection.
% VS: the sentense above does not make sense. 

%If we have weak dependence across time dimensions, we obtain sharper concentration.

\begin{lemma}[Concentration of Products of sub-Gaussian Random Variables under Weak Dependence]
\label{lem:covmat}
Suppose Assumption \ref{ass:sampling} (1) holds, and let
$\varphi_{nj}(\cdot): \mathcal{W} \rightarrow \mathrm{R}$ be a deterministic function. Suppose that $\varphi_{nj}(W_{it})$ are uniformly $\bar{\sigma}_{n}$-sub-Gaussian as in  \eqref{eq:subg} for $n=1,2,\dots, \bar{N}$ ($\bar{N} \geq 2$ is fixed and finite) and $j=1,2,\dots, d$ and any $i,t$.  Then, 
\begin{align}
\label{eq:constanta}
    \max_{j,i,t} \left | \Ep \left [ \prod_{n=1}^{\bar{N}} \varphi_{nj}(W_{it}) \right ]  \right|  \leq C_A \prod_{n=1}^{\bar N} \bar{\sigma}_n,
\end{align}
for some positive constant $C_A$ that depends on $\bar N$ and with probability approaching 1,
\begin{align}
\label{eq:bound2}
\| S \|_{\infty} :&=\max_{1 \leq j \leq d} \left | (NT)^{-1} \sum_{i=1}^N \sum_{t=1}^T \left [\prod_{n=1}^{\bar{N}} \varphi_{nj}(W_{it}) - \Ep \left [ \prod_{n=1}^{\bar{N}} \varphi_{nj}(W_{it}) \right] \right] \right|\\
&\leq \bar C_V  \sqrt{\log^{\bar{N}+1} (d \log (NT)) \log (NT)/NT} \prod_{n=1}^{\bar N} \bar{\sigma}_n. \nonumber
\end{align}
for some positive constant $\bar C_V$ that depends only on $\bar N$. 
\end{lemma}

\begin{proof}[Proof of Lemma \ref{lem:covmat}] Define
$$
\phi(W_{it}):= \{ \phi_j(W_{it})\}_{j=1}^d, \quad  \phi_j(W_{it}):= \prod_{n=1}^{\bar{N}} \varphi_{nj}(W_{it}).
$$
Let the block size $q$, the odd blocks, even blocks, and remainder blocks, and events $\mathcal{I}_1$ and $\mathcal{I}_2$ be as  defined in the proof of Lemma \ref{lem:matrixerrorinf}. Likewise, let $S^{*}_{\text{odd}}(q)$ be as in \eqref{eq:soddq2}, that is,
\begin{align*}
& \phi(B^{*}_{i (2l-1)}):= \sum_{t=(2l-2)q+1}^{t=(2l-2)q+q} \phi(W^{*}_{it}), \quad  S^{*}_{\text{odd}}(q):=(NT)^{-1} \sum_{i=1}^N \sum_{l=1}^L \phi (B^{*}_{i (2l-1)}),
\end{align*}
$S^{*}_{\text{even}}(q)$ be its analog for the even-numbered blocks, and $S^{*}_{\text{rem}}(q)$ be as in \eqref{eq:remblockq}.
The first claim (\ref{eq:constanta}) is immediate from the previous lemma. Lemma \ref{lem:proposition3kocktang} with $L \geq 2$ and  $M=NL$ and $V=q$ implies that wp $1-o(1)$
 \begin{align*}
\| S^{*}_{\text{odd}}(q)\|_{\infty}:&= (Lq/T) \left \| (NLq)^{-1} \sum_{i=1}^N \sum_{l=1}^L \{\varphi(B^{*}_{i (2l-1)}) - \Ep \varphi(B^{*}_{i (2l-1)})\} \right \|_{\infty} \\
&\leq  (Lq/T) C_V \left(\sqrt{\log^{\bar{N}+1} (dq)/NL}\right) 
\leq^{i} C_V \left(\sqrt{\log^{\bar{N}+1} (dq) q/NT}\right),
\end{align*}
where (i) follows from $L = \lfloor T/2q \rfloor \leq T/2q$ and  $L \geq \lfloor T/2q \rfloor \geq T/2q - 1 \geq  T/4q$. A similar bounds holds for $S^{*}_{\text{even}}(q)$. If $T_{\text{rem}} \neq 0$, Lemma \ref{lem:proposition3kocktang} with $M=N$ and $V=T_{\text{rem}} \leq q$ implies  that wp $1- o(1)$:
 \begin{align*}
 \| S_{\text{rem}}(q)\|_{\infty} &:= T_{\text{rem}}/T \left \| (NT_{\text{rem}})^{-1} \sum_{i=1}^N (\varphi(B_{ir})  - \Ep \varphi(B_{ir}) ) \right \|_{\infty} \\
 & \leq  T_{\text{rem}}/T  C_V \left(\sqrt{\log^{\bar{N}+1} (d T_{\text{rem}} )/N} \right) \\
 &\leq q/T C_V \left(\sqrt{\log^{\bar{N}+1} (d q )/N}  \right).
\end{align*}
Plugging $q^2 /NT^2 \leq q/NT$ into the R.H.S above gives the bound $C_V \left(\sqrt{\log^{\bar{N}+1} (dq) q/NT}\right)$. Let $NT$ be large enough so that $L = \lfloor T/2q \rfloor \geq 2$ and $(2/\kappa) \leq  \log (NT)$ so that $q \leq \log^2 (NT)$ and $dq \leq (d \log (NT))^2$. Collecting the bounds gives \eqref{eq:bound2}. Adding up the bounds and plugging choice of $q = (2/\kappa) \log (NT) $ as in \eqref{eq:defq} and noting that $L = \lfloor T/2q \rfloor \geq 2$ for $T$ large enough   gives \eqref{eq:bound2}. 
\end{proof}

\begin{corollary}\label{lem:cartesian}
Suppose Assumption \ref{ass:sampling} (1) holds. Suppose $Z_{1,nit}$ and $Z_{2,nit}$ are $d$-vectors obtained as (measurable) transformations of $W_{it}$, whose entries are uniformly $\bar{\sigma}_1^2$ and $\bar{\sigma}_2^2$-sub-Gaussian for $n=1,2,\dots, \bar{N}$. Let $U_{it}$ be uniformly $\bar{\sigma}^2$-sub-Gaussian and $g \geq 0$ be a finite power. Then, 
\begin{align}
    &\max_{1 \leq k,j \leq d} \max_{i,t} \left | \Ep \left [ \prod_{n=1}^{\bar{N}} Z_{1,nitk} Z_{2,nitj} U^{2g}_{it} \right]  \right |  \leq C_A  (\bar \sigma_1 \bar \sigma_2 \bar \sigma^{2g})^{\bar N} \label{eq:constanta2} \\
 &\max_{1 \leq k,j \leq d}
 \left | (NT)^{-1} \sum_{i=1}^N \sum_{t=1}^T \left [ \prod_{n=1}^{\bar{N}} Z_{1,nitk} Z_{2,nitj} U^{2g}_{it} - {\Ep} \left [ \prod_{n=1}^{\bar{N}} Z_{1,nitk} Z_{2,nitj} U^{2g}_{it}\right ] \right]   \right|\nonumber\\
 & \leq \bar C_V    (\bar \sigma_1 \bar \sigma_2 \bar \sigma^{2g})^{\bar N}\bigg(\sqrt{\log^{2 \bar{N}+2g+1} (d^2 \log (NT)) \log (NT)/NT} \bigg) \label{eq:covmat2}
\end{align}

\end{corollary}

\begin{remark}
\label{rm:cartesian}
Suppose Assumptions \ref{ass:sampling} and \ref{ass:subgauss} hold. Invoking \eqref{eq:constanta2} with $\bar N=1$ and $Z_{1,it}=Z_{2,it}=V_{it}$  implies for some finite $\sigma_V< \infty$, 
$$
\max_{it} \| \Ep V_{it} V_{it}' \|_{\infty} \leq \max_{itj} \Ep V^2_{itj} \leq \sigma_V^2. 
$$ 
Likewise, Assumption \ref{ass:subgauss} implies for some finite $\sigma_{VU}< \infty$,
$$
\sup_{it} \Ep [U^2_{it} \mid V_{it}] \leq \sigma^2_{VU} \text{ a.s. }.
$$
\end{remark}

\subsection{Some Technical Lemmas}
Here we provide technical extensions of the results in  \cite{KockTang}, keeping the notation as in the original \cite{KockTang} and references therein.  

\begin{lemma}[Theorem 2.1 in \cite{FanGramaLiu}, Proposition F.1 in \cite{KockTang} ]\label{lem:fan}. Let $\alpha \in(0,1)$. Assume that $\left(X_i, \mathcal{F}_i\right)_{i=1}^n$ is a sequence of supermartingale differences satisfying $\sup _i \mathrm{E}\left[e^{\left|X_i\right|^{\frac{2 \alpha}{1-\alpha}}}\right] \leq C_1$ for some constant $C_1 \in(0, \infty)$. Define $S_k:=\sum_{i=1}^k X_i$. Then, for all $\epsilon>0$,
$$
\mathrm{P}\left(\max _{1 \leq k \leq n} S_k \geq n \epsilon\right) \leq C(\alpha, n, \epsilon) e^{-(\epsilon / 4)^{2 \alpha} n^\alpha}
$$
\end{lemma}

where
$$
C(\alpha, n, \epsilon):=2+35 C_1\left[\frac{1}{16^{1-\alpha}\left(n \epsilon^2\right)^\alpha}+\frac{1}{n \epsilon^2}\left(\frac{3(1-\alpha)}{2 \alpha}\right)^{\frac{1-\alpha}{\alpha}}\right] .
$$

\vspace{ .2in}

\begin{lemma}[Proposition F.2 in \cite{KockTang}]\label{lem:KTF.2} Let $\alpha \in(0,1)$. Assume that $\left(X_i, \mathcal{F}_i\right)_{i=1}^n$ is a sequence of martingale differences satisfying $\sup _i \mathrm{E}\left[e^{D\left|X_i\right|^{\frac{2 \alpha}{1-\alpha}}}\right] \leq C_1$ for some positive constant $D$, where $C_1 \geq 1$ can change with the sample size $n$. Then, for all $\epsilon \geq {1/\sqrt{n}}$,
$$
\mathrm{P}\left(\left|\sum_{i=1}^n X_i\right| \geq n \epsilon\right) \leq 
C_1 A(\alpha) e^{-K\left(\epsilon^2 n\right)^\alpha}, \quad K= (D^{\frac{1-\alpha}{2 \alpha}}/4)^{2\alpha}.
$$
where $$A=A(\alpha)=
2+35\left[\frac{1}{16^{1-\alpha}}+\left(\frac{3(1-\alpha)}{2 \alpha}\right)^{\frac{1-\alpha}{\alpha}}\right]. $$  \qed
\end{lemma}
Lemma \ref{lem:KTF.2} restates Proposition F.2 in \cite{KockTang} with explicit constants in Tail Bounds.

Proof. Note that for some positive constant $D$,
$$
\mathrm{P}\left(\sum_{i=1}^n X_i \geq n \epsilon\right)=\mathrm{P}\left(\sum_{i=1}^n D^{\frac{1-\alpha}{2 \alpha}} X_i \geq n D^{\frac{1-\alpha}{2 \alpha}} \epsilon\right)=\mathrm{P}\left(\sum_{i=1}^n Y_i \geq n \delta\right),
$$
where $Y_i:=D^{\frac{1-\alpha}{2 \alpha}} X_i$ and $\delta:=D^{\frac{1-\alpha}{2 \alpha}} \epsilon$. Now $\left(Y_i\right)_{i=1}^n$ is a sequence of martingale differences satisfying $\sup _i \mathrm{E}\left[e^{\left|Y_i\right|^{\frac{2 \alpha}{1-\alpha}}}\right] \leq C_1$. Invoking the preceding theorem, we have
$$
\mathrm{P}\left(\sum_{i=1}^n Y_i \geq n \delta\right) \leq C(\alpha, n, \delta) e^{-(\delta / 4)^{2 \alpha} n^\alpha} .
$$
$\left(-Y_i\right)_{i=1}^n$ is also a sequence of martingale differences satisfying the same exponential moment condition. Thus,
$$
\mathrm{P}\left(\left|\sum_{i=1}^n X_i\right| \geq n \epsilon\right)=\mathrm{P}\left(\left|\sum_{i=1}^n Y_i\right| \geq n \delta\right) \leq 2 C(\alpha, n, \delta) e^{-(\delta / 4)^{2 \alpha} n^\alpha}
$$
$$
=2 C\left(\alpha, n, D^{\frac{1-\alpha}{2 \alpha}} \epsilon\right) e^{-\left(D^{\frac{1-\alpha}{2 \alpha}} \epsilon / 4\right)^{2 \alpha} n^\alpha} \leq C_1 A(\alpha)  e^{-K \epsilon^{2 \alpha} n^\alpha},
$$
where we can select 
$$A=A(\alpha)=
2+35\left[\frac{1}{16^{1-\alpha}}+\left(\frac{3(1-\alpha)}{2 \alpha}\right)^{\frac{1-\alpha}{\alpha}}\right] $$ and $K$ as defined above. \qed

The following Lemma is inspired by Prop F.3 of \cite{KockTang}. The difference is that the constants are made explicit to make the results applicable to arrays; and part of the proof was replaced by another argument (as we were not able to follow one step in their proof).\footnote{KT's proof uses the inequality $(x-(y\wedge x) )^{2 / L} \leq x^{2 / L}-(y\wedge x)^{2 / L}$, for $x>0$ and $y> 0$. This inequality is not true (for example, with $x=10$, $y=1$, $L=4$, the inequality implies $3< 2.163$.), so we changed the middle part of the proof; the end result is preserved; none of conclusions in KT are affected.} 

\begin{lemma}\label{KT:extend} Suppose we have random variables $Z_{l, i, t, j}$ uniformly $(K,\sigma^2_l)>0$ subgaussian for $l=1, \ldots, L(L \geq 2$ fixed $), i=1, \ldots, N, t=1, \ldots, T$ 
and $j=1, \ldots, p$ that is,
$$
\Pr (|\sigma^{-1}_l Z_{l,t,i,j}| \geq \varepsilon) \leq K \exp(-\varepsilon^2),
$$
and $Z_{l_2, i_2, t_2, j_2}$ are independent as long as $i_1 \neq i_2$ regardless of the values of other subscripts. Then, we have that (1)
$$
\max _{j,t,i}  \mathrm{E}\left| \prod_{l=1}^L  Z_{l, i, t, j}\right| \leq  \left( L !(\log 2)^{-1 / 2} (1+K)^{1 / 2} \right) \prod_{l=1}^L \sigma_l ,
$$
and (2) with probability $1-
A^{\prime}(p T)^{-1/2}$,
$$
\max_{1 \leq j \leq d}\left|\frac{1}{N T} \sum_{i=1}^N \sum_{t=1}^T\left(\prod_{l=1}^L Z_{l, i, t, j}-\mathrm{E} \left[ \prod_{l=1}^L Z_{l, i, t, j}\right]\right)\right|\leq M\left(\sqrt{\frac{(\log (p T))^{L+1}}{N}}\right) \prod_{l=1}^L \sigma_l,
$$
for $M> M'$, and some positive constants $A'$ and $M'$ that only depend on $L$ and $K$.
\end{lemma}

\begin{proof}
Hölder's inequality gives
$$
\max _{j,t,i} \mathrm{E}\left|\prod_{l=1}^L \sigma^{-1}_l Z_{l, i, t, j}\right| \leq \max _{j,t,i} \prod_{l=1}^L\left(\mathrm{E}\left|\sigma^{-1}_l Z_{l, i, t, j}\right|^L\right)^{\frac{1}{L}},
$$
where
\begin{align*}
&\left(\mathrm{E}\left|\sigma^{-1}_l Z_{l, i, t, j}\right|^L\right)^{\frac{1}{L}} \leq L !\left\| \sigma^{-1}_l Z_{l, i, t, j}\right\|_{\psi_1} \\
& \leq L !(\log 2)^{-1 / 2}\left\|\sigma^{-1}_l Z_{l, i, t, j}\right\|_{\psi_2} \leq L !(\log 2)^{-1 / 2}\left(1+K\right)^{1 / 2} =:A,
\end{align*}
where the first two inequalities are from \cite{vdvwellner}, p.95 and the third inequality from Lemma $2.2.1$ in \cite{vdvwellner}. Thus,
$$
\max _{j,t,i}  \mathrm{E}\left|\prod_{l=1}^L \sigma^{-1}_l Z_{l, i, t, j}\right| \leq  \left( L !(\log 2)^{-1 / 2} (1+K)^{1 / 2} \right) =:A.
$$
This implies the first claim, after multiplying both sides by $\prod_{l=1}^L \sigma^{}_l$. Let 
\begin{align*}
X_{i,t,j} =\prod_{l=1}^L \sigma^{-1}_l Z_{l, i, t, j}-\mathrm{E}\left[\prod_{l=1}^L \sigma^{-1}_l Z_{l, i, t, j}
\right].
\end{align*}
For every $\epsilon \geq 0$,
\begin{align*}
\mathrm{P}\left(\left|X_{i,t,j}\right|   \geq 2\epsilon\right) 
& \leq \mathrm{P}\left(  \left| \prod_{l=1}^L \sigma^{-1}_l Z_{l, i, t, j} \right| \geq \epsilon\right)
+ 
\mathrm{P}\left(  \left| \Ep \prod_{l=1}^L \sigma^{-1}_l  Z_{l, i, t, j} \right| \geq \epsilon \right)\\
& \leq \sum_{l=1}^L \mathrm{P}\left( \sigma^{-1}_l \left|Z_{l, i, t, j}\right| \geq \varepsilon^{1/L} \right)
+ 1( \epsilon \leq A)
\\
& \leq L K e^{-\epsilon^{2 / L}} + 1( \epsilon^{2/L} \leq A^{2/L})
\\
&\leq L K e^{-\epsilon^{2 / L}} + e^{A^{2/L}} e^{-\epsilon^{2 / L}}
= K' e^{-\epsilon^{2 / L}}. \quad (K^{\prime}:=(LK + e^{A^{2/L}})).
\end{align*}
Let $$X_{i, j} := \frac{1}{T} \sum_{t=1}^T\left(\prod_{l=1}^L \sigma^{-1}_l Z_{l, i, t, j}-\mathrm{E}\left[\prod_{l=1}^L \sigma^{-1}_l Z_{l, i, t, j}\right]\right).$$ 
For every $\epsilon \geq 0$,
$$
\begin{aligned}
\mathrm{P}\left(\left|X_{i, j}  \right| \geq 2 \epsilon\right) & \leq \mathrm{P}\left(\max _{1 \leq t \leq T}\left|X_{i,t,j}\right| \geq 2\epsilon\right)  \leq T K' e^{-\epsilon^{2 / L}}.
\end{aligned}
$$
Consider some positive constant $D<1$, then as \cite{vdvwellner}, p.96,  using Fubini and change of order of integration:
\begin{align*}
&\mathrm{E}\left[e^{ D \left|X_{i, j}/2\right|^{2 / L}}\right]=\int_{x \in \mathbb{R}} \int_0^{|x/2|^{2 / L}} D e^{D s} d s P(d x)+1=\int_0^{\infty} D e^{D s} \mathrm{P}\left(\left|X_{i, j}\right|>2 s^{L / 2}\right) d s+1.
\end{align*}
This is further bounded by
$$
\int_0^{\infty} T K^{\prime} D e^{(D-1) s} d s+1=\frac{T K^{\prime} D}{1-D}+1 \leq B T; \quad (B:= \frac{K^{\prime} D}{1-D}+1).
$$

Then we can use independence across $i$ to invoke the previous Lemma \ref{lem:KTF.2}  with $\alpha=\frac{1}{L+1}$ and $C_1=B T$, for $\epsilon \geq \frac{1}{\sqrt{N}}$
$$
\mathrm{P}\left(\left|\sum_{i=1}^N \frac{1}{T} \sum_{t=1}^T X_{i,t,j}\right| \geq 2 N \epsilon\right) \leq A^{\prime} T e^{-K''\left(\epsilon^2 N\right)^{\frac{1}{L+1}}}
$$
for positive constants $A^{\prime}$ and $K''$
that depend only on $K$, $L$, and $D$.

Setting $$\epsilon=\sqrt{\frac{M(\log (p T))^{L+1}}{N}}$$ for some $M\geq 1$, we have
$$
\mathrm{P}\left(\max _{1 \leq j \leq p}\left|\sum_{i=1}^N \frac{1}{T} \sum_{t=1}^T X_{i,t,j}\right| \geq 2 N \epsilon\right) \leq p A^{\prime} T e^{-K''\left(\epsilon^2 N\right)^{\frac{1}{L+1}}}=A^{\prime}(p T)^{1-K'' M \frac{1}{L+1}} .
$$
Therefore, with probability $1-
A^{\prime}(p T)^{1-K''M \frac{1}{L+1}}$,
$$
\max _{j}\left|\frac{1}{N T} \sum_{i=1}^N \sum_{t=1}^T\left( X_{i,t,j}\right)\right|\leq 2 M\left(\sqrt{\frac{(\log (p T))^{L+1}}{N}}\right),
$$
for any $M \geq 1$. Setting $M$ large enough such that
$$
1-K''M \frac{1}{L+1}< -\frac{1}{2}, 
$$
guarantees that the bound holds with probability at most
$$
A^{\prime}(p T)^{-1/2},
$$
which decreases to zero if $T \to 0$. The bounds can be then be restated as in the statement of the theorem. \qed
\end{proof}

\newpage

\setcounter{definition}{0}    
\setcounter{section}{0}    
\setcounter{lemma}{0}    
\setcounter{equation}{0}
\setcounter{remark}{0}
\setcounter{theorem}{0}

\renewcommand{\theequation}{C.\arabic{equation}}
\renewcommand{\thelemma}{C.\arabic{lemma}}
\renewcommand{\thecorollary}{C.\arabic{corollary}}
\renewcommand{\thetable}{C.\arabic{table}}

\renewcommand{\theassumption}{C.\arabic{assumption}}
\renewcommand{\thesection}{C}

\section{Tools: High-Dimensional Central Limit Theorems for Weakly Dependent Data}

Let $ \{X_m\}_{m=1}^M$ be a  weakly dependent martingale difference sequence (m.d.s.) with respect to natural filtration. Define its $\beta$-mixing  coefficient
\begin{align*}
\gamma_X(q) =   \sup_{m \leq M} \gamma((X_1, \dots, X_{m-1}, X_m), (X_{m+q},X_{m+q+1},\dots)).
\end{align*}
The scaled sum
$$
S_X= M^{-1/2} \sum_{m=1}^M X_{m}
 $$
has the variance 
\begin{align}
\Sigma_G &:=M^{-1} \sum_{m=1}^M \Ep X_m X_m'. \label{eq:sigmag} 
\end{align}
The distribution of the scaled sum  over the cubes can be approximated by the Gaussian distribution $N(0, \Sigma_{G})$ over the cubes, as shown in the lemma below.

We will introduce the following notation. For some numbers $\bar{r}=\bar{r}_{NT}, \bar{q}=\bar{q}_{NT}$ and $ L = \lfloor M/(\bar{q}+\bar{r}) \rfloor$,  define Bernstein's "large" and "small" blocks of size $\bar q$ and $\bar r$:

\begin{align*}
P_l &= \{ (l-1) (\bar{q}+\bar{r}) +1, \dots, (l-1) (\bar{q}+\bar{r}) +\bar{q}\}, \quad l=1,2, \dots, L \\
 Q_l &= \{ (l-1) (\bar{q}+\bar{r}) +1 +\bar{q}, \dots, l (\bar{q}+\bar{r})\}
\end{align*}
and let
\begin{align*}
S_{l} := \sum_{m \in P_l} X_m, \quad  U_l :=  \sum_{m \in Q_l} X_m, \quad U_{L+1} :=\sum_{m=L(\bar{q}+\bar{r})+1}^{M}  X_m.
\end{align*}
Denote
\begin{align}
\Sigma_P &:= (L\bar{q})^{-1} \sum_{l=1}^L \Ep S_l S_l' = (L\bar{q})^{-1} \sum_{l=1}^L \sum_{m \in P_l} \Ep X_m X_m' \label{eq:sigmap}
\end{align}
and observe that
$$ \Sigma_G =   (L\bar{q}/M) \Sigma_P + M^{-1} \sum_{l=1}^{L+1} \Ep U_l U_l' .$$

The following result is useful both in the proof below and also for performing Gaussian inference, where we replaced unknown variance-covariance matrix by an estimated one.

\begin{lemma}[Comparison of distributions]
\label{lem:comp}
Let $ X \sim N(0, \Sigma_X)$ and $ Y \sim N(0, \Sigma_Y)$ be centered normal $d$-vectors, and let  $\Delta_{XY}:= \| \Sigma_X - \Sigma_Y \|_{\infty}$.  Suppose  $\min_{1 \leq j \leq d } (\Sigma_Y)_{jj}>0$. Then,
\begin{align}
\label{eq:prtf}
\sup_{t \geq 0} | \Pr (   \| X \|_{\infty} \leq t ) - \Pr (    \| Y \|_{\infty}  \leq t ) | \leq  C' (\Delta_{XY} \log^2 (2d))^{1/2},
\end{align}
where $C'>0$ depends only on $\min_{1 \leq j \leq d } (\Sigma_Y)_{jj}$ and $\max_{1 \leq j \leq d } (\Sigma_Y)_{jj}$.
\end{lemma}
Lemma \ref{lem:comp} follows from  Proposition 2.1 in \cite{CCKY} for vectors $\overline{X} = (X, - X)$ and $\overline{Y}  = (Y, -Y)$ and  $$\Sigma_{\overline{X}} =  \begin{pmatrix} \Sigma_X & - \Sigma_X \\
- \Sigma_X & \Sigma_X \end{pmatrix}, \quad \Sigma_{\overline{Y}} =  \begin{pmatrix} \Sigma_Y & - \Sigma_Y \\
- \Sigma_Y & \Sigma_Y \end{pmatrix}, \quad \| \Sigma_{\overline{X}} - \Sigma_{\overline{Y}} \|_{\infty} = \Delta_{XY}. $$
\\

Another result is the following anti-concentration property. This result is useful for showing that linearization errors do not impact the behavior of the key statistics. The statistics are approximate means, namely  averages of some centered influence functions plus linearization errors. 

\begin{lemma}[Anti-concentration]
\label{lem:anti}
Let $X=(X_1,X_2, \dots, X_d)' \sim N(0, \Sigma_X)$ be a centered Gaussian random vector in $\mathrm{R}^d$. Assume $\min_{1 \leq j \leq d } (\Sigma_X)_{jj}>0$.  Then, 
\begin{align}
\label{eq:anticont}
\sup_{t \in \mathrm{R} } \Pr (| \|  X \|_{\infty} - t |\leq \epsilon) \leq C \epsilon \sqrt{1 \vee \log (2d/\epsilon)},
\end{align}
where $C>0$ depends on $\min_{1 \leq j \leq d } (\Sigma_X)_{jj}$ and $\max_{1 \leq j \leq d } (\Sigma_X)_{jj}$.

\end{lemma}

Lemma \ref{lem:anti} follows from  Corollary 1 in \cite{PRTF} with $\overline{X} = (X, -X)$.

The following result is a consequence of Theorem E.1 in \cite{CCK} for martingale difference sequence. 

\begin{lemma}[High-dimensional CLT for martingale difference sequence under weak dependence]
\label{lem:cck}
Let $ \{X_m\}_{m=1}^M$ be a  weakly dependent m.d.s. of $d$-vectors  obeying for $D_M \geq 1$:
\begin{align*}
\sup_{ m \leq M} \| X_m \|_{\infty} \leq D_M \text{ a.s.}
\end{align*}
Suppose there exist constants $0<a_1 \leq A_1$ and $0<c_2 <1/4$ such that
\begin{align}
\label{eq:varcond}
a_1 \leq  \min_{1 \leq j \leq d} \min_{1 \leq m \leq M}  \text{Var} X_{mj}  \leq  \max_{1 \leq j \leq d} \sup_{1 \leq m \leq M}    \text{Var} X_{mj} \leq A_1,
\end{align}
and let $\bar{r}$ and $\bar{q}$ be such that $\bar{r}/\bar{q} \leq A_1  M^{-c_2} \log^{-2} d$ and
\begin{align}
\label{eq:theoreme1}
\max \{ \bar{r} D_M \log^{3/2} d ,  \bar{q} D_M \log^{1/2} d, \sqrt{\bar{q}} D_M \log^{7/2} (dM) \} \leq A_1 M^{1/2-c_2}.
\end{align} 
Then, there exist constants $c_X, C_X>0$ depending only on $a_1, A_1, c_2$ such that
\begin{align}
\label{eq:fbound}
\sup_{t \geq 0} \bigg| \Pr ( \|  S_X \|_{\infty}  \leq t ) -   \Pr ( \| G_P \|_{\infty} < t)  \bigg| \leq 2  \frac{M}{\bar{q}+\bar{r}} \gamma_X(\bar{r})  + C_X M^{-c_X},
\end{align}
where $G_P  \sim N(0, \Sigma_{P})$ is a centered normal $d$-vector. 
\end{lemma}

Note that this result uses $\Sigma_{P}$ as the variance in the Gaussian approximation. In our application, we will be using $\Sigma_G$ in place of $\Sigma_P$ (i.e., Lemma \ref{lem:cck2}) so as not to worry about omitting small blocks. Therefore below, we will provide a sequence of the results that allow this replacement.

\begin{proof}[Proof of Lemma \ref{lem:cck}]
Let $$\overline{X}_m:= (X_m, -X_m), \quad m=1,2,\dots, M$$ be a sequence of $2d$-vectors. Observe that $\{ \overline{X}_m \}^M_{ m =1 }$ is an m.d.s. It obeys 
\begin{align*}
\sup_{ m \leq M} \| \overline{X}_m \|_{\infty} \leq D_M, \text{ a.s.}, \quad \gamma_{\bar{X}}(q) = \gamma_X(q) \quad \forall q. 
\end{align*}
By construction, for any integer $r$
\begin{align*}
\bar{\sigma}^2(r) = \max_{1 \leq j \leq d} \max_{I}  \text{Var} (r^{-1/2} \sum_{m \in I} X_{mj}) = \max_{1 \leq j \leq 2d} \max_{I}  \text{Var} (r^{-1/2} \sum_{m \in I} \overline{X}_{mj}), \\
\underline{\sigma}^2(r) = \min_{1 \leq j \leq d} \min_{I}  \text{Var} (r^{-1/2} \sum_{m \in I} X_{mj}) = \min_{1 \leq j \leq 2d} \min_{I}  \text{Var} (r^{-1/2} \sum_{m \in I} \overline{X}_{mj}),
\end{align*}
where $\max_{I} { }$ and $\min_{I} { }$ are taken over the sets $I = \{i+1, i+2, \dots, i+r\}$ of size $r$. Theorem E.1 in \cite{CCK} requires
\begin{align}
\label{eq:c1c1}
a_1 \leq \underline{\sigma}^2(\bar{q})  \leq  \bar{\sigma}^2(\bar{q}) \vee  \bar{\sigma}^2(\bar{r})  \leq A_1.
\end{align}
Because $ \{X_m\}_{m=1}^M$ is an m.d.s, $$\text{Cov} (X_{m_1}, X_{m_2}) =0 \in \mathrm{R}^{d \times d} \text{ for } m_1 \neq m_2.$$ Therefore,  for any $r$ and any $I = \{i+1, i+2, \dots, i+r\}$,
\begin{align*}
a_1 \leq \text{Var} (r^{-1/2} \sum_{m \in I} X_{mj} ) = r^{-1} \sum_{m \in I} \text{Var} (X_{mj} ) \leq A_1, \quad 1 \leq j \leq d,
\end{align*}
which implies \eqref{eq:c1c1}. All other conditions of Theorem E.1 in \cite{CCK} are satisfied. Invoking Theorem E.1 in \cite{CCK} with
$$
T := \max_{1 \leq j \leq 2d}   M^{-1/2} \sum_{m=1}^M \overline{X}_{mj} = \| S_X \|_{\infty}
$$
and 
$$
\overline{G}_P \sim N(0, \Sigma_{G_P})
$$
being a centered normal $(2d)$-vector with 
$$
\Sigma_{G_P}= \begin{pmatrix} \Sigma_P & - \Sigma_P \\
 - \Sigma_P & \Sigma_P \\
 \end{pmatrix}.
$$
gives \eqref{eq:fbound}.

\end{proof}

\begin{lemma}[Comparison  of distributions, cont.]
\label{lemma:comparison}
Consider the setup above with $\Sigma_X=\Sigma_G$ and $\Sigma_Y=\Sigma_P$, where $\Sigma_G$ and $\Sigma_P$ are as in \eqref{eq:sigmap} and \eqref{eq:sigmag} where $$\sup_{1 \leq m \leq M} \| \Ep X_m X_m'\|_{\infty}  \leq \sup_{1 \leq m \leq M} \sup_{1 \leq j \leq d} \text{Var} (X_{mj}) \leq A_1.$$
For some $c_2 \in (0, 1/4)$, assume that the growth condition holds:
$$
D_M \log d  \log M   \log^{7/2} (dM) \lesssim M^{1/2-2c_2}.
$$
and $\log^4 d \log^2 M = o(\sqrt{M})$.  Then, the max distance
$
\Delta_{GP} := \|\Sigma_G - \Sigma_P\|_\infty 
$
obeys
$$(\Delta_{GP} \log^2 d)^{1/2}  \lesssim M^{-c_2/2}.
$$
\end{lemma}

\begin{proof}[Proof of Lemma \ref{lemma:comparison}]
Observe that
\begin{align*}
\Sigma_G - \Sigma_P = (L\bar{q}/M-1) \Sigma_P + M^{-1} \sum_{l=1}^{L+1} \Ep U_l U_l'.
\end{align*} Since $L= \lfloor M/(\bar{q}+\bar{r}) \rfloor, L \geq M/(\bar{q}+\bar{r})-1$. Therefore,
\begin{align*}
1-L\bar{q}/M &\leq 1- \bar{q}/(\bar{q}+\bar{r}) + \bar{q}/M = \bar{r}/(\bar{q}+\bar{r}) + \bar{q}/M \leq \bar{r}/\bar{q} + \bar{q}/M.
\end{align*}
Furthermore, $(L+1)/M \leq 2L/M \leq 2/\bar{q}$.  The following bound holds
\begin{align*}
\Delta_{GP} &\leq  ((1-L\bar{q}/M) + (L+1)/M) \sup_{1 \leq m \leq M} \| \Ep X_m X_m' \|_{\infty} = O (  \bar{r}/\bar{q} \vee \bar{q}/M \vee 1/\bar{q}),
\end{align*}
Taking  $\bar{q} =M^{c_2} \log^2 d  \log^2 M$  and $\bar r = (2/\kappa) \log M$    gives:
\begin{align*}
 \bar{r}/\bar{q} &=  (2/\kappa) M^{-c_2} \log^{-2} d \log^{-1} M = o (M^{-c_2} \log^{-2} d) \\
    \bar{q}/M &= M^{c_2-1} \log^2 d  \log^2 M  =^{i} o (M^{-c_2} \log^{-2} d) \\
    1/\bar{q} &= M^{-c_2} \log^{-2} d \log^{-2} M = o (M^{-c_2} \log^{-2} d),
\end{align*}
where (i) follows from $c_2<1/4$ and
$$
\log^4 d \log^2 M = o (M^{1-2 \cdot 1/4})= o(M^{1-2c_2})$$
Plugging $\Delta_{GP} = o (M^{-c_2} \log^{-2} d)$ into $(\Delta_{GP} \log^2 (2d))^{1/2}$ gives 
\begin{align*}
(\Delta_{GP}  \log^2 (2d))^{1/2} = o (M^{-c_2/2}).
\end{align*}
\end{proof}

\begin{remark}[Sufficient Growth Condition]
\label{rm:cck}
If the growth condition holds 
\begin{align}
    \label{eq:maincond} 
    D_M \log d  \log M   \log^{7/2} (dM)\lesssim  M^{1/2-2c_2}
\end{align}
holds, then 
\begin{align}
\label{eq:qr}
\bar{r} = \log M, \quad \bar q=M^{c_2} \log^2 d  \log^2 M
\end{align}
obeys \eqref{eq:theoreme1} and $\bar{r}/\bar{q} \leq A_1 M^{-c_2} \log^{-2} d$ for $M$ large enough.
\end{remark}

\begin{proof}[Proof of Remark \ref{rm:cck}]
Let $M$ be large enough such that $M^{-c_2/2} \leq A_1$ and $(2/\kappa) \log^{-1} M \leq A_1$. Then,  the growth condition 
$$
D_M \log d  \log M   \log^{7/2} (dM) \lesssim M^{1/2-2c_2} \leq A_1 {M^{1/2-3/2c_2}}
$$
implies the third inequality in \eqref{eq:theoreme1}
$$\sqrt{\bar{q}} D_M \log^{7/2} (dM) \leq A_1 {M^{1/2-c_2}}.$$
Next, for $d \geq e$ such that $\log d \geq 1$, and 
\begin{align*}
M^{-c_2} D_M  \bar{q} \log^{1/2} d &= D_M  \log^{5/2} d \log^2 M \\
&\leq D_M \log^{5/2} (dM) \log M \log (dM) \log d \leq A_1 M^{1/2-3/2c_2}.
\end{align*}
Multiplying both sides by $M^{c_2}$ gives
$$
D_M  \bar{q} \log^{1/2} d \leq A_1 M^{1/2-c_2},
$$
which coincides with the second inequality in \eqref{eq:theoreme1}. Finally,
$$
D_M \bar{r} \log^{3/2} d = (2/\kappa) D_M \log M  \log^{3/2} d  \leq D_M \bar{q} \log^{1/2} d,
$$
as long as $(2/\kappa) \leq \log M$, which verifies \eqref{eq:theoreme1}. For $M$ large enough $\bar{r}/\bar{q} = 2/ \kappa M^{-c_2} \log^{-2} d  \log^{-1} M \leq A_1 $.

\end{proof}

\begin{lemma}[Summary]
\label{lem:cck2}
Let $ \{X_m\}_{m=1}^M$ be a  weakly dependent m.d.s. of $d$-vectors  obeying for $D_M \geq 1$:
\begin{align*}
\sup_{ m \leq M} \| X_m \|_{\infty} \leq D_M \text{ a.s.}
\end{align*}
Suppose there exist constants $0<a_1 \leq A_1$ such that
\begin{align*}
a_1 \leq  \min_{1 \leq j \leq d} \min_{1 \leq m \leq M}  \text{Var} X_{mj}  \leq  \max_{1 \leq j \leq d} \sup_{1 \leq m \leq M}    \text{Var} X_{mj} \leq A_1.
\end{align*}
For some constant $c_2 \in (0,1/4)$, the growth condition  \eqref{eq:varcond} holds, namely
$$
D_M \log d  \log M   \log^{7/2} (dM)  \lesssim M^{1/2-2c_2}.
$$
and $\log^4 d \log^2 M = o (M^{1/2})$.  Then, there exist constants $c_X, C_X>0$ depending only on $a_1, A_1, c_2$ such that for 
$\bar{r} = (2/ \kappa \log M)$
and $\bar q=M^{c_2} \log^2 d  \log^2 M$
\begin{align}
\label{eq:fboundsummary}
&\sup_{t \geq 0} | \Pr ( \|  S_X \|_{\infty}  \leq t ) -   \Pr ( \| G_\Sigma \|_{\infty} < t)  | \lesssim C_X M^{-c_X} + M^{-c_2/2}, 
\end{align}
where $G_\Sigma \sim N(0, \Sigma_{G})$ is a centered normal $d$-vector.
\end{lemma}
Triangular inequality gives
\begin{align}
    &\sup_{t \geq 0} | \Pr ( \|  S_X \|_{\infty}  \leq t ) -   \Pr ( \| G_\Sigma \|_{\infty} < t)  | \\
&\leq \sup_{t \geq 0} | \Pr ( \|  S_X \|_{\infty}  \leq t ) -   \Pr ( \| G_P \|_{\infty} < t)  | + 
\sup_{t \geq 0} |\Pr ( \| G_P \|_{\infty} < t) - \Pr ( \| G_\Sigma \|_{\infty} < t) | \nonumber  \\
&\lesssim 2  \frac{M}{\bar{q}+\bar{r}} \gamma(\bar{r}) + C_X M^{-c_X} + M^{-c_2/2} = o(M^{-c_2/2} +M^{-c_X} ).
\end{align}

\setcounter{definition}{0}    
\setcounter{section}{0}    
\setcounter{lemma}{0}    
\setcounter{equation}{0}
\setcounter{remark}{0}
\setcounter{theorem}{0}

\renewcommand{\theequation}{D.\arabic{equation}}
\renewcommand{\thelemma}{D.\arabic{lemma}}
\renewcommand{\thecorollary}{D.\arabic{corollary}}
\renewcommand{\thetable}{D.\arabic{table}}

\renewcommand{\theassumption}{D.\arabic{assumption}}
\renewcommand{\thesection}{D}

\section{Proofs for Section \ref{sec:theory}}
\label{sec:app:proofs}
 
\subsection{Bounds on Errors for Estimating $Q$ and  Gradient $S$.}\label{sec:boundsOnamfe}
Below, we define the following terms that appear in the analysis of $\widehat{Q}$ and the least squares gradient $S$.  In what follows we use the notations defined in the main text heavily, without further warning.

%VS: so decomposition of \widehat{Q}-Q has moved to the proof of Lemma A.16 

Define the first-stage approximation error as a function of $d(\cdot)$ and $l(\cdot)$:
\begin{align}
\label{eq:ritalt}
R_{it}(\mathbf{d}, \mathbf{l}):= l_{i0}(X_{it}) -l_i(X_{it}) - (d_{i0}(X_{it}) - d_i(X_{it}))'\beta_0.
\end{align}
Define the first-order error terms
\begin{align}
  \bar{a} &:= \ENT V_{it} (d_{i0} (X_{it}) - \widehat{d}_i (X_{it})) = \ENT V_{it}  (\widehat{V}_{it} - V_{it})  \label{eq:ak} \\
\bar{m} &= \ENT  V_{it}  (l_{i0}(X_{it}) - \widehat{l}_i(X_{it})) = \ENT V_{it}  (\widehat{\widetilde{Y}}_{it} - \widetilde{Y}_{it}) \label{eq:mk} \\
\bar{f} &= \ENT  U_{it} (d_{i0}(X_{it}) - \widehat{d}_i(X_{it})) = \ENT  U_{it} ( \widehat{V}_{it} - V_{it}) , \label{eq:fk}\\
\bar{e} &= \ENT V_{it} R_{it}(\widehat{\mathbf{d}}, \widehat{\mathbf{l}}) =  \bar{m}-\bar{a}' \beta_0. \label{eq:ek} 
\end{align}
the second-order error terms 
\begin{align}
\bar{b} &= \ENT  (d_{i0}(X_{it}) - \widehat{d}_i(X_{it})) (d_{i0}(X_{it}) - \widehat{d}_i(X_{it}))'  \label{eq:bk} \\
\bar{z} &= \ENT  (d_{i0}(X_{it}) - \widehat{d}_i(X_{it})) (l_{i0}(X_{it}) - \widehat{l}_i(X_{it}))\label{eq:zk} \\
\bar{g} &= \ENT  (d_{i0}(X_{it}) - \widehat{d}_i(X_{it})) R_{it}(\widehat{\mathbf{d}}, \widehat{\mathbf{l}}) =\bar{z}- \bar{b}'\beta_0.   \label{eq:gk} 
\end{align}

\begin{lemma}[First-Order Terms]
\label{lem:firstchunk}
Under Assumptions \ref{ass:sampling}--\ref{ass:smallbiashds}, we have that
\begin{align}
\| \bar{a} \|_{\infty}&\lesssim_P  \left(  \textbf{d}_{NT,\infty} \sqrt{\log (dNT)  /NT}\right)  \label{eq:akproofinf}\\
\| \bar{m} \|_{\infty}&\lesssim_P  \left(  \textbf{l}_{NT,\infty} \sqrt{\log (dNT)  /NT}\right)  \label{eq:mkproofinf}\\
\| \bar{f} \|_{\infty}&\lesssim_P (  \textbf{d}_{NT,\infty} \sqrt{\log (dNT) /NT}  ) \label{eq:fkproofinf} \\
\| \bar{e} \|_{\infty}&\lesssim_P  ( \sqrt{\log (dNT)/NT} (\textbf{d}_{NT,\infty} \| \beta_0 \|_1 +  \textbf{l}_{NT,\infty} ) ). \label{eq:ekproofinf}
\end{align}

\end{lemma}
\begin{proof}[Proof of Lemma \ref{lem:firstchunk}]
Define $$\zeta^V_{NT} :=  \textbf{d}_{NT, \infty} \sqrt{ \log (d NT)/NT}, \quad \zeta^B_{NT} =0,$$ 
and the $A$-function as\begin{align*}
    A (W_{it},\eta) =V_{it} (d_{i0}(X_{it}) - d_i(X_{it})).
\end{align*}
Define $B_{Ak}(\eta)$ and $V_{Ak}(\eta)$ with $\eta = \mathbf{d}$ as in \eqref{eq:genericbias}--\eqref{eq:genericdem}. 

Consider any $\eta=\eta_{NT} \in D_{NT}$ in what follows. Since $V_{it}$ obeys the martingale difference property by assumption, we have that
\begin{align}
\label{eq:mdsvit}
 \Ep [ V_{it} \mid \cup_{t' \leq t, t' \in \mathcal{M}_{k}} (V_{it'}, X_{it'}) ] =0,
\end{align}
and it follows that $\| B_{Ak} (\eta_{NT}) \|_{\infty}=0$. By  Assumption \ref{ass:subgauss} and Lemma \ref{lem:subgauss}, each entry of $V_{it} (d_{i0}(X_{it}) - d_i(X_{it}))$ is $\bar{\sigma}^2 \textbf{d}^2_{NT, \infty}$-sub-Gaussian. 
Invoking Lemma \ref{lem:mds1} gives  $$\| V_{Ak} (\eta_{NT}) \|_{\infty} \lesssim_P (\bar{\sigma} \textbf{d}_{NT, \infty} \sqrt{\log d /N T_k}) =o_P (\zeta^V_{NT})$$ since $T_k \asymp T$ (as we keep number of blocks $K$ fixed).  By Assumption \ref{ass:smallbiashds}, we have that
$\Pr (\widehat{\mathbf{d}}_k \in D_{NT}, \ \forall k=1,...,K) \to 1$. We conclude by Lemma \ref{cor:remterms} that  \eqref{eq:akproofinf} holds.   Repeating the same argument for 
$$A(W_{it}, \eta) = V_{it} (l_{i0}(X_{it}) - l_i(X_{it}))
\text { and } A(W_{it}, \eta) = U_{it} (d_{i0}(X_{it}) - d_i(X_{it}))$$ 
establishes claims \eqref{eq:mkproofinf} and \eqref{eq:fkproofinf}.  Finally, \eqref{eq:ekproofinf} holds by definition of $\bar{e}=\bar{m}-\bar{a}'\beta_0$ and Holder inequalities.\end{proof}

\begin{lemma}[Second-Order Term]
\label{lem:demeaned}
Under Assumptions \ref{ass:sampling}--\ref{ass:smallbiashds}, 
we have that  
\begin{align}
\| \bar{z} \|_{\infty}&\lesssim_P \left( \textbf{d}_{NT}\textbf{l}_{NT}  +    \textbf{d}_{NT,\infty} \textbf{l}_{NT,\infty} \sqrt{ (NT)^{-1}  \log (NT) \log d  } \right) \label{eq:zkproofinf}  \\
\| \bar{b} \|_{\infty}&\lesssim_P \left( \textbf{d}_{NT}^2  +   \textbf{d}_{NT,\infty}^2 \sqrt{ (NT)^{-1} \log (NT) \log d  } \right) \label{eq:bkproofinf} \\
\| \bar{g}\|_{\infty} &\lesssim_P
\bigg (\| \beta_0 \|_1  \textbf{d}_{NT}^2 + \textbf{d}_{NT}\textbf{l}_{NT}
+ ( \| \beta_0 \|_1 
\textbf{d}_{NT,\infty}^2 +
\textbf{d}_{NT,\infty} \textbf{l}_{NT,\infty}) \sqrt{ (NT)^{-1}  \log (NT) \log d  }  \bigg)\label{eq:gkproofinf}
\end{align}
\end{lemma}
%\textsc{[VC: the previous bound on $g$ was stated incorrectly. Please check if there are any consequences for the results that make use of this bound in the appendix and in the main text.]}

\begin{proof}[Proof of Lemma \ref{lem:demeaned}]
Define the $A$-function as
$$A (W_{it},\eta) = (d_{i0}(X_{it}) - d_i(X_{it}))(l_{i0}(X_{it}) - l_i(X_{it})), \quad \eta =(\textbf{d}, \textbf{l}).$$ 
Let  $B_{Ak}(\eta)$ and $V_{Ak}(\eta)$ be defined according to \eqref{eq:genericbias}--\eqref{eq:genericdem}. Let $$\zeta^B_{NT}=\mathbf{l} _{NT} \textbf{d}_{NT}, \quad  \zeta^V_{NT}=\sqrt{\mathbf{l}^2_{NT, \infty} \textbf{d}^2_{NT,\infty} \log d \log NT/NT }.$$  For any $i$ and $t$, $m$ and $j$, the Cauchy-Schwarz inequality gives 
\begin{align*}
  & \Ep [|(d_{i0}(X_{it}) - d_i(X_{it}))_m  (l_{i0}(X_{it}) - l_i(X_{it})) |] \\
    & \quad \leq \sqrt{\Ep (d_{i0}(X_{it}) - d_i(X_{it}))_m^2  \Ep (l_{i0}(X_{it}) - l_i(X_{it}))^2}  =: \sqrt{ a^2_{it} b^2_{it}} = |a_{it}||b_{it}|
\end{align*}
Another application of the Cauchy-Schwarz gives
\begin{align}
     (T_kN)^{-1} \sum_{i} \sum_{t \in \mathcal{M}_k} |a_{it}||b_{it}|  &\leq \sqrt{ (T_kN)^{-1} \sum_{i=1}^N \sum_{t \in \mathcal{M}_k} a^2_{it} } \sqrt{ (T_kN)^{-1} \sum_{i=1}^N \sum_{t \in \mathcal{M}_k}   b^2_{it}  } \nonumber \\
   &\leq \sqrt{ (T_kN)^{-1} \sum_{i=1}^N \sum_{t=1}^T  a^2_{it} } \sqrt{ (T_kN)^{-1} \sum_{i=1}^N \sum_{t=1}^T   b^2_{it}  } \nonumber   \leq  \textbf{d}_{NT} \mathbf{l} _{NT} T/T_k. \nonumber
\end{align}
Therefore $\| B_{Ak} (\eta_{NT}) \|_{\infty} = O(
\zeta^B_{NT})$. Furthermore, each entry of $A (W_{it},\eta)$
is bounded by $\textbf{d}_{NT, \infty} \mathbf{l} _{NT, \infty}$, and, therefore, is $\textbf{d}^2_{NT, \infty} \mathbf{l}^2_{NT, \infty}$-sub-Gaussian.  By Lemma \ref{lem:matrixerrorinf}, $$\| V_{Ak}(\eta_{NT}) \|_{\infty} \lesssim_P (\zeta^V_{NT}),$$
since $T_k \asymp T$.  Furthermore, by Assumption \ref{ass:smallbiashds}
$\Pr ( (\widehat{\mathbf{d}}_k, \widehat{\mathbf{l}}_k)\in D_{NT} \times L_{NT}, \forall k=1,...,K) \to 1.
$
We conclude by Lemma \ref{cor:remterms} that  
\eqref{eq:zkproofinf} holds.  The bound \eqref{eq:bkproofinf} follows from the same argument. 
Finally, the bound \eqref{eq:rnterror} follows from the definition $\bar{g} =   \bar{z}-\bar{b}'\beta_0$
and Holder inequality and union bounds. We obtain
\begin{align*}
\| \bar{g}\|_{\infty} &\lesssim_P
\bigg (   \| \beta_0 \|_1  \left( \textbf{d}_{NT}^2 +   \textbf{d}_{NT,\infty}^2 \sqrt{ (NT)^{-1} \log (NT) \log d  } \right)  \nonumber \\ 
& \quad + \textbf{d}_{NT}\textbf{l}_{NT}  +   \textbf{d}_{NT,\infty} \textbf{l}_{NT,\infty} \sqrt{ (NT)^{-1}  \log (NT) \log d  } 
\bigg) 
\end{align*}
Then we rewrite the bound as in \eqref{eq:rnterror}.
\end{proof}
Define 
\begin{align*}
  \widehat{Q}&= \mathbb{E}_{NT} \widehat{V}_{it} \widehat{V}_{it}', \quad \widetilde{Q}= \mathbb{E}_{NT} {V}_{it} {V}_{it}', \quad \widehat S := \mathbb{E}_{NT} \widehat{V}_{it} (\widehat {\widetilde{Y}}_{it}-\widehat{V}_{it}'\beta_0), \quad  S:= \mathbb{E}_{NT} V_{it} U_{it}
\end{align*}
and the following rates
\begin{align}
\kappa_{N T} & :=  \sqrt{\log ^3\left(d^2 \log (N T)\right) \log (N T) / N T} \\
q_{NT} &:= \textbf{d}_{NT,\infty}  \sqrt {\log (dNT) /NT}  +\textbf{d}_{NT}^2 + 
\textbf{d}_{NT,\infty}^2 \sqrt {\log (NT) \log (d) /NT} \label{eq:qnt}   
\end{align}
We will also use the following rates defined in the Section \ref{sec:theory} of main text 
\begin{align*}
\rho_{NT} & :=  \textbf{d}_{NT,\infty} \sqrt{\log (dNT) /NT} +  \sqrt{\log (dNT)/NT} (\textbf{d}_{NT,\infty} \| \beta_0 \|_1 + \textbf{l}_{NT,\infty} ) 
+ r_{NT} \\
r_{NT} & :=  \| \beta_0 \|_1  \textbf{d}_{NT}^2 + \textbf{d}_{NT}\textbf{l}_{NT}
+ ( \| \beta_0 \|_1 
\textbf{d}_{NT,\infty}^2 +
\textbf{l}_{NT,\infty}) \sqrt{ (NT)^{-1}  \log (NT) \log d  }
\end{align*}

\begin{lemma}[Summary of Gram Matrix and Gradient Error Bounds]
\label{lem:matrixerrorinf2}
 Suppose Assumptions \ref{ass:sampling}--\ref{ass:smallbiashds} hold. Then, the following bounds hold wp $1-o(1)$
\begin{align}
%& \| \widetilde{Q} - Q \|_{\infty} \lessim_P  C \kappa_{NT} = o(1) \label{eq:covmat} \\
& \| \widetilde{Q} - Q \|_{\infty}  \lesssim_P o(\kappa_{NT} \log (d^2NT)). \label{eq:covmat} \\
& \| \widetilde{Q} - \widehat{Q} \|_{\infty} \lesssim_P (q_{NT}) = o_P( (NT)^{-1/2}) \label{eq:tildeqinf} \\
 &\| \widehat{Q} - Q \|_{\infty} \lesssim_P o (\kappa_{NT} \log (d^2NT)) \label{eq:covmat22}\\
& \|\widehat S - S\|_{\infty} \lesssim_P (\rho_{NT}) = o_P( (NT)^{-1/2}),\label{eq:approxboundinf}
\end{align}
%for some large enough constant $C$. 
\end{lemma}

\begin{proof}[Proof of Lemma \ref{lem:matrixerrorinf2}]
 Decompose matrix first-stage estimation error gives
\begin{align*}
    \widehat{Q}&= \mathbb{E}_{NT} \widehat{V}_{it} \widehat{V}_{it}' \\
    &= \mathbb{E}_{NT} (V_{it} + (d_{i0}(X_{it}) - \widehat{d}_i(X_{it})))(V_{it} + (d_{i0}(X_{it}) - \widehat{d}_i(X_{it})))' \\
    &= \mathbb{E}_{NT} V_{it} V_{it}' \\
    &+ \mathbb{E}_{NT} V_{it} (d_{i0}(X_{it}) - \widehat{d}_i(X_{it}))' + (\mathbb{E}_{NT} V_{it} (d_{i0}(X_{it}) - \widehat{d}_i(X_{it}))' )'\\
    &+ \mathbb{E}_{NT} (d_{i0}(X_{it}) - \widehat{d}_i(X_{it})) (d_{i0}(X_{it}) - \widehat{d}_i(X_{it}))'= \widetilde{Q} + \bar{a} + \bar{a}' + \bar{b}.
\end{align*}
Then, an application of Lemma
\ref{lem:covmat} with $\bar N=2$ gives wp $1-o(1)$
$
\|\widetilde{Q}-Q\|_{\infty}\leq \bar C_{\kappa} \kappa_{NT}
$ for large enough $\bar C_{\kappa} $. The bounds on $\| \bar{a} \|_{\infty}$ and  $\| \bar{b} \|_{\infty}$ are given in \eqref{eq:akproofinf} and \eqref{eq:bkproofinf}, respectively. Collecting terms gives the bound  \eqref{eq:tildeqinf}. The \eqref{eq:covmat22}  follows from the triangle inequality and  $q_{NT} = o_P(\kappa_{NT}).$
We can decompose the gradient error $\widehat S -S$ as follows.
Note that
\begin{align*}
\widehat{\widetilde{Y}}_{it}-\widetilde{Y}_{it} &= Y_{it} - \widehat{l}_i (X_{it}) - ( Y_{it} - l_{i0} (X_{it})) = l_{i0} (X_{it}) - \widehat{l}_i (X_{it}) \\
    \widehat{V}_{it} - V_{it} &= D_{it} - \widehat{d}_i (X_{it}) - (D_{it} - d_{i0} (X_{it})) = d_{i0} (X_{it}) - \widehat{d}_i (X_{it}) 
\end{align*}
The difference of the two equations is
\begin{align*}
\widehat{\widetilde{Y}}_{it}-\widetilde{Y}_{it} - (\widehat{V}_{it} - V_{it})' \beta_0 = R_{it}(\widehat{\mathbf{d}}, \widehat{\mathbf{l}}).
\end{align*}
Therefore, 
\begin{align}
\label{eq:uitrit}
\widehat{\widetilde{Y}}_{it} - \widehat{V}_{it}' \beta_0 &= (\widetilde{Y}_{it} - V_{it}' \beta_0) + ((\widehat{\widetilde{Y}}_{it}-\widetilde{Y}_{it}) - (\widehat{V}_{it} - V_{it})' \beta_0) =U_{it} + R_{it}(\widehat{\mathbf{d}}, \widehat{\mathbf{l}}).
\end{align}
Decompose the gradient:
\begin{align*}
    \widehat{S} &= \mathbb{E}_{NT} \widehat{V}_{it} (\widehat {\widetilde{Y}}_{it}-\widehat{V}_{it}'\beta_0) = \mathbb{E}_{NT} V_{it} (\widehat {\widetilde{Y}}_{it}-\widehat{V}_{it}'\beta_0) +  \mathbb{E}_{NT} (\widehat V_{it} - V_{it}) (\widehat {\widetilde{Y}}_{it}-\widehat{V}_{it}'\beta_0) = \widehat{S}_1 + \widehat{S}_2,
    \end{align*}
    where
\begin{align*}
 \widehat{S}_1 &=\mathbb{E}_{NT} V_{it} U_{it} + \mathbb{E}_{NT} V_{it} R_{it}(\widehat{\mathbf{d}}, \widehat{\mathbf{l}}) = S + \bar{e}, \\
    \widehat{S}_2 & = \mathbb{E}_{NT} (d_{i0} (X_{it} - \widehat{d}_i (X_{it})) U_{it} + \mathbb{E}_{NT}   (d_{i0} (X_{it}) - \widehat{d}_i (X_{it})) R_{it}(\widehat{\mathbf{d}}, \widehat{\mathbf{l}}) 
    = \bar{f} + \bar{g}. 
\end{align*}
Invoking bounds on $\bar e$, $\bar f$, and $\bar g$ in \eqref{eq:ekproofinf}--\eqref{eq:gkproofinf} gives the result.

\end{proof}

\newpage

 \subsection{Proof of Orthogonal Lasso Rate: Theorem \ref{thrm:ortholasso}}

 \paragraph{Group Sparsity Notation.} We use the same notation as \cite{GeerGroup}. Consider a generic covariate vector of size $g \cdot d$, where $d$ is the number of groups and $g$ is the group size. Partition the set of indices $\{1,2,\dots, gd\}$ into $d$ groups of size $g$:
\begin{align*}
J_j:=   \{j, d+j, \dots, (g-1) d +j\}, \quad j=1,2, \dots, d, \quad | J_j | = g.
\end{align*}
For a group index $j$ and a subset of group indices $\mathcal{T}$, and vector $\Delta \in \mathrm{R}^{gd}$, denote
\begin{align*}
 \Delta^j = (\Delta_{m})_{m \in J_j} \in \mathrm{R}^g, \quad  \Delta^{\mathcal{T}}= (\Delta_{m})_{\{m \in J_j,  j \in \mathcal{T}\}} \in \mathrm{R}^{|\mathcal{T}| \cdot g}.
\end{align*}
For any $\Delta \in \mathrm{R}^{gd}$, define the group-vector norms
\begin{align*}
\| \Delta  \|_{2, \infty} &= \max_{1 \leq j \leq d} \| \Delta^j \|_2,  \quad \| \Delta  \|_{2, 1}  = \sum_{j=1}^d \| \Delta^j \|_2.
\end{align*}
% VC: WHAT IS k?
% VS: fixed.
For a symmetric matrix $M$, define
$$
\| M \|_{2, \infty} = \| M' \|_{2, \infty} = \max_{1 \leq i \leq dg} \max_{1 \leq j \leq d} \left(\sum_{k \in J_j} M_{i, k}^2 \right)^{1/2}.
$$ 
Define the group restricted cone as 
\begin{align*}
\mathsf{REG}(\bar{c}) := \left \{ \Delta \in \mathrm{R}^{gd}: \sum_{j \in \mathcal{T}^c} \| \Delta^j \|_2 \leq \bar{c} \sum_{j \in \mathcal{T}} \| \Delta^j \|_2 , \quad \Delta \neq 0 \right \}.
\end{align*}
Given a matrix $M \in \mathrm{R}^{gd} \times \mathrm{R}^{gd}$, define the restricted group-sparse eigenvalue 
\begin{align*}
\kappa_g(M, \mathcal{T},\bar{c}) = \min_{\Delta \in \mathsf{REG}(\bar{c})} \dfrac{\sqrt{s} (\Delta' M \Delta)^{1/2} }{\| \Delta^{\mathcal{T}} \|_{2,1}}.
\end{align*}
When the group size $g$ is equal to 1, the objects above reduce to the following quantities:
\begin{align*}
& \Delta^j = \Delta_j, \Delta^{\mathcal{T}} = \Delta_{\mathcal{T}} = (\Delta_m)_{\{m \in \mathcal{T}\}}  , \| M \|_{2, \infty} = \| M \|_{ \infty}, 
\end{align*}
the group restricted cone is regular restricted cone
\begin{align*}
\mathsf{REG}(\bar{c})  = \mathsf{RE}(\bar{c}) = \{ \Delta \in \mathrm{R}^{d}:  \| \Delta_{\mathcal{T}^c} \|_1 \leq \bar{c}\| \Delta_{\mathcal{T}} \|_1 , \quad \Delta \neq 0\}, \end{align*}
and the the restricted group-sparse eigenvalue  reduces to restricted eigenvalue
\begin{align*}
\kappa_1(M, \mathcal{T},\bar{c}) = \kappa (M, \mathcal{T}, \bar{c}) =  \min_{\Delta \in \mathsf{RE}(\bar{c})} \dfrac{\sqrt{s} (\Delta' M \Delta)^{1/2} }{\| \Delta_{\mathcal{T}} \|_{1}}.
\end{align*}
Let $\bar{X}_{it} \in \mathrm{R}^{gd}$ be a generic covariate $(dg)$-vector and $\bar{Y}_{it}$ be a generic outcome. Given a parameter $\bar{\beta}_0 $, decompose
\begin{align*}
    \bar{Y}_{it} = \bar{X}_{it}'\bar{\beta}_0 + U_{it}.
\end{align*}
The least squares loss function is
\begin{align*}
    \mathcal{Q}(\bar{\beta}):=1/2 (NT)^{-1} \sum_{i=1}^N \sum_{t=1}^T (\bar{Y}_{it} - \bar{X}_{it}' \bar{\beta} )^2.
\end{align*}
The group lasso estimator is 
\begin{align}
    \label{eq:grouplasso}
    \widehat{\bar{\beta}}:= \arg \min_{\bar{\beta}}  \mathcal{Q}(\bar{\beta}) + \lambda \| \bar{\beta} \|_{2,1}.
\end{align}
The least squares gradient is
\begin{align*}
    \mathcal{S}(\bar{\beta}_0):=\nabla_{\bar{\beta}_0} \mathcal{Q}(\bar{\beta}_0) =  (NT)^{-1} \sum_{i=1}^N \sum_{t=1}^T  (\bar{Y}_{it} -\bar{X}_{it}' \bar{\beta}_0) \bar{X}_{it}.
\end{align*}
and the Hessian is
\begin{align*}
    \mathcal{H}(\bar{\beta}_0):= (NT)^{-1} \sum_{i=1}^N \sum_{t=1}^T  \bar{X}_{it} \bar{X}_{it}'.
\end{align*}

\begin{lemma}[Grouped Norm Inequalities]
For any two vectors $a, b \in \mathrm{R}^{gd}$ and matrix $M  \in \mathrm{R}^{gd} \bigtimes \mathrm{R}^{gd}$, the following inequalities hold
\begin{align}
| a' b | &\leq  \| a \|_{2,1} \| b \|_{2,\infty}  \label{eq:groupsparsity1} \\
| v' M v | & \leq \sqrt{g}  \| v \|_{2, 1}^2 \cdot \| M  \|_{2, \infty}  \label{eq:groupsparsity2}  \\
\label{eq:groupsparsity5}
\| M \|_{2, \infty} &\leq \| M \|_{\infty} \sqrt{g}
\end{align}
\end{lemma}
\begin{proof}
For each group $j=1,2,\dots, d$, Cauchy inequality gives
\begin{align*}
\left | \sum_{k \in J_j} a_k b_k \right| \leq \Bigg(\sum_{k \in J_j} a_k^2 \Bigg)^{1/2}  \Bigg(\sum_{k \in J_j} b_k^2\Bigg)^{1/2} \leq  \max_{1 \leq j \leq d}  \Bigg(\sum_{k \in J_j} a_k^2 \Bigg)^{1/2} \| b^j \|_2 = \left (\max_{1 \leq j \leq d} \| a^j \|_2 \right) \| b^j \|_2,
 \end{align*}
 which implies
 \begin{align*}
 | a' b | \leq \sum_{j=1}^d \left | \sum_{k \in J_j} a_k b_k \right | \leq \left (\max_{1 \leq j \leq d} \| a^j \|_2\right)  \sum_{j=1}^d \| b^j \|_2 = \| a \|_{2, \infty} \| b \|_{2,1}
  \end{align*}
For each index $i, 1 \leq i \leq kg$, the following bound holds:
\begin{align*}
\left | \sum_{k=1}^{gd} M_{i, k} v_k \right| & \leq \sum_{j=1}^d \left | \sum_{k \in J_j} M_{i, k} v_k \right | \leq \sum_{j=1}^d \left (\sum_{k \in J_j} M_{i, k}^2 \right )^{1/2} \left (\sum_{k \in J_j}  v_k^2 \right)^{1/2} \\
&\leq  \max_{1 \leq j \leq d} \left(\sum_{k \in J_j} M_{i, k}^2 \right)^{1/2} \sum_{j=1}^d  \left (\sum_{k \in J_j}  v_k^2 \right)^{1/2} \leq  \| M \|_{2,\infty} \| v \|_{2, 1}.
\end{align*}
Then,
\begin{align*}
\| M v \|_{2, \infty} &= \max_{1 \leq j \leq d}   \|(Mv)^{j}  \|_2 = \max_{1 \leq j \leq d}   \left(\sum_{i \in J_j}
|Mv|_i^2\right)^{1/2} \\
& \leq 
\max_{1 \leq j \leq d}   \left(\sum_{i \in J_j}
\| v \|^2_{2, 1} \| M \|^2_{2,\infty}\right)^{1/2}
\leq \sqrt{g} \| v \|_{2, 1} \| M \|_{2,\infty}.
\end{align*}
Therefore, we obtain  \eqref{eq:groupsparsity2} by combining inequalities above:
\begin{align*}
| v' M v | & \leq \|v\|_{2,1} \|Mv\|_{2,\infty} \leq \sqrt{g}  \| M  \|_{2, \infty}  \cdot \| v \|_{2, 1}^2.
\end{align*}
Finally, the bound \eqref{eq:groupsparsity5} follows from
$$
M_{2,\infty}= \max_{1 \leq j \leq d}   \| M^{j}  \|_2 
\leq \max_{1 \leq j \leq d}   \sqrt{g} \| M^{j}  \|_\infty 
$$
using the fact that $\|v\|_2 \leq \sqrt{\mathrm{dim}(v)} \|v\|_\infty$.

\end{proof}

\begin{lemma}[First-Stage Effect on the Curvature]    \label{smallmultapp}
 Let $\mathbf{M}_1, \mathbf{M}_2  \in \mathrm{R}^{gd \times gd} $ be two matrices. Let  $\lambda^{M}_{NT}:=\| \mathbf{M}_1- \mathbf{M}_2 \|_{\infty}$.  On the event $\kappa^2_g(\mathbf{M}_2, \mathcal{T},\bar{c})> 0$, for any $\Delta \in \mathsf{REG}(\bar{c})$,
 \begin{align}
 \label{eq:smallmult}
    | \kappa^2_g(\mathbf{M}_2,\mathcal{T},\bar{c}) -  \kappa^2_g(\mathbf{M}_1,\mathcal{T},\bar{c})| \leq \lambda^{M}_{NT}(1 + \bar{c})^2 s g.
     \end{align}
\end{lemma}

\begin{proof}[Proof of Lemma \ref{smallmultapp}]
 For any $\Delta \in \mathrm{R}^{gd}$, the  difference can be bounded as
    \begin{align}
         |\Delta' ( \mathbf{M}_1 - \mathbf{M}_2 ) \Delta|  &\leq^{i} \sqrt{g} \| \mathbf{M}_1 - \mathbf{M}_2   \|_{2, \infty}   \| \Delta \|_{2,1}^2          \leq^{ii}  g \lambda^{M}_{NT} \| \Delta \|_{2, 1}^2,    \label{eq:deltaany}
    \end{align} 
    where $i$ follows from \eqref{eq:groupsparsity2} and $ii$ from \eqref{eq:groupsparsity5}.      For any $\Delta \in \mathsf{REG}(\bar{c})$,
    \begin{align}
    \label{eq:deltarec}
    \| \Delta \|_{2, 1}^2  &\leq   (1 + \bar{c})^2  \| \Delta^{\mathcal{T}} \|_{2, 1}^2   \leq   \frac{(1 + \bar{c})^2 s}  {\kappa^2_g(\mathbf{M}_2, \mathcal{T},\bar{c}) } \Delta' \mathbf{M}_2 \Delta  =: \gamma \cdot \Delta' \mathbf{M}_2 \Delta.
    \end{align}
Combining  \eqref{eq:deltaany} and \eqref{eq:deltarec} gives
 \begin{align}
         |\Delta' ( \mathbf{M}_1 - \mathbf{M}_2 ) \Delta| \leq (g \lambda^M_{NT} \gamma) \cdot \Delta' \mathbf{M}_2 \Delta.
\end{align}
Noting that $x \leq |x|$ gives
\begin{align*}
 \Delta' ( \mathbf{M}_1 - \mathbf{M}_2 ) \Delta \leq  |\Delta' ( \mathbf{M}_1 - \mathbf{M}_2 ) \Delta|  \leq (g \lambda^M_{NT} \gamma) \cdot \Delta' \mathbf{M}_2 \Delta,
\end{align*}
which implies
\begin{align}
\label{eq:smallmultapp}
  \Delta' \mathbf{M}_1 \Delta \leq   \Delta' \mathbf{M}_2 \Delta (1 + g \lambda^M_{NT} \gamma).
  \end{align}
Noting that $-x \leq |x|$ gives
\begin{align}
\label{eq:smallmultapp2}
 \Delta' ( \mathbf{M}_2 - \mathbf{M}_1 ) \Delta \leq (g \lambda^M_{NT} \gamma) \cdot   (\Delta' \mathbf{M}_2 \Delta) 
\end{align}
which implies
\begin{align}
\label{eq:smallmultapp3}
 \Delta'   \mathbf{M}_1 \Delta  \geq \Delta' \mathbf{M}_2 \Delta \cdot (1-g \lambda^M_{NT} \gamma).
\end{align}
Rearranging \eqref{eq:smallmultapp} gives an upper bound on $\kappa_g(\mathbf{M}_1,\mathcal{T},\bar{c})$:
    \begin{align*}
        \kappa_g(\mathbf{M}_1,\mathcal{T},\bar{c}) :&= \min_{\Delta \in \mathsf{REG}(\bar{c})} \frac{\sqrt{s} (\Delta' \mathbf{M}_1 \Delta)^{1/2}  }{\| \Delta^{\mathcal{T}}\|_{2,1}}  \\
        &\leq \min_{\Delta \in \mathsf{REG}(\bar{c})} \frac{\sqrt{s} (\Delta' \mathbf{M}_2 \Delta)^{1/2}  }{\| \Delta^{\mathcal{T}}\|_{2,1}}  \sqrt{1+g \lambda^M_{NT} \gamma} \\
        &= \kappa_g(\mathbf{M}_2,\mathcal{T},\bar{c}) \sqrt{1+ g \lambda^M_{NT} \gamma }.
    \end{align*} 
A lower bound on $\kappa_g(\mathbf{M}_1,\mathcal{T},\bar{c})$ follows analogously, that is,
   \begin{align*}
        \kappa_g(\mathbf{M}_1,\mathcal{T},\bar{c}) 
        &\geq \min_{\Delta \in \mathsf{REG}(\bar{c})} \frac{\sqrt{s} (\Delta' \mathbf{M}_2 \Delta)^{1/2}  }{\| \Delta^{\mathcal{T}}\|_{2,1}}  \sqrt{1-g \lambda^M_{NT} \gamma} \\
        &= \kappa_g(\mathbf{M}_2,\mathcal{T},\bar{c}) \sqrt{1- g \lambda^M_{NT} \gamma }.
    \end{align*} 
Taking the squares of both sides of the inequality and rearranging gives \eqref{eq:smallmult}.

\end{proof}

\begin{lemma}[Oracle Inequality for Group Lasso]
\label{lem:el3}
On the event $\mathcal{G}_1: = \{ \lambda \geq c \sqrt{g} \| \mathcal{S} (\bar{\beta}_0) \|_{\infty} \}$,
the error vector $\Delta =\widehat{\bar{\beta}} -\bar{\beta}_0$ belongs to the restricted set:
$$
\Delta \in \mathsf{REG}(\bar{c})
$$
and  obeys the bound
\begin{align}
( \Delta' \mathcal{H}(\bar{\beta}_0) \Delta) &\leq 2 \lambda \bar{c} \| \Delta^{\mathcal{T}} \|_{2,1}, \label{eq:fourthgroupineq}
\end{align}
where $\bar{c}:=(c+1)/(c-1)$.

\end{lemma}
\begin{proof} [Proof of Lemma \ref{lem:el3}]

Assume the event $\mathcal{G}_1$ holds throughout, which implies $ \lambda \geq c \| \mathcal{S} (\bar{\beta}_0) \|_{2,\infty}$. \cite{Negahban}  establishes
\begin{align}
\label{eq:inequality}
\| \bar{\beta}_0 \|_{2,1}  - \| \widehat{\bar{\beta}} \|_{2,1}  &\leq \| \Delta^{\mathcal{T}}\|_{2,1} -  \| \Delta^{\mathcal{T}^c} \|_{2,1},
\end{align}
and shows that $\Delta \in \mathsf{REG}(\bar{c})$, which implies 
\begin{align}
\label{eq:step1}
\| \Delta \|_{2,1} \leq (1 + \bar{c}) \| \Delta^{\mathcal{T}}  \|_{2,1}.
\end{align}
Note that  $\widehat{\bar{\beta}}$ solves group lasso minimization problem \eqref{eq:grouplasso}, so that
\begin{align*}
   \mathcal{Q}(\widehat{\bar{\beta}}) + \lambda \| \widehat{\bar{\beta}} \|_{2,1}  \leq  \mathcal{Q}({\bar{\beta}}_0) + \lambda \| \bar{\beta}_0 \|_{2,1}.
\end{align*}
Expanding the least squares criterion gives
\begin{align*}
    \mathcal{Q}(\widehat{\bar{\beta}})-\mathcal{Q}({\bar{\beta}}_0) = \mathcal{S}( \bar{\beta}_0)' \Delta + 1/2 (\Delta' \mathcal{H}(\bar{\beta}_0) \Delta) \leq \lambda (\| \bar{\beta}_0 \|_{2,1} -\| \bar{\widehat{\beta}} \|_{2,1}) 
\end{align*}
Invoking inequality \eqref{eq:groupsparsity1} for $\mathcal{S}( \bar{\beta}_0)' \Delta$ gives
 \begin{align*}
 1/2 (\Delta' \mathcal{H}(\bar{\beta}_0) \Delta)  &\leq  \lambda (\| \bar{\beta}_0 \|_{2,1} -\| \bar{\widehat{\beta}} \|_{2,1})   +  \| \mathcal{S}( \bar{\beta}_0) \|_{2,\infty} \| \Delta \|_{2,1}.
\end{align*}
Then \begin{align}
1/2 (\Delta' \mathcal{H}(\bar{\beta}_0) \Delta)  &\leq^{i}  \lambda   ( \| \Delta^{\mathcal{T}} \|_{2,1}  -  \| \Delta^{\mathcal{T}^c} \|_{2,1}) +  \lambda/c \| \Delta \|_{2,1}  \nonumber \\
 &\leq  \lambda  \| \Delta^{\mathcal{T}} \|_{2,1}  + 0 +  \lambda /c \| \Delta \|_{2,1}  \nonumber \\
  &\leq^{ii}  \lambda   \| \Delta^{\mathcal{T}} \|_{2,1}  + (\lambda/c) (1+\bar{c})  \| \Delta^{\mathcal{T}} \|_{2,1} \nonumber \\
  &=^{iii} \lambda \bar{c} \| \Delta^{\mathcal{T}} \|_{2,1}, \label{eq:finalboundapp}
 \end{align} 
where (i) follows from \eqref{eq:inequality}, (ii) from \eqref{eq:step1}, and (iii) from $$1+c^{-1}(\bar{c} +1)  = (c+(c+1)/(c-1))/c =
(c+1)/(c-1) = \bar{c}.$$  
Since $\Delta \in \mathsf{REG}(\bar{c})$, \eqref{eq:fourthgroupineq} follows. 
\end{proof}

\begin{proof}[Proof of Theorem \ref{thrm:ortholasso}]
We invoke Lemma \ref{lem:el3} with the group size $g=1$,  $\bar{\beta}_0 = \beta_0$ and $\bar{U}_{it} = U_{it} + R_{it}(\widehat{\mathbf{d}}, \widehat{\mathbf{l}})$. The gradient $\mathcal{S}(\beta_0) = \widehat{S}$, the Hessian 
is $\mathcal{H}(\beta_0) =\widehat{Q}$ and the penalty $\lambda = \lambda_{\beta}$. Note that $\delta \in \mathsf{RE}(\bar{c})$ has been established in the proof of Lemma \ref{lem:el3}.

\textbf{Step 1. } Union bound implies
\begin{align*}
    \Pr (\lambda_{\beta} \leq c \sqrt{g}  \| \widehat{S} \|_{\infty} ) &\leq  \Pr ( \lambda_{\beta} /2\leq   c \sqrt{g}  \| S  \|_{\infty} ) + \Pr (\lambda_{\beta}/2 \leq  c \sqrt{g} \| \widehat{S}  -S  \|_{\infty}) \\
    & = P_S + P_{\widehat{S}-S} \leq o(1)+ o(1),
\end{align*}
where $P_S \leq 2/d=o(1)$ is given in \eqref{eq:mdseq} and  $P_{\widehat{S}-S}=o(1)$ since $$ \| \widehat{S}  -S  \|_{\infty}  \lesssim_P (\rho_{NT}) =o_P (\sqrt{\log d /NT}).$$

\textbf{Step 2. } Let  $\mathbf{M}_2:=Q = (NT)^{-1} \sum_{i=1}^N \sum_{t=1}^T \Ep V_{it} V_{it}'$ and $\mathbf{M}_1:=\widetilde{Q} = \mathbb{E}_{NT} V_{it} V_{it}'$. Observe that
\begin{align}
\label{eq:kappabound1}
\kappa^2(Q, \mathcal{T}, \bar{c}) = \min_{\delta \in \mathsf{RE}(\bar{c})} \dfrac{s \delta' Q \delta }{\| \delta_{\mathcal{T}} \|_{1}^2}  \geq \min_{\delta \in \mathsf{RE}(\bar{c})} \dfrac{s \min \eig (Q) \| \delta \|_2^2}{\| \delta_{\mathcal{T}} \|_{1}^2} \geq^{i}    \min \eig (Q),
\end{align}
where (i) follows from 
$$
s \| \delta \|_2^2 \geq s \| \delta_{\mathcal{T}}\|_2^2 \geq \| \delta_{\mathcal{T}}\|_1^2 \quad \forall \delta \in \mathrm{R}^{d}.
$$
The bounds \eqref{eq:smallmult} and \eqref{eq:covmat} imply
\begin{align*}
|\kappa^2 (\widetilde{Q}, \mathcal{T}, \bar{c})  -  \kappa^2(Q, \mathcal{T}, \bar{c}) | \leq s  \| \widetilde{Q}- Q\|_{\infty} (1 + \bar{c})^2 \lesssim_P (s \kappa_{NT}).
\end{align*}
Therefore, the event $\mathcal{G}_2:=\{ \kappa^2 (\widetilde{Q}, \mathcal{T}, \bar{c}) > C_{\text{min}}/2\}$ holds w.p. $1-o(1)$.

\textbf{Step 3. } Invoke Lemma \ref{smallmultapp} on  the event $\mathcal{G}_2$ with $\mathbf{M}_2:=\widetilde{Q}$ and $\mathbf{M}_1:=\widehat{Q}$. \eqref{eq:smallmult} gives
\begin{align*}
|\kappa^2 (\widehat{Q}, \mathcal{T}, \bar{c})  -  \kappa^2(\widetilde{Q}, \mathcal{T}, \bar{c}) | 
\leq s  \| \widehat{Q}- \widehat{Q}\|_{\infty} (1 + \bar{c})^2 \lesssim_P (s q_{NT}),
\end{align*}
which implies
\begin{align*}
    |\kappa^2 (\widehat{Q}, \mathcal{T}, \bar{c})  -  \kappa^2(Q, \mathcal{T}, \bar{c}) | \lesssim_P (s (q_{NT}+\kappa_{NT})).
\end{align*}
Therefore, the event $\{ \kappa^2 (\widehat{Q}, \mathcal{T}, \bar{c}) > C_{\text{min}}/2\}$ holds w.p. $1-o(1)$. Thus, the event $$
\mathcal{G}_{3}:= s \| \widehat{Q} -  \widetilde{Q} \|_{\infty} (1+\bar{c})^2/ \kappa^2(\widehat{Q}, \mathcal{T}, \bar{c}) < 1/2
$$
 is well-defined and holds w.p. $1-o(1)$.

 \textbf{Step 4. } On the event $\mathcal{G}_1 \cap \mathcal{G}_2 \cap \mathcal{G}_3$, invoking \eqref{eq:smallmultapp3} with $\mathbf{M}_2=\widetilde{Q}$ and $\mathbf{M}_1=\widehat{Q}$ gives
\begin{align*}
    \delta' \widehat{Q} \delta \geq (1/2) \cdot   \delta' \widetilde{Q} \delta
\end{align*}
 Combining inequality above with  \eqref{eq:fourthgroupineq} gives
 \begin{align*}
     \delta' \widetilde{Q} \delta \leq 2 \delta' \widehat{Q} \delta  \leq 4 \lambda_{\beta} \bar{c} \| \delta_{\mathcal{T}} \|_{1} \leq   \sqrt{s} \lambda_{\beta} \frac{   4 \bar{c} (\delta' \widetilde{Q} \delta)^{1/2}  } {\kappa(\widetilde{Q}, \mathcal{T},\bar{c}) }.
 \end{align*}
Dividing LHS and RHS by $(\delta' \widetilde{Q} \delta)^{1/2}$ gives   
 \begin{align*}
(\delta' \widetilde{Q} \delta)^{1/2} \leq \sqrt{s}\lambda_{\beta} \frac{  4  \bar{c}  } {\kappa(\widetilde{Q}, \mathcal{T},\bar{c}) } \lesssim_P (\sqrt{s} \lambda_{\beta})
 \end{align*}
and 
\begin{align*}
\| \delta \|_1 \leq (1+\bar{c}) \| \delta_{\mathcal{T}} \|_{1} \leq (1+\bar{c}) \frac{\sqrt{s} ( \delta' \widetilde{Q} \delta)^{1/2}  } {\kappa(\widetilde{Q}, \mathcal{T},\bar{c}) } \leq 4  (1+\bar{c}) \frac{s \lambda_{\beta} \bar{c} } {\kappa^2(\widetilde{Q}, \mathcal{T},\bar{c}) }.
\end{align*}

\end{proof}

\subsection{Proof of Theorem \ref{thrm:DOL}}

In what follows, we use the notation $Q^{-1} = (\omega^0_{ij})$ and $Q^{-1}_{\cdot,j} :=\omega^0_j$. Define the following quantities
\begin{align*}
s_{j} (\lambda):=  \| 1 \{ |\omega^0_j| \geq  \lambda \} \|_1, \quad r_{j}(\lambda):=   \| (\omega^0_j) 1 \{ |\omega^0_j| \leq  \lambda \} \|_1.
\end{align*}

\begin{remark}
\label{rm:qinverse}
  Assumption \ref{ass:approx} implies the following bounds
\begin{align}
\label{eq:qinverse1}
 \|Q^{-1}\|_{1, \infty} = \max_{1 \leq j \leq d} \| \omega^0_j \|_1 &\leq A_Q \sum_{j=1}^p j^{-a_Q}
\leq A_Q \int_1^\infty j^{-a_Q} dj \leq A_Q/(a_Q-1).
\end{align}
Furthermore, if $A_Q j^{-a_Q} \leq \lambda$, then $j \geq j_Q^{*}:= (A_Q/\lambda)^{1/a_Q}$. This implies 
\begin{align*}
s_j (\lambda):= \| 1 \{ |\omega^0_j| \geq  \lambda \} \|_1 \leq \| 1 \{ A_Q j^{-a_Q} \geq  \lambda \} \|_1  \leq \sum_{j=1}^{j_Q^{*}} 1 =
j_Q^{*}=(A_Q/\lambda)^{1/a_Q}.
\end{align*}
\begin{align*}
 r_j(\lambda)  \leq \int^{\infty}_{j_Q^{*}} A_Q j^{-a_Q} dj = A_Q \dfrac{(j_Q^{*})^{1-a_Q}}{a_Q-1} 
 =  A_Q \dfrac{(A_Q/\lambda)^{(1-a_Q)/a_Q}}{a_Q-1}
=\frac{A_Q^{1/a_Q}}{(a_Q-1)} \lambda^{1-1/a_Q}.
\end{align*}

\end{remark}

\begin{proof}[Proof of Lemma \ref{lem:climerate}]

\textbf{Step 0.}  Suppose Assumptions \ref{ass:sampling}--\ref{ass:approx} hold.  We claim that the event
\begin{align}
\label{eq:g4}
    \mathcal{G}_Q:=  \left \{    \|  \widehat{Q} - Q \|_{\infty} \| Q^{-1} \|_{1, \infty} \leq  \lambda_Q \right \},
\end{align}
holds w.p. $1-o(1)$.   On this event $\mathcal{G}_Q$, by definition of $\widehat{\Omega}$, we have 
\begin{align}
\label{eq:g5}
\| \widehat{\Omega} \|_{1, \infty} \leq \| Q^{-1} \|_{1, \infty},
\end{align}
and, therefore, 
\begin{align}\label{Omegabound} \| \widehat{\Omega}^{\text{CLIME}} \|_{1, \infty}  \leq \| \widehat{\Omega} \|_{1, \infty} \leq \| Q^{-1} \|_{1, \infty}.
\end{align}
To show that $\Pr ( \mathcal{G}_Q) = 1-o(1)$, decompose
\begin{align*}
    \widehat{Q} Q^{-1} - I_d  &= \widehat{Q} Q^{-1} - Q Q^{-1} = (\widehat{Q}-Q) Q^{-1}.
\end{align*}
By Lemma \ref{lem:matrixerrorinf2} for some $\bar C_{\kappa} >0$, w.p. $1-o(1)$,
\begin{align*}
    \|  \widehat{Q} - Q \|_{\infty} \leq  \bar C_{\kappa} \kappa_{NT}.
\end{align*}
Therefore, w.p. $1-o(1)$, 
\begin{align}
\label{eq:climebound3}
     \| \widehat{Q} Q^{-1} - I_d \|_{\infty} \leq \| \widehat{Q} - Q \|_{\infty}  \| Q^{-1} \|_{1, \infty} \leq \bar C_{\kappa} 2 \kappa_{NT} \| Q^{-1} \|_{1, \infty} \leq  \lambda_Q
\end{align}
as long as $C_Q \geq  2 \bar C_{\kappa}  \| Q^{-1} \|_{1, \infty}$. Since $\| Q^{-1} \|_{1, \infty} \leq A_Q/(a_Q-1)$, $C_Q \geq 2 \bar C_{\kappa}  \| Q^{-1} \|_{1, \infty}$ holds by Assumption \ref{ass:approx}.

\textbf{ Step 1.} We establish  \eqref{eq:omegaqinv1}. Specifically, we show that, on the event $\mathcal{G}_Q$,  we have 
$$
\left\|\widehat{\Omega}^{\text {CLIME }}-Q^{-1}\right\|_{\infty}\leq \left\|\widehat{\Omega}-Q^{-1}\right\|_{\infty}\leq \frac{4A_Q}{(a_Q-1)} \lambda_Q.
$$
The argument repeats the proof of equation (13) in \cite{CLIME}, Theorem 6.   On the event $\mathcal{G}_Q$, the bound holds
\begin{align*}
 \| Q \widehat{\Omega}-  I_d \|_{\infty} =  \| Q (\widehat{\Omega}-   Q^{-1}) \|_{\infty}    &\leq \| (Q - \widehat{Q}) (\widehat{\Omega} - Q^{-1}) \|_{\infty} + \|\widehat{Q}  (\widehat{\Omega} - Q^{-1}) \|_{\infty} \\
 &\leq \| Q - \widehat{Q} \|_{\infty} \| \widehat{\Omega} - Q^{-1} \|_{1, \infty} + \| \widehat{Q} \widehat{\Omega}  - I_d \|_{\infty} + \| I_d - \widehat{Q} Q^{-1} \|_{\infty} \\
 &\leq \| Q - \widehat{Q} \|_{\infty} ( \| Q^{-1} \|_{1, \infty} + \| \widehat{\Omega}  \|_{1, \infty} )  + \lambda_Q + \|\widehat{Q} Q^{-1}-I_d \|_{\infty}.  
\end{align*}
Invoking \eqref{eq:g5} and \eqref{eq:climebound3} gives
\begin{align}
\label{eq:inclusion}
    \| Q \widehat{\Omega}-  I_d \|_{\infty} &\leq 2 \| Q - \widehat{Q} \|_{\infty} \| Q^{-1} \|_{1, \infty} + \lambda_Q  + \lambda_Q   \leq 2 \lambda_Q + 2 \lambda_Q = 4 \lambda_Q.
\end{align}
Pre-multiplying $Q \widehat{\Omega}-  I_d$ by $Q^{-1}$ and invoking \eqref{eq:qinverse1} gives
$$
 \|  \widehat{\Omega}-  Q^{-1} \|_{\infty} =  \|  Q^{-1} (Q\widehat{\Omega}-  I_d) \|_{\infty}
    \leq \|Q^{-1} \|_{\infty, 1}  \| Q \widehat{\Omega}-  I_d \|_{\infty} \leq 4 \frac{A_Q}{a_Q-1} \lambda_Q.
$$
Since $Q$ is a symmetric matrix, so is $Q^{-1}$, and 
$$
|\widehat{\Omega}_{mj}^{\text{CLIME}}-  Q^{-1}_{mj}| \leq \max ( |\widehat{\Omega}_{mj}-  Q^{-1}_{mj}|, |\widehat{\Omega}_{jm}-  Q^{-1}_{jm}| ) \leq \| \widehat{\Omega} - Q^{-1} \|_{\infty},
$$
which implies \eqref{eq:omegaqinv1}.

\textbf{Step 2.} We show that \eqref{eq:debiasing1} holds. Specifically, we show that on the event $\mathcal{G}_Q$  we have that
$$
   \| \widehat{\Omega}^{\text{CLIME}} - Q^{-1} \|_{1, \infty} \leq \bar{C}_Q {\lambda_Q}^{1-1/a_Q},
$$
for some constant $\bar{C}_Q$ that depends on $Q$. We closely follow the proof of (14), page 605 in \cite{CLIME}. Using their notation, let 
\begin{align*}
t_n &:= \| \widehat{\Omega}^{CLIME} - Q^{-1} \|_{\infty}, \quad \omega^0_j:=Q^{-1}_{\cdot, j}\\
    h_j &:= \widehat{\Omega}_{\cdot, j}^{\text{CLIME}} - \omega^0_j, \quad 
    h^1_j:=(\widehat{\omega}_{ij} 1\{ |\widehat{\omega}_{ij}| \geq 2 t_n \})_{i=1}^p  - \omega^0_j, \quad
    h^2_j:=h_j - h_j^1.
\end{align*}
By definition of CLIME, on the event $\mathcal{G}_G$, 
\begin{align*}
&\| \omega^0_j \|_1 - \| h^1_j \|_1  + \| h^2_j \|_1  \leq \|  h^1_j + \omega^0_j \|_1 + \| h^2_j \|_1 \\
&=^i  \| h^2_j +   h^1_j + \omega^0_j \|_1  =  \| \widehat{\Omega}_{\cdot,j}^{\text{CLIME}} \|_1 \leq \| \widehat{\Omega}_{\cdot,j} \|_1 \leq \| \omega^0_j \|_1 ,
\end{align*}
where $(i)$ follows from $h^1_j + \omega^0_j$
and $h^2_j$ having non-overlapping support.
This implies $$\| h_j - h_j^1 \|_1 := \| h^2_j \|_1  \leq \| h^1_j \|_1, \quad  \| h_j \|_1 \leq 2 \| h^1_j \|_1.$$ 
Then, the following bound holds:
\begin{align*}
     \| h^1_j \|_1  &= \sum_{i=1}^d | \widehat{\omega}_{ij} 1\{ |\widehat{\omega}_{ij}| \geq 2 t_n \} - \omega^0_{ij} | \\
     &\leq \sum_{i=1}^d  |\omega^0_{ij}|  1\{ \omega^0_{ij} \leq 2 t_n \}  + \sum_{i=1}^d | \widehat{\omega}_{ij}   1\{ |\widehat{\omega}_{ij}| \geq 2 t_n \}  - \omega^0_{ij}  1\{ |\omega^0_{ij}| \geq 2 t_n \}  |   \\
     &\leq   r_{j}(2t_n) +  t_n {\sum_{i=1}^d  1\{ |\widehat{\omega}_{ij}| \geq 2 t_n \}|} +\sum_{i=1}^d |\omega^0_{ij}| |( 1\{ |\widehat{\omega}_{ij}| \geq 2 t_n \}     -   1\{ |\omega^0_{ij}| \geq 2 t_n \} ) | \\
     & \leq r_{j}(2t_n) +t_n \sum_{i=1}^d 1\left\{\left|\omega_{i j}^0\right| \geq t_n\right\} +\sum_{i=1}^d\left|\omega_{i j}^0\right| I\left\{|| \omega_{i j}^0\left|-2 t_n\right| \leq\left|\hat{\omega}_{i j}-\omega_{i j}^0\right|\right\}\\
     & \leq r_{j} (2t_n) +  t_n s_{j}(t_n) + \sum_{i=1}^d |\omega^0_{ij}|   1\{ \omega^0_{ij} \leq 3 t_n \}   \\
     &\leq  r_{j} (2t_n) + t_n s_{j}(t_n) + r_{j}(3t_n) \\
     &\leq C_Q' t_n^{1- 1/a_Q}. \quad \quad (C_Q':=\frac{A_Q^{1/a_Q}}{(a_Q-1)} ( 2^{1-1/a_Q} + (a_Q-1)  +  3^{1-1/a_Q})).
\end{align*}
Since $t_n \leq \| \widehat{\Omega} - Q^{-1} \|_{\infty}$ from Step 1, we have
\begin{align*}
\|(\widehat{\Omega}^{\text{CLIME}} - Q^{-1})' \|_{1, \infty} &=  \|\widehat{\Omega}^{\text{CLIME}} - Q^{-1} \|_{1, \infty}  :=  \max_{1 \leq j \leq d}  \| h_j \|_1 \leq C_Q' (\| \widehat{\Omega} - Q^{-1} \|_{\infty})^{1-1/a_Q} \\
&\leq \bar{C}_Q  {\lambda_Q}^{1-1/a_Q}
\end{align*}
where $\bar{C}_Q=C_Q' (4 A_Q/ (a_Q-1))^{1-1/a_Q}$ is a constant that depends on $Q$. Thus, \eqref{eq:debiasing1} follows.

\textbf{ Step 3. } We show \eqref{eq:debiasing2}. Specifically, we show that on the event $\mathcal{G}_Q$ and $\| \widehat Q - Q \| \leq 1$ and once $\lambda_Q \leq 1$, we have that
$$ \|  I_d-  \widehat{\Omega}^{\text{CLIME}}\widehat{Q}  \|_{\infty} =  \|  I_d- \widehat{Q} \widehat{\Omega}^{\text{CLIME}}  \|_{\infty} \leq C'_Q  {\lambda_Q}^{1-1/a_Q}, $$
for some constant $C'_Q$ that depends only on $Q$. Indeed,
\begin{align}
\| I_d - \widehat{Q} \widehat{\Omega}^{\text{CLIME}} \|_{\infty} &\leq \| I_d - \widehat{Q} Q^{-1} \|_{\infty} + \|  \widehat{Q}  (Q^{-1}- \widehat{\Omega}^{\text{CLIME}} ) \|_{\infty} \nonumber \\
&\leq \| I_d - \widehat{Q} Q^{-1}\|_{\infty} +(\| Q\|_{\infty} + 1) \|  \widehat{\Omega}^{\text{CLIME}} - Q^{-1} \|_{1,\infty}\\  & \leq \lambda_Q + 
(\|Q\|_\infty +  1 ) \bar{C}_Q \lambda_Q^{1- 1/a_Q} \leq C'_Q \lambda_Q^{1- 1/a_Q} \nonumber
\end{align}
for example, taking $C'_Q$ to bound: 
$$
(\lambda_Q^{1- 1/a_Q} +(\|Q\|_\infty + 1 ) C_Q) \leq (1+ (\|Q\|_\infty + 1) C_Q) =: C'_Q  \
$$

\end{proof}

\begin{lemma}[Linearization in Sup-Norm]
Suppose Assumptions \ref{ass:sampling}--\ref{ass:approx} hold.  Then, the debiased estimator $\widehat{\beta}_{DL}$   is asymptotically linear
\begin{align}
        \label{eq:dolclt}
        & \sqrt{NT} (\widehat{\beta}_{DL} - \beta_{0}) =  Q^{-1} \G_{NT} V_{it} U_{it} +R_{NT},\\
        & \| R_{NT}\|_{\infty} \lesssim_P 
        {\lambda_Q}^{1-1/a_Q} \sqrt{ s^2 \log d} + \sqrt{NT} \rho_{NT} =o_{P} (1).    \label{eq:dolclt2}
\end{align}
\label{lem:uLLN}
\end{lemma}

\begin{proof}[Proof of Lemma \ref{lem:uLLN}]

\textbf{ Step 1. }  Recall that
$$
R_{it}(\mathbf{d}, \mathbf{l}):= l_{i0}(X_{it}) -l_i(X_{it}) - (d_{i0}(X_{it}) - d_i(X_{it}))'\beta_0.
$$
and invoking \eqref{eq:uitrit}, which states that
$$
\widehat{\widetilde{Y}}_{it} - \widehat{V}_{it}' \beta_0 = (\widetilde{Y}_{it} - V_{it}' \beta_0) + ((\widehat{\widetilde{Y}}_{it}-\widetilde{Y}_{it}) - (\widehat{V}_{it} - V_{it})' \beta_0) =U_{it} + R_{it}(\widehat{\mathbf{d}}, \widehat{\mathbf{l}}).
$$
we can see that \begin{align*}
& \widehat{\widetilde{Y}}_{it} - \widehat{V}_{it}' \widehat{\beta}_{L} = \widehat{\widetilde{Y}}_{it} - \widehat{V}_{it}' \beta_0 + \widehat{V}_{it}' (\beta_0 - \widehat{\beta}_{L}) \\
& \mathbb{E}_{NT} \widehat{V}_{it} (\widehat{\widetilde{Y}}_{it} - \widehat{V}_{it}'  \widehat{\beta}_{L}) = \mathbb{E}_{NT} \widehat{V}_{it} (U_{it} + R_{it}(\widehat{\mathbf{d}}, \widehat{\mathbf{l}})) + \widehat{Q} (\beta_0 - \widehat{\beta}_{L}).
\end{align*}
Since 
$$ \widehat{\beta}_{DL} - \beta_0 = \widehat{\beta}_{L}-\beta_0 + \widehat{\Omega}^{\text{CLIME}} (\ENT \widehat{V}_{it} (\widehat{\widetilde{Y}}_{it} - \widehat{V}_{it}' \widehat{\beta}_L))
$$
we have that \begin{align*}
  \widehat{\beta}_{DL} - \beta_{0}   &=\widehat{\Omega}^{\text{CLIME}} ( \mathbb{E}_{NT}   \widehat{V}_{it}  (U_{it} +  R_{it}(\widehat{\mathbf{d}}, \widehat{\mathbf{l}})   )
  + \widehat{\Omega}^{\text{CLIME}} \widehat{Q} (\beta_0 - \widehat{\beta}_L) + \widehat{\beta}_L - \beta_0 \\
  &= \widehat{\Omega}^{\text{CLIME}} ( \mathbb{E}_{NT}   \widehat{V}_{it}  (U_{it} +  R_{it}(\widehat{\mathbf{d}}, \widehat{\mathbf{l}})   ) + \underbracket{(I_d - \widehat{\Omega}^{\text{CLIME}} \widehat{Q} ) (\widehat{\beta}_L-\beta_0)}_{L_3} \\
  &=  Q^{-1} \mathbb{E}_{NT} V_{it} U_{it} +
  ( \widehat{\Omega}^{\text{CLIME}}  - Q^{-1}) \mathbb{E}_{NT} V_{it} U_{it} \\
  &+\widehat{\Omega}^{\text{CLIME}} ( \mathbb{E}_{NT}  [ \widehat{V}_{it}  (U_{it} +  R_{it}(\widehat{\mathbf{d}}, \widehat{\mathbf{l}})  - V_{it} U_{it} ] )+L_3\\
&=  Q^{-1}  \mathbb{E}_{NT} V_{it} U_{it} + L_1 + L_2 + L_3,
\end{align*}
where 
\begin{align*}
L_1 &=  (\widehat{\Omega}^{\text{CLIME}} - Q^{-1}) \mathbb{E}_{NT} V_{it} U_{it} \\
L_2 &= \widehat{\Omega}^{\text{CLIME}} \mathbb{E}_{NT} [ V_{it}  R_{it}(\widehat{\mathbf{d}}, \widehat{\mathbf{l}})  + (\widehat{V}_{it} - V_{it} ) (U_{it} +  R_{it}(\widehat{\mathbf{d}}, \widehat{\mathbf{l}})   )] \\
L_3 &=  (I_d -\widehat{\Omega}^{\text{CLIME}}\widehat{Q} )(\widehat{\beta}_{L} - \beta_0).
\end{align*}
\textbf{ Term $L_1$.  } The bounds \eqref{eq:debiasing1}  and  \eqref{eq:mdseq} imply
\begin{align}
\label{eq:terml1}
 \| L_1 \|_{\infty} &\leq   \| \widehat{\Omega}^{\text{CLIME}} - Q^{-1} \|_{1,\infty}   \sqrt{NT} \| \mathbb{E}_{NT} V_{it} U_{it} \|_{\infty}  \\
&\lesssim_P {\lambda_Q}^{1-1/a_Q} \sqrt{NT} \sqrt{\log d /NT}   = o_P(1), \nonumber 
\end{align}
because ${\lambda_Q}^{1-1/a_Q} = o(s^{-1} \log^{-1/2} d) = o (\log^{-1/2} d)$ as assumed in \eqref{eq:debiasing1}.

\textbf{ Term $L_2$.  } The bounds \eqref{Omegabound}  and  the gradient error bound \eqref{eq:approxboundinf}  imply
% YES I CHECKED
\begin{align*}
  \| L_2 \|_{\infty} &\leq \|\widehat{\Omega}^{\text{CLIME}} \|_{\infty,1} \|    \sqrt{NT}   \mathbb{E}_{NT} [ V_{it}  R_{it}(\widehat{\mathbf{d}}, \widehat{\mathbf{l}})  + (\widehat{V}_{it} - V_{it} ) (U_{it} +  R_{it}(\widehat{\mathbf{d}}, \widehat{\mathbf{l}})   )]   \|_{\infty} \\
 &\lesssim_{P} 1 \sqrt{NT} \rho_{NT}= o(1).
\end{align*}
because $\sqrt{NT} \rho_{NT} = o(1)$ is implied by our assumption Assumptions \ref{ass:sampling}--\ref{ass:smallbiashds}.

\textbf{ Term $L_3$.  } The conditions \eqref{eq:debiasing2}  and \eqref{eq:lassoboundmain} imply
\begin{align*}
\sqrt{NT} \| L_3 \|_{\infty} &= \sqrt{NT} \|  (I_d -\widehat{\Omega}^{\text{CLIME}} \widehat{Q} )(\widehat{\beta}_{L} - \beta_0) \|_{\infty} \\& \leq \sqrt{NT} \| I_d -\widehat{\Omega}^{\text{CLIME}} \widehat{Q}  \|_{\infty} \| \widehat{\beta}_{L} - \beta_0 \|_1 \\
& \lesssim_P ( {\lambda_Q}^{1-1/a_Q} \sqrt{NT} \sqrt{ s^2 \log d/NT} ) = o(1),
\end{align*}
where $\| I_d -\widehat{\Omega}^{\text{CLIME}} \widehat{Q}  \|_{\infty} \lesssim_P {\lambda_Q}^{1-1/a_Q} = o (s^{-1} \log^{-1/2} d)$ as assumed in \eqref{eq:debiasing1}.
\end{proof}

\begin{proof}[Proof of Theorem \ref{thrm:DOL}]
\textbf{Step 1.} 
Let $\alpha \in \mathrm{R}^{d}$ be such that  $\| \alpha \|_1 = K_{\alpha}=O(1)$ and $\|\alpha\|_2 = 1$.  Lemma \ref{lem:uLLN} implies
$$
 \alpha'(\alpha' \Sigma \alpha)^{-1/2} 
 ( \sqrt{NT} (\widehat{\beta}_{DL} - \beta_{0})) = 
\alpha'(\alpha' \Sigma \alpha)^{-1/2}  Q^{-1}   \sqrt{NT} \mathbb{E}_{NT}  V_{it} U_{it} + o_P(1),
$$
where $(\alpha' \Sigma \alpha)^{-1/2} = O(1)$
because 
\begin{align}
\label{eq:infbound}
\alpha' \Sigma \alpha \geq \ubar{\sigma}^2 \alpha' Q^{-1} \alpha \geq \ubar{\sigma}^2  C^{-1}_{\text{max}}>0    
\end{align}
by the assumptions of the  Theorem, so that
\begin{align}
\label{eq:mainbound2}
|  \alpha' (\alpha' \Sigma \alpha)^{-1/2} R_{NT}|  \leq  O(1) K_{\alpha} \|R_{NT}\|_\infty = o_P(1).
\end{align}
Consider a sequence $$\xi_{m}(\alpha) := \alpha' Q^{-1} (\alpha' \Sigma \alpha)^{-1/2} V_{m} U_{m}, \quad m=1,2,\dots, M$$ with $$ m=m(i,t) = T(i-1)+t, \quad 1 \leq t \leq T, 1 \leq i \leq N. $$ As shown in Corollary \ref{lem:mds},   $\{ \xi_{m}(\alpha) \}_{m=1}^M$ is a martingale difference sequence w.r.t. natural filtration with $M=NT$.   By Law of Large Numbers in \cite{HansenLLN} and the assumed Lindeberg condition
$$\frac{1}{NT} \sum_{m=1}^{NT} \xi_{m}^2 (\alpha) \rightarrow_{p} \frac{ \alpha' Q^{-1} \Gamma Q^{-1} \alpha  }{\alpha ' \Sigma \alpha }=1.$$ 

As discussed in \cite{McLeish1974}, the Lindeberg condition assumed in the Theorem  \ref{thrm:DOL} implies conditions (i) and (ii) in Theorem 2.3 of \cite{McLeish1974}, which implies the first part  of the Theorem:
$$
 \Pr( \alpha'(\alpha' \Sigma \alpha)^{-1/2} 
  \sqrt{NT} (\widehat{\beta}_{DL} - \beta_{0}) \leq t ) \to \Phi(t).
$$
By Polya's theorem, the convergence is uniform in $t \in \mathrm{R}$. Since the result holds for any sequence $\{\alpha\}$ (indexed by $N,T$ obeying conditions above, the convergence is uniform over such sequences.)

\textbf{Step 2.} Let $K_{\alpha}$ be a finite constant in the statement of the theorem. Thus,
\begin{align*}
   \sup_{\alpha: \| \alpha \|_2 \leq 1, \| \alpha\|_1 \leq K_{\alpha} } |\alpha' ( \widehat{\Sigma} - \Sigma) \alpha |  &\leq K_{\alpha}^2 \|  \widehat{\Sigma} - \Sigma  \|_{\infty}  =o_P(1)
\end{align*}
by assumption. Since $
\min_{\|\alpha\|_2 = 1} \alpha'  \Sigma \alpha
\geq \ubar{\sigma}^2  C^{-1}_{\text{max}},
$ by assumption, 
we conclude that, for $N$ and $T$ large enough, the event 
\begin{align*}
  \mathcal{G}_K:= \Big \{ \inf_{\|\alpha\|_2 = 1, \|\alpha\|_1 \leq K_{\alpha}}
 \alpha'  \widehat{\Sigma} \alpha > \ubar{\sigma}^2  C^{-1}_{\text{max}}/2 \Big\}
 \end{align*}
 occurs wp $1-o(1)$. Hence wp $1-o(1)$.  
\begin{align}
\label{eq:alphant}
\alpha_{NT} - 1:= \dfrac{(\alpha' \Sigma \alpha)^{1/2}} {(\alpha'  \widehat{\Sigma} \alpha)^{1/2}} - 1
\end{align}
obeys
$$
|\alpha_{NT} -1 | \leq (\ubar{\sigma}^2  C^{-1}_{\text{max}}/2)^{-1}  K_{\alpha}^2 \|  \widehat{\Sigma} - \Sigma  \|_{\infty} =o_P(1)
$$
which follows from the inequality 
\begin{align*}
    &\bigg|1 -\frac{\sqrt{x} }{\sqrt{y}} \bigg|= \frac{| \sqrt{x} - \sqrt{y} |}{\sqrt{y}} = \frac{|x-y|}{\sqrt{y} (\sqrt{x} + \sqrt{y})}; \quad  x>0, y>0
\end{align*}
Then
$$| (\alpha_{NT}-1) \alpha' (\alpha' \Sigma \alpha)^{-1/2} \sqrt{NT} (\widehat{\beta}_L - \beta_0)| \leq |\alpha_{NT}-1| | \alpha' (\alpha' \Sigma \alpha)^{-1/2} \sqrt{NT} (\widehat{\beta}_L - \beta_0)|
= o_p(1) O_P(1).$$
Therefore, 
$$
\alpha'(\alpha' \widehat\Sigma \alpha)^{-1/2} 
  \sqrt{NT} (\widehat{\beta}_{DL} - \beta_{0}) =  \alpha' (\alpha' \Sigma \alpha)^{-1/2} \sqrt{NT} (\widehat{\beta}_L - \beta_0) + o_P(1).
$$
Then convergence in distribution for the left side follows by Slutsky's lemma and Step 1.

\end{proof}

\subsection{Estimation of $\Sigma$: Proof of Lemma \ref{lem:sigmaolsbound}}
Define the following terms 
\begin{align}
\label{eq:higherorder1}
      \bar{b}_1 &=  \mathbb{E}_{NT} V_{it} V_{it}' U^2_{it} - \Gamma\\
    \bar{b}_2 &= \ENT  (d_{i0}(X_{it}) - \widehat{d}_i(X_{it})) V_{it} U^2_{it} \label{eq:higherorder2} \\
    \bar{b}_3 &= \ENT  (d_{i0}(X_{it}) - \widehat{d}_i(X_{it})) (d_{i0}(X_{it}) - \widehat{d}_i(X_{it}))' U^2_{it} \label{eq:higherorder3} \\
        \bar{b}_4 &= \ENT V_{it} V_{it}' (\widehat{U}^2_{it} - U^2_{it})\label{eq:higherorder4} \\
    \bar{b}_5 &= \ENT  (d_{i0}(X_{it}) - \widehat{d}_i(X_{it})) V_{it} (\widehat{U}^2_{it} - U^2_{it})  \label{eq:higherorder5} \\
     \bar{b}_6 &= \ENT  (d_{i0}(X_{it}) - \widehat{d}_i(X_{it}))(d_{i0}(X_{it}) - \widehat{d}_i(X_{it}))' (\widehat{U}^2_{it} - U^2_{it}) \label{eq:higherorder6}
     \end{align}

The following Lemma establishes tail bound on $\bar{b}_1$.  Recall that $\kappa_{NT}$ from Lemma \ref{lem:matrixerrorinf2} is 
$$
\kappa_{NT}:= \sqrt{\log^3 (d^2 \log (NT)) \log NT /NT }.
$$

\begin{lemma}[Higher-Order Term $\bar{b}_1$] 
\label{lem:gamma1}
Under Assumptions \ref{ass:sampling}--\ref{ass:smallbiashds} and \ref{ass:sigma}, 
\begin{align}
\label{eq:gammainfmean}
&    \| \Gamma \|_{\infty} = \max_{1 \leq m,j \leq d} |\Ep V_{itj} V_{itm} U^2_{it}| = O(1)\\
\label{eq:gamma1}
&    \| \bar{b}_1 \|_{\infty} \lesssim_P \sqrt{\log^5 (d^2 \log (NT)) \log NT /NT } \leq \kappa_{NT} \log (d^2 NT) = o(1).
\end{align}
\end{lemma}
Proof. The bound \eqref{eq:gammainfmean} and \eqref{eq:gamma1}  follow from \eqref{eq:constanta2} and \eqref{eq:covmat2} with 
$Z_{1,nit}=Z_{2,nit}=V_{it}$,  $\bar{N}=1$ and $g=1$. \qed

\begin{lemma}[Higher-Order Term $\bar{b}_2$]
\label{lem:ALTpower}
Under Assumptions \ref{ass:sampling}--\ref{ass:smallbiashds} and \ref{ass:sigma},
\begin{align}
    \| \bar{b}_2 \|_{\infty} \lesssim_P   (NT)^{-1/4}. \label{eq:maincauchy44}
\end{align}
\end{lemma}

\begin{proof}[ Proof of Lemma \ref{lem:ALTpower}] 
\textbf{ Step 1.}  For $\bar{b}$ as in \eqref{eq:bk} and $q_{NT}$ as in \eqref{eq:qnt},
\begin{align*}
    P^2_1:= \max_{1 \leq j \leq d} \ENT (d_{i0} (X_{it}) - \widehat{d}_i (X_{it}))_j^{2}   \leq \| \bar{b} \|_{\infty} \lesssim_P q_{NT}.
\end{align*}
Invoking the convergence requirement \eqref{eq:kappad2nt} gives
$$
\bigg(1 +  \sqrt{ \log^{7} (d^2 \log (NT)) \log (NT)/NT} \bigg) \lesssim 1 + \kappa_{NT} \log^2 (d^2NT) \lesssim 1.
$$
Invoking the bounds \eqref{eq:constanta2}--\eqref{eq:covmat2}  with $Z_{1,nit}=1$ and $Z_{2,nit}=V_{it}$ and  $\bar{N}=2$ and $g=2$  gives
\begin{align*}
 P^2_2:=  \max_{1 \leq m \leq d}   \ENT V^{2}_{itm} U^{4}_{it} \lesssim_P \bigg(1 +  \sqrt{ \log^{7} (d^2 \log (NT)) \log (NT)/NT} \bigg) \lesssim_P 1.
\end{align*}
Cauchy inequality implies
\begin{align*}
     &\max_{1 \leq m,j \leq d}    |\ENT  |(d_{i0} (X_{it}) - \widehat{d}_i (X_{it}))_j|   |V_{itm}| U^{2}_{it} | \\
     &\leq      \max_{1 \leq j \leq d} 
     (\ENT  (d_{i0} (X_{it}) - \widehat{d}_i (X_{it}))_j^{2})^{1/2}   \max_{1 \leq m \leq d}   (\ENT V_{itm}^{2} U^{4}_{it})^{1/2}\\
     &\lesssim_P \sqrt{q_{NT} \cdot 1} = o_P ((NT)^{-1/4}),
\end{align*}
where the last bound is established in \eqref{eq:tildeqinf}. 

\end{proof}

\begin{lemma}[Higher-Order Term $\bar{b}_3$]
\label{lem:powerbound}
Under Assumptions \ref{ass:sampling}--\ref{ass:smallbiashds} and \ref{ass:sigma},
\begin{align}
\label{eq:maincauchy}
   \| \bar{b}_3 \|_{\infty}  \lesssim_P   o ((NT)^{-1/4}).
\end{align}
\end{lemma}
\begin{proof}[Proof of Lemma \ref{lem:powerbound}] 
On the event $\sup_{it} |d_{i0} (X_{it}) - \widehat{d}_i (X_{it}) | \leq \mathbf{d}_{NT, \infty} \leq 1$, which happens with probability $1-o(1)$, 
\begin{align*}
    &\max_{1 \leq m,j \leq d}    \ENT  |(d_{i0} (X_{it}) - \widehat{d}_i (X_{it}))_j|   |(d_{i0} (X_{it}) - \widehat{d}_i (X_{it}))_m|  U^{2}_{it} \\
    &\leq 
    \max_{1 \leq j \leq d}    \ENT   |(d_{i0} (X_{it}) - \widehat{d}_i (X_{it}))_j|  U^{2}_{it} \\
    &\leq  \max_{1 \leq j \leq d} (\ENT   (d_{i0} (X_{it}) - \widehat{d}_i (X_{it}))^2_j)^{1/2} (\ENT U^{4}_{it})^{1/2} \\
    &\leq \sqrt{P_1^2}  (\ENT U^{4}_{it})^{1/2} \lesssim_P   o ((NT)^{-1/4}).
\end{align*}
\end{proof}

Recall that the first-order estimation error is 
$$
R_{it}(\mathbf{d}, \mathbf{l}):= l_{i0}(X_{it}) -l_i(X_{it}) - (d_{i0}(X_{it}) - d_i(X_{it}))'\beta_0.
$$
\begin{lemma}[Squared Error] 
\label{lem:auxiliary}
Under Assumptions \ref{ass:sampling}--\ref{ass:smallbiashds}, we have that 
\begin{align}
    \label{eq:squarederror}
    \ENT R^2_{it}(\widehat{\mathbf{d}}, \widehat{\mathbf{l}})     \lesssim_P \mathbf{l}^2_{NT} + o((NT)^{-1/2}). 
\end{align}
\end{lemma}

\begin{proof}[Proof of Lemma \ref{lem:auxiliary}]
 \textbf{Step 1.} Consider a term $\bar{z}$ in \eqref{eq:zk} in a special case when 
 $$d_{i0}(X_{it}):=l_{i0}(X_{it}) \cdot (1,1), \quad d=2.$$
 Then, $\bar{z}$ reduces to a $2$-vector
 $$
 \bar{z} :=   \ENT (l_{i0}(X_{it}) - \widehat{l}_i(X_{it}))^2 \cdot (1,1),
 $$
 and 
 $$
 \| \bar{z} \|_{\infty} =  \ENT (l_{i0}(X_{it}) - \widehat{l}_i(X_{it}))^2.
 $$
Invoking \eqref{eq:zkproof} with $\mathbf{d}_{NT}$ and  $\mathbf{d}_{NT, \infty}$ replaced by $\mathbf{l}_{NT}$ and $\mathbf{l}_{NT, \infty}$ gives the bound.

\textbf{Step 2.} The following bound holds
\begin{align*}
    \ENT R^2_{it}(\widehat{\mathbf{d}}, \widehat{\mathbf{l}}) &\leq 2 \ENT ((d_{i0}(X_{it}) - \widehat{d}_i(X_{it}))'\beta_0)^2 + 2 \ENT (l_{i0}(X_{it}) - \widehat{l}_i(X_{it}))^2 \\
    &=2 \beta_0' \bar{b} \beta_0 + 2\ENT (l_{i0}(X_{it}) - \widehat{l}_i(X_{it}))^2 \\
    &\leq 2 \| \bar{b} \|_{\infty} \| \beta_0 \|^2_1 + 2\ENT (l_{i0}(X_{it}) - \widehat{l}_i(X_{it}))^2 \\
    &\lesssim_P \|\beta_0\|_1^2\left (\textbf{d}_{NT}^2  +   \textbf{d}_{NT,\infty}^2 \sqrt{ (NT)^{-1} \log (NT) \log d}\right) + \mathbf{l}^2_{NT} + o((NT)^{-1/2}) \\
    &\lesssim_P^{i} o((NT)^{-1/2})+  \mathbf{l}^2_{NT} + o((NT)^{-1/2}),
\end{align*}
where (i) follows combining $\| \beta_0 \|_1 \leq \bar C_\beta$ assumed in  Assumption \ref{ass:smallbiashds} (a) and $q_{NT} = o ((NT)^{-1/2})$, established in \eqref{eq:tildeqinf}. 

\end{proof}

\begin{lemma}[Higher-Order Terms $\bar{b}_4,\bar{b}_5,\bar{b}_6$ with $\widehat{U}^2_{it}-U^2_{it}$]
\label{lem:gamma46}
Under Assumptions \ref{ass:sampling}--\ref{ass:sigma},
\begin{align}
\label{eq:gamma46}
   \sum_{k=4}^{6} \| \bar{b}_{k} \|_{\infty} \lesssim_P ((NT)^{-1/4} + \mathbf{l}_{NT} + \sqrt{s  \log d/NT} + \mathbf{l}^2_{NT} \log (d^2 NT)) =: \gamma_{NT}.
\end{align}
\end{lemma}

\begin{proof}[Proof of Lemma \ref{lem:gamma46}]

\textbf{Step 1.} Decompose 
\begin{align*}
    \widehat{U}^2_{it} - U^2_{it} = (\widehat{U}_{it} - U_{it} + U_{it})^2 - U^2_{it} =    2 U_{it} (\widehat{U}_{it} - U_{it})+ (\widehat{U}_{it} - U_{it})^2.
\end{align*}
Invoking \eqref{eq:uitrit} gives 
\begin{align*}
    \widehat{U}_{it} =\widehat{\widetilde{Y}}_{it} - \widehat{V}_{it}'\widehat{\beta}_L  &= (\widehat{\widetilde{Y}}_{it} - \widehat{V}_{it}'\beta_0) + (\widehat{V}_{it}'\beta_0 - \widehat{V}_{it}'\widehat{\beta}_L) \\
    &= U_{it} + R_{it} (\widehat{\mathbf{d}}, \widehat{\mathbf{l}}) + \widehat{V}_{it}' (\beta_0 - \widehat{\beta}_L).
\end{align*}
Cauchy inequality implies
\begin{align*}
  \ENT (\widehat{U}_{it} - U_{it})^2  &\leq 2 \ENT R^2_{it}(\widehat{\mathbf{d}}, \widehat{\mathbf{l}}) + 2\ENT (\widehat{V}_{it}' (\widehat{\beta}_L- \beta_0))^2 \\
  &= 2 \ENT R^2_{it}(\widehat{\mathbf{d}}, \widehat{\mathbf{l}})  + 2(\widehat{\beta}_L- \beta_0)' \widehat{Q} (\widehat{\beta}_L- \beta_0) =: U_1 + U_2, 
\end{align*}
where $U_1   \lesssim_P  o(NT)^{-1/2} + \mathbf{l}^2_{NT}$ in established in Lemma \ref{lem:auxiliary} and $U_2 \lesssim_P s \log d /NT$ is Theorem \ref{thrm:ortholasso}.

\textbf{Step 2.}   Let $C(W_{it}, \eta)= (C_{mj} (W_{it}, \eta))$ be a $d \times d$ matrix. For any coordinates $m$ and $j$, decompose
\begin{align*}
\ENT C_{mj}(W_{it}, \widehat{\eta}) (\widehat{U}^2_{it} - U^2_{it}) &= 2\ENT C_{mj}(W_{it}, \widehat{\eta})  U_{it} (\widehat{U}_{it} - U_{it})  +  \ENT C_{mj}(W_{it}, \widehat{\eta}) (\widehat{U}_{it} - U_{it})^2 \\
&=: 2 D_{1mj} (\widehat{\eta}) + D_{2mj} (\widehat{\eta}).
\end{align*}
Cauchy inequality gives
\begin{align*}
   | D_{1mj} (\widehat{\eta}) | &\leq (\ENT C^2_{mj} (W_{it}, \widehat{\eta}) U^2_{it} )^{1/2} (\ENT (\widehat{U}_{it} - U_{it})^2)^{1/2} \\
   &\leq \max_{1 \leq m, j \leq d } (\ENT C^2_{mj} (W_{it}, \widehat{\eta}) U^2_{it} )^{1/2}  (\ENT (\widehat{U}_{it} - U_{it})^2)^{1/2}.
\end{align*}
Maximal inequality gives
\begin{align}
\label{eq:d2eta2}
   | D_{2mj} (\widehat{\eta}) | &\leq \max_{it} \max_{mj}  |C_{mj}(W_{it}, \widehat{\eta})| \ENT (\widehat{U}_{it} - U_{it})^2.
\end{align}
\textbf{Step 3.} If one can verify 
\begin{align}
\label{eq:d1eta}
    \max_{1 \leq m,j \leq d} \ENT C^2_{mj} (W_{it}, \widehat{\eta}) U^2_{it} \lesssim_P 1,
\end{align}
 we have that 
\begin{align*}
    \| D_1 (\widehat{\eta}) \|_{\infty} &= \max_{1 \leq m,j \leq d}  | D_{1mj} (\widehat{\eta}) | \\
    &\leq  O_P (1) \cdot O_P ( (NT)^{-1/4} + \mathbf{l}_{NT}+ \sqrt{ s \log d /NT} )\\
    &\lesssim_P ( (NT)^{-1/4} + \mathbf{l}_{NT}+ \sqrt{ s \log d /NT}).
\end{align*}
If one can verify another  condition
\begin{align}
\label{eq:d2eta}
   \max_{it} \max_{mj}  |C_{mj}(W_{it}, \widehat{\eta}| \lesssim_P ( \log (d^2 NT)),
\end{align}
we have that
\begin{align*}
    \| D_2 (\widehat{\eta}) \|_{\infty} =  O_P ( \log (d^2 NT)) \cdot  O_P (s \log d /NT  + \mathbf{l}^2_{NT}+  (NT)^{-1/2}).
\end{align*}

\textbf{Step 4.1}  Take $C (W_{it}, \widehat{\eta}) = V_{it} V_{it}'$, which corresponds to $\bar{b}_4$ in \eqref{eq:higherorder4}. Invoking \eqref{eq:constanta2} and \eqref{eq:covmat2} with $Z_{1,nit}=Z_{2,nit}=V_{it}$ and  $\bar{N}=2$ and $g=1$ as well the assumed bound \eqref{eq:kappad2nt} gives $$\max_{jk} |\ENT V^2_{itk} V^2_{itj} U^2_{it}| \lesssim_P (1 + \kappa_{NT} \log^2 (d^2 NT)) \lesssim_P 1,$$
which verifies  \eqref{eq:d1eta}. By Lemma \ref{lem:subgauss} (6),  $$ \max_{it} \max_{mj} |V_{itk} V_{itj}| \lesssim_P (\log (d^2 NT)), $$ which verifies \eqref{eq:d2eta}. 

\textbf{Step 4.2} Take $C (W_{it}, \widehat{\eta}) = V_{it} (d_{i0}(X_{it})-\widehat{d}_i(X_{it}))$, which corresponds to $\bar{b}_5$ in \eqref{eq:higherorder5}.  In what follows, we focus on the event $\sup_{it} |d_{i0} (X_{it}) - \widehat{d}_i (X_{it}) | \leq \mathbf{d}_{NT, \infty} \leq 1$. Invoking \eqref{eq:bound2} with $\bar{N}=2$ and $g=1$ gives
\begin{align*}
&\max_{1 \leq m,j \leq d} | \ENT V^2_{itm} (d_{i0}(X_{it})-\widehat{d}_i(X_{it}))^2_{j} U^2_{it}| \\
&\lesssim_P   \mathbf{d}^2_{NT, \infty} \max_{1 \leq m \leq d} | \ENT V^2_{itm} U^2_{it}| \\
&\lesssim_P 
d_{NT, \infty} \left (1+\sqrt{ \log^{5}( d \log NT) \log(NT)/NT}\right) \\
&= \mathbf{d}_{NT, \infty} ( 1+ \kappa_{NT}  \log( d \log NT))) \lesssim  \mathbf{d}_{NT, \infty}.
\end{align*}
Likewise,
$$\max_{it} \max_{mj} | C (W_{it}, \widehat{\eta}) |  \leq \textbf{d}_{NT, \infty} \max_{it} \max_{1 \leq j \leq d} |V_{itj}|  \lesssim_P ( \log (d^2NT)  \textbf{d}_{NT, \infty} )
$$
verifies \eqref{eq:d2eta}. 

\textbf{Step 4.3} Take $C (W_{it}, \widehat{\eta}) =(d_{i0}(X_{it})-\widehat{d}_i(X_{it})) (d_{i0}(X_{it})-\widehat{d}_i(X_{it}))'$, which corresponds to $\bar{b}_6$ in \eqref{eq:higherorder6}. On the event $\sup_{it} |d_{i0} (X_{it}) - \widehat{d}_i (X_{it}) | \leq \mathbf{d}_{NT, \infty} \leq 1$, the condition \eqref{eq:d1eta} becomes
\begin{align*}
&\max_{1 \leq m,j \leq d}  \ENT (d_{i0}(X_{it})-\widehat{d}_i(X_{it}))^2_{m} (d_{i0}(X_{it})-\widehat{d}_i(X_{it}))^2_{j} U^2_{it} \leq  \ENT U^2_{it} \lesssim_P  1.
\end{align*} 
 Noting that $\max_{it} \max_{mj} | C (W_{it}, \widehat{\eta}) | \lesssim_P \textbf{d}^2_{NT, \infty} \lesssim_P 1$ verifies \eqref{eq:d2eta}. 

\textbf{Step 5. (Conclusion).} Collecting the bounds and invoking $(s \vee 1) \kappa_{NT} = o(1)$ gives 
\begin{align}
\label{eq:gamma4ntpre}
    o(NT)^{-1/4} + \mathbf{l}_{NT}+ \sqrt{s \log d/NT} + (\log (d^2 NT) (o(NT)^{-1/2} + \mathbf{l}^2_{NT}+ s \log d/NT)  ) \lesssim \gamma_{NT}.
\end{align}
For $N$ and $T$ large enough, \begin{align*}
  \log (d^2 NT)/ (NT)^{-1/4} \leq 1, 
\end{align*}
which implies $\log (d^2 NT) (NT)^{-1/4}=  o ( (NT)^{-1/4})$. Likewise, $$  \log (d^2 NT)   \sqrt{s \log d /NT} \leq s \sqrt{\log^2 (d^2 NT)  \log d /NT}  \leq (s \vee 1) \kappa_{NT} = o(1), $$
which gives \eqref{eq:gamma4ntpre}.

\end{proof}

\begin{lemma}[Bound on $\| \widehat{\Gamma} (\widehat{\beta}_L) - \Gamma \|_{\infty}$]
Under Assumptions \ref{ass:sampling}--\ref{ass:smallbiashds} and \ref{ass:sigma}, we have that:
\begin{align}
\label{eq:gammantbound}
\| \widehat{\Gamma} (\widehat{\beta}_L) - \Gamma \|_{\infty} \lesssim_P (\gamma_{NT} + \kappa_{NT} \log (d^2 NT)) = o_p(1).
\end{align}

\end{lemma}
\begin{proof}
 Decompose the matrix first-stage error
\begin{align*}
   \widehat{\Gamma} (\widehat{\beta}_L) - \Gamma &=
   \mathbb{E}_{NT} \widehat{V}_{it} \widehat{V}_{it}' \widehat{U}^2_{it} - \Gamma \\
   &=\mathbb{E}_{NT} \widehat{V}_{it} \widehat{V}_{it}' (\widehat{U}^2_{it} - U^2_{it}) + \mathbb{E}_{NT} (\widehat{V}_{it} \widehat{V}_{it}' - V_{it} V_{it}') U^2_{it} + \mathbb{E}_{NT} V_{it} V_{it}' U^2_{it}
   - \Gamma \\
   &= \bar{b}_6  + \bar{b}_5 +  \bar{b}_5' + \bar{b}_4 + \bar{b}_3 + \bar{b}_2 + \bar{b}_2' + \bar{b}_1.
\end{align*}
The bound on $\bar{b}_1$ is given in \eqref{eq:gamma1}, Lemma \ref{lem:gamma1} . The bound on $\bar{b}_2$ is given in \eqref{eq:maincauchy44}, Lemma \ref{lem:ALTpower}. The bound on $\bar{b}_3$ is given in \eqref{eq:maincauchy}, Lemma \ref{lem:powerbound}. The bounds on $\bar{b}_4-\bar{b}_6$ are given in \eqref{eq:gamma46}, Lemma \ref{lem:gamma46}. Summing the bounds gives \eqref{eq:gammantbound}.

\end{proof}

\begin{proof}[Proof of Lemma \ref{lem:sigmaolsbound}]
\textbf{Step 1.} Define the following bounds
\begin{align*}
    \Sigma_1:&= \| \widehat{\Omega}^{\text{CLIME}} - Q^{-1} \|_{\infty,1} \| \widehat{\Gamma} (\widehat{\beta}_L)   \|_{\infty} \| \widehat{\Omega}^{\text{CLIME}}  \|_{1, \infty} \\
     \Sigma_2:&=  \| Q^{-1} \|_{\infty, 1} \| \widehat{\Gamma} (\widehat{\beta}_L) -\Gamma  \|_{\infty} \| \widehat{\Omega}^{\text{CLIME}}  \|_{1, \infty}  \\
     \Sigma_3:&=\| \widehat{\Omega}^{\text{CLIME}} - Q^{-1} \|_{\infty,1}  \| \Gamma \|_{\infty} \| Q^{-1} \|_{ 1, \infty}
\end{align*}
and note that
$$
\| \widehat{\Sigma}(\widehat{\beta}_L) - \Sigma \|_{\infty} = \| \widehat{\Omega}^{\text{CLIME}}  \widehat{\Gamma} (\widehat{\beta}_L) \widehat{\Omega}^{\text{CLIME}}   - Q^{-1} \Gamma Q^{-1} \|_{\infty} \leq \Sigma_1 + \Sigma_2 + \Sigma_3.
$$

\textbf{Step 2.} Invoking \eqref{eq:gammainfmean} and \eqref{eq:gammantbound} gives
$$\| \widehat{\Gamma} (\widehat{\beta}_L) \|_{\infty} \leq \| \Gamma \|_{\infty} +  \| \widehat{\Gamma} (\widehat{\beta}_L)-\Gamma \|_{\infty}   \lesssim_P 1 +\gamma_{NT}+ \kappa_{NT} \log (d^2 NT) \lesssim_P 1.$$
Invoking \eqref{Omegabound} and \eqref{eq:qinverse1} gives $$\| \widehat{\Omega}^{\text{CLIME}} \|_{1,\infty} \leq  \| \widehat{\Omega} \|_{1,\infty}  \leq  \| Q^{-1} \|_{1, \infty} \leq (A_Q/(a_Q-1)).$$
As a result, invoking \eqref{eq:debiasing2} gives
$$
\Sigma_1  = O_P ({\lambda_Q}^{1-1/a_Q}) \cdot O_P(1) \cdot O_P(1).
$$
Likewise, 
\begin{align*}
  \Sigma_2:=  \| Q^{-1} \|_{\infty, 1} \| \widehat{\Gamma} (\widehat{\beta}_L) -\Gamma  \|_{\infty} \| \widehat{\Omega}^{\text{CLIME}}  \|_{1, \infty} &= O(1) \cdot O_P (\gamma_{NT}  + \kappa_{NT} \log (d^2NT)) \cdot O_P(1) \\
  &\lesssim_P (\gamma_{NT}  + \kappa_{NT} \log (d^2NT)) \\
  \Sigma_3:=\| \widehat{\Omega}^{\text{CLIME}} - Q^{-1} \|_{\infty,1}  \| \Gamma \|_{\infty} \| Q^{-1} \|_{ 1, \infty} &= O(1) \cdot O_P (\gamma_{NT}) \cdot O_P(1) \lesssim_P ({\lambda_Q}^{1-1/a_Q}).
\end{align*}
Collecting the terms gives
\begin{align*}
\| \widehat{\Sigma}(\widehat{\beta}_L) - \Sigma \|_{\infty} &= \| \widehat{\Omega}^{\text{CLIME}}  \widehat{\Gamma} (\widehat{\beta}_L) \widehat{\Omega}^{\text{CLIME}}   - Q^{-1} \Gamma Q^{-1} \|_{\infty} \\
&\leq  \Sigma_1 + \Sigma_2 + \Sigma_3 \\
&\lesssim_P  {\lambda_Q}^{1-1/a_Q} + \gamma_{NT}  + \kappa_{NT} \log (d^2NT)+  {\lambda_Q}^{1-1/a_Q}.
\end{align*}

\end{proof}

\subsection{Proof of Theorem \ref{thrm:manycoef}}

The proof is divided in several steps. Step 1 outlines the proof. Step 2-5  establish \eqref{eq:rbound}. Steps 6-8 establish \eqref{eq:rbound2}. 

\begin{proof}
\textbf{Step 1. (Outline)} Let  $Z \sim N(0, \mathcal{C})$ and  $\widehat{Z} \mid \mathcal{\widehat C}   \sim N(0,\widehat {\mathcal{C}})$ be as defined in the Theorem. Define 
$$
 T_{\Sigma, \beta}:= \sqrt{NT}  \Sigma^{-1/2}_{jj} (\widehat{\beta}_{DL,j} - \beta_0), \qquad   T_{ \widehat \Sigma, \beta}:= \sqrt{NT}  {\widehat \Sigma}^{-1/2}_{jj} (\widehat{\beta}_{DL,j} - \beta_0)
$$
and
$$
T_{\Sigma} := \Sigma^{-1/2}_{jj}  \GNT V_{itj} U_{it}.
$$
Define 
\begin{align*}
    O_1 (t):&=\Pr ( \|  T_{\Sigma, \beta} \|_{\infty} < t) - \Pr ( \| T_{\Sigma}  \|_{\infty}   < t + \delta_1   ) \\
    O_2 (t):&= \Pr (  \| T_{\Sigma} \|_{\infty}   \leq t + \delta_1   )   -  \Pr ( \| Z \|_{\infty}   < t+\delta_1) \\
    O_3 (t):&= \Pr ( \| Z \|_{\infty}   < t+\delta_1) -\Pr (   \| Z \|_{\infty}  < t)  
\end{align*}
and note that for each $t$
\begin{align*}
  \Pr (   \| T_{\Sigma, \beta} \|_{\infty}  < t) -\Pr (   \| Z \|_{\infty}  < t)  =\sum_{k=1}^{3} O_k(t).
\end{align*}
Likewise, define 
\begin{align*}
    O_4(t):=\Pr (  \| T_{\widehat \Sigma, \beta}\|_{\infty}   < t)  -\Pr (   \| T_{\Sigma, \beta}\|_{\infty}    < t)
\end{align*}
and 
\begin{align*}
    O_5(t):=\Pr (\|Z \|_{\infty} < t) - \Pr ( \|  \widehat{Z} \|_{\infty}  < t  \mid \mathcal{\widehat C} ).
\end{align*}
Note that  for each $t$,
\begin{align*}
    \Pr ( \| T_{\widehat \Sigma, \beta}\|_{\infty}    < t)  -  \Pr ( \|  \widehat{Z} \|_{\infty}  < t  \mid \mathcal{\widehat C} ) = O_4(t)+\sum_{k=1}^3 O_k(t)+O_5(t).
\end{align*}
Then,  \eqref{eq:rbound} is equivalent to 
\begin{align}
\label{eq:rboundeq}
\sup_{t \geq 0} |\Pr (   \| T_{\Sigma, \beta}\|_{\infty}    < t)  - \Pr ( \| Z \|_{\infty}  < t )| \to 0
\end{align}
and  \eqref{eq:rbound2} is equivalent to 
\begin{align}
\label{eq:rbound2eq}
\sup_{t \geq 0} |\Pr ( \| T_{\widehat \Sigma, \beta} \|_{\infty}  < t)  - \Pr ( \|  \widehat{Z} \|_{\infty}  < t  \mid \mathcal{\widehat C} )| \to_P 0.
\end{align}

\textbf{Step 2.} We show that the elements of $\mathrm{diag} \Sigma $ are bounded from above and below. By Assumption \ref{ass:subgauss} (2), there exists a finite $\bar{\sigma}_{UV}$ such that $\max_{it} \Ep [U^2_{it} \mid V_{it}] \leq \bar{\sigma}_{UV}^2 \text{a.s.}$. As a result, Assumption \ref{ass:subgauss} gives
$$
0< \ubar{\sigma}^2  \leq \min_{it} \Ep [U^2_{it} \mid V_{it}] \leq \max_{it} \Ep [U^2_{it} \mid V_{it}] \leq \bar{\sigma}_{UV}^2 < \infty \text{ a.s. },
$$
which implies $\ubar{\sigma}^2 Q  \preceq \Gamma \preceq \bar{\sigma}_{UV}^2  Q $, and $\ubar{\sigma}^2 Q^{-1} \preceq  \Sigma \preceq \bar{\sigma}_{UV}^2 Q^{-1}$.  As a result, 
\begin{align*}
    0< \ubar{\sigma}^2  C_{\text{max}}^{-1}  \leq  c_{\Sigma}= \min_{1 \leq j \leq d} \Sigma_{jj}  \leq  C_{\Sigma}=  \max_{1 \leq j \leq d} \Sigma_{jj} \leq \bar{\sigma}_{UV}^2  C_{\text{min}}^{-1} < \infty.
\end{align*}
Likewise, the elements of $(\text{diag} \Sigma)^{-1/2}$ are bounded from above by $c_{\Sigma}^{-1/2}$ and from below by
 $ C_{\Sigma}^{-1/2}$.
 
\textbf{Step 3.} We bound $  \sup_{t \geq 0} |O_1(t)|$ with $\delta_1=\log^{-1/2} d \log^{-1/2} NT$. Decomposition \eqref{eq:dolclt} implies
\begin{align*}
 \| T_{\Sigma}\|_{\infty} - \| R_{NT} \|_{\infty} \leq  \| T_{\Sigma, \beta} \|_{\infty} \leq \| T_{\Sigma}\|_{\infty} + \| R_{NT} \|_{\infty},
\end{align*}
and union bound gives
\begin{align*}
  \Pr (   \| T_{\Sigma, \beta}\|_{\infty}    < t) \leq  \Pr (  \| T_{\Sigma}\|_{\infty}   \leq t + \delta_1   ) + \Pr ( \| R_{NT} \|_{\infty} \geq \delta_1) \\
    \Pr (   \| T_{\Sigma}\|_{\infty} < t) \leq  \Pr (  \| T_{\Sigma, \beta} \|_{\infty}  \leq t + \delta_1   ) + \Pr ( \| R_{NT} \|_{\infty} \geq \delta_1) 
\end{align*}
which gives 
\begin{align}
\label{eq:o1tbound}
     \sup_{t \geq 0} |O_1(t)| \leq \Pr ( \| R_{NT} \|_{\infty} \geq \delta_1) =^{i} o(1),
\end{align}
where (i) follows from 
$$ \| R_{NT} \|_{\infty} \lesssim_P  \lambda_Q^{1-1/a_Q} s \log^{1/2} d + \sqrt{NT} \rho_{NT} = o_P (\log^{-1/2} d \log^{-1/2} NT)$$ given in  \eqref{eq:dolclt2} and \eqref{eq:o1tcond} .

\textbf{Step 4.} We verify the conditions of Lemma  \ref{lem:cck2} for  the m.d.s. with
$$
m=m(i,t) = T (i-1)+t, \quad M=NT
$$
and
$$
X_m:= (\text{diag} \Sigma)^{-1/2} V_m U_m, \quad m=1,2,\dots, M, \quad D_M = c_{\Sigma}^{-1} \pi^{VU}_{M}.
$$
To verify the condition \eqref{eq:varcond}, we invoke Assumption \ref{ass:manycoef} which gives
$$
\text{Var} (X_{mj}) = \Sigma^{-1/2}_{jj} \Ep V_{mj}^2 U^2_{m} \Sigma^{-1/2}_{jj} \geq \ubar{\sigma}^2  \min_{it} \| \Ep V_{it} V_{it}' \|_{\infty} C_{\Sigma}^{-1} =: a_1> 0
$$
and Remark \ref{rm:cartesian}
$$
\text{Var} (X_{mj}) = \Sigma^{-1/2}_{jj} \Ep V_{mj}^2 U^2_{m} \Sigma^{-1/2}_{jj} \leq \bar{\sigma}^2 \max_{it} \| \Ep V_{it} V_{it}'  \|_{\infty} c_{\Sigma}^{-1} =: A_1 < \infty.
$$
By Assumption \ref{ass:manycoef}, 
$$\bar{r}:= (2/\kappa) \cdot \log (NT),  \quad \bar{q}:= (NT)^{c_2}  \log^2 d \log^2 (NT)$$ obey \eqref{eq:theoreme2}, which implies  \eqref{eq:maincond}.
By Lemma \ref{lem:cck2}, there exist constants $c_2 \in (0,1/4)$ and $c_X$ and $C_X$ depending on $\ubar{\sigma}, \bar{\sigma}, c_2, C_{\text{min}}, C_{\text{max}}$ such that
\begin{align}
\label{eq:fbound2}
\sup_{t \geq 0} |O_2(t) | = \sup_{t \geq 0} | \Pr ( \| T_{\Sigma} \|_{\infty} \leq t ) -  \Pr (    \|  Z  \|_{\infty}  \leq t ) | \lesssim  C_X (NT)^{-c_X} + (NT)^{-c_2/2}.
\end{align}

\textbf{Step 4.} Bound on $  \sup_{t \geq 0} |O_3(t)|$. Invoking Lemma \ref{lem:anti}  gives
\begin{align*}
&\sup_{t \geq 0}| O_3(t)| \\
&\sup_{t \geq 0} | \Pr ( \| Z \|_{\infty}   < t+\delta_1) -\Pr (   \| Z \|_{\infty}  < t)  |  \\
&\leq  \sup_{t \geq 0} | \Pr ( \| Z \|_{\infty}   < t+\delta_1) -\Pr (   \| Z \|_{\infty}  < t-\delta_1)  | \\
&= \sup_{t \geq 0} \Pr ( | \| Z \|_{\infty} -t | \leq \delta_1) \leq C \delta_1 \sqrt{1 \vee \log (2d/\delta_1)}.
\end{align*}
Notie that the R.H.S is a non-decreasing function of $\delta_1$ in some neighborhood of $0$ and that $\sqrt{1 \vee (x + y)} \leq 1+\sqrt{x} + \sqrt{y}$ for $x,y>0$. Plugging in $\delta_1 = \log^{-1/2} d \log^{-1/2} NT$ gives 
\begin{align}
 \sup_{t \geq 0} | O_3(t)| &\leq  C \log^{-1/2} d \log^{-1/2} NT \sqrt{1 \vee (\log (2d) + \log (\log^{1/2} d \log^{1/2} NT) ) } \nonumber  \\
 &\lesssim \log^{-1/2} NT + \log^{-1/2} d \log^{-1/2} NT \log^{1/2} \log (NT) = o(1). \label{eq:o4}
\end{align}
Combining \eqref{eq:o1tbound} and \eqref{eq:fbound2} and \eqref{eq:o4} gives \eqref{eq:rboundeq}.  By a standard calculation we have $\Ep \| Z \|_{\infty}  \lesssim \sqrt{\log 2d}$. Invoking Gaussian  concentration inequality (see, e.g., \cite{Ledoux}, Theorem 7.1 or Comment 4 in \cite{PRTF}, page 56) implies 
$$
\| Z \|_{\infty}  \lesssim_P \log^{1/2} (2d) + \log^{1/2} (NT).
$$
Since $\| Z \|_{\infty}  $ and $\| T_{\Sigma, \beta} \|_{\infty}$ converge in distribution to the same limit, 
\begin{align}
\label{eq:opbound}
     \| T_{\Sigma, \beta} \|_{\infty} \lesssim_P \log^{1/2} (2d) + \log^{1/2} (NT).
\end{align}

\textbf{ Step 5.1. }  We bound $\sup_{t \geq 0} |O_4(t)|$. 
Take $\rho_j = \Sigma^{1/2}_{jj}/\widehat{\Sigma}^{1/2}_{jj} $  and let $\rho:=(\rho_1, \rho_2, \dots, \rho_d)'$ be a $d$-vector.  Note that all Eucledian $j$-vectors $e_j$ vectors obey $\| e_j \|_2 = \|e_j \|_1=1$ and therefore belong to the set in Theorem \ref{thrm:DOL} with $K_{\alpha}=1$. Let $\alpha_{NT} = (\alpha \Sigma \alpha)^{1/2} /(\alpha \widehat \Sigma \alpha)^{1/2}$ be as in \eqref{eq:alphant}. Invoking \eqref{eq:mainbound2}  and the bound \eqref{eq:zetant} in Lemma \ref{lem:sigmaolsbound} gives
$$
\max_{j} | \rho_j - 1| \leq \sup_{\alpha: \| \alpha \|_2 = \| \alpha \|_1 =1} | \alpha_{NT} -1 | \lesssim_P \gamma_{NT}.
$$
In particular, it implies that the even
$$
 \min_{1 \leq j \leq d} \rho_j > 1/2
$$
occurs wp $1-o(1)$.  For any $\rho_j>1/2$, 
$$
| \rho_j^{-1} -1 | = |1-\rho_j|/|\rho_j| \leq 2 | \rho_j-1|.
$$
Combining the bounds above on the event $ \min_{1 \leq j \leq d} \rho_j > 1/2$ gives
\begin{align}
\label{eq:sigmabound}
\max_{1 \leq j \leq d} | \widehat{\Sigma}^{-1/2}_{jj}/\Sigma^{-1/2}_{jj} -1 | = 
\max_{1 \leq j \leq d } | \rho^{-1}_j - 1|  \leq 2 
\max_{1 \leq j \leq d } | \rho_j - 1| \lesssim_P \gamma_{NT}.
\end{align}

\textbf{ Step 5.2. } Let $v_1 \cdot v_2$ denote $(v_1 \cdot v_2)_j = v_{1j} \cdot v_{2j}$ for $j=1,2,\dots, d$. Note that 
$$
 T_{{\widehat \Sigma} , \beta}  = T_{\Sigma , \beta}  \cdot \rho^{-1}, 
$$
or, equivalently, 
$$
 T_{{\widehat \Sigma} , \beta} -T_{\Sigma , \beta} = (\rho^{-1} -1) T_{\Sigma, \beta}.
$$
Invoking \eqref{eq:sigmabound} and  \eqref{eq:opbound}  gives
\begin{align*}
\| T_{{\widehat \Sigma} , \beta} -T_{\Sigma , \beta} \|_{\infty} &\leq \max_{1 \leq j \leq d} | \rho^{-1}_j -1 | \| T_{\Sigma, \beta} \|_{\infty}
&= O_P (\zeta_{NT}) \cdot O_P (\log^{1/2} d + \log^{1/2} NT) =^{i} o_P(1),
\end{align*}
where (i) follows from \eqref{eq:zetantcond}.  Thus, $\| T_{{\widehat \Sigma} , \beta} \|_{\infty}$ and $ \| T_{\Sigma , \beta} \|_{\infty}$ converge to the same limit in distribution.

\textbf{ Step 6. }    We bound $\sup_{t \geq 0} |O_5(t)|$. 
Invoking Lemma \ref{lem:comp} with $X  \sim N(0, \mathcal{C}) | \widehat{\mathcal{C}}$ and $Y \sim N(0, \mathcal{C})$ and $\widehat \Delta= \| \mathcal{C}  - \widehat{\mathcal{C}} \|_{\infty}$
$$
\sup_{t \geq 0} |O_5(t)| \leq   C' (\widehat{\Delta} \log^2 (2d))^{1/2},
$$
where $C$ depends only on the constants defined in Assumptions \ref{ass:identification} and \ref{ass:subgauss}. In Step 7, we show that for $\zeta_{NT}$ in \eqref{eq:zetant}, 
\begin{align}
   \widehat{\Delta}:= \| \mathcal{C}  - \widehat{\mathcal{C}} \|_{\infty} \lesssim_P^{i} \zeta_{NT} =^{ii} o_P (\log^{-2} d \log^{-1} NT), \label{eq:step6}
\end{align}
where (i) is verified in Steps 7-8 and (ii) is directly assumed in   \eqref{eq:zetantcond}.

\textbf{ Step 7. } Note that
\begin{align*}
    \| \Sigma \|_{\infty} = \|Q^{-1} \Gamma Q^{-1} \|_{\infty} \leq \| Q^{-1} \|_{\infty,1 } \| \Gamma \|_{\infty} \| Q^{-1} \|_{1, \infty} \leq (A_Q/(a_Q-1))^2 \| \Gamma \|_{\infty} = O(1).
\end{align*}
As a result, $$ \|\widehat{\Sigma} \|_{\infty} \leq \|\widehat{\Sigma} - \Sigma  \|_{\infty} + \| \Sigma \|_{\infty} \lesssim_P 1+ \gamma_{NT} \lesssim_P 1.$$  Likewise,
\begin{align*}
    \|  (\mathrm{diag}  \widehat{\Sigma})^{-1/2}  \|_{\infty, 1} =  \|  (\mathrm{diag}  \widehat{\Sigma})^{-1/2}  \|_{1, \infty} =  \max_{1 \leq j \leq d}\widehat{\Sigma}^{-1/2}_{jj} \lesssim_P \zeta_{NT} + c_{\Sigma}^{-1/2} \lesssim_P 1.
\end{align*}
\textbf{ Step 8. }  Define 
\begin{align*}
C_1 :&= \max_{1 \leq j \leq d}| \widehat{\Sigma}^{-1/2}_{jj} - \Sigma^{-1/2}_{jj}|  \| \widehat{\Sigma} \|_{\infty} \| (\text{diag} \widehat{\Sigma})^{-1/2} \|_{1, \infty} \\
C_2 :&= \| (\mathrm{diag}  \Sigma)^{-1/2} \|_{\infty, 1} \| \widehat{\Sigma}  - \Sigma \|_{\infty} \|  (\mathrm{diag}  \widehat{\Sigma})^{-1/2}  \|_{1,\infty} \\
C_3:&= \| (\mathrm{diag}  \Sigma)^{-1/2} \|_{\infty, 1} \| \Sigma \|_{\infty}  \max_{1 \leq j \leq d}| \widehat{\Sigma}^{-1/2}_{jj} - \Sigma^{-1/2}_{jj}|
\end{align*}
and note that
\begin{align*}
    \| \widehat{\mathcal{C}} - \mathcal{C} \|_{\infty} &= \|  (\mathrm{diag}  \widehat \Sigma)^{-1/2}  \widehat{\Sigma} (\mathrm{diag}  \widehat \Sigma)^{-1/2} - (\mathrm{diag}  \Sigma)^{-1/2} \Sigma (\mathrm{diag}   \Sigma)^{-1/2} \|_{\infty} \leq C_1 + C_2 + C_3.
\end{align*}
Invoking \eqref{eq:sigmabound} and \eqref{eq:zetant} 
\begin{align*}
    \max_{1 \leq j \leq d}| \widehat{\Sigma}^{-1/2}_{jj} - \Sigma^{-1/2}_{jj}|  \lesssim_P \zeta_{NT}, \qquad \| \widehat \Sigma - \Sigma \|_{\infty} \lesssim_P \zeta_{NT}
\end{align*}
implies that each term $C_j$ is a product of two $O_P(1)$ terms and a single $O_P (\zeta_{NT})$ term. Thus, $C_1 + C_2 + C_3 \lesssim_P \zeta_{NT} $  verifies (i) in \eqref{eq:step6}.

\end{proof}

\begin{proof}[Proof of Lemma \ref{lem:extended}]
We invoke  Lemma \ref{lem:el3} with 
$\bar{V}_{it} = D_{it} -d_{i0}(Z_{it})$  and $\bar{\widetilde{Y}}_{it} = Y_{it} -l_{i0}(Z_{it})$ and $\bar{\beta}_0 = (\beta_0, \rho_0)$ and $g=2$. 
Steps 1, 2 and 3 are established similarly to the  proof of Theorem \ref{thrm:ortholasso}. Thus, the bounds \eqref{eq:lassoboundgroup} hold for the Orthogonal Group Lasso. As a result, $\| \widehat{\beta}_{L} - \beta_0 \|_1 \leq \sqrt{s^2 \log d /NT}$ wp $1-o(1)$. As a result, the debiased Orthogonal Group Lasso obeys the uniform linearization result \eqref{eq:dolclt}, and Theorems \ref{thrm:DOL} and \ref{thrm:manycoef} hold.

\end{proof}

\setcounter{definition}{0}    
\setcounter{section}{0}    
\setcounter{lemma}{0}    
\setcounter{equation}{0}
\setcounter{remark}{0}
\setcounter{theorem}{0}
\renewcommand{\theequation}{E.\arabic{equation}}
\renewcommand{\thelemma}{E.\arabic{lemma}}
\renewcommand{\thecorollary}{E.\arabic{corollary}}
\renewcommand{\thetable}{E.\arabic{table}}
\renewcommand{\thesection}{E}

\section{ Proofs for Section \ref{sec:fs}}
\label{sec:proofs:fs}

\begin{proof}[Proof of Remark \ref{rm:subgaussPO}]
To prove this, let $\| \cdot \|_{\psi_2}$ denote the Orlizs sub-Gaussian norm under the probability measure $\Pr$ (see \cite{vdvwellner}). Then
\begin{align*}
\| \|F_{it}\|\|_{\psi_2}
& \leq \| \|\Pi_{it} F_{i,t-1}\| \|_{\psi_2}
+ \|\|Q T_{it}\|\|_{\psi_2}  \leq (1- \delta) \| \|F_{i,t-1}\| \|_{\psi_2}
+ A' \bar \sigma^2,
\end{align*}
where $A'$ is a numerical constant.
Iterating on this inequality exactly $t$ times we obtain
$$
\| \|F_{it}\|\|_{\psi_2}
\leq (1- \delta)^{t} \| \|F_{i,0}\| \|_{\psi_2}
+ A' \sum_{\bar t =1}^{t-1}(1- \delta)^{\bar t} \bar \sigma^2 \leq A'\frac{ 
\bar \sigma^2}{1- \delta}.
$$
\qed 
\end{proof}

\begin{proof}[Proof of Remark \ref{rm:realizationTR}]
Step 1 shows that $\mathbf{p}_{NT} \leq N^{-1/2} (2(B_{\text{max}} + 1))^{1/2} \zeta_{NT,\infty}$. Step 2 shows that wp $1-o(1)$, 
$$
 \sup_{it} | p_{i} (X_{it}) - p_{i0} (X_{it}) | \leq 2 \zeta_{NT, \infty}.
$$
\textbf{ Step 1.} For any $\delta^P$ and $\xi \in \bar{P}_{NT}$, 
\begin{align}
\| \delta^P - \delta^P_0 \|_2 \leq \| \delta^P - \delta^P_0 \|_1 \leq N^{-1/2}\zeta_{NT, \infty} \label{eq:delta1pntbound} \\
\| \xi - \xi_0 \|_2 \leq \| \xi - \xi_0 \|_1 \leq \zeta_{NT, \infty}.  \label{eq:delta1pbound}
\end{align}
Cauchy inequality gives
\begin{align*}
(p_{i}(X_{it}) - p_{i0}(X_{it}))^2 = (X_{it}' (\delta^P - \delta^P_0) + \xi_i - \xi_{i0})^2 \leq 2 (X_{it}' (\delta^P - \delta^P_0))^2 + 2 (\xi_i - \xi_{i0})^2.
\end{align*} 
Summing over $i=1,2,\dots, N$ and $t=1,2,\dots, T$ gives
\begin{align*}
\mathbf{p}^2_{NT} &\leq 2(NT)^{-1} \sum_{i=1}^N \sum_{t=1}^T \Ep (X_{it}' (\delta^P - \delta^P_0))^2  +2 N^{-1} \| \xi-\xi_0 \|^2 \\
&\leq 2 B_{\text{max}} N^{-1} \zeta^2_{NT, \infty} + 
2 N^{-1} \zeta^2_{NT, \infty}.
\end{align*}
Wp $1-o(1)$, $\max_{1 \leq i \leq N, 1 \leq t \leq T} \| X_{it} \|_{\infty} \leq C_X \sqrt{ \log d_X NT}$ for some finite $C_X$ by Lemma \ref{lem:subgauss}.  

\textbf{ Step 2.} The following bound holds wp $1-o(1)$,
\begin{align*}
 \sup_{it} | p_{i} (X_{it}) - p_{i0} (X_{it}) | &\leq \sup_{it} | X_{it}' (\delta^P -\delta^P_0) | + | \xi_i - \xi_{i0} | \\
 &\leq \sup_{it} \| X_{it} \|_{\infty}  \| \delta^P -\delta^P_0 \|_1  + \| \xi - \xi_0 \|_1 \\
 &\leq C_X \sqrt{\log(d_X NT)} N^{-1/2} \zeta_{NT, \infty} +\zeta_{NT, \infty} \\
 &\leq 2 \zeta_{NT, \infty},
\end{align*}
where the last step holds assuming $N$ is large enough and 
$C_X \sqrt{\log(d_X NT)/N} \leq 1$.

\end{proof}

\begin{proof}[Proof of Remark \ref{rm:realization2}]
Step 1 shows that $\mathbf{l}_{NT} = O( N^{-1/2}  (\zeta_{NT,\infty}+\zeta^E_{NT,\infty}))$. Step 2 shows that wp $1-o(1)$, 
$$
 \sup_{it} | l_{i} (X_{it}) - l_{i0} (X_{it}) | \leq 2 \bar{K} \| \beta_0 \|_1 \zeta_{NT, \infty} + 2 \zeta^E_{NT, \infty}.
$$
\textbf{ Step 1}. Decompose 
\begin{align*}
 l_{i} (X_{it}) - l_{i0} (X_{it}) = (d_i(X_{it}) - d_{i0}(X_{it}))' \beta_0 + X_{it}' (\delta^E - \delta^E_0) + \xi^E_i - \xi^E_{i0} + d_i(X_{it})' (\beta - \beta_0)
\end{align*}
Cauchy inequality gives
\begin{align*}
(l_i(X_{it}) - l_{i0}(X_{it}))^2  &\leq 4 \bigg( ( (d_{i} (X_{it}) - d_{i0}(X_{it}))' \beta_0)^2 \\
&+ (X_{it}' (\delta^E- \delta^E_0))^2 + (\xi^E_i - \xi^E_{i0} )^2 + (d_i(X_{it})' (\beta - \beta_0))^2  \bigg).
\end{align*}
Note that $d_i(X_{it})=K (X_{it}) p_i (X_{it})= K(X_{it})(X_{it}'\delta^P + \xi_i)$. Summing over $i=1,2,\dots, N$ and $t=1,2,\dots, T$ gives
\begin{align*}
\textbf{l}^2_{NT} &\leq  4 \underbracket{(NT)^{-1} \sum_{i=1}^N \sum_{t=1}^T  (\delta^P - \delta^P_0)'  \Ep [(K_{it}'\beta_0)^2 X_{it} X_{it}' ] (\delta^P - \delta^P_0) }_{(\delta^P - \delta^P_0)' \Psi_{D} (\delta^P - \delta^P_0)} \\
&+4(NT)^{-1} \sum_{i=1}^N \sum_{t=1}^T \Ep (X_{it}' (\delta^E - \delta^E_0))^2 + 4 N^{-1} \| \xi^E - \xi^E_0 \|^2_2 + 4 (NT)^{-1} \sum_{i=1}^N \sum_{t=1}^T  \Ep  \| d_i (X_{it})\|^2_{\infty} \| \beta - \beta_0 \|^2_1 \\
& \leq 4 (B_{\text{max}} N^{-1} \zeta^2_{NT, \infty}
+ B_{\text{max}} N^{-1} (\zeta^E_{NT, \infty})^2 
+ N^{-1} (\zeta^E_{NT, \infty})^2 
+ B_4 N^{-1} (\zeta^E_{NT, \infty})^2) \\
&\leq 4 (B_{\text{max}}+1+B_4) N^{-1} (\zeta^E_{NT, \infty})^2 + 4 N^{-1}B_{\text{max}} \zeta_{NT, \infty}^2.
\end{align*}
Note that for $N,T$ large enough such that $\| \delta^P \|^2_2 \leq 2 \| \delta^P_0  \|^2_2 \leq 2 \| \delta^P_0 \|^2_1$, which is bounded,
\begin{align*}
   B_4 &= (NT)^{-1} \sum_{i=1}^N \sum_{t=1}^T  \Ep \| d_i (X_{it}) \|^2_{\infty}  \\
   &\leq \bar{K}^2 (NT)^{-1} \sum_{i=1}^N \sum_{t=1}^T \Ep (X_{it}'\delta^P+ \xi_i)^2 \leq 2 \bar{K}^2 ({\delta^P}' \Psi_X \delta^P +  N^{-1} \| \xi \|^2) \\
   &\leq 2 \bar{K}^2 (B_{\text{max}} \| \delta^P \|^2_2 +  N^{-1} \| \xi \|_2^2) \\
   &\leq 4 \bar{K}^2 B_{\text{max}} \| \delta_0^P \|^2_2 +  1.
\end{align*}
\textbf{ Step 2}. For $N,T$ large enough, wp $1-o(1)$,
\begin{align*}
    &\sup_{it} | l_{i} (X_{it}) - l_{i0} (X_{it}) | \leq 
      \sup_{it} | K_{it}'\beta_0 | \sup_{it} |p_i (X_{it}) - p_{i0} (X_{it}) |\\
     &+    \sup_{it}  \| X_{it} \|_{\infty} \| \delta^E -\delta^E_0\|_1  \\
     &+ \sup_{i} | \xi_i - \xi_i^E | \\
     &+  \sup_{it} \| K(X_{it}) \|_{\infty}|  X_{it}'\delta^E | \| \beta - \beta_0 \|_1  \\
     &\leq 2 \bar{K} \| \beta_0 \|_1 \zeta_{NT, \infty}+
     \| X_{it} \|_{\infty} (  N^{-1/2} \zeta^E_{NT, \infty} + \bar{K} \| \delta^E \|_1 N^{-1/2} \zeta^E_{NT, \infty} ) + \| \xi^E - \xi^E_0 \|_1 \\
     &\leq 2 \bar{K} \| \beta_0 \|_1 \zeta_{NT, \infty}+ 
     C_X \sqrt{\log(d_X NT)}  N^{-1/2} (  1 + \bar{K} \| \delta^E \|_1  ) \zeta^E_{NT, \infty} + \zeta^E_{NT, \infty} \\
     &\leq  2 \bar{K} \| \beta_0 \|_1 \zeta_{NT, \infty} + 2 \zeta^E_{NT, \infty},
\end{align*}
where the last step holds assuming $N$ is large enough and
$\| \delta^E \|_1 \leq \| \delta_0^E \|_1$ and
$$C_X  \sqrt{\log(d_X NT)/N} (  1 + 2\bar{K} \| \delta_0^E \|_1  ) \leq 1.$$

\end{proof}

\begin{proof}[Proof of Remark \ref{rm:verification}]
Invoking Remark \ref{rm:realizationTR} and the bound \eqref{eq:zetantp} on $\zeta^P_{NT, \infty}$ in Lemma \ref{lem:kocktang1}  gives \begin{align*}
 \sqrt{NT} \mathbf{p}^2_{NT}  \lesssim \sqrt{NT} N^{-1}  (\zeta^P_{NT, \infty})^2 &\lesssim (S^P)^2 N^{-1/2} T^{1/2} T^{\nu-1} \log^{3 (1-\nu)} (d_X +N) \\
    &\lesssim (S^P)^2 N^{-1/2} T^{\nu-1/2}    \log^{3 (1-\nu)} (d_X +N)   = o (1).\end{align*} In addition, the bound \eqref{eq:zetante} on $\zeta^E_{NT, \infty}$ in Lemma \ref{lem:kocktang2} gives
    \begin{align*}
    \sqrt{NT} \mathbf{p}_{NT} \mathbf{l}_{NT}  &\lesssim \sqrt{NT} N^{-1}  \zeta^P_{NT, \infty} \zeta^E_{NT, \infty} \\
    &\lesssim S^P \cdot S^E  N^{-1/2} T^{(\nu + \nu^E)/2 -1} \log^{3 ( 1 - (\nu + \nu^E)/2 )} (d_X +N) = o(1).
\end{align*} 
\qed
\end{proof}

\setcounter{definition}{0}    
\setcounter{section}{0}    
\setcounter{lemma}{0}    
\setcounter{equation}{0}
\setcounter{remark}{0}
\setcounter{theorem}{0}

\renewcommand{\theequation}{F.\arabic{equation}}
\renewcommand{\thelemma}{F.\arabic{lemma}}
\renewcommand{\thecorollary}{F.\arabic{corollary}}
\renewcommand{\thetable}{F.\arabic{table}}
\renewcommand{\theremark}{F.\arabic{remark}}
\renewcommand{\theassumption}{F.\arabic{assumption}}
\renewcommand{\thesection}{F} 

\section{Tools: Tails Bounds for Empirical Rectangular Matrices under Weak Dependence}
\label{sec:toolslows}

\begin{lemma}[Rectangular Matrix Bernstein, Theorem $1.6$ in Tropp (2012)]
\label{lem:tropp}
Consider a finite sequence $\left\{\Xi_m\right\}_{m=1}^M$ of independent, random  matrices with dimensions $d_1 \times d_2$. Assume that there exist constants $R_{\Xi}$ and $\sigma_{\Xi}$ such that
\begin{align}
\label{eq:infcond0}
\mathrm{E} \Xi_m=0, \quad\left\|\Xi_m\right\| \leq R_{\Xi} \text { a.s. }.
\end{align}
Define
\begin{align}
\label{eq:sigma2tropp}
  \sigma_{\Xi}^2=\max \left(\left\|\mathrm{E} \sum_{m=1}^M \Xi_m^{\prime} \Xi_m\right\|,\left\|\mathrm{E} \sum_{m=1}^M \Xi_m \Xi_m^{\prime}\right\|\right) .
\end{align}
Then, for all $t \geq 0$,
\begin{align}
\label{eq:tropp}
    \mathrm{P}\left(\left\|\sum_{m=1}^M \Xi_m\right\| \geq t\right) \leq\left(d_1+d_2\right) e^{-\left(t^2 /2 \left(\sigma_{\Xi}^2+R_{\Xi} t / 3\right) \right)} .
\end{align}
\end{lemma}

\begin{lemma}[Tail Bounds for Weakly Dependent Sums, Operator Norm]
\label{lem:matrixberbeeweakdep}
Consider the setup of  Lemma \ref{lem:matrixerrorinf} with weakly dependent data $\{W_{it}\}$ and matrix-valued functions $\{ \phi_i (\cdot) \}_{i=1}^N: \mathcal{W} \rightarrow \mathrm{R}^{d_1 \times d_2}$.
Let $q= \lfloor (2/ \kappa) \log (NT) \rfloor$ be as in \eqref{eq:defq} and $L= \lfloor T/2q \rfloor $. For $i=1,2,\dots, N$ and $l=1,2,\dots, L$,  let the data blocks $B_{i(2l-1)}$, $B_{i2l}$ and $B_{ir}$ be as in \eqref{eq:oddblock}--\eqref{eq:remblock}. Let the full-sized odd-block sums $\phi_i (B_{i(2l-1)}) $ be as in \eqref{eq:soddq2}, that is,
$$
\phi_i(B_{i(2l-1)}) = \sum_{t=(2l-2)q+1}^{t=(2l-2)q+q} \phi_i (W_{it}), \quad \phi_i(B_{i(2l)}) = \sum_{t=(2l-1)q+1}^{t=(2lq)} \phi_i(W_{it})
$$
and let $\phi_i(B^{*}_{i(2l-1)}) $ and $\phi_i(B^{*}_{i(2l)}) $ be their Berbee copies.  In  case $T \neq 2Lq$, the remainder block $\phi_i(B_{ir}) $ as in \eqref{eq:remblockq}, that is
$$
\phi_i(B_{ir}):= \sum_{t=2Lq+1}^{T} \phi_i(W_{it})
$$

Suppose that there exist constants $R$ and $\sigma$ such that the following conditions hold 
\begin{align}
\label{eq:infcond1}
 \Ep \phi_i(W_{it}) = 0, \quad \sup_{it} \| \phi_i(W_{it}) \| \leq R \text{ a.s. }
\end{align}
and
\begin{align}
\label{eq:ineq1}
 &\max \left ( \| \sum_{i=1}^N \sum_{l=1}^L   \Ep \phi_i (B^{*}_{i(2l)})' \phi_i(B^{*}_{i(2l)}) \|,  \| \sum_{i=1}^N \sum_{l=1}^L   \Ep \phi_i(B^{*}_{i(2l)}) \phi_i (B^{*}_{i(2l)})' \| \right) \leq q NT \sigma^2 \\
\label{eq:ineq2} & \max \left ( \| \sum_{i=1}^N \sum_{l=1}^L   \Ep \phi_i(B^{*}_{i(2l-1)})' \phi_i(B^{*}_{i(2l-1)}) \|,  \| \sum_{i=1}^N \sum_{l=1}^L   \Ep \phi_i(B^{*}_{i(2l-1)}) \phi_i(B^{*}_{i(2l-1)})' \| \right) \leq q NT \sigma^2 \\
\label{eq:ineq3} &  \max \left (  \| \sum_{i=1}^N \Ep \phi_i (B_{ir}) \phi_i(B_{ir})' \|,  \| \sum_{i=1}^N \Ep \phi_i (B_{ir})' \phi_i(B_{ir}) \| \right) \leq q NT \sigma^2
\end{align}
Then, for any $t \geq 0$,
\begin{align}
\label{eq:meanweakdepbound}
    \Pr (\| (NT)^{-1} \sum_{i=1}^N \sum_{t=1}^T \phi_i(W_{it}) \| \geq  3t) \leq 3 (d_1+d_2) e^{-t^2 NT /2\left(q  \sigma^2+q R t / 3\right)}  +2NL \gamma(q)
\end{align}
and  under geometric beta-mixing condition  \eqref{eq:expmix},
\begin{align}\label{eq:meanweakdeprate}
    \left \|\frac{1}{NT} \sum_{i=1}^N \sum_{t=1}^T \phi_i (W_{it}) \right \| \lesssim_P \frac{1}{\sqrt{NT}} \left( \sigma \sqrt{\log (NT) \log (d_1+d_2)} +  \frac{1}{\sqrt{NT}}  \log (NT) R \log (d_1+d_2)\right) .\end{align}

\end{lemma}

\begin{remark}
In what follows, we write $\phi (W_{it})$ in place of $\phi_i(W_{it})$, but subsume the dependence on $i$. 
\end{remark}
\begin{proof}[Proof of Lemma \ref{lem:matrixberbeeweakdep}]

Union bound gives
\begin{align}
\label{eq:unionboundop}
    &\Pr ( \| \sum_{i=1}^N \sum_{t=1}^T \phi (W_{it}) \| \geq 3t) \\
    &\leq \Pr ( \| \sum_{i=1}^N \sum_{l=1}^L \phi (B^{*}_{i(2l-1)}) \| \geq t) \nonumber  \\
    &+ \Pr ( \| \sum_{i=1}^N \sum_{l=1}^L \phi (B^{*}_{i(2l)}) \| \geq t) + \Pr ( \| \sum_{i=1}^N \phi (B_{ir}) \| \geq t) +2 NL \gamma(q). \nonumber
\end{align}
We first establish the bound for the odd-block sums. Define $$m=m(i,l)=L\cdot (i-1)+l, \quad M=NL, \quad \Xi_m:=\phi (B^{*}_{i(2l-1)}).$$
Since $\phi (B^{*}_{i(2l-1)})$ consists of $q$ summands and $W^{*}_{it}$ and $W_{it}$ have the same marginal distributions, the bound \eqref{eq:infcond1} gives
\begin{align*}
    \| \phi (B^{*}_{i(2l-1)}) \|  \leq q R \text{ a.s. },
\end{align*}
which verifies \eqref{eq:infcond0} with $R_{\Xi}=qR$. Likewise, \eqref{eq:ineq1} directly verifies \eqref{eq:sigma2tropp} with the bound $\sigma^2_{\Xi}=q NT \sigma^2$. Invoking Lemma \ref{lem:tropp} gives
\begin{align*}
    \Pr \left( \| \sum_{i=1}^N \sum_{l=1}^L \phi (B^{*}_{i(2l-1)}) \| \geq t \right) \leq (d_1+d_2) e^{-t^2/2\left(q NT \sigma^2+q R t / 3\right)}.
\end{align*}
A similar bound holds for the even-numbered sums. For the remainder blocks, we take
$$ m=i, \quad M=N, \quad \Xi_m = \phi (B_{ir}).$$
Since the remainder block has at most $q$ elements,
\begin{align*}
    \| \phi (B_{ir}) \| \leq  q R \text{ a.s. }
\end{align*}
which implies \eqref{eq:infcond0} with $R_{\Xi}=qR$. Likewise, \eqref{eq:ineq3} directly verifies \eqref{eq:sigma2tropp} with the bound $\sigma^2_{\Xi}=q NT \sigma^2$. 
Therefore, 
\begin{align*}
    \Pr \left( \| \sum_{i=1}^N  \phi (B_{ir}) \| > t \right) \leq (d_1+d_2) e^{-t^2/2\left(q NT \sigma^2+q R t / 3\right)}.
\end{align*}
Invoking union bound \eqref{eq:unionboundop} gives 
\begin{align*}
    \Pr \left (\|  \sum_{i=1}^N \sum_{t=1}^T  \phi (W_{it}) \| \geq 3t \right) \leq 3 (d_1+d_2) e^{-t^2 / 2\left(q NT\sigma^2+q R t / 3\right)}  + 2 NL \gamma(q).
\end{align*}
Plugging $t(NT)$ in place of $t$ gives and dividing each side by $NT$ gives
\begin{align*}
    \Pr \left (\| \frac{1}{NT} \sum_{i=1}^N \sum_{t=1}^T \phi (W_{it}) \| \geq 3t \right) &\leq 3 (d_1+d_2) e^{-t^2 (NT)^2 /2\left(q NT \sigma^2+q R NT t / 3\right)}  + 2 NL \gamma(q) \\
    &= 3 (d_1+d_2) e^{-t^2 NT /2\left(q  \sigma^2+q R t / 3\right)}  + 2 NL \gamma(q),
\end{align*}
which coincides with \eqref{eq:meanweakdepbound}.  For geometric mixing, taking $q$ as in \eqref{eq:defq} gives $NL \gamma (q)=o(1)$. Noting that
\begin{align*}
    &3 (d_1+d_2) e^{-t^2 NT /2\left(q \sigma^2 +q R t / 3\right)}  \\
    &\leq \max \left (3 (d_1+d_2) e^{-(t^2 NT /4q \sigma^2 )},  3 (d_1+d_2) e^{-(t NT /4qR/3 )}  \right ).
\end{align*}
Plugging $t=C' \sigma \sqrt{ q \log (d_1+d_2)/NT}$ and taking $C'$ large makes the first term in the max as small as desired.  Plugging $t=C' \log (d_1+d_2) q R/NT$ and taking $C'$ large makes the second terms in the max as small as desired. Therefore, 
\begin{align*}
    \left \|\frac{1}{NT} \sum_{i=1}^N \sum_{t=1}^T \phi (W_{it}) \right \| \lesssim_P \frac{1}{\sqrt{NT}} \left( \sigma \sqrt{\log (NT) \log (d_1+d_2)} +  \frac{1}{\sqrt{NT}}  \log (NT) R \log (d_1+d_2)\right) .\end{align*}
\end{proof}

\begin{lemma}
\label{lem:covmatmompre}
Let $ \gamma(X): \mathcal{X}  \rightarrow \mathrm{R}^{d_1 \bigtimes d_2}$ be a fixed matrix-valued function of a random vector $X$. Define the functional 
\begin{align}
\label{eq:phix}
\phi(X)&=  \gamma(X) - \Ep [\gamma(X)].
\end{align}
Let  $\gamma^{\infty}_{NT}$ and $\gamma_{NT}$ be numeric sequences obeying
\begin{align}
    \sup_{it} \|    \gamma(X_{it})  \|  \leq \gamma^{\infty}_{NT} \text{ a.s. }, \quad   (NT)^{-1} \sum_{i=1}^N \sum_{t=1}^T \Ep\|   \gamma(X_{it})  \|^2  \leq \gamma^2_{2NT}  \label{eq:infcond} %\\
 %   (NT)^{-1} \sum_{i=1}^N \sum_{t=1}^T \Ep\|   \gamma(X_{it})  \|^2  \leq \gamma^2_{2NT} \label{eq:4powercond}
\end{align}
Then, the conditions \eqref{eq:infcond1} and \eqref{eq:ineq1}--\eqref{eq:ineq3} hold with
\begin{align}
& R=  2 \gamma_{NT}^{\infty},  \quad   \sigma^2 =  2\gamma_{2NT}^2. \label{eq:meanbound1}
\end{align}
As a result, the  bound \eqref{eq:meanweakdeprate} in Lemma \ref{lem:matrixberbeeweakdep} reduces to 
\begin{align}
\label{eq:meanweakdeprate1}
    \left \| \frac{1}{NT} \sum_{i=1}^N \sum_{t=1}^T \phi (X_{it}) \right \| \lesssim_P  \frac{1}{\sqrt{NT}} \left( \gamma_{2NT} \sqrt{\log (NT) \log (d_1+d_2)} + \frac{1}{\sqrt{NT}} \log (NT) \gamma^{\infty}_{NT} \log (d_1+d_2) \right).
\end{align}

\end{lemma}

\begin{proof}[Proof of Lemma \ref{lem:covmatmompre} ]

 \textbf{Step 1.} Let  $X$ and $\bar X$ be two random vectors, and $\gamma(X)$ and $\gamma(\bar X)$ be $d_1 \times d_2$ matrices.  The following inequalities hold
 \begin{align}
 \|     \Ep \gamma(X) \gamma(\bar X)' \| &\leq^{i} \Ep \|  \gamma(X) \gamma(\bar X)' \|  \leq^{ii} \Ep \| \gamma(X) \| \| \gamma(\bar X)' \| \nonumber \\
 &\leq^{iii} \sqrt{\Ep \| \gamma(X) \|^2 \Ep \| \gamma(\bar X)' \|^2 } \nonumber \\
 &\leq^{iv} 1/2 ( \Ep \| \gamma(X) \|^2 + \Ep \| \gamma(\bar X)' \|^2)=^{v}1/2 ( \Ep \| \gamma(X) \|^2 + \Ep \| \gamma(\bar X) \|^2), \label{eq:xxbar1}   
  \end{align}
where (i) follows from the convexity of the norm and  Jensen's inequality,   (ii) from sub-multiplicativity of operator norm $\| A B \| \leq \| A \| \| B \|$, (iii)-(iv) from Cauchy inequalities and (v) from $\| A' \|=\| A \|$.  Likewise,
   \begin{align}
     \| \Ep \gamma(X) \Ep \gamma(\bar X)' \| &\leq^{i} \| \Ep \gamma(X) \| \| \Ep \gamma(\bar X)' \| \nonumber \\
     &\leq^{ii} 1/2 ( (\| \Ep \gamma(X) \|)^2 + (\| \Ep \gamma(\bar X)' \|)^2) \nonumber \\
     &\leq^{iii} 1/2 ( \Ep \| \gamma(X) \|^2 + \Ep \| \gamma(\bar X)' \|^2)=^{iv} 1/2 ( \Ep \| \gamma(X) \|^2 + \Ep \| \gamma(\bar X) \|^2). \label{eq:xxbar2}, 
  \end{align}
where (i) follows from $\| A B \| \leq \| A \| \| B \|$, (ii) from Cauchy inequality, (iii) from the convexity of composition $t \rightarrow t^2$ and $\cdot \rightarrow \| \cdot \|$ and Jensen's inequality and (iv) from $\| A' \|= \| A \|$.
Finally, since the RHS of \eqref{eq:xxbar1} and \eqref{eq:xxbar2} is invariant under transposition, 
the same bound holds for the transposed quantities:
 \begin{align*}
\max ( \|     \Ep \gamma(X)' \gamma(\bar X) \|, \| \Ep \gamma(X) \gamma(\bar X)' \|) \leq 1/2 ( \Ep \| \gamma(X) \|^2 + \Ep \| \gamma(\bar X) \|^2) \\
\max ( \|     \Ep \gamma(X)' \Ep \gamma(\bar X) \|, \| \Ep \gamma(X) \Ep \gamma(\bar X)' \|) \leq 1/2 ( \Ep \| \gamma(X) \|^2 + \Ep \| \gamma(\bar X) \|^2).
 \end{align*}

\textbf{Step 2.}  For $\phi(X) = \gamma(X) - \Ep \gamma(X)$, 
\begin{align}
\label{eq:cov}
    \Ep \phi (X) \phi (\bar X)' &= \Ep \gamma(X)  \gamma (\bar X)' - \Ep \gamma(X) \Ep \gamma(\bar X)' -\Ep \gamma(X) \Ep \gamma (\bar X)' + \Ep \gamma(X) \Ep \gamma(\bar X)' \\
    &=  \Ep \gamma(X)  \gamma (\bar X)' - \Ep \gamma(X) \Ep \gamma(\bar X)'. \nonumber
\end{align}
Let $\{ X_{mz} \}_{m,z=1}^{M,Z}$ be a double-indexed sequence. For every value of $m$,
$$
 (\sum_{z=1}^{Z} \gamma(X_{mz}) ) (\sum_{z'=1}^{Z} \gamma(X_{mz'}) )'=\sum_{1 \leq z,z' \leq Z}  \gamma(X_{mz})  \gamma(X_{mz'})'.
$$
Define
\begin{align*}
    M_1:= \Ep \sum_{m=1}^M (\sum_{z=1}^{Z} \gamma(X_{mz}) ) (\sum_{z=1}^{Z} \gamma(X_{mz'}) )' = \sum_{m=1}^M \sum_{1 \leq z,z' \leq Z} \Ep  \gamma(X_{mz})  \gamma(X_{mz'})'  \\
    M_2:=\sum_{m=1}^M (\Ep \sum_{z=1}^{Z} \gamma(X_{mz}) ) (\Ep \sum_{z=1}^{Z} \gamma(X_{mz'}) )' = \sum_{m=1}^M \sum_{1 \leq z,z' \leq Z} \Ep  \gamma(X_{mz})  \Ep \gamma(X_{mz'})' 
\end{align*}
Invoking \eqref{eq:cov} gives 
\begin{align*}
     \sum_{m=1}^M \Ep [\sum_{z=1}^{Z} \phi(X_{mz})][ \sum_{z=1}^{Z} \phi(X_{mz'})]' = M_1 - M_2.
\end{align*}

\textbf{Step 3.} The bound on  $\| M_1 \|$ is
\begin{align}
\| M_1 \| &\leq  \sum_{m=1}^M \sum_{1 \leq z,z' \leq Z} \| \Ep \gamma(X_{mz})  \gamma(X_{mz'})' \| \leq 1/2 \sum_{m=1}^M \sum_{z=1}^Z \sum_{z'=1}^Z   (\Ep \|  \gamma(X_{mz})   \|^2 + \Ep \|  \gamma(X_{mz'})   \|^2)  \nonumber \\
&= Z/2( \sum_{m=1}^M \sum_{z=1}^Z \Ep \|  \gamma(X_{mz})   \|^2 +  \sum_{m=1}^M \sum_{z'=1}^Z \Ep \|  \gamma(X_{mz'})   \|^2) = Z \sum_{m=1}^M \sum_{z=1}^Z \Ep \|  \gamma(X_{mz})   \|^2. \label{eq:m1bound}
\end{align}
Likewise,  
\begin{align}
\| M_2 \| &\leq  \sum_{m=1}^M \sum_{1 \leq z,z' \leq Z} \| \Ep \gamma(X_{mz})  \Ep \gamma(X_{mz'})' \| \nonumber \\
&\leq 1/2 \sum_{m=1}^M \sum_{z=1}^Z \sum_{z'=1}^Z   (\Ep \|  \gamma(X_{mz})   \|^2 + \Ep \|  \gamma(X_{mz'})   \|^2) \nonumber \\
&= Z \sum_{m=1}^M \sum_{z=1}^Z \Ep \|  \gamma(X_{mz})   \|^2. \label{eq:m2bound}
\end{align}
As a result, 
\begin{align*}
    \| M_1 - M_2 \| \leq \| M_1 \| + \| M_2 \| \leq 2 Z \sum_{m=1}^M \sum_{z=1}^Z \Ep \|  \gamma(X_{mz})   \|^2.
\end{align*}
Because the bounds \eqref{eq:xxbar1} and \eqref{eq:xxbar2} are invariant to transpositions of $\gamma(X)$ and/or $\gamma(\bar X)$, 
\begin{align}
\label{eq:ineqgeneral}
  \left \| \sum_{m=1}^M \Ep [\sum_{z=1}^{Z} \phi(X_{mz})'][ \sum_{z=1}^{Z} \phi(X_{mz'})] \right \|  \leq 2 Z \sum_{m=1}^M \sum_{z=1}^Z \Ep \|  \gamma(X_{mz})   \|^2.
\end{align}

\textbf{Step 4.} We first verify the condition \eqref{eq:ineq1} for the odd-numbered full-sized blocks. We note that the L.H.S of \eqref{eq:ineq1} is a special case of the L.H.S of \eqref{eq:ineqgeneral} with 
\begin{align*}
m=m(i,l) &= L \cdot (i-1)+l, \quad M = NL, \quad Z=q \\  X_{mz}:&= X_{i,(2l-2)q+z}\\
\phi (B_{i(2l-1)}) &= \sum_{t=(2l-2)q+1}^{t=(2l-2)q+q} \phi (X_{it})= \sum_{z=1}^q \phi (X_{mz}).
\end{align*}
As a result, 
\begin{align}
  \left \|  \sum_{i=1}^N \sum_{l=1}^L   \Ep \phi (B^{*}_{i(2l-1)}) \phi (B^{*}_{i(2l-1)})'  \right \|  & \leq 2 q \sum_{i=1}^N \sum_{l=1}^{L} \sum_{z=1}^q \Ep \|  \gamma(X^{*}_{i (2l-1),z})   \|^2 \\
  &=  2 q \sum_{i=1}^N \sum_{l=1}^{L} \sum_{z=1}^q \Ep \|  \gamma(X_{i (2l-1),z})   \|^2 \\
  & \leq 2q \sum_{i=1}^N \sum_{t=1}^T \Ep \|  \gamma(X_{i t})   \|^2  
  = 2q NT \gamma^2_{2NT}.
  \label{eq:ineqgeneral3}
\end{align}
A similar argument for even-numbered full-sized blocks and $\phi (B^{*}_{i(2l)})=\sum_{t=(2l-1)q+1}^{t=(2l)q} \phi(X^{*}_{it})$ verifies condition \eqref{eq:ineq2} of Lemma \ref{lem:matrixberbeeweakdep}. Finally, if the remainder block is non-empty, i.e., $T-2Lq \neq 0$, invoking \eqref{eq:ineqgeneral} with 
$$
m=i, \quad M=N, \quad Z:=T-2Lq
$$
and noting that
\begin{align}
\label{eq:ineqgeneral2}
  \left \|  \sum_{i=1}^N   \Ep \phi (B_{ir}) \phi (B_{ir})'  \right \|  \leq 2 q \sum_{i=1}^N \sum_{z=1}^{T-2Lq} \Ep \|  \gamma(X_{i z})   \|^2 \leq 2q \sum_{i=1}^N \sum_{t=1}^T \Ep \|  \gamma(X_{i t})   \|^2 = 2q NT \gamma^2_{2NT},
\end{align}
which verifies condition \eqref{eq:ineq3} of Lemma \ref{lem:matrixberbeeweakdep}. Finally, the condition \eqref{eq:infcond1} follows from 
$$
\|  \phi (B^{*}_{i(2l-1)}) \| \leq qR \text{ a.s. }, \|  \phi (B^{*}_{i(2l)}) \| \leq qR \text{ a.s. }, \| \phi (B_{ir})  \| \leq q R,
$$
since each block has at most $q$ summands. Plugging $R= 2 \gamma^{\infty}_{NT}$ and $q=2\gamma^2_{2NT}$   into \eqref{eq:meanweakdeprate} gives \eqref{eq:meanweakdeprate1}.
\end{proof}

Corollaries \ref{lem:covmatmom} and \ref{lem:matrixmoment2} are special cases of Lemma \ref{lem:covmatmompre}   with various cases of $\gamma$-function.

\begin{corollary}[Covariance Matrix Moments]
\label{lem:covmatmom}
Let $ \psi(X): \mathcal{X}  \rightarrow \mathrm{R}^{d \bigtimes 1}$ be a fixed vector function of a random vector $X$. Define
\begin{align}
\label{eq:ksix}
\gamma(X)&=  \psi(X)   \psi(X)'
\end{align}
and the $\phi$-function
\begin{align*}
    \phi(X):=\gamma(X) - \Ep [\gamma(X)] = \psi(X)   \psi(X)'- \Ep[  \psi(X)   \psi(X)' ].
\end{align*}
Let the numeric sequences $\psi^{\infty}_{NT}$ and $\psi_{4NT}$ obey
\begin{align}
    \sup_{it} \|    \psi(X_{it})  \|  \leq \psi^{\infty}_{NT} \text{ a.s. } \label{eq:infcondpsi} \\
    (NT)^{-1} \sum_{i=1}^N \sum_{t=1}^T \Ep\|   \psi(X_{it})  \|^4  \leq \psi^4_{4NT} \label{eq:4powercond}
\end{align}
Then, the bound \eqref{eq:infcond} hold with
$\gamma^{\infty}_{NT}:=(\psi^{\infty}_{NT})^2$ and $\gamma^2_{2NT}:=\psi^4_{4NT}$.   As a result, the rate \eqref{eq:meanweakdeprate1} reduces to
\begin{align}
\label{eq:meanweakdeprate2}
    \| (NT)^{-1} \sum_{i=1}^N \sum_{t=1}^T \phi (X_{it}) \| \lesssim_P  \sqrt{\psi^4_{4NT} \log (NT) \log (2d)/NT} + 
\log (NT) (\psi^{\infty}_{NT})^2 \log (2d)/ NT.
\end{align}
  
\end{corollary}

\begin{proof}[Proof of Corollary \ref{lem:covmatmom}]
Noting that 
$$
\| \gamma(X_{it}) \|_{\infty} \leq \| \psi(X_{it})\|_{\infty}^2 \leq (\psi^{\infty}_{NT})^2
$$
and
$$
\| \gamma(X_{it}) \|^2 = \| \psi(X_{it}) \psi(X_{it}) ' \|^2 \leq \| \psi(X_{it}) \|^2 \| \psi(X_{it}) '\|^2 = \| \psi(X_{it}) \|^4
$$
Therefore,
\begin{align*}
   (NT)^{-1}   \sum_{i=1}^N \sum_{t=1}^T \Ep \| \gamma(X_{it}) \|^2 \leq (NT)^{-1}  \sum_{i=1}^N \sum_{t=1}^T \Ep \| \psi(X_{it}) \|^4 \leq  \psi^4_{4NT}.
\end{align*}
Application of Lemma  \ref{lem:covmatmompre} yields the result.
\end{proof}

\begin{corollary}[Product Moments]
\label{lem:matrixmoment2}
Let $ \psi(X): \mathcal{X}  \rightarrow \mathrm{R}^{d \bigtimes 1}$ be a fixed vector function of a random vector $X$ and $\xi(X)$ be a random variable. Define  
\begin{align*}
 \gamma(X):=\psi(X) \cdot \xi(X)
\end{align*}
Let the numeric sequences $\psi^{\infty}_{NT}, \xi^{\infty}_{NT}$ and $\psi_{4NT}, \xi_{4NT}$ obey 
\begin{align}
    \sup_{it} \|    \psi(X_{it})  \|  &\leq \psi^{\infty}_{NT} \text{ a.s. },  \quad  \sup_{it} |    \xi(X_{it})  |  \leq \xi^{\infty}_{NT} \text{ a.s. } \label{eq:infcondl} \\
    (NT)^{-1} \sum_{i=1}^N \sum_{t=1}^T \Ep\|   \psi(X_{it})  \|^4  &\leq \psi^4_{4NT}, \quad 
    (NT)^{-1} \sum_{i=1}^N \sum_{t=1}^T \Ep   \xi^4(X_{it})    \leq \xi^4_{4NT}. \label{eq:4powercondl} 
\end{align}
Then, the bound \eqref{eq:infcond} holds with
$$\gamma^{\infty}_{NT}:=(\psi^{\infty}_{NT})\cdot \xi^{\infty}_{NT}, \quad \gamma^2_{2NT}:=1/2(\psi^4_{4NT} + \xi^4_{4NT}).$$   As a result, the rate \eqref{eq:meanweakdeprate1} reduces to
\begin{align}
\label{eq:meanweakdeprate3}
    \| (NT)^{-1} \sum_{i=1}^N \sum_{t=1}^T \phi (X_{it}) \| & \lesssim_P  \sqrt{ (\psi^4_{4NT} + \xi^4_{4NT}) \log (NT) \log (d+1)/NT} \\
    & + 
\log (NT) \psi^{\infty}_{NT} \xi^{\infty}_{NT} \log (d+1)/ NT. \nonumber 
\end{align}

\end{corollary}
\begin{proof}[Proof of Corollary \ref{lem:matrixmoment2} ]
Noting that 
$$
\| \gamma(X_{it}) \|_{\infty} \leq \| \psi(X_{it})\|_{\infty} | \xi(X_{it}) | \leq \psi^{\infty}_{NT} \xi^{\infty}_{NT}
$$
and
$$
\| \gamma(X_{it}) \|^2 = \| \psi(X_{it})  \|^2  \xi^2(X_{it}).
$$
Cauchy inequality gives
$$
\Ep \| \gamma(X_{it}) \|^2 = \Ep \| \psi(X_{it})  \|^2  \xi^2(X_{it}) \leq 1/2 (\Ep \| \psi(X_{it})  \|^4 + \Ep  \xi^4(X_{it}) ).
$$
Therefore,
\begin{align*}
  (NT)^{-1}  \sum_{i=1}^N \sum_{t=1}^T \Ep \| \gamma(X_{it}) \|^2 \leq      (NT)^{-1} (\sum_{i=1}^N \sum_{t=1}^T \Ep \| \psi(X_{it})  \|^4  + 
    \sum_{i=1}^N \sum_{t=1}^T \Ep \xi^4 (X_{it})) \leq (\psi^4_{4NT} + \xi^4_{4NT})/2.
\end{align*}
Application of Lemma  \ref{lem:covmatmompre} yields the result.
\end{proof}

\setcounter{definition}{0}    
\setcounter{section}{0}    
\setcounter{lemma}{0}    
\setcounter{equation}{0}
\setcounter{remark}{0}
\setcounter{theorem}{0}
\renewcommand{\theequation}{G.\arabic{equation}}
\renewcommand{\thelemma}{G.\arabic{lemma}}
\renewcommand{\thecorollary}{G.\arabic{corollary}}
\renewcommand{\thetable}{G.\arabic{table}}
\renewcommand{\theremark}{G.\arabic{remark}}
\renewcommand{\thetheorem}{G.\arabic{theorem}}

\renewcommand{\theassumption}{G.\arabic{assumption}}
\renewcommand{\thesection}{G}
 
\section{Additional Results on Orthogonal OLS}
\label{sec:lowd}

\begin{assumption}[Tail Bound on Empirical Covariance Matrix in $\ell_2$ norm]
\label{ass:growth}
For some sequence $v_{NT}=o(1)$, in the regime where $d \to \infty$, we have that 
\begin{align}\label{eq:vnt}
\| \widetilde{Q}- Q \| \lesssim_P v_{NT}.
\end{align}
\end{assumption}

\begin{remark}
Suppose Assumptions \ref{ass:sampling}--\ref{ass:subgauss} hold and $\sup_{it} \| V_{it} \|_{\infty} \leq R \text{ a.s. }$ and 
$$
\max_{itj} \Ep V^4_{itj} \leq \sigma^4_{4V}
$$
We invoke Corollary \ref{lem:covmatmom} with $\psi(W_{it})=V_{it}$ and $\psi^{\infty}_{NT}:=\sqrt{d} R$ and  \eqref{eq:4powercond} with $\psi^4_{4NT} = d^2  \sigma^4_{4V}$. As a result, the rate bound \eqref{eq:meanweakdeprate2} reduces to
\begin{align*}
    v_{NT} =  \sqrt{d^2 \log (2d) \log (NT) /NT} + d R \log (2d) \log (NT)/NT.
\end{align*}
Further improvement of this rate may be possible under additional structure on $V_{it}$,  see, e.g., Theorem 1 and Corollary 3 in \cite{MerlevedeYoussef}.
\end{remark}

Let $D_{NT}\times L_{NT}$ be a sequence of realization sets such that the following conditions hold.    Let $\textbf{d}_{NT}, \textbf{l}_{NT},  \textbf{d}_{NT,4},  \textbf{l}_{NT,4}$ be the numeric sequences obeying the following bounds 
  \begin{align*}
    \sup_{\mathbf{d} \in D_{NT}} (NT)^{-1} \sum_{i=1}^N \sum_{t=1}^T (\Ep  \|  d_i(X_{it}) - d_{i0}(X_{it}) \|^2 )^{1/2} \leq     \textbf{d}_{NT}, \\
     \sup_{\mathbf{d} \in D_{NT}} (NT)^{-1} \sum_{i=1}^N \sum_{t=1}^T (\Ep  \|  d_i(X_{it}) - d_{i0}(X_{it}) \|^4 )^{1/4} \leq    \textbf{d}_{NT,4}  \\
    \sup_{\mathbf{l} \in L_{NT}}    (NT)^{-1} \sum_{i=1}^N \sum_{t=1}^T  (\Ep ( l_i(X_{it}) - l_{i0}(X_{it}) )^4 )^{1/4} \leq  \textbf{l}_{NT,4}.
 \end{align*}
Define the following rates
\begin{align}
r_{2NT}:&=  \textbf{d}_{NT} \textbf{l}_{NT} + \sqrt{\dfrac{  (\textbf{d}^4_{NT,4} + \textbf{l}^4_{NT,4}) \log (NT) \log (d+1) }{NT}} + \sqrt{d}  \log (d+1) \log (NT)/NT \\
\chi_{NT}:&= \textbf{d}_{NT}^2  +  \sqrt{\textbf{d}^4_{NT,4} \log (2d) \log (NT)/NT} + d \log (2d) \log (NT)/NT
\end{align}

\begin{assumption}
\label{ass:smallbiasOLS}
We suppose that the true parameter vector has bounded $\ell_2$-norm: $$\|\beta_0\|_2 \leq \bar C_\beta$$ for some finite constant $\bar C_\beta$;  We suppose that the reduced form estimators obey: $\widehat{\mathbf{l}}(\cdot) \in L_{NT}$ and $\widehat{\mathbf{d}}(\cdot) \in D_{NT}$ if such that
$\textbf{d}_{NT}$,  $\textbf{d}_{NT,4}$, $\textbf{l}_{NT}, \textbf{l}_{NT,4}$  decay sufficiently fast:
\begin{equation}
r_{2NT} + \chi_{NT}= o ( (NT)^{-1/2})
\end{equation}
Furthermore, the reduced form estimates are bounded as
\begin{align*}
   & \sup_{\mathbf{d} \in D_{NT}} \| d_i (X_{it}) \| \leq \sqrt{d} D, \quad \sup_{\mathbf{l} \in L_{NT}} |l_i (X_{it})| \leq L \quad \forall i.
\end{align*}
\end{assumption}

\begin{theorem}[Orthogonal Least Squares] Suppose Assumptions \ref{ass:sampling}--\ref{ass:subgauss}, \ref{ass:growth} and \ref{ass:smallbiasOLS} hold. Then, the following statements hold. 
	\begin{enumerate}
	\item The  Orthogonal Least Squares estimator converges at the rate $\sqrt{d/NT}$:
	
     \begin{align}   \label{eq:olsrate} \| \widehat{\beta}_{OLS}- \beta_0 \|_{2}  \lesssim_{P}  \sqrt{d/NT}. \end{align}
     
    \item    For any  deterministic sequence $\{\alpha\}=\{\alpha_{N,T}\}$ with $\| \alpha_{N,T}\| =1$, the estimator $\alpha' \widehat{\beta}_{OLS}$ of $\alpha'\beta_0$ is asymptotically linear:
        \begin{align}
        \label{eq:olspointwise}
       \sqrt{NT} \alpha' (\widehat{\beta}_{OLS}- \beta_0)  = \alpha' Q^{-1}  \G_{NT} V_{it} U_{it} + o_P(1),
    \end{align}

	\item If the Lindeberg condition holds for each $M>0$: $$\limsup_{NT \to \infty}\sup_{\|\alpha\|_2 =1}  (NT)^{-1} \sum_{i=1}^N \sum_{t=1}^T \Ep [(\alpha' V_{it} U_{it})^2  1 \{  |\alpha' V_{it} U_{it}|   > M \sqrt{NT}\}  = 0,$$
then the Orthogonal Least Squares estimator is asymptotically Gaussian:
	\begin{align}
        \lim_{NT \rightarrow \infty} \sup_{\| \alpha\|_2 =1}\sup_{t \in \mathrm{R}} \left| \Pr \left(\frac{\sqrt{NT} \alpha' (\widehat{\beta}_{OLS} - \beta_0)}{\| \alpha' \Sigma \|^{1/2}}  < t \right) - \Phi(t) \right| = 0.
	\end{align}

	\end{enumerate}
	\label{thrm:orthools}
\end{theorem}

\begin{lemma}[First-Order Terms, $\ell_2$-norm]
\label{lem:errortermsel2}
Let $\bar{a}, \bar{m}, \bar{f}, \bar{e}$ be as in \eqref{eq:ak}--\eqref{eq:ek}. Under Assumptions \ref{ass:sampling}--\ref{ass:subgauss}, the following bounds hold
\begin{align}
\| \bar{a} \|&\lesssim_P  \left(  \sqrt{d/NT} \textbf{d}_{NT} \right)  \label{eq:akproof}\\
\| \bar{m} \|&\lesssim_P  \left( \sqrt{d/NT} \textbf{l}_{NT} \right) \label{eq:mkproof}\\
\| \bar{f} \|&\lesssim_P (\textbf{d}_{NT} (NT)^{-1/2}) \label{eq:fkproof} \\
\| \bar{e} \|&\lesssim_P  (  \sqrt{d/NT} (\textbf{d}_{NT}  + \textbf{l}_{NT})  ) \label{eq:ekproof}
\end{align}
\end{lemma}

\begin{proof}[Proof of Lemma \ref{lem:errortermsel2}]
Define 
\begin{align*}
    \xi^V_{NT}:=\sqrt{d/NT} \textbf{d}_{NT}, \quad \xi^B_{NT} = 0,
\end{align*}
and the $A$-function as
\begin{align}
    A(W_{it}, \eta) = V_{it} (d_{i0}(X_{it}) - d_i(X_{it})).
\end{align}
Define $B_{Ak}(\eta)$ and $V_{Ak}(\eta)$ with $\eta = \mathbf{d}$ as in \eqref{eq:genericbias}--\eqref{eq:genericdem}.  Consider any $\eta = \eta_{NT} \in D_{NT}$ in what follows. Since $V_{it}$ obeys the martingale difference property \eqref{eq:mdsvit}, it follows that $\| B_{Ak} (\eta_{NT}) \| =0$. Furthermore, for any $1 \leq j,j' \leq d$,  
\begin{align}
\label{eq:mdsel2}
&\Ep [ (\alpha' V_{it}) (\alpha' V_{(it')} ) (d_{i0}(X_{it}) - d_i(X_{it}))_{j}  (d_{i0}(X_{it'}) - d_i(X_{it'}))_{j'} ] =0.
\end{align}
Combining \eqref{eq:mdsel2} and Assumption \ref{ass:subgauss} 

\begin{align*}
&\Ep [\| \alpha' V_{Ak} (\eta_{NT})  \|^2  ] =^{i} (NT_k)^{-2} \sum_{i=1}^N \sum_{t \in \mathcal{M}_k} \sum_{j=1}^d \Ep [(\alpha' V_{it})^2 (d_{i0}(X_{it}) - d_i(X_{it}))_{j}^2 ]\\
&\leq  (NT_k)^{-2}\sup_{\mathbf{d} \in D_{NT}} \sum_{i=1}^N   \sum_{t \in \mathcal{M}_k}  \sum_{j=1}^d \Ep \bigg[ \Ep [\| V_{it} \|^2 | \Phi_{it}, X_{it}  ]  (d_{i0}(X_{it}) - d_i(X_{it}))_{j}^2 \bigg] \\
 &\leq  (NT_k)^{-2}\sup_{\mathbf{d} \in D_{NT}} \sum_{i=1}^N \sum_{t=1}^T \Ep  \| d_{i0}(X_{it}) - d_i(X_{it}) \|^2  d  \sigma_V^2 \\
&\leq^{ii} (d/NT_k) \sigma_V^2 (T/T_k) \textbf{d}_{NT}^2,
\end{align*}
where $(i)$ follows from \eqref{eq:mdsel2} and (ii) follows from definition of $\mathbf{d}_{NT}$. By Assumption \ref{ass:smallbiashds}, we have that
$\Pr (\widehat{\mathbf{d}}_k \in D_{NT}, \ \forall k=1,...,K) \to 1$. Moreover, since the number of cross-fit folds is finite, the size $T_k$ of each fold obeys $$1 \lesssim T_k/T \leq  1.$$ We conclude by Lemma \ref{cor:remterms} that  \eqref{eq:akproof} holds.  
Repeating the same argument for 
$$A(W_{it}, \eta) = V_{it} (l_{i0}(X_{it}) - l_i(X_{it}))
\text { and } A(W_{it}, \eta) = U_{it} (d_{i0}(X_{it}) - d_i(X_{it}))$$ 
establishes claims \eqref{eq:mkproof} and \eqref{eq:fkproof}.  Finally, \eqref{eq:ekproofinf} holds by definition of $\bar{e}=\bar{m}-\bar{a}'\beta_0$ and Holder inequalities.

\end{proof}

In the Lemma below, abusing the notation, we treat $l_i$ as some generic vector-valued function.
\begin{lemma}[Second-Order Bias]
\label{lem:errortermsel22bias}
Let $d_{i0}(X_{it})$ be a $d_1$-vector and $l_{i0}(X_{it})$ be a $d_2$-vector. Suppose that
\begin{align*}
     \sup_{\mathbf{d} \in D_{NT}} (NT)^{-1} \sum_{i=1}^N \sum_{t=1}^T (\Ep  \|  d_i(X_{it}) - d_{i0}(X_{it}) \|^2 )^{1/2} \leq     \textbf{d}_{NT}, \\
       \sup_{\mathbf{l} \in D_{NT}} (NT)^{-1} \sum_{i=1}^N \sum_{t=1}^T (\Ep  \|  l_i(X_{it}) - l_{i0}(X_{it}) \|^2 )^{1/2} \leq     \textbf{l}_{NT}
\end{align*}
Consider the $A$-function as
\begin{align}
    A(W_{it}, \eta) = (d_{i0}(X_{it}) - d_i(X_{it})) (l_{i0}(X_{it}) - l_i(X_{it})), \quad \eta = (\mathbf{d}, \mathbf{l})
\end{align}
and its bias $B_{Ak}(\eta)$ as in  \eqref{eq:genericbias}. Then, we have the bias bound:
\begin{align*}
   \sup_{ \eta \in (D_{NT}, L_{NT})} \| B_{Ak}(\eta) \|_2 \leq \textbf{d}_{NT} \textbf{l}_{NT} (T/T_k).
\end{align*}

\end{lemma}

\begin{proof}[Proof of Lemma \ref{lem:errortermsel22bias}]
Take $\alpha \in \mathcal{S}^{d-1}$. Let 
$X_{it}(\alpha):=\alpha' (d_{i0}(X_{it}) - d_i(X_{it}))$ and $Y_{itj}:=(l_{i0}(X_{it}) - l_i(X_{it}))_j$ and $$a^2_{it}=\| d_{i0}(X_{it}) - d_i(X_{it}) \|^2, \quad b^2_{it} = \| l_{i0}(X_{it}) - l_i(X_{it}) \|^2 = \sum_{j=1}^d b^2_{itj} $$ Recognize that 
\begin{align*}
 \alpha' B_{Akj}(\eta) &=  (NT_k)^{-1} \sum_{i=1}^N \sum_{t \in {\mathcal{M}}_k}  \Ep  \alpha'(d_{i0}(X_{it}) - d_i(X_{it}))  (l_{i0}(X_{it}) - l_i(X_{it}))_j \\
 &= (NT_k)^{-1} \sum_{i=1}^N \sum_{t \in {\mathcal{M}}_k}  \Ep X_{it} (\alpha) Y_{itj}.
\end{align*}
 Cauchy inequality gives
\begin{align*}
    |\Ep  X_{it} (\alpha) Y_{itj} | &\leq \sqrt{ \Ep X^2_{it} (\alpha) \Ep Y^2_{itj}} \leq \sqrt{ \Ep \| d_{i0}(X_{it}) - d_i(X_{it}) \|^2 \Ep Y^2_{itj}} =: \sqrt {a^2_{it} b^2_{itj}}.
\end{align*}
Summing over $i$ and $t$ and invoking Cauchy inequality gives $$  \alpha' B_{Akj}(\eta) \leq (NT_k)^{-1} \sum_{i=1}^N \sum_{t \in {\mathcal{M}}_k}  \sqrt{ a^2_{it} b^2_{itj} } \leq (NT_k)^{-1} \sqrt{ \sum_{i=1}^N \sum_{t \in {\mathcal{M}}_k} a^2_{it} } \sqrt{ \sum_{i=1}^N \sum_{t \in {\mathcal{M}}_k} b^2_{itj}  }.$$
\begin{align*}
 \| \alpha' B_{Ak}(\eta) \|^2 &=    \sum_{j=1}^d | \alpha' B_{Akj}(\eta)|^2  \leq (NT_k)^{-1} (\sum_{i=1}^N \sum_{t \in {\mathcal{M}}_k} a^2_{it}) (NT_k)^{-1} ( \sum_{j=1}^d  \sum_{i=1}^N \sum_{t \in {\mathcal{M}}_k} b^2_{itj}) \\
 &\leq (NT_k)^{-1} (\sum_{i=1}^N \sum_{t=1}^T a^2_{it}) (NT_k)^{-1} (  \sum_{i=1}^N \sum_{t=1}^T b^2_{it}) \\
 &\leq  \textbf{l}^2_{NT} \textbf{d}^2_{NT} (T/T_k)^2.
\end{align*}

\end{proof}

Next, we invoke Lemmas \ref{lem:matrixberbeeweakdep} and \ref{lem:covmatmompre} and Corollary \ref{lem:covmatmom}-\ref{lem:matrixmoment2} from Appendix \ref{sec:toolslows}. 

\begin{lemma}[Second-Order Covariance Term]
\label{lem:errortermsel22}
Define
$$\zeta^B_{NT}=\textbf{d}_{NT}^2, \quad  \zeta^V_{NT}=\sqrt{ \textbf{d}^4_{NT,4} \log (2d) \log NT/NT } + d D \log (2d) \log (NT)/NT.$$
Under Assumptions \ref{ass:sampling}--\ref{ass:subgauss}, the following bounds hold for the term $\bar b$ defined in \eqref{eq:bk}
\begin{align}
\| \bar{b} \|& \lesssim_P \zeta^B_{NT}+ \zeta^V_{NT} = \chi_{NT}. \label{eq:bkproof} 
\end{align}

\end{lemma}

\begin{proof}[Proof of Lemma \ref{lem:errortermsel22}]
Define the $A$-function as
$$A (W_{it},\eta) = (d_{i0}(X_{it}) - d_i(X_{it}))(d_{i0}(X_{it}) - d_i(X_{it}))', \quad \eta = \mathbf{d} = \mathbf{l}.$$ 
Let  $B_{Ak}(\eta)$ and $V_{Ak}(\eta)$ be defined according to \eqref{eq:genericbias}--\eqref{eq:genericdem}.
Invoking Lemma \ref{lem:errortermsel22bias} with $\mathbf{l}=\mathbf{d}$ gives   $\| B_{Ak} (\eta_{NT}) \|_{\infty} = O(
\zeta^B_{NT})$ for any partition $k$. Note that
\begin{align*}
    V_{Ak}(\eta_{NT}) &= (NT_k)^{-1} \sum_{i=1}^N \sum_{t \in \mathcal{M}_k} \bigg( (d_{i0}(X_{it}) - d_i(X_{it})) (d_{i0}(X_{it}) - d_i(X_{it}))'\\
    &- \Ep [(d_{i0}(X_{it}) - d_i(X_{it})) (d_{i0}(X_{it}) - d_i(X_{it}))']    \bigg) \\
    &=: (NT_k)^{-1} \sum_{i=1}^N \sum_{t \in \mathcal{M}_k} \phi_i (X_{it}). 
\end{align*}
Define
\begin{align*}
\psi_i (X_{it}) &= (d_{i0}(X_{it}) - d_i(X_{it})) \\
\gamma_i (X_{it}) &= \psi_i (X_{it})\psi_i (X_{it})'= (d_{i0}(X_{it}) - d_i(X_{it}))(d_{i0}(X_{it}) - d_i(X_{it}))' \\
\phi_i (X_{it}) &= \gamma_i (X_{it})-\Ep [ \gamma_i (X_{it})].
\end{align*}
Note that $\psi_{i}(X_{it})=(d_{i0}(X_{it}) - d_i(X_{it}))$ obeys the conditions \eqref{eq:infcondpsi} and \eqref{eq:4powercond}  with
$$ \psi_{NT}^{\infty}:=\sqrt{d} D, \quad  \psi_{NT,4}:= \mathbf{d}_{NT,4}.$$
As a result, the bound \eqref{eq:meanweakdeprate2} reduces to $\zeta^V_{NT}$ for each partition $k$ and $T=T_k$. Since $T_k /T \asymp 1$, the bound follows. 

\end{proof}

\begin{lemma}[Second-Order Covariance Term, cont.]
\label{lem:errortermsel222}
Suppose Assumptions \ref{ass:sampling}--\ref{ass:subgauss} and \ref{ass:smallbiasOLS} hold. Let $\bar z$ and $\bar g$ be as defined in \eqref{eq:zk} and \eqref{eq:gk}. Then,
\begin{align}
\| \bar{z} \|&\lesssim_{P} r_{2NT} \label{eq:zkproof}  \\
\| \bar{g} \|& \lesssim_P r_{2NT} + \chi_{NT}. \label{eq:gkproof} 
\end{align}
\end{lemma}

\begin{proof}[Proof of Lemma \ref{lem:errortermsel222}]
Define the $A$-function as
$$A (W_{it},\eta) = (d_{i0}(X_{it}) - d_i(X_{it}))(l_{i0}(X_{it}) - l_i(X_{it})), \quad \eta = (\mathbf{d}, \mathbf{l}).$$ 
Let  $B_{Ak}(\eta)$ and $V_{Ak}(\eta)$ be defined according to \eqref{eq:genericbias}--\eqref{eq:genericdem}. Let $$\zeta^B_{NT}=\textbf{d}_{NT} \textbf{l}_{NT}, \quad \zeta^V_{NT}= \sqrt{\dfrac{  (\textbf{d}^4_{NT,4} + \textbf{l}^4_{NT,4} ) \log (NT) \log (d+1) }{NT}} + \sqrt{d} D \log (2d) \log (NT)/NT.$$
Invoking Lemma \ref{lem:errortermsel22bias} with $d_1=d$ and $d_2=1$ gives   $\| B_{Ak} (\eta_{NT}) \|_{\infty} = O(\zeta^B_{NT})$ for any partition $k$. Note that
\begin{align*}
    V_{Ak}(\eta_{NT}) &= (NT_k)^{-1} \sum_{i=1}^N \sum_{t \in \mathcal{M}_k} \bigg( (d_{i0}(X_{it}) - d_i(X_{it})) (l_{i0}(X_{it}) - l_i(X_{it}))\\
    &- \Ep [(d_{i0}(X_{it}) - d_i(X_{it})) (l_{i0}(X_{it}) - l_i(X_{it}))']    \bigg) \\
    &=: (NT_k)^{-1} \sum_{i=1}^N \sum_{t \in \mathcal{M}_k} \phi_i (X_{it}). 
\end{align*}
Define
\begin{align*}
\psi_i (X_{it}) &= (d_{i0}(X_{it}) - d_i(X_{it})), \quad \xi_i (X_{it}) = l_{i0}(X_{it}) - l_i(X_{it}) \\
\gamma_i (X_{it}) &= \psi_i (X_{it})\xi_i (X_{it})= (d_{i0}(X_{it}) - d_i(X_{it}))(l_{i0}(X_{it}) - l_i(X_{it})) \\
\phi_i (X_{it}) &= \gamma_i (X_{it})-\Ep [ \gamma_i (X_{it})].
\end{align*}
Note that $\psi_{i}(X_{it})=(d_{i0}(X_{it}) - d_i(X_{it}))$ and  $\xi_i (X_{it}) = l_{i0}(X_{it}) - l_i(X_{it})$ obey the conditions \eqref{eq:infcondl} and \eqref{eq:4powercondl}  with
\begin{align*} \psi_{NT}^{\infty}:&=\sqrt{d} D, \quad  \psi_{NT,4}:= \mathbf{d}_{NT,4}, \quad \xi_{NT}^{\infty}:=L, \quad  \xi_{NT,4}:= \mathbf{l}_{NT,4}.
\end{align*}
As a result, the bound \eqref{eq:meanweakdeprate3} reduces to $\zeta^V_{NT}$ for each partition $k$ and $T=T_k$. Since $T_k /T \asymp 1$, the bound \eqref{eq:zkproof} follows. Recognizing that $\bar{g} =   \bar{z}-\bar{b}'\beta_0$ and invoking $\| \beta_0 \| \leq C_{\beta}$ as in Assumption \ref{ass:smallbiasOLS} gives
$$
\| \bar{g} \| \leq \| \bar{z} \| +  \| \bar{b} \beta_0 \| \leq \| \bar{z} \| + \| \bar{b} \| \| \beta_0 \|,
$$
\eqref{eq:gkproof} follows.
\end{proof}

\begin{proof}[Proof of Theorem  \ref{thrm:orthools} ]

\textbf{Step 0.} Let $R_{it}(\widehat{d}, \widehat{l})$ be as defined in \eqref{eq:ritalt}. Let $\bar{a}, \bar{b}, \bar{e}, \bar{f},\bar{g}$ be as defined in \eqref{eq:ak}, \eqref{eq:bk}, \dots, \eqref{eq:gk}.  As shown in the proof of Lemma \ref{lem:matrixerrorinf2}, the Gram matrix estimation error
$$
\widehat{Q} - \widetilde{Q}= \mathbb{E}_{NT}\widehat{V}_{it}\widehat{V}_{it}' - \mathbb{E}_{NT} V_{it} V_{it}' = \bar{a} + \bar{a}' + \bar{b}
$$
and gradient estimation error
$$
\widehat{S} - S =  \mathbb{E}_{NT}\widehat{V}_{it} (U_{it} + R_{it}(\widehat{\mathbf{d}}, \widehat{\mathbf{l}})) - V_{it} U_{it}  ]= \bar{e} + \bar{f}+\bar{g}.
$$
We have that $$\| \widehat Q - Q \| \leq \| \widehat Q - \widetilde Q \| + \| \widetilde Q - Q \|  
\lesssim_P^{i} (\chi_{NT}+ \sqrt{d/NT}\mathbf{d}_{NT} + v_{NT}   ) = o(1),$$
where (i) follows from Lemmas \ref{lem:errortermsel2}-\ref{lem:errortermsel222} and Assumption \ref{ass:growth}.Furthermore, by Lemmas 
\begin{align*}
\| \widehat S - S \| = \| \bar{e} + \bar{f} + \bar{g} \| \lesssim_P (r_{2NT} + \chi_{NT}) = o (1/\sqrt{NT}),
 \end{align*}
where  we used Assumptions \ref{ass:sampling}--\ref{ass:subgauss} and \ref{ass:smallbiasOLS} to conclude that  $r_{2NT} + \chi_{NT} = o (1/\sqrt{NT})$.

\textbf{Step 1.}  Since $Q$ is invertible by assumption, $\widehat{Q}$ is also invertible wp $1-o(1)$ by Step 0.  Therefore, we can decompose $ \widehat{\beta}_{OLS}- \beta_0$ as
  \begin{align*}
    \widehat{\beta}_{OLS}- \beta_0 &=  \widehat{Q}^{-1} \mathbb{E}_{NT}[\widehat{V}_{it} \widehat{\widetilde{Y}}_{it} ] -  \widehat{Q}^{-1}  \widehat{Q}'\beta_0  = \widehat{Q}^{-1} \mathbb{E}_{NT}[\widehat{V}_{it} \widehat{\widetilde{Y}}_{it} ] -  \widehat{Q}^{-1}  (\mathbb{E}_{NT}\widehat V_{it} \widehat V_{it}') \beta_0 \\
    &=   \widehat{Q}^{-1}  \mathbb{E}_{NT}[\widehat{V}_{it} (\widehat{\widetilde{Y}}_{it} -  \widehat{V}_{it}'\beta_0)]  \\
    &=  \widehat{Q}^{-1} \mathbb{E}_{NT}[\widehat{V}_{it} (U_{it} + R_{it}(\widehat{\mathbf{d}}, \widehat{\mathbf{l}}))] \\
    &=   \widehat{Q}^{-1} \mathbb{E}_{NT}V_{it} U_{it} +  \widehat{Q}^{-1}  \mathbb{E}_{NT}[\widehat{V}_{it} (U_{it} + R_{it}(\widehat{\mathbf{d}}, \widehat{\mathbf{l}})) - V_{it} U_{it}  ].
      \end{align*}
Therefore the following bound holds by triangle and Holder inequalities:
      \begin{align*}
    \|   \widehat{\beta}_{OLS}- \beta_0  \|  &\leq \| \widehat{Q}^{-1} \| \| \mathbb{E}_{NT}V_{it} U_{it} \| +  \| \widehat{Q}^{-1} \| \| \widehat S -S \| =: \| \widehat{Q}^{-1} \|( L_1 + L_2).
  \end{align*}
The first term $L_1$ is bounded as
 \begin{align*}
\Ep \| \mathbb{E}_{NT}V_{it} U_{it} \|^2 &= \sum_{j=1}^d \Ep (\mathbb{E}_{NT}(V_{it})_j U_{it})^2 \\
&=^{i} (NT)^{-2} \sum_{j=1}^d \sum_{i=1}^N \sum_{t=1}^T  \Ep ((V_{it})_j U_{it})^2 \\
 &\leq  (NT)^{-2} \sum_{i=1}^N \sum_{t=1}^T \Ep \| V_{it} \|^2 \sup_{it} \Ep [U_{it}^2 \mid V_{it}]  \\
 & \leq \bar{\sigma}^2 (NT)^{-1} \sum_{i=1}^N \sum_{t=1}^T \text{trace} (\Ep V_{it} V_{it}') \\
 &= \bar{\sigma}^2  \text{trace} (Q)  \leq^{ii} (d/NT) C_{\text{max}},
 \end{align*}
 where (i) follows from  the m.d.s property in Lemma \ref{lem:mds},  and (ii) from $\max \eig (Q) \leq C_{\text{max}}$. Markov inequality gives $L_1 \lesssim_P (\sqrt{d/NT})$.   The second term $L_2:=\| \widehat S - S \|$ is $o_P (1/\sqrt{NT})$ by Step 0.
 Step 0 implies $\max \eig (\widehat Q^{-1}) < 2C^{-1}_{\text{min}}$ w.p. $1-o(1)$.  Therefore, the rate bound  \eqref{eq:olsrate} follows.

\textbf{ Step 2. }  From Step 1,
 \begin{align*}
\alpha' (\widehat{\beta}_{OLS}- \beta_0 ) &= \alpha'  \widehat{Q}^{-1}  \mathbb{E}_{NT}[\widehat{V}_{it} (U_{it} + R_{it}(\widehat{d}, \widehat{l})] \\
 &=\alpha'  Q^{-1}  \mathbb{E}_{NT}V_{it} U_{it} \\
 &  \quad +\alpha'  (\widehat{Q}^{-1} - Q^{-1} )\mathbb{E}_{NT}V_{it} U_{it} + \alpha'  \widehat{Q}^{-1}[\mathbb{E}_{NT}[\widehat{V}_{it} (U_{it} + R_{it}(\widehat{d}, \widehat{l}) - V_{it} U_{it}  ]  ] \\
 &=:\alpha'  Q^{-1}  \mathbb{E}_{NT}V_{it} U_{it} + S_1(\alpha) + S_2(\alpha).
    \end{align*}
    The bound on $S_1 (\alpha)$ follows
    \begin{align*}
    |S_1 (\alpha)| &\leq  \| \widehat{Q}^{-1} - Q^{-1} \| \| \mathbb{E}_{NT} V_{it} U_{it} \| \\
    &\leq  \| \widehat{Q}^{-1} \| \| \widehat{Q}-Q \| \|Q^{-1} \|  \| \mathbb{E}_{NT} V_{it} U_{it} \| \\
    &= O_P (1) \cdot o_P(1) \cdot O_P(1) \cdot O_P ( (NT)^{-1/2}) =  o_{P}((NT)^{-1/2}),
    \end{align*}
where $O_P(\cdot)$ and $o_P(\cdot)$ bounds are established in Steps 0-2.
       The bound on $S_2(\alpha)$ 
       follows from:
    \begin{align*}
      |S_2 (\alpha)| &\leq    \| \alpha \| C_{\min}^{-1} \| \widehat S - S \| \lesssim_P (r_{2NT} + \chi_{NT}) = o_P ( (NT)^{-1/2}),
    \end{align*}
where we are using the results of Step 0. As a result, 
\begin{align*}
  \sqrt{NT}  \alpha' (\widehat{\beta}_{OLS}- \beta_0 ) =\alpha'  Q^{-1}  \mathbb{G}_{NT}V_{it} U_{it} + o_P(1),
\end{align*}
which gives \eqref{eq:olspointwise}.

\textbf{ Step 3. } The proof of pointwise normality follows similarly to Step 1 of the proof of Theorem \ref{thrm:DOL}, where the step \eqref{eq:mainbound2} is replaced by
$$
| \alpha' (\alpha' \Sigma \alpha)^{-1/2} R_{NT} | \leq \| \alpha \|_2 O(1) \|R_{NT} \|_2 \lesssim_P \sqrt{NT}(r_{2NT} + \chi_{NT} ) = o_P(1).
$$

\end{proof}

\bibliography{my_new_bibtex.bib}
\bibliographystyle{apalike}
\end{document}